%% file: ms.tex
\documentclass[letterpaper, 10 pt, conference]{IEEEtran}  % Comment this line out if you need a4paper

\IEEEoverridecommandlockouts                              % This command is only needed if 
\usepackage[noend]{algorithmic}
\usepackage{balance}
\usepackage{hyperref}
\usepackage{subcaption}
\usepackage{graphicx}
\usepackage{mathematix}
\usepackage{amsmath}
\usepackage{bm}
\usepackage[capitalize]{cleveref}
\usepackage{todonotes}
\usepackage{mathdots}

\newcommand{\ccirc}{\mathbin{\mathchoice
  {\xcirc\scriptstyle}
  {\xcirc\scriptstyle}
  {\xcirc\scriptscriptstyle}
  {\xcirc\scriptscriptstyle}
}}
\newcommand{\xcirc}[1]{\vcenter{\hbox{$#1\circ$}}}

\usepackage{pgf}
\input{ourmacros.tex}

\title{\LARGE \bf
PV-OSIMr: A Lowest Order Complexity Algorithm for Computing the Delassus Matrix
}

\author{Ajay Suresha Sathya$^{1}$, Wilm Decr\'e$^{2}$ and Jan Swevers$^{2}$% <-this % stops a space
\thanks{The authors gratefully acknowledge funding from Research Foundation Flanders (FWO) (Grant agreement No. G0D1119N) and Flanders Make SBO project ARENA.}% <-this % stops a space
\thanks{$^{1}$~Inria - Département d’Informatique de l’École normale supérieure, PSL Research University.  $^{2}$~Division of Robotics, Automation and
Mechatronics in the Department of Mechanical Engineering, KU Leuven, and FlandersMake@KULeuven, Leuven, Belgium
        {\tt\small ajay.sathya@inria.fr}, {\tt\small wilm.decre@kuleuven.be}, {\tt\small jan.swevers@kuleuven.be}}%
}

\begin{document}

\maketitle
\thispagestyle{empty}
\pagestyle{empty}

%%%%%%%%%%%%%%%%%%%%%%%%%%%%%%%%%%%%%%%%%%%%%%%%%%%%%%%%%%%%%%%%%%%%%%%%%%%%%%%%
\begin{abstract}

   We present PV-OSIMr, an efficient algorithm for computing the Delassus matrix (also known as the inverse operational space inertia matrix) for a kinematic tree, with the lowest order computational complexity known in literature. PV-OSIMr is derived by optimizing the recently proposed PV-OSIM algorithm using the compositionality of the force and motion propagators. It has a computational complexity of $O(n+m^2)$ compared to $O(n + m^2d)$ of the PV-OSIM algorithm and $O(n+md +m^2)$ of the extended force propagator algorithm (EFPA), where $n$ is the number of joints, $m$ is the number of constraints and $d$ is the depth of the kinematic tree. Since the Delassus matrix is an $m \times m$ sized matrix and its computation must consider all the $n$ joints, PV-OSIMr's asymptotic computational complexity is optimal. We further benchmark our algorithm and find it to be often more efficient than the PV-OSIM and EFPA in practice.

\end{abstract}

%%%%%%%%%%%%%%%%%%%%%%%%%%%%%%%%%%%%%%%%%%%%%%%%%%%%%%%%%%%%%%%%%%%%%%%%%%%%%%%%
\section{Introduction}

% \textcolor{red}{Picture of a humanoid robot with shadow hands doing push-ups.}
\input{introduction.tex}

\section{Background}
\input{PV_OSIM.tex}

\section{PV-OSIMr algorithm}
\input{low_complexity.tex}

\section{Results and discussions} \label{sec:results}
\input{results.tex}

\section{Conclusions} \label{sec:conclusion}
\input{conclusions.tex}


%%%%%%%%%%%%%%%%%%%%%%%%%%%%%%%%%%%%%%%%%%%%%%%%%%%%%%%%%%%%%%%%%%%%%%%%%%%%%%%%

%%%%%%%%%%%%%%%%%%%%%%%%%%%%%%%%%%%%%%%%%%%%%%%%%%%%%%%%%%%%%%%%%%%%%%%%%%%%%%%%
\balance

\bibliographystyle{IEEEtran}
\bibliography{references}

\end{document}

%% file: ourmacros.tex
\newcommand{\pkg}[1]{\textsc{#1}}

%% file: introduction.tex
%\textcolor{red}{Central to mechanics.}

The Delassus matrix~\cite{brogliato2016nonsmooth}, also known as the inverse operational space inertia matrix (OSIM), represents the inertial coupling between different constraints on a system of rigid bodies. It is a fundamental physical quantity with applications in robotics and computer graphics such as simulating constrained dynamics~\cite{featherstone2014rigid,carpentier2021proximal}, simulating contact dynamics~\cite{duriez2005realistic,felis2017rbdl,raisim}, operational space control~\cite{khatib1987unified}, computing the dynamically consistent pseudo-inverse~\cite{khatib1987unified,dietrich2015overview}  as well as solving inverse dynamics of floating-base robots~\cite{righetti2013optimal}. The Delassus matrix is named after Etienne Delassus \cite{delassus1917memoire} who ``first studied the unilateral contact problem with multiple constraints''~\cite{brogliato2016nonsmooth}. The Delassus matrix and inverse OSIM terms have different origins in constrained dynamics~\cite{brogliato2016nonsmooth} and operational space control (where constraints are virtual) literature respectively~\cite{khatib1987unified}. These terms refer to the apparent inertia in a given space (whether operational space or constraint space), and are therefore mathematically identical. In this paper, where the main concern is algorithms to compute this quantity, we will use the terms Delassus matrix and inverse OSIM interchangeably.

%The Delassus matrix is named ``in honor of Etienne Delassus \cite{delassus1917memoire} who was the first to deeply analyze the unilateral contact problem with multiple constraints''\cite{brogliato2016nonsmooth}.

Computing the Delassus matrix naively from its mathematical definition~\cite{khatib1987unified} through dense factorization of the joint space inertia matrix (JSIM) has a high computational complexity of $O(n^3 + m^2n)$, where $n$ is the degrees-of-freedom (DoF) of the mechanism in generalized coordinates and $m$ is the constraint dimension. Considering the centrality of the Delassus matrix in computationally demanding applications like model predictive control (MPC) and reinforcement learning (RL), where it is often the most expensive step \cite{sathya_constrained_dynamics} of dynamics computation, an  efficient Delassus algorithm is highly desirable and there exists a long tradition of research in this direction.

\begin{table}[t]
    \caption{Computational complexity of various algorithms for inverse OSIM computation.}
    \label{table_complexity}
    \begin{center}
    \begin{tabular}{|c||c|}
    \hline
    Algorithm & Complexity\\
    \hline
    Naive & $O(n^3 + m^2n)$\\
    LTL-OSIM \cite{featherstone2010exploiting} & $O(nd^2 + m^2d + md^2)$\\
    KJR \cite{kreutz1992recursive} & $O(n + m^2d)$\\
    EFPA \cite{wensing2012reduced} & $O(n + md + m^2)$\\
    PV-OSIM \cite{sathya_constrained_dynamics} & $O(n + m^2d)$\\
    \textbf{PV-OSIMr} (this paper) & $\boldsymbol{O(n + m^2)}$\\
    \hline
    \end{tabular}
    \end{center}
    \end{table}
    
The Kreutz-Jain-Rodriguez (KJR) algorithm~\cite{kreutz1992recursive} represented a major advance by proposing an $O(n + m^2d)$ complexity recursive Delassus algorithm, where $d$ is the kinematic tree's depth. This algorithm was further improved in~\cite{chang2001efficient} by re-using certain computations to obtain $O(n + mn + m^2)$ complexity. By exploiting extended force propagators (EFP), the EFP algorithm (EFPA)~\cite{wensing2012reduced} significantly speeds up computing the off-diagonal block of the Delassus matrix, and has the lowest known computational complexity of $O(n + md + m^2)$ in literature. Compared to the complex recursive algorithms mentioned above, \cite{featherstone2010exploiting} presents the simpler LTL-OSIM algorithm that improves upon the naive method by exploiting the branching-induced sparsity in both the JSIM and the constraint Jacobian. Despite its worse complexity of $O(nd^2 + m^2d + md^2)$, the LTL-OSIM algorithm was found~\cite{featherstone2010exploiting} to be significantly faster than KJR and competitive with EFPA~\cite{wensing2012reduced,sathya_constrained_dynamics} even for the ASIMO robot (a high DoF humanoid robot with favorable branching structure). Recently, the LTL-OSIM was extended to support kinematic closed loops in \cite{carpentier2021proximal}, which also provided an efficient \pkg{C++} implementation in the \pkg{Pinocchio} library~\cite{carpentier2019pinocchio}. %This sparsity exploiting method was found to be more efficient than the EFPA as well in , for the ASIMO robot, while being less efficient on longer robots due to its worse complexity.  

In our recent work~\cite{sathya_constrained_dynamics}, we showed that an existing constrained dynamics algorithm due to Popov and Vereshchagin (PV)~\cite{popov1978manipuljacionnyje,vereshchagin1989modeling} computed the Delassus matrix as an intermediate quantity, thereby providing another Delassus algorithm that we called PV-OSIM. The PV-OSIM algorithm  has an efficient computation structure requiring only two sweeps compared to three sweeps required by KJR and EFPA. Though PV-OSIM has a higher complexity of $O(n + m^2d)$ compared to EFPA's $O(n + md + m^2)$ complexity, it was found~\cite{sathya_constrained_dynamics} to be significantly more efficient than the KJR, EFPA and LTL-OSIM for a wide range of robots such as quadrupeds (Go1) and humanoids (Atlas and Talos). 

However, because of its lower complexity, EFPA scales better than PV-OSIM for long mechanisms with many constraints that need to be propagated through most of the tree's depth. Then the $m^2d$ term in PV-OSIM loses out EFPA's $md$ term. This letter explores whether the PV-OSIM's efficiency can be retained for smaller robots while still scaling better than EFPA to higher DoF robots with many constraints. %Other potential approaches that we mention for the sake of completeness can, in theory, compute OSIM efficiently by leveraging parallelism include the divide and conquer methods proposed in \cite{bhalerao2014distributed, featherstone1999divide, yamane2009comparative}. However, they are typically very complex to implement and are not faster than the single threaded algorithms 

\subsection{Approach and contributions}

By using the compositionality of EFP, we optimize the PV-OSIM algorithm~\cite{sathya_constrained_dynamics} to propose a new $O(n + m^2)$ complexity Delassus algorithm called PV-OSIMr. This is the lowest order algorithm that we are aware of in the literature (see~\cref{table_complexity}).  Presenting this algorithm and computationally benchmarking it is the main contribution of this paper.

PV-OSIMr (also EFPA and PV-OSIM) however assumes that the constraints are imposed on individual links and would require non-trivial modification for constraints involving multiple links, e.g. a constraint on the center-of-mass (CoM). Supporting such a constraint is relatively easier for the joint-space algorithms like LTL-OSIM~\cite{featherstone2010exploiting}.

\subsubsection*{Organization} We introduce the preliminaries and kinematic tree notation in the rest of this section. Then we review LTL-OSIM, extended propagators, PV-OSIM and EFPA in \cref{sec:background} before presenting the PV-OSIMr algorithm in \cref{sec:PV-OSIMr}. Identical notation used for all the three recursive algorithms (PV-OSIM, EFPA and PV-OSIMr) facilitates comparing them. Finally, we present numerical benchmarking results in \cref{sec:results} and conclude in \cref{sec:conclusion}.

% add a table with computational complexity of various algorithms

\subsection{Preliminaries} \label{sec:prelim}

Constrained dynamics involves solving the equations
\begin{subequations}
    \begin{align}
M(\mathbf{q}) \boldsymbol{\dot{\nu}} + \mathbf{c}(\mathbf{q}, \boldsymbol{\nu}) + J(\mathbf{q})^T \boldsymbol{\lambda} = \boldsymbol{\tau}, \label{eq:eom_js}\\
J(\mathbf{q}) \boldsymbol{\dot{\nu}} + \dot{J}(\mathbf{q}, \boldsymbol{\nu}) \boldsymbol{\nu} = \mathbf{k}(\mathbf{q}, \boldsymbol{\nu}), \label{eq:eom_cons}
    \end{align}
\end{subequations}
where $M(\mathbf{q}) \in \mathbb{R}^{n \times n}$ is the symmetric positive definite JSIM, $\mathbf{c}(\mathbf{q}, \boldsymbol{\nu}) \in \mathbb{R}^n$ includes the Coriolis, centrifugal and gravitational force  terms, $\boldsymbol{\tau} \in \mathbb{R}^n$ is joint torques, $J(\mathbf{q}) \in \mathbb{R}^{m \times n}$ is the constraint Jacobian matrix, $\boldsymbol{\lambda} \in \mathbb{R}^m$ is Lagrange multipliers denoting the constraint force magnitude, $\mathbf{k} \in \mathbb{R}^m$ is the desired constraint acceleration and $\dot{J}(\mathbf{q}, \boldsymbol{\nu}) \in \mathbb{R}^{m \times n}$ is $J(\mathbf{q})$'s time-derivative. The functional dependence of the expressions will be dropped for brevity from now on.

Solving for $\boldsymbol{\dot{\nu}}$ in \cref{eq:eom_js} and substituting in \cref{eq:eom_cons} gives
\begin{equation}
    \Lambda^{-1} \boldsymbol{\lambda} = \dot{J}\boldsymbol{\nu} - \mathbf{k} + J M^{-1} (\boldsymbol{\tau} - \mathbf{c}), \label{eq:cons_dyn}
\end{equation}
where $\Lambda(\mathbf{q})^{-1} := J M^{-1} J^T$ is the Delassus matrix, which maps constraint force magnitudes to constraint accelerations and captures the inertial coupling between constraints.

While the equations above were described in the joint-space, they can also be equivalently described at the link level in `maximal' coordinates. We use the standard Featherstone's spatial algebra~\cite{featherstone2014rigid} to refer to a rigid-body's physical quantities. Let $\mathbf{a}_i \in \mathbb{M}^{6}$ and $H_i \in \mathbb{I}^{6}$ represent the spatial acceleration and spatial inertia of the $i$th link respectively. $S_i$ is the $i$th joint's motion subspace matrix of size ${6 \times n_i}$, with each column in $\mathbb{M}^6$, where $n_i$ the $i$th joint's DoF. The constraint on a link $i$ can be expressed as
\begin{equation}
    K_i(\mathbf{q}) \mathbf{a}_i = \mathbf{k}_i(\mathbf{q}, \boldsymbol{\nu}), \label{eq:link_cons} 
\end{equation}
where $K_i(\mathbf{q}) \in \mathbb{R}^{m_i \times 6}$ is the $i$th link's constraint matrix and $\mathbf{k}_i(\mathbf{q}, \boldsymbol{\nu}) \in \mathbb{R}^{m_i}$ is the desired $i$th link's constraint acceleration vector. For more information on spatial algebra and for solving constrained dynamics expressed at the link level, readers are referred to \cite{featherstone2014rigid} and \cite{sathya_constrained_dynamics} respectively. 

\subsection{Kinematic tree notation}

%We mean 6D spatial forces and spatial accelerations whenever we refer to link accelerations and forces in this paper. $:=$ operator defines the left hand side symbol with the right hand side expression. $\vert . \vert$ denotes the cardinality of a set. 

Let all the links of a kinematic tree with $n_b$ bodies be indexed from 1 to $n_b$, with $i < j$ if the $i$th link is an ancestor of the $j$th link in the tree. The $i$th link's parent joint is also indexed with the number $i$. The inertial world frame is indexed to be the $0$th link. For floating-base trees, the $1$st joint, connecting the 0th and 1st links, is a free-flyer joint that permits motion in all 6 directions. Let $\pi(i)$ index the $i$th link's parent link and $\gamma(i)$ is the set of all the $i$th link's children links' indices. Let $m_b$ denote the number of motion constraints. To facilitate future notation, we adopt the clever approach from~\cite{wensing2012reduced} and introduce a fictitious child link for each constraint. This fictitious link is assumed to be rigidly attached to the parent constrained link with coinciding frames. Let these fictitious links, which we call end-effectors, be indexed as $\mathcal{E} = \{n_b + 1, n_b + 2, \hdots n_b+m_b\}$.

Let $\mathcal{S}=\{1, 2, \hdots n_b \}$ be an ordered set and $\mathcal{S}_r$ its reverse. Let anc$(i)$ and desc$(i)$ be the set of all link indices that are ancestors and descendants of the $i$th link in the tree respectively. Let ES$(i) = \{j \in \mathcal{E} | j \in \mathrm{desc}(i)\}$, be the set indexing all end-effectors supported by the $i$th link. Let cca$(i,j) = \mathrm{\textbf{max}} \left(\left(\mathrm{anc}(i) \cup \{i\}\right)\cap\left(\mathrm{anc}(j) \cup \{j\}\right)\right)$, define the closest common ancestor of the $i$th and $j$th links. Let $\mathcal{N} = \{0\} \cup \mathcal{E} \cup \{i \in \mathcal{S} | \mathrm{ES}(j) \subset \mathrm{ES}(i), \forall j \in \gamma(i)\}$, define a set of `branching links'. Let $\mathcal{A}(i) = \textrm{\textbf{max}}  \,\left(\mathrm{anc}(i) \cap \mathcal{N}\right)$ index the closest ancestor branching link, and likewise let $\mathcal{D}(i) = \textrm{\textbf{min}}  \,\left(\left(\mathrm{desc}(i)\cup \{i\}\right) \cap \mathcal{N}\right)$ index the closest descendant branching link. Let path($i, j$) be the set of all joints between $i$th and $j$th links.

\begin{figure}
    \vspace{0.4cm}
        \centering
        \includegraphics[width=0.3\textwidth]{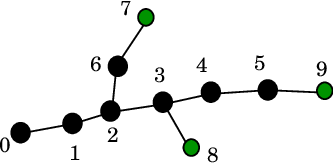} 
        \caption{Schematic diagram of a kinematic tree, nodes are the links, edges are the joints. The green nodes are the fictitious links representing end-effectors.}
        \label{fig:tree_schematic}
\end{figure}

We will illustrate the terms introduced above using a 6-link tree shown in~\cref{fig:tree_schematic}. Three of its links are constrained, to represent which, three fictitious end-effectors $\mathcal{E} = \{7, 8, 9\}$ are introduced (indicated by green nodes). For this tree, $\mathcal{S} = \{1,2,3,4,5,6\}$, anc$(8) = \{ 0, 1, 2, 3\}$. As examples of ES, ES$(5) = \{9\}$, ES$(3) = \{ 8, 9\}$ and ES$(2) = \{ 7, 8,9\}$. cca$(8,9) = 3$ and cca$(7,9) = 2$. The branching links are $\mathcal{N} = \{ 0, 2, 3, 7, 8, 9\}$. The branching link ancestors and descendants of each link are as follows
\begin{equation*}
\begin{bmatrix} i \\ \mathcal{D}(i) \\ \mathcal{A}(i) \end{bmatrix} :    
\begin{bmatrix} 
1 & 2 & 3 & 4 & 5 & 6 & 7 & 8 & 9 \\ 
2 & 2 & 3 & 9 & 9 & 7 & 7 & 8 & 9 \\
0 & 0 & 2 & 3 & 3 & 2 & 2 & 3 & 3 
\end{bmatrix}.
\end{equation*}

%% file: PV_OSIM.tex
\label{sec:background}
We now introduce the LTL-OSIM, PV-OSIM and EFPA algorithms and relevant physical quantities along the way. All the spatial motion vectors, force vectors and inertia matrices are expressed in and with respect to a fixed inertial frame to make the algorithm presentation less cluttered. 
\subsection{LTL-OSIM algorithm}

The LTL-OSIM algorithm \cite{featherstone2010exploiting} is the simplest among the existing efficient OSIM algorithms. It exploits the branching-induced sparsity in $M$ and $J$ matrices.  Let 
\begin{equation}
L^TL = M,
\end{equation}
be the LTL Cholesky decomposition of the JSIM $M$, where $L$ is a lower triangular matrix. Note that the LTL decomposition above differs from the traditional LLT Cholesky decomposition. LTL  preserves $M$'s sparsity pattern in $L$ \cite{featherstone2005efficient} when joints are numbers from root to leaves. 
Let 
\begin{equation}
    Y = JL^{-1},
\end{equation}
where $Y$ shares $J$'s sparsity pattern. The inverse OSIM $\Lambda^{-1}$ is given by 
\begin{equation}
    \Lambda^{-1} = YY^T.
\end{equation}
Despite  $\Lambda^{-1}$ being a dense matrix in general, exploiting sparsity in $M$, $L$, $J$ and $Y$ matrices makes  LTL-OSIM efficient and competitive with lower complexity algorithms for kinematic trees up to a certain size.

\subsection{Extended propagators}

%\textcolor{red}{Constraint forces of the link $i \in \mathcal{E}$ are $\boldsymbol{\lambda}$ while the constraint accelerations are $K_i\mathbf{a}_i$}

We now review the force and motion propagators used later in this paper. Let $H_i^A$ denote the articulated body inertia (ABI) matrix \cite{featherstone1983calculation}, which is the apparent inertia felt at the $i$th link considering the subtree rooted at the $i$th link with all its descendant joints being free to move. Projecting the ABI onto the $i$th joint's motion subspace yields
\begin{equation}
    D_i := S_i^TH_i^AS_i,
\end{equation}
which is the $i$th link's apparent inertia felt at the $i$th joint. Let
\begin{equation}
    ^{\pi(i)}P_i := (I_{6\times 6} - H_i^AS_i(D_i)^{-1} S_i^T),
\end{equation}
denote the projection matrix, that that transmits spatial forces acting on the $i$th link to its parent link after removing the force component that causes the $i$th joint's motion. Trivially, $^iP_i := I_{6 \times 6}$. For any end-effector link $i \in \mathcal{E}$, $^{\pi(i)}P_i := I_{6 \times 6}$ since the virtual end-effector link is assumed to be rigidly attached to its parent link.

It is long known \cite{lilly1989efficient} that $^{\pi(i)}P_i$'s transpose $^{\pi(i)}P_i^{T}$ (also a projection matrix) is the motion propagator, that transmits accelerations acting on the parent link $\pi(i)$ to the child link $i$ through the $i$th joint. The force propagators can be composed~\cite{chang2001efficient} to define extended force propagators (EFP)
\begin{equation}
    ^jP_i := {^jP_{\pi \ccirc \hdots \ccirc \pi(i)}} \hdots {^{\pi(\pi(i))}P_{\pi(i)}} {^{\pi(i)}P_{i}}, \label{eq:extended_proj_operator}
\end{equation}
that directly transmit spatial forces backward from the $i$th link to an ancestral link $j \in \mathrm{anc}(i)$. Analogously, $^jP_i^T$ is the extended motion propagator (EMP) that transmits spatial accelerations forward from the $j$th link to the $i$th link. Let  
\begin{equation}
    ^{\pi(i)}\Omega_i := S_i(D_i)^{-1}S_i^T, \label{eq:Omega_i}
\end{equation}
where ${^{\pi(i)}}\Omega_i \in \mathbb{R}^{6 \times 6}$ denotes the $i$th link's apparent spatial inverse inertia if the $\pi(i)$ is grounded. Analogously to the EFPs, we now denote $^j\Omega_i$ as the  $i$th link's apparent spatial inverse inertia when an ancestral link $j \in \mathrm{anc}(i)$ is grounded, computing which requires accumulating contributions to the $i$th link's acceleration due to the motion of all joints in $k \in \mathrm{path}(j,i)$ caused by the force acting on the $i$th link. Each such joint $k$'s contribution is obtained by back propagating the $i$th link's force to the $k$th link using EFP, multiplying it with $^{\pi(k)}\Omega_k$ to  compute the resulting $k$th link's acceleration and forward propagating this acceleration term to the $i$th link using EMP. Adding these terms gives 
\begin{equation} \label{eq:apparent_spatial_inertia}
    ^{j}\Omega_i :=   \sum_{k \in \text{path}(j, i)}{^kP_i^T} \ ^{\pi(k)}\Omega_k \ {^kP_i}.
\end{equation}

To facilitate Delassus matrix computation, which is the constraint space inverse inertia, we now introduce constraint space force and motion propagators.
Let 
 \begin{equation}
    ^{j}K_i := K_i \  {{^j}P_i^T}, \label{eq:extended_Kpi}
\end{equation}
where, $^jK_i$ is the constraint space EMP (CEMP) that forward propagates the $j$th link's spatial acceleration to the $i$th link's (an end-effector) constraint space accelerations. Analogously, $^{j}K_i^{T}$ is the constraint space EFP (CEFP), which back propagates the $i$th end-effector's  constraint force magnitudes ($\boldsymbol{\lambda}_i$) to the $j$th link's spatial forces. Let
\begin{equation}
    ^{k}\Lambda^{-1}_{i, j} :=  {^{\text{cca}(i, j)}{K_i}} \ ^k\Omega_{\text{cca}(i,j)} \ ^{\text{cca}(i, j)}{K_j}^{T}, \label{eq:extended_Lpij}
\end{equation}
where $^{k}\Lambda^{-1}_{i, j} \in \mathbb{R}^{m_i \times m_j}$, denotes the cross-coupling constraint space inverse inertia mapping $\boldsymbol{\lambda}_j$ to the $i$th link's spatial acceleration, if a link $k \in \mathrm{anc}(\mathrm{cca}(i,j))$ is grounded. %In other words, $^{k}L_{i, j}^A$ maps the $j$th link's constraint force magnitudes $\boldsymbol{\lambda}^A_j$ to the $i$th link's constraint space accelerations $K_i^A \mathbf{a}_i$ considering the subtree rooted at the $k$th link. % due to the motion the joints in path($k$, $\text{cca}(i, j)$).

%of all  constraints on end effector links supported by the $i$th link denoted by $ES(i)$, $^{j}K_i^A \in \mathbb{R}^{m_i \times 6}$, is the CEMP that transmits $j$th link's spatial accelerations ($j \in \text{anc(i)}$) to constraint space accelerations of the constraints supported by the $i$th link. 

We now introduce notation used by PV-OSIM that aggregates terms related to all end-effectors in ES$(i)$. Let $K_i^A = [\hdots {^iK_j^T} \hdots]^T, \, \mathrm{for} \, j \in \mathrm{ES}(i)$.  Similarly, let \begin{equation} \label{eq:inverse_OSIM}
    L_i^A := \begin{bmatrix}
     %& \hdots &  & \hdots  &  \\
    %\vdots & 
    \ddots & \vdots & \iddots  \\
    % & 
    \hdots & ^i\Lambda^{-1}_{j, k} & \hdots\\
    %\vdots &
     \iddots & \vdots & \ddots \\
    %& \hdots & & \hdots & 
    \end{bmatrix}
\end{equation}
aggregate all the constraint space inverse inertias for all end-effectors $j,k \in \mathrm{ES}(i)$ for a mechanism grounded at the $i$th link, with its matrix blocks being the $^i\Lambda^{-1}_{j, k}$ terms. Considering the whole tree (where only the ground inertial link $0$ is grounded) gives the Delassus matrix 
\begin{equation}
\Lambda^{-1} = L_0^A. 
\end{equation}
\subsection{PV-OSIM algorithm}

Algorithm~\ref{alg:PV_OSIM} lists the PV-OSIM \cite{sathya_constrained_dynamics} algorithm for completeness. 
%$K_i^A$, computed in line~\ref{line:Kpi_recursion} is the row concatenation of $^iK_j^A \ \forall j \in ES(i)$. $L_i^A$ is the inverse OSIM for constraints $ES(i)$ and the subtree rooted at the $i$th link, consisting of $^iL_{j, k}^A \ \forall j,k \in ES(i)$ as its subblocks. It is computed recursively in lines \ref{eq:Lpi_recursion} by computing the constraint accelerations due to each of the constraint forces. It achieves this by computing the spatial force on the $i$th joint due to constraint forces using CEFP, computes the $i$th link's accelerations due to these forces and propagates this accelerations to the end-effectors to obtain the constraint acclerations using CEMP. \textcolor{red}{This is done at each step to accumulate the inverse OSIM.}
The main computational bottleneck of PV-OSIM is the line~\ref{eq:Lpi_recursion} which requires $O(m^2d)$ number of operations, while line~\ref{line:Kpi_recursion} requires $O(md)$ operations, because computing lines~\ref{eq:Lpi_recursion} and~\ref{line:Kpi_recursion} for all links at a given depth $k$ from the root requires at most $O(m^2)$ and $O(m)$ operations respectively. All the other lines of the algorithm require $O(n)$ or $O(m)$ number of operations, bringing the total complexity  to $O(n + m^2d)$.

\begin{algorithm}
    \caption{The PV-OSIM algorithm}\label{alg:PV_OSIM}%
    \begin{algorithmic}[1]
      \REQUIRE
      \ $\mathbf{q}$,\
      \ $K_i$s,\
      \ robot model\
  
      \textbf{First forward sweep}
      \FOR {$i$ in $\mathcal{S}$}
      \STATE $ H_i^A \gets H_i; \, K_i^A \gets \left[ \quad \right]; \, L_i^A \gets \left[ \quad \right];$
      \ENDFOR{}
      \textbf{Backward sweep}
      \FOR {$i$ in $\mathcal{E}_r $}
      %\STATE $^{\pi(i)}P_i = I_{6\times 6}$
      \STATE $K_{\pi(i)}^A \gets \begin{bmatrix} K_{\pi(i)}^A \\  K_i \end{bmatrix}$
      \STATE $L^A_{\pi(i)} \gets \begin{bmatrix} L^A_{\pi(i)} & \\ & 0_{m_i \times m_i} \end{bmatrix}$ 
      \ENDFOR{}
      \FOR {$i$ in $\mathcal{S}_r $}
      \STATE $D_i = S_i^TH_i^AS_i; \, ^{\pi(i)}\Omega_i = S_iD_i^{-1}S_i^T $
      \IF {$\pi(i) > 0$}
      \STATE $^{\pi(i)}P_i = I_{6\times 6} - H_i^A \, ^{\pi(i)}\Omega_i$
      \STATE $H_{\pi(i)}^A \gets H_{\pi(i)}^A + {^{\pi(i)}P_iH_{i}^A}$
      \ENDIF{}
      \IF {ES$(i) \neq \varnothing $} 
      \IF {$\pi(i) > 0$}
      \STATE $K_{\pi(i)}^A \gets \begin{bmatrix} K_{\pi(i)}^A \\  K_i^A \, {^{\pi(i)}P_i^T} \end{bmatrix}$ \label{line:Kpi_recursion}
      \ENDIF{}
      \STATE $L^A_{\pi(i)} \gets \begin{bmatrix} L^A_{\pi(i)} & \\ & L^A_{i}  + K_i^A \,  ^{\pi(i)}\Omega_i K_i^{AT} \end{bmatrix}$ \label{eq:Lpi_recursion}
      \ENDIF {}
      \ENDFOR{}
      \STATE $\Lambda^{-1} = L_0^A$\\
    \end{algorithmic}
\end{algorithm}

\subsection{Extended force propagator algorithm (EFPA)}

Algorithm~\ref{alg:EFPA} presents the EFPA. While the recursive CEMP computation in EFPA in line~\ref{line:EFPA_Kpi} appears different from PV-OSIM's line~\ref{line:Kpi_recursion}, these two lines are computationally identical. Only the EFPA maintains each end-effector's CEMP separately instead of concatenating them to enable efficient Delassus matrix assembly later. The main difference between PV-OSIM and EFPA arises in line~\ref{line:EFPA_Lambdaij}, where EFPA recursively computes the term $\left(^0\Omega_i {^iK}^{T}_j\right) \in \mathbb{R}^{6 \times m_j}$, which maps each $j$th end-effector's constraint force magnitudes $\boldsymbol{\lambda}_j$ to the $i$th link's spatial acceleration for all ancestors $i \in \mathrm{anc}(j)$. This requires $O(md)$ operations. Then $\frac{m_b (m_b -1)}{2}$ number of $^0\Lambda^{-1}_{i,j}$ blocks are computed in line~\ref{line:EFPA_offdiagonal} using EFP, bringing EFPA's total computational complexity to $O(n + md + m^2)$, because $m \propto m_b$. 

\begin{algorithm}
    \caption{The EFPA algorithm}\label{alg:EFPA}%
    \begin{algorithmic}[1]
      \REQUIRE
      \ $\mathbf{q}$,\
      \ $K_i$s,\
      \ robot model\
  
      \textbf{First forward sweep}
      \FOR {$i$ in $\mathcal{S}$}
      \STATE $  H_i^A \gets H_i; $
      \ENDFOR{}
      \textbf{Backward sweep}
      \FOR {$i$ in $\mathcal{E}_r $}
      %\STATE $^{\pi(i)}P_i = I_{6\times 6}$
      \STATE $^{\pi(i)}K_i \gets K_i $

      \ENDFOR{}
      \FOR {$i$ in $\mathcal{S}_r $}
      \STATE $D_i = S_i^TH_i^AS_i; \ ^{\pi(i)}\Omega_i = S_iD_i^{-1}S_i^T$
      \IF {$\pi(i) > 0$}
      \STATE $^{\pi(i)}P_i = I_{6\times 6} - H_i^AS_i(D_i)^{-1} S_i^T$
      \STATE $H_{\pi(i)}^A \gets H_{\pi(i)}^A + {^{\pi(i)}P_i}H_{i}^A$ \label{line:EFPA_Hpi}
      \FOR {$j \in ES(i)$}
        \STATE ${^{\pi(i)}K_j} = {^iK_j} \ {^{\pi(i)}P_i^T} $ \label{line:EFPA_Kpi}
      \ENDFOR{}
      \ENDIF {}
      \ENDFOR{}
      \textbf{Second forward sweep}%forward pass
        \FOR {$i \in \mathcal{S}$}
        \FOR {$j \in ES(i)$}
            \IF {$\pi(i) \neq 0$}
            \STATE $\left(^0\Omega_i {^iK_j^T}\right) = {^{\pi(i)}P^T_i} (^0\Omega_{\pi(i)} {^{\pi(i)}K_j^T}) + {^{\pi(i)}\Omega_i} ^iK_j^{T}$ \label{line:EFPA_Lambdaij}
            \ELSE 
            \STATE $\left(^0\Omega_i {^iK_j^T}\right) = {^{0}\Omega_i} \ ^iK_j^{T}$
            \ENDIF {}
        \ENDFOR{}
        \ENDFOR{}
        \textbf{Assembling the Delassus matrix}
        \FOR{$i, j \in \mathcal{E}$, if $i \leq j$}
          \STATE $^0\Lambda_{i,j}^{-1} = {^{cca(i, j)}K_i} \left(^0\Omega_{cca(i, j)} {^{cca(i, j)}K_j^T}\right) $ \label{line:EFPA_offdiagonal}
        \ENDFOR{}
    \end{algorithmic}
\end{algorithm}

%% file: low_complexity.tex
\label{sec:PV-OSIMr}

Equipped with the previous section's notation, we now adapt the PV-OSIM algorithm to obtain the PV-OSIMr algorithm. We first describe the insight that reduces  PV-OSIM's complexity and follow that with the PV-OSIMr algorithm.

\noindent \textbf{Insight behind the reduced complexity:} \label{sec:insight}
The lines~\ref{line:Kpi_recursion} and~\ref{eq:Lpi_recursion} of \cref{alg:PV_OSIM} are evaluated over \text{all} end-effector supporting links wastefully leading to $O(md)$ and  $O(m^2d)$ operations respectively. However, from the line~\ref{line:EFPA_offdiagonal} from~\cref{alg:EFPA}, we see that we need to compute $^iK_j$ for end-effectors $j$ only at select ancestral branching links $i$ ($i \in \mathcal{N} \cap \mathrm{anc}(j)$), and likewise we need inverse inertias $^0\Omega_i$ only at the same select branching links $i$.  % if the total dimensionality of constraints supported by the $i$th link (number of rows in $K_i^A$) is more than six, one can simply compute the EFPs $^jP_i$ for $i$th link's ancestors instead of the CEFPs, which requires only $O(d)$ operations. The CEFP can then be recovered from EFP using \cref{eq:extended_Kpi}.

\begin{figure}[h]
  \begin{subfigure}[h]{0.49\linewidth} 
    \centering
    \includegraphics[width=\textwidth]{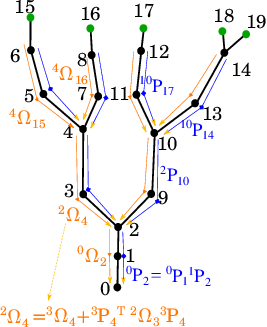}  
    \caption{First backward sweep (lines~\ref{line:first_backward_sweep_start} to~\ref{line:first_backward_sweep_end} of~\cref{alg:PV_OSIMr}). The spatial inverse inertias and EMPs are computed between the branching links in $O(n)$ operations using~\cref{eq:Omegai_recursion} and~\cref{eq:EMP_recursion} respectively.}\label{fig:backward_sweeps}
  \end{subfigure}
  \begin{subfigure}[h]{0.49\linewidth} 
    \centering
    \includegraphics[width=\textwidth]{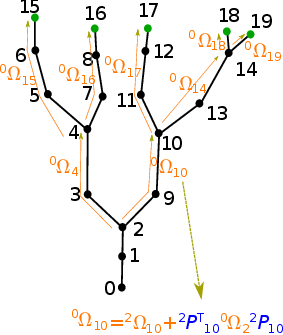}  
    \caption{Forward sweep over only the branching links $\mathcal{N}$ to compute the total spatial inverse inertia of the branching links using~\cref{eq:total_spatial_inv_inertia} in $O(n)$ operations.}\label{fig:forward_sweep_inv_inertias}
  \end{subfigure}
  \begin{subfigure}[h]{0.49\linewidth} 
    \centering
    \includegraphics[width=\textwidth]{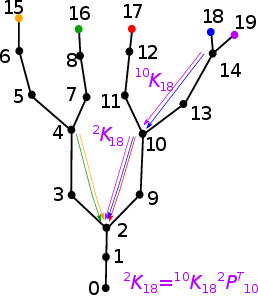}  
    \caption{Backward sweep over $\mathcal{N}$.}\label{fig:backward_K_sweep}
  \end{subfigure}
  \begin{subfigure}[h]{0.49\linewidth} 
    \centering
    \includegraphics[width=\textwidth]{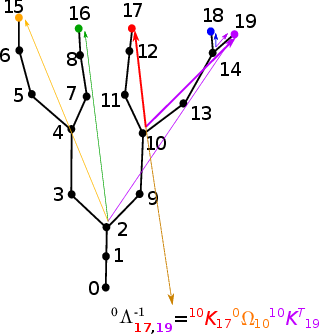}  
    \caption{Assemble Delassus blocks.}\label{fig:assemble_delassus}
  \end{subfigure}
  \caption{Graphical illustration of the different computation sweeps in~\cref{alg:PV_OSIMr}.}\label{fig:sweeps}
\end{figure}

To compute these terms efficiently, we first recursively compute the EFPs and spatial inverse inertia between only the branching links in a backward sweep over all the links $k \in \mathcal{S}_r$ in $O(n)$ operations, as shown in~\cref{fig:backward_sweeps}  using 
\begin{align} \label{eq:Omegai_recursion}
    ^{\pi(k)}\Omega_{\mathcal{D}(k)} &= {^{k}\Omega_{\mathcal{D}(k)}} + {^kP_{\mathcal{D}(k)}^T} S_k\left(D_k^{-1}\right)S_k^T \ {^kP_{\mathcal{D}(k)}}, \\
    ^{\pi(k)}P_{\mathcal{D}(k)} &= {^{\pi(k)}P_{k}} {^{k}P_{\mathcal{D}(k)}}. \label{eq:EMP_recursion}
\end{align}

Then in a limited forward sweep over branching links $i \in \mathcal{N}$ as shown in~\cref{fig:forward_sweep_inv_inertias}, the $^0\Omega_i$ terms are computed using 
\begin{equation}\label{eq:total_spatial_inv_inertia}
  ^0\Omega_i = {^{\mathcal{A}(i)}\Omega_i} + {^{\mathcal{A}(i)}P_i^T \, {^0\Omega_{\mathcal{A}(i)}} \, {^\mathcal{A}(i)}P_i },
\end{equation}
in $O(n)$ operations. Then for each end-effector $j$, the CEMPs $^iK_j$  are computed from ancestral branching links $i  \in \left(\mathcal{N} - \{0\}\right) \cap \mathrm{anc}(j)$ recursively in a backward sweep over branching links as shown in~\cref{fig:backward_K_sweep} using 
\begin{equation}
  ^{\mathcal{A}(i)}K_j = {^{i}K_j} {^{\mathcal{A}(i)}P^T_{i}},
\end{equation}
in at most $O(m^2)$ operations, because the number of branching links (excluding end-effectors and the root link) cannot exceed the number of end-effectors. Finally, $\Lambda^{-1}$ is assembled by calling~\cref{eq:extended_Lpij} $m_b^2$ times to compute each of its blocks ($^0\Lambda_{i,j}^{-1}$) in $O(m^2)$ as shown in~\cref{fig:assemble_delassus}. This brings the total complexity of PV-OSIMr to $O(n + m^2)$ operations. %Each block $^{0}L_{i, j}^A, \ \forall i, j \in \mathcal{E}$ of the inverse OSIM blocks can then be recovered using \cref{eq:extended_Lpij}
% \begin{equation} \label{eq:off_diagonal_terms}
%     ^{0}L_{i, j}^A =  \ ^{\text{cca}(i,j)}K_{i}^A \ ({^{0}}\Omega_{\text{cca}(i,j)}) \  ^{\text{cca}(i,j)}K_{j}^{AT},
% \end{equation}
% in further $O(m^2)$ operations. The equation above implies that the CEFP $^{\text{cca}(i,j)}K_{i}^A$ and the spatial inverse inertias $({^{0}}\Omega_{\text{cca}(i,j)})$ need to be computed only for the links that are $\mathrm{cca}(i,j), \forall i,j \in \mathcal{E}$, which is the same as the list $\mathcal{N}$. 

%\subsection{Algorithmic form}

The PV-OSIMr algorithm is presented in~\cref{alg:PV_OSIMr}. The algorithm is slightly optimized in line~\ref{line:losim_notation_abuse} with an abuse of notation by assigning the CEMP to EMP at the end-effector links. This avoids the unnecessary computation of EMP between an end-effector and its ancestral branching link. Furthermore, CEMP and constraint-space inverse inertias are directly available at the end-effectors in lines~\ref{line:EE_CEMP} and \ref{line:EE_delassus_block} respectively, thereby avoiding computations for transforming EMPs and spatial inverse inertias to the constraint space.

\begin{algorithm}
  \vspace{0.3cm}
    \caption{The PV-OSIMr algorithm}\label{alg:PV_OSIMr}%
    \begin{algorithmic}[1]
      \REQUIRE
      \ $\mathbf{q}$,\
      \ $K_i$s,\
      \ robot model\
  
      \textbf{First forward sweep}
      \FOR {$i$ in $\mathcal{S}$}
      \STATE $H_i^A \gets H_i; $ \label{line:first_forward_sweep}
      \ENDFOR{}
      \FOR {$i \in \mathcal{E}$}
      \STATE $^{\pi(i)}P_i \gets K_i^T$; \ $^{\pi(i)}\Omega_i \gets 0_{m_i \times m_i}$~\label{line:losim_notation_abuse}
      \ENDFOR{}
      \textbf{Backward sweep}
      \FOR {$i$ in $\mathcal{S}_r $} \label{line:first_backward_sweep_start}
      \IF {$i \in \mathcal{N}$}
      \STATE $ ^iP_i = I_{6 \times 6}; \ ^i\Omega_i \gets 0_{6 \times 6}$ 
      \ENDIF{}
      \STATE $D_i = S_i^TH_i^AS_i; \, ^{\pi(i)}\Omega_i = S_iD_i^{-1}S_i^T $
      \IF {$\pi(i) > 0$}
      \STATE $^{\pi(i)}P_i = I_{6\times 6} - H_i^A \, ^{\pi(i)}\Omega_i$ 
      \STATE $H_{\pi(i)}^A \gets H_{\pi(i)}^A + {^{\pi(i)}P_i}H_{i}^A$
      \IF {ES$(i) \neq \varnothing$}
      \STATE ${^{\pi(i)}P_{\mathcal{D}(i)}} = {^{\pi(i)}P_i} \, {^iP_{\mathcal{D}(i)}}$ 
      \ENDIF {}
      \ENDIF {}
      \IF {ES$(i) \neq \varnothing$}
      \STATE $^{\pi(i)}\Omega_{\mathcal{D}(i)} = {^{i}\Omega_{\mathcal{D}(i)}} + {^iP^T_{\mathcal{D}(i)}} \ ^{\pi(i)}\Omega_i \ {^iP_{\mathcal{D}(i)}}$
      \ENDIF {}
      \ENDFOR{} \label{line:first_backward_sweep_end}
      \textbf{Forward sweep over $\mathcal{N}$}
      \FOR{$i$ in $\mathcal{N} - \{0\}$ if $\mathcal{A}(i) \neq 0$}
      \STATE $^0\Omega_i = {^{\mathcal{A}(i)}\Omega_i} + {^{\mathcal{A}(i)}P_i^T} \ ^0\Omega_{\mathcal{A}(i)} \ {^{\mathcal{A}(i)}P_i}$ \label{line:forward_Omega}
      \ENDFOR{}
      \textbf{Backward sweep over $\mathcal{N}$}
      \FOR{$i \in \mathcal{E}$}
      \IF{$\mathcal{A}\left(i\right) \neq 0$}
      \STATE $^{\mathcal{A}(i)}K_i = {^{\mathcal{A}(i)}P^T_i}$ \label{line:EE_CEMP}
        \FOR{$j$ in $(\mathcal{N}_r - \{0\})$ if $j \in \mathrm{anc}(i), \,\mathcal{A}(j) \neq 0,$}
        \STATE $^{\mathcal{A}(j)}K_{i} = \ ^jK_{i} \ {^{\mathcal{A}(j)}P_j^T}$ \label{line:backward_K}
        \ENDFOR{}
        \ENDIF{}
      \ENDFOR{}
      \textbf{Assemble the Delassus matrix}
      \FOR{$i, j \in \mathcal{E}, \ i \leq j$}
      \IF{ $i < j$}
        \STATE $^{0}\Lambda^{-1}_{i, j} =  \ ^{\text{cca}(i,j)}K_{i} \ {^{0}}\Omega_{\text{cca}(i,j)} \  ^{\text{cca}(i,j)}K_{j}^{T}$ \label{line:off_diagonal}
       \STATE $^{0}\Lambda^{-1}_{j, i} = {^{0}\Lambda^{-T}_{i, j}}$
        \ELSE
        \STATE $^{0}\Lambda^{-1}_{i, i} \gets {^{0}\Omega_{i}}$ \label{line:EE_delassus_block}
        \ENDIF {}
      \ENDFOR{}
    \end{algorithmic}
\end{algorithm}

%% file: results.tex
\subsection{Implementation}

Similarly to \cite{sathya_constrained_dynamics}, we have implemented the PV-OSIMr algorithm using the \pkg{CasADi} \cite{andersson2019casadi} library. Apart from enabling efficient C code generation for the dynamics algorithms, \pkg{CasADi} implementation enables us to count the number of operations in a given algorithm. We will use this operation count to benchmark PV-OSIMr with our implementation of PV-OSIM, EFPA and LTL-OSIM algorithms. 
The algorithms presented in this paper assumed that all the quantities are expressed in an inertial frame to avoid the clutter due to spatial transforms. However, the OSIM algorithms (as well as the ABA algorithm \cite{featherstone2014rigid}, \cite{otter1987algorithm}) are more efficiently implemented in the local frame. Therefore, our \pkg{CasADi} implementation is also in the local frame. Readers are referred to the source code\footnote{https://github.com/AjSat/spatial\_V2} of the implementation for further details on the local frame implementation. Note that a \pkg{C++} implementation of PV-OSIMr is also made available in the \pkg{Pinocchio} library \cite{carpentier2019pinocchio} and can be found in the {\tt computePvDelassusMatrix()} function of the {\tt delassus.hxx} file\footnote{\href{https://github.com/stack-of-tasks/pinocchio/blob/master/include/pinocchio/algorithm/delassus.hxx}{https://github.com/stack-of-tasks/pinocchio/blob/master/include\\/pinocchio/algorithm/delassus.hxx}}. However, the {\tt C++} implementation represents physical quantities in inertial frame instead of the more efficient local frame, and improving its efficiency by switching to local frames will be our future work.

\subsection{Computational scaling study}

\begin{figure*}[h!] 
  \begin{subfigure}[h]{0.32\linewidth} 
    \centering
    \includegraphics[width=\textwidth]{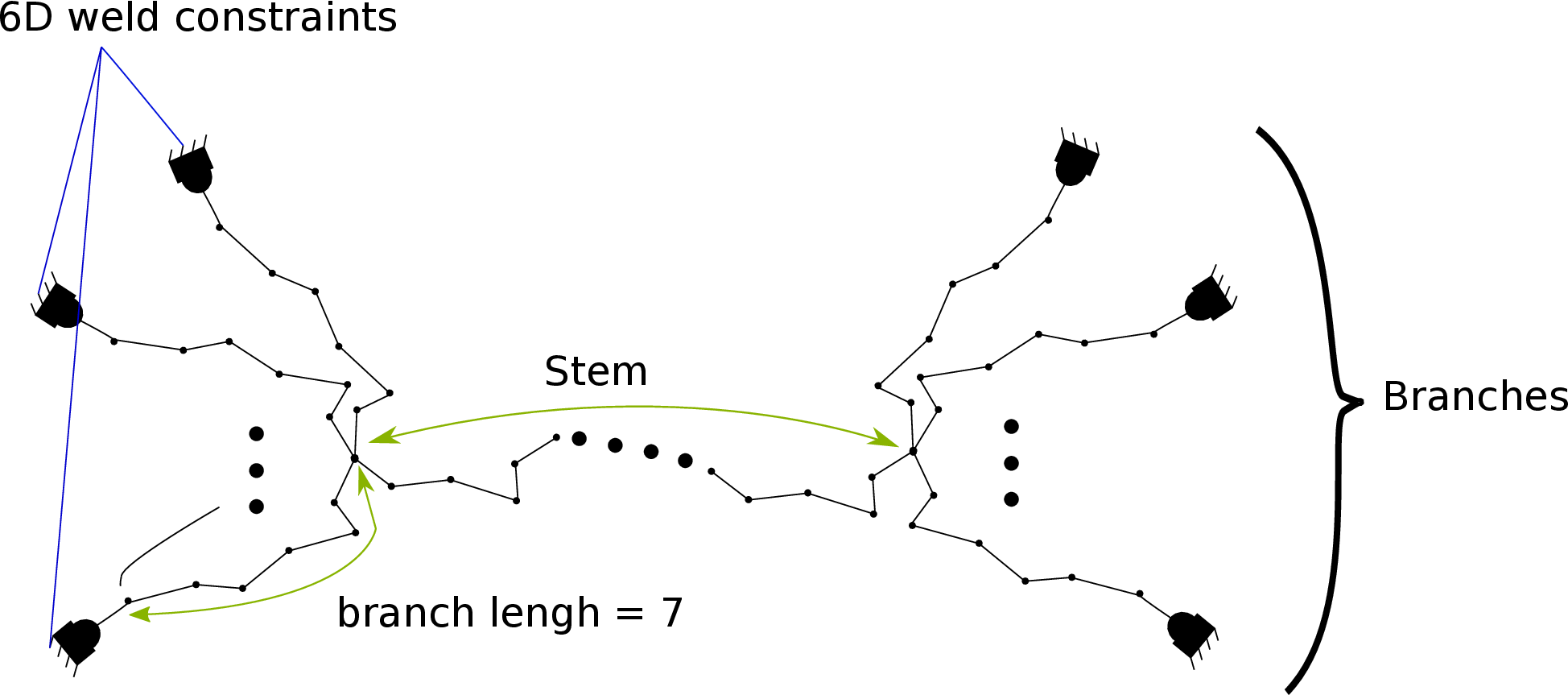} 
    \caption{The mechanism on which the OSIM scaling is benchmarked.}
    \label{fig:stalk}
\end{subfigure} 
    \begin{subfigure}[h]{0.32\linewidth} 
        \centering
        \resizebox{\linewidth}{!}{\input{graphics/tree_comparison1.pgf}} 
        \caption{One branch on each side (a chain structure).}
        \label{fig:OSIM_stalk1}
    \end{subfigure} 
        % \hspace{-1.05cm}
    \begin{subfigure}[h]{0.32\linewidth}
          \centering
          \resizebox{\linewidth}{!}{\input{graphics/tree_comparison2.pgf}}
          \caption{Two branches on either side.}
          \label{fig:OSIM_stalk2}
    \end{subfigure}
    \begin{subfigure}[h]{0.32\linewidth}
            \centering
            \resizebox{\linewidth}{!}{\input{graphics/tree_comparison4.pgf}}   
            \caption{Four branches on either side.}
            \label{fig:OSIM_stalk4}
    \end{subfigure} 
    \begin{subfigure}[h]{0.32\linewidth}
        \centering
        \resizebox{\linewidth}{!}{\input{graphics/tree_comparison7.pgf}}
        \caption{Seven branches on either side.}
        \label{fig:OSIM_stalk7}
      % \end{center}
    \end{subfigure}
        \begin{subfigure}[h]{0.32\linewidth}
          \centering
          \resizebox{\linewidth}{!}{\input{graphics/tree_comparison10.pgf}}
          \caption{Ten branches on either side.}
          \label{fig:OSIM_stalk10}
        % \end{center}
          \end{subfigure}
        %   \begin{subfigure}[h]{0.45\linewidth}
        %     \centering
        %     \includegraphics[width=\textwidth]{graphics/stalk.eps}
        %     \caption{The long stalk on which the computational scaling is studied.}
        %     \label{fig:OSIM_stalk}
        %   % \end{center}
        %     \end{subfigure} \hspace{1.7cm}
        %   \begin{subfigure}[h]{0.45\linewidth}
        %     \centering
        %     \resizebox{\linewidth}{!}{\input{graphics/OSIM_long_tree.pgf}}
        %     \caption{Computational scaling of PV-OSIM vs EFPA}
        %     \label{fig:OSIM_long_tree}
        %   % \end{center}
        %     \end{subfigure}  
      \caption{Benchmarking the number of operations of the different OSIM algorithms.}
      \label{fig:OSIM_benchmarks}
    \end{figure*}

We now analyze the computational scaling of the recursive Delassus algorithms (PV-OSIM, EFPA and PV-OSIMr) w.r.t. increasing tree depth, number of branches and constraints. Consider a mechanism (see \cref{fig:stalk}) with a long stem and branches on either side, with each branch consisting of 7 links. Each branch tip is rigidly constrained with a {\tt weld} constraint (6D constraint). This mechanism is chosen because it is a simple mechanism with the property that path lengths, along which the computation sweep over, increases linearly with the stem length for any choice of floating base link. The operation count of the different algorithms for mechanisms with different number of branches is plotted in \cref{fig:OSIM_benchmarks}, with the x-axis denoting the stem length. 

When there is only one branch, the mechanism is a serial chain with 6D constraints on each end, and there is no difference between the PV-OSIMr and PV-OSIM algorithms as the dimensionality of the propagated constraints does not exceed 6 to benefit from PV-OSIMr's optimization w.r.t to PV-OSIM. For two or more branches, PV-OSIMr's advantage becomes apparent as it scales significantly better than PV-OSIM due to its lower computational complexity.  EFPA is more expensive than both PV-OSIMr and PV-OSIM for nearly all the considered mechanisms. As discussed extensively in~\cite{sathya_constrained_dynamics}, despite having a lower asymptotic complexity than PV-OSIM, EFPA is more expensive for small mechanisms with few constraints ($< \sim 50$) because it has a three sweep structure compared to the two sweep structure of PV-OSIM, and it computes inverse inertia in each link's frame (see line~\ref{line:EFPA_Lambdaij} in \cref{alg:EFPA}) which requires an expensive similarity transformation during the second forward sweep. %The similarity transformation is avoided in both PV-OSIM and the PV-OSIMr which compute the inverse inertias already in the task-frame. 
However, for long enough mechanisms with sufficiently many constraints the EFPA eventually becomes faster than the PV-OSIM (see \cref{fig:OSIM_stalk10}) due to its lower computational complexity. %Finally, we note that PV-OSIMr was more efficient than both PV-OSIM and EFPA, which are the state-of-the-art inverse OSIM algorithms.

\begin{figure}[h]
  \begin{subfigure}[h]{0.49\linewidth} 
    \centering
    \resizebox{\linewidth}{!}{\input{graphics/chain_md.pgf}} 
    \caption{Studying scaling when the $md$ term dominates.}\label{fig:chain_md}
  \end{subfigure}
  \begin{subfigure}[h]{0.49\linewidth} 
    \centering
    \resizebox{\linewidth}{!}{\input{graphics/chain_all_con.pgf}} 
    \caption{Comparing algorithms when every link is an end-effector.}\label{fig:chain_all_con}
  \end{subfigure}
  \caption{Comparing the scaling of the recursive algorithms on more mechanisms.}\label{fig:chain}
\end{figure}
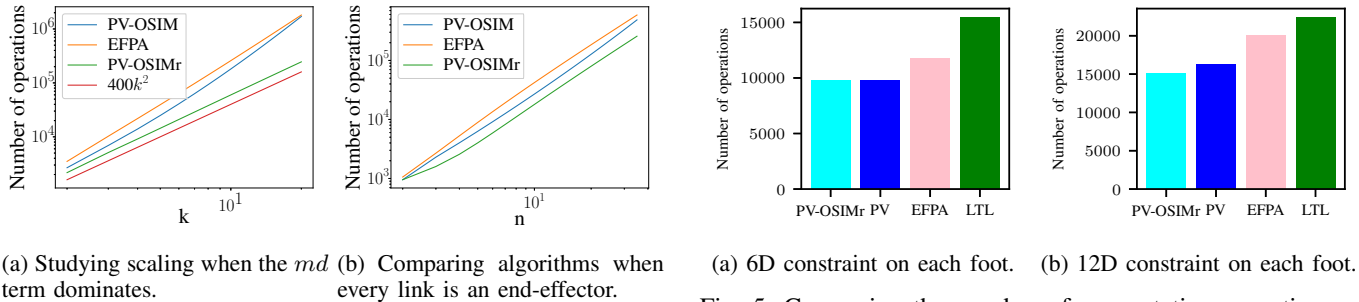
\noindent \textbf{Scaling when the $md$ term dominates:}

While the examples considered above showcase PV-OSIMr's efficiency, they do not empirically demonstrate its lower complexity. Therefore, we now consider mechanisms where the $md$ term dominates. Consider a serial chain with  $n_b = k^2$ bodies and let the links $\{k, 2k \hdots k^2\}$ be constrained with 6D constraints. For this mechanism, $n = k^2$, $d = k^2$ and $m = 6k$, $md = 6k^3$ and $m^2 = 36k^2$. Therefore the $md$ term dominates as $k$ increases and PV-OSIMr (with $O(n + m^2)$) scales as $O(k^2)$, EFPA (with $O(n + md + m^2)$) scales as $O(k^3)$, and PV-OSIM(with $O(n + m^2d)$) scales as $O(k^4)$. The number of operations required by different algorithms for increasing values of $k$ is plotted in a log-log plot in~\cref{fig:chain_md}. PV-OSIMr clearly has the same slope as the $O(k^2)$ function while EFPA has a higher slope due to the $md$ term.

\noindent \textbf{Scaling when $m \propto d$:}

We now study the undesirable scenario for PV-OSIMr, when all the links are in $\mathcal{N}$. Consider a chain with a 6D constraint on all its links. In this case ($n=d=\frac{m}{6}$), both EFPA and PV-OSIMr have the same asymptotic complexity and scale quadratically with $n$, while PV-OSIMr scales cubically. The results for different algorithms for a chain of increasing sizes is plotted in a log-log plot in~\cref{fig:chain_all_con}. PV-OSIMr and EFPA are found to have the same slope as expected. However, PV-OSIMr remains significantly faster than EFPA even in this example, mainly because of the way EFPA  propagates a separate inverse inertia term for every supported end-effector (see line~\ref{line:EFPA_Lambdaij} in \cref{alg:EFPA}) requiring $O(md)$ operations. In contrast, PV-OSIMr computes all inverse inertia terms in only $O(n)$ operations. 
\subsection{Humanoid robot standing constraints}\label{sec:humanoid_standing}

% \textcolor{red}{Provide a computational breakdown for the different sub parts of the algorithms for a select robot examples.}

% add graphics/atlas_comparison.pgf plot from graphics folder below
\begin{figure}[h]
  \begin{subfigure}[h]{0.49\linewidth} 
    \centering
    \resizebox{\linewidth}{!}{\input{graphics/Atlas_6D.pgf}} 
    \caption{6D constraint on each foot.}\label{fig:one_foot}
  \end{subfigure}
  \begin{subfigure}[h]{0.49\linewidth} 
    \centering
    \resizebox{\linewidth}{!}{\input{graphics/Atlas_12D.pgf}} 
    \caption{12D constraint on each foot.}\label{fig:both_feet}
  \end{subfigure}
  \caption{Comparing the number of computation operations of the OSIM algorithms for the Atlas robot.}\label{fig:atlas}
\end{figure}

We now compare the different OSIM algorithms (PV-OSIMr, PV-OSIM, EFPA and LTL) for a humanoid robot Atlas and the results are tabulated in \cref{fig:atlas}. The number of operations required to compute the inverse OSIM when the Atlas robot is standing, which is modeled as each of its feet being rigidly connected to the world with a 6D {\tt weld} constraint is plotted in \cref{fig:one_foot}. The hip link to which both the legs are connected is selected as the floating base. There is no difference between PV-OSIM and PV-OSIMr in this case as the two 6D constraints meet only at the floating-base link during the backward propagation and therefore does not benefit from the PV-OSIMr's improvement.  

Next, we consider the example where the foot-ground constraint is modelled as {\tt connect}-type 3D constraints at four contact points as is commonly done to model foot contact wrench cone~\cite{caron2015stability}. PV-OSIM propagates 12D constraints from each foot while PV-OSIMr only propagates 6D constraints leading to a reduction in computation cost with PV-OSIM requiring 19.9\% more operations than PV-OSIMr. For both examples, the PV-OSIM and PV-OSIMr significantly outperform EFPA and LTL. 

\subsection{Humanoid robot with shadow hands}

% add graphics/atlas_comparison.pgf plot from graphics folder below
\begin{figure*}[h!]
  \begin{subfigure}[h]{0.24\linewidth} 
    \centering
    \resizebox{\linewidth}{!}{\input{graphics/SR.pgf}} 
    \caption{Considering a single shadow hand, a 24 d.o.f system. Connect constraint (3D) on each fingertip leading to 15 constraints.}\label{fig:SR}
  \end{subfigure}
    \begin{subfigure}[h]{0.24\linewidth} 
      \centering
      \resizebox{\linewidth}{!}{\input{graphics/atlas_left_hand.pgf}} 
      \caption{Shadow hand connected to Atlas's left wrist with same constraint as in \cref{fig:SR} (total 15 constraints on a 60 d.o.f system).}\label{fig:single_hand}
    \end{subfigure}
    \begin{subfigure}[h]{0.24\linewidth} 
      \centering
      \resizebox{\linewidth}{!}{\input{graphics/atlas_both_hands.pgf}} 
      \caption{Shadow hand connected to both wrists of Atlas with each hand having the same constraints as in \cref{fig:SR} (total 30 constraints on an 84 d.o.f system).}\label{fig:atlas_both_hands}
    \end{subfigure}\label{fig:both_hands}
    \begin{subfigure}[h]{0.24\linewidth} 
      \centering
      \resizebox{\linewidth}{!}{\input{graphics/atlas_standing_both_hands.pgf}} 
      \caption{Atlas with both hands constrained as in \cref{fig:atlas_both_hands} along with 6D {\tt weld} constraint on each foot for standing. (total 42 constraints on an 84 d.o.f system).}\label{fig:both_hands_standing}
    \end{subfigure}
    \caption{Comparing the number of computation operations of the OSIM algorithms for the Atlas robot equipped with the shadow hand.}\label{fig:shadow}
\end{figure*}
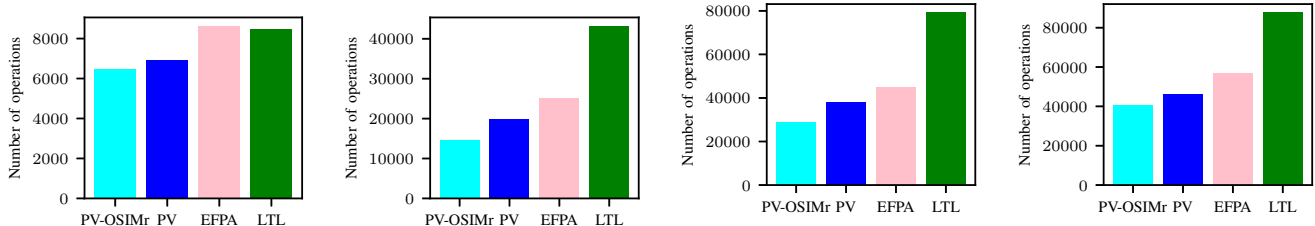

Finally, we compare the OSIM algorithms for an Atlas robot equipped with the Shadow hand  in \cref{fig:shadow}. We impose a 3D {\tt connect} type constraint on each fingertip of the hand leading to a total of 15D constraints. We plot the OSIM results for i) just a single Shadow hand, ii) an Atlas robot with a single shadow hand connected to its left wrist, iii) an Atlas robot with a shadow hand connected to each wrist, and iv) finally an Atlas robot with both its feet constrained with 6D {\tt weld} constraints on each foot in addition to  the 15D constraint on each shadow hand in \cref{fig:SR}, \cref{fig:single_hand}, \cref{fig:atlas_both_hands} and \cref{fig:both_hands_standing} respectively.

The typically expensive LTL-OSIM is computationally more efficient than EFPA for just the Shadow Hand in \cref{fig:SR} because the LTL-OSIM benefits from the sparsity induced by the extensive branching of the hand's mechanism. When the Shadow hand is connected to the Atlas robot in the remaining three examples, the performance of the high complexity LTL-OSIM significantly deteriorates compared to the lower complexity algorithms PV-OSIMr, PV-OSIM and EFPA. The PV-OSIMr algorithm is found to be significantly more efficient than the other algorithms with up to $\sim$34\% fewer operations needed than the next most efficient algorithm PV-OSIM.

\subsection{Caveats}

While PV-OSIMr outperforms PV-OSIM in all the examples presented so far, we note that there were some cases when PV-OSIM was more efficient. This occurs especially when more than six constraints need to be propagated in PV-OSIM through only a few ($\leq 3$) single DoF joints, because for single DoF joints PV-OSIM recursively computes the OSIM through efficient symmetric rank-1 (SR1)  updates per joint (by computing the vector $K_i^A S_i$, see line~\ref{eq:Lpi_recursion} in \cref{alg:PV_OSIM}). PV-OSIMr instead performs two symmetric rank-6 computations (see line~\ref{line:off_diagonal} and line~\ref{line:forward_Omega} in \cref{alg:PV_OSIMr}), which are more expensive than SR1 updates over a few joints. For example, if a 6D {\tt weld} constraint is imposed on both the hands (without the Shadow Hand) in addition to the 6D constraints on the feet of Atlas considered in \cref{sec:humanoid_standing}, the PV-OSIM requires $\sim$1.5\% fewer operations than PV-OSIMr, because PV-OSIM's simpler backward pass remains more efficient when performed over just three joints  (from  sternum to the floating base hip link).

% \subsection{Limitations}

% One of the limitations of the presented algorithm is that it is not fully parallelizable because the sweeps within a branch are computed recursively and hence depend on sequential computation. This is unlike the DCA/ADA algorithms that are fully parallelizable. However, the computations of the presented algorithm over different branches can be parallelized.

%% file: graphics/tree_comparison1.pgf
%% Creator: Matplotlib, PGF backend
%%
%% To include the figure in your LaTeX document, write
%%   \input{<filename>.pgf}
%%
%% Make sure the required packages are loaded in your preamble
%%   \usepackage{pgf}
%%
%% Figures using additional raster images can only be included by \input if
%% they are in the same directory as the main LaTeX file. For loading figures
%% from other directories you can use the `import` package
%% and then include the figures with
%%   \import{<path to file>}{<filename>.pgf}
%%
%% Matplotlib used the following preamble
%%
\begingroup%
\makeatletter%
\begin{pgfpicture}%
\pgfpathrectangle{\pgfpointorigin}{\pgfqpoint{6.400000in}{4.800000in}}%
\pgfusepath{use as bounding box, clip}%
\begin{pgfscope}%
\pgfsetbuttcap%
\pgfsetmiterjoin%
\definecolor{currentfill}{rgb}{1.000000,1.000000,1.000000}%
\pgfsetfillcolor{currentfill}%
\pgfsetlinewidth{0.000000pt}%
\definecolor{currentstroke}{rgb}{1.000000,1.000000,1.000000}%
\pgfsetstrokecolor{currentstroke}%
\pgfsetdash{}{0pt}%
\pgfpathmoveto{\pgfqpoint{0.000000in}{0.000000in}}%
\pgfpathlineto{\pgfqpoint{6.400000in}{0.000000in}}%
\pgfpathlineto{\pgfqpoint{6.400000in}{4.800000in}}%
\pgfpathlineto{\pgfqpoint{0.000000in}{4.800000in}}%
\pgfpathclose%
\pgfusepath{fill}%
\end{pgfscope}%
\begin{pgfscope}%
\pgfsetbuttcap%
\pgfsetmiterjoin%
\definecolor{currentfill}{rgb}{1.000000,1.000000,1.000000}%
\pgfsetfillcolor{currentfill}%
\pgfsetlinewidth{0.000000pt}%
\definecolor{currentstroke}{rgb}{0.000000,0.000000,0.000000}%
\pgfsetstrokecolor{currentstroke}%
\pgfsetstrokeopacity{0.000000}%
\pgfsetdash{}{0pt}%
\pgfpathmoveto{\pgfqpoint{1.501490in}{0.964913in}}%
\pgfpathlineto{\pgfqpoint{6.100000in}{0.964913in}}%
\pgfpathlineto{\pgfqpoint{6.100000in}{4.500000in}}%
\pgfpathlineto{\pgfqpoint{1.501490in}{4.500000in}}%
\pgfpathclose%
\pgfusepath{fill}%
\end{pgfscope}%
\begin{pgfscope}%
\pgfsetbuttcap%
\pgfsetroundjoin%
\definecolor{currentfill}{rgb}{0.000000,0.000000,0.000000}%
\pgfsetfillcolor{currentfill}%
\pgfsetlinewidth{0.803000pt}%
\definecolor{currentstroke}{rgb}{0.000000,0.000000,0.000000}%
\pgfsetstrokecolor{currentstroke}%
\pgfsetdash{}{0pt}%
\pgfsys@defobject{currentmarker}{\pgfqpoint{0.000000in}{-0.048611in}}{\pgfqpoint{0.000000in}{0.000000in}}{%
\pgfpathmoveto{\pgfqpoint{0.000000in}{0.000000in}}%
\pgfpathlineto{\pgfqpoint{0.000000in}{-0.048611in}}%
\pgfusepath{stroke,fill}%
}%
\begin{pgfscope}%
\pgfsys@transformshift{1.668286in}{0.964913in}%
\pgfsys@useobject{currentmarker}{}%
\end{pgfscope}%
\end{pgfscope}%
\begin{pgfscope}%
\definecolor{textcolor}{rgb}{0.000000,0.000000,0.000000}%
\pgfsetstrokecolor{textcolor}%
\pgfsetfillcolor{textcolor}%
\pgftext[x=1.668286in,y=0.867691in,,top]{\color{textcolor}\fontsize{20.000000}{24.000000}\selectfont \(\displaystyle {0}\)}%
\end{pgfscope}%
\begin{pgfscope}%
\pgfsetbuttcap%
\pgfsetroundjoin%
\definecolor{currentfill}{rgb}{0.000000,0.000000,0.000000}%
\pgfsetfillcolor{currentfill}%
\pgfsetlinewidth{0.803000pt}%
\definecolor{currentstroke}{rgb}{0.000000,0.000000,0.000000}%
\pgfsetstrokecolor{currentstroke}%
\pgfsetdash{}{0pt}%
\pgfsys@defobject{currentmarker}{\pgfqpoint{0.000000in}{-0.048611in}}{\pgfqpoint{0.000000in}{0.000000in}}{%
\pgfpathmoveto{\pgfqpoint{0.000000in}{0.000000in}}%
\pgfpathlineto{\pgfqpoint{0.000000in}{-0.048611in}}%
\pgfusepath{stroke,fill}%
}%
\begin{pgfscope}%
\pgfsys@transformshift{2.723959in}{0.964913in}%
\pgfsys@useobject{currentmarker}{}%
\end{pgfscope}%
\end{pgfscope}%
\begin{pgfscope}%
\definecolor{textcolor}{rgb}{0.000000,0.000000,0.000000}%
\pgfsetstrokecolor{textcolor}%
\pgfsetfillcolor{textcolor}%
\pgftext[x=2.723959in,y=0.867691in,,top]{\color{textcolor}\fontsize{20.000000}{24.000000}\selectfont \(\displaystyle {25}\)}%
\end{pgfscope}%
\begin{pgfscope}%
\pgfsetbuttcap%
\pgfsetroundjoin%
\definecolor{currentfill}{rgb}{0.000000,0.000000,0.000000}%
\pgfsetfillcolor{currentfill}%
\pgfsetlinewidth{0.803000pt}%
\definecolor{currentstroke}{rgb}{0.000000,0.000000,0.000000}%
\pgfsetstrokecolor{currentstroke}%
\pgfsetdash{}{0pt}%
\pgfsys@defobject{currentmarker}{\pgfqpoint{0.000000in}{-0.048611in}}{\pgfqpoint{0.000000in}{0.000000in}}{%
\pgfpathmoveto{\pgfqpoint{0.000000in}{0.000000in}}%
\pgfpathlineto{\pgfqpoint{0.000000in}{-0.048611in}}%
\pgfusepath{stroke,fill}%
}%
\begin{pgfscope}%
\pgfsys@transformshift{3.779631in}{0.964913in}%
\pgfsys@useobject{currentmarker}{}%
\end{pgfscope}%
\end{pgfscope}%
\begin{pgfscope}%
\definecolor{textcolor}{rgb}{0.000000,0.000000,0.000000}%
\pgfsetstrokecolor{textcolor}%
\pgfsetfillcolor{textcolor}%
\pgftext[x=3.779631in,y=0.867691in,,top]{\color{textcolor}\fontsize{20.000000}{24.000000}\selectfont \(\displaystyle {50}\)}%
\end{pgfscope}%
\begin{pgfscope}%
\pgfsetbuttcap%
\pgfsetroundjoin%
\definecolor{currentfill}{rgb}{0.000000,0.000000,0.000000}%
\pgfsetfillcolor{currentfill}%
\pgfsetlinewidth{0.803000pt}%
\definecolor{currentstroke}{rgb}{0.000000,0.000000,0.000000}%
\pgfsetstrokecolor{currentstroke}%
\pgfsetdash{}{0pt}%
\pgfsys@defobject{currentmarker}{\pgfqpoint{0.000000in}{-0.048611in}}{\pgfqpoint{0.000000in}{0.000000in}}{%
\pgfpathmoveto{\pgfqpoint{0.000000in}{0.000000in}}%
\pgfpathlineto{\pgfqpoint{0.000000in}{-0.048611in}}%
\pgfusepath{stroke,fill}%
}%
\begin{pgfscope}%
\pgfsys@transformshift{4.835304in}{0.964913in}%
\pgfsys@useobject{currentmarker}{}%
\end{pgfscope}%
\end{pgfscope}%
\begin{pgfscope}%
\definecolor{textcolor}{rgb}{0.000000,0.000000,0.000000}%
\pgfsetstrokecolor{textcolor}%
\pgfsetfillcolor{textcolor}%
\pgftext[x=4.835304in,y=0.867691in,,top]{\color{textcolor}\fontsize{20.000000}{24.000000}\selectfont \(\displaystyle {75}\)}%
\end{pgfscope}%
\begin{pgfscope}%
\pgfsetbuttcap%
\pgfsetroundjoin%
\definecolor{currentfill}{rgb}{0.000000,0.000000,0.000000}%
\pgfsetfillcolor{currentfill}%
\pgfsetlinewidth{0.803000pt}%
\definecolor{currentstroke}{rgb}{0.000000,0.000000,0.000000}%
\pgfsetstrokecolor{currentstroke}%
\pgfsetdash{}{0pt}%
\pgfsys@defobject{currentmarker}{\pgfqpoint{0.000000in}{-0.048611in}}{\pgfqpoint{0.000000in}{0.000000in}}{%
\pgfpathmoveto{\pgfqpoint{0.000000in}{0.000000in}}%
\pgfpathlineto{\pgfqpoint{0.000000in}{-0.048611in}}%
\pgfusepath{stroke,fill}%
}%
\begin{pgfscope}%
\pgfsys@transformshift{5.890977in}{0.964913in}%
\pgfsys@useobject{currentmarker}{}%
\end{pgfscope}%
\end{pgfscope}%
\begin{pgfscope}%
\definecolor{textcolor}{rgb}{0.000000,0.000000,0.000000}%
\pgfsetstrokecolor{textcolor}%
\pgfsetfillcolor{textcolor}%
\pgftext[x=5.890977in,y=0.867691in,,top]{\color{textcolor}\fontsize{20.000000}{24.000000}\selectfont \(\displaystyle {100}\)}%
\end{pgfscope}%
\begin{pgfscope}%
\definecolor{textcolor}{rgb}{0.000000,0.000000,0.000000}%
\pgfsetstrokecolor{textcolor}%
\pgfsetfillcolor{textcolor}%
\pgftext[x=3.800745in,y=0.556068in,,top]{\color{textcolor}\fontsize{20.000000}{24.000000}\selectfont Number links in the stem}%
\end{pgfscope}%
\begin{pgfscope}%
\pgfsetbuttcap%
\pgfsetroundjoin%
\definecolor{currentfill}{rgb}{0.000000,0.000000,0.000000}%
\pgfsetfillcolor{currentfill}%
\pgfsetlinewidth{0.803000pt}%
\definecolor{currentstroke}{rgb}{0.000000,0.000000,0.000000}%
\pgfsetstrokecolor{currentstroke}%
\pgfsetdash{}{0pt}%
\pgfsys@defobject{currentmarker}{\pgfqpoint{-0.048611in}{0.000000in}}{\pgfqpoint{-0.000000in}{0.000000in}}{%
\pgfpathmoveto{\pgfqpoint{-0.000000in}{0.000000in}}%
\pgfpathlineto{\pgfqpoint{-0.048611in}{0.000000in}}%
\pgfusepath{stroke,fill}%
}%
\begin{pgfscope}%
\pgfsys@transformshift{1.501490in}{1.506940in}%
\pgfsys@useobject{currentmarker}{}%
\end{pgfscope}%
\end{pgfscope}%
\begin{pgfscope}%
\definecolor{textcolor}{rgb}{0.000000,0.000000,0.000000}%
\pgfsetstrokecolor{textcolor}%
\pgfsetfillcolor{textcolor}%
\pgftext[x=0.743731in, y=1.406920in, left, base]{\color{textcolor}\fontsize{20.000000}{24.000000}\selectfont \(\displaystyle {20000}\)}%
\end{pgfscope}%
\begin{pgfscope}%
\pgfsetbuttcap%
\pgfsetroundjoin%
\definecolor{currentfill}{rgb}{0.000000,0.000000,0.000000}%
\pgfsetfillcolor{currentfill}%
\pgfsetlinewidth{0.803000pt}%
\definecolor{currentstroke}{rgb}{0.000000,0.000000,0.000000}%
\pgfsetstrokecolor{currentstroke}%
\pgfsetdash{}{0pt}%
\pgfsys@defobject{currentmarker}{\pgfqpoint{-0.048611in}{0.000000in}}{\pgfqpoint{-0.000000in}{0.000000in}}{%
\pgfpathmoveto{\pgfqpoint{-0.000000in}{0.000000in}}%
\pgfpathlineto{\pgfqpoint{-0.048611in}{0.000000in}}%
\pgfusepath{stroke,fill}%
}%
\begin{pgfscope}%
\pgfsys@transformshift{1.501490in}{2.201992in}%
\pgfsys@useobject{currentmarker}{}%
\end{pgfscope}%
\end{pgfscope}%
\begin{pgfscope}%
\definecolor{textcolor}{rgb}{0.000000,0.000000,0.000000}%
\pgfsetstrokecolor{textcolor}%
\pgfsetfillcolor{textcolor}%
\pgftext[x=0.743731in, y=2.101973in, left, base]{\color{textcolor}\fontsize{20.000000}{24.000000}\selectfont \(\displaystyle {40000}\)}%
\end{pgfscope}%
\begin{pgfscope}%
\pgfsetbuttcap%
\pgfsetroundjoin%
\definecolor{currentfill}{rgb}{0.000000,0.000000,0.000000}%
\pgfsetfillcolor{currentfill}%
\pgfsetlinewidth{0.803000pt}%
\definecolor{currentstroke}{rgb}{0.000000,0.000000,0.000000}%
\pgfsetstrokecolor{currentstroke}%
\pgfsetdash{}{0pt}%
\pgfsys@defobject{currentmarker}{\pgfqpoint{-0.048611in}{0.000000in}}{\pgfqpoint{-0.000000in}{0.000000in}}{%
\pgfpathmoveto{\pgfqpoint{-0.000000in}{0.000000in}}%
\pgfpathlineto{\pgfqpoint{-0.048611in}{0.000000in}}%
\pgfusepath{stroke,fill}%
}%
\begin{pgfscope}%
\pgfsys@transformshift{1.501490in}{2.897045in}%
\pgfsys@useobject{currentmarker}{}%
\end{pgfscope}%
\end{pgfscope}%
\begin{pgfscope}%
\definecolor{textcolor}{rgb}{0.000000,0.000000,0.000000}%
\pgfsetstrokecolor{textcolor}%
\pgfsetfillcolor{textcolor}%
\pgftext[x=0.743731in, y=2.797026in, left, base]{\color{textcolor}\fontsize{20.000000}{24.000000}\selectfont \(\displaystyle {60000}\)}%
\end{pgfscope}%
\begin{pgfscope}%
\pgfsetbuttcap%
\pgfsetroundjoin%
\definecolor{currentfill}{rgb}{0.000000,0.000000,0.000000}%
\pgfsetfillcolor{currentfill}%
\pgfsetlinewidth{0.803000pt}%
\definecolor{currentstroke}{rgb}{0.000000,0.000000,0.000000}%
\pgfsetstrokecolor{currentstroke}%
\pgfsetdash{}{0pt}%
\pgfsys@defobject{currentmarker}{\pgfqpoint{-0.048611in}{0.000000in}}{\pgfqpoint{-0.000000in}{0.000000in}}{%
\pgfpathmoveto{\pgfqpoint{-0.000000in}{0.000000in}}%
\pgfpathlineto{\pgfqpoint{-0.048611in}{0.000000in}}%
\pgfusepath{stroke,fill}%
}%
\begin{pgfscope}%
\pgfsys@transformshift{1.501490in}{3.592098in}%
\pgfsys@useobject{currentmarker}{}%
\end{pgfscope}%
\end{pgfscope}%
\begin{pgfscope}%
\definecolor{textcolor}{rgb}{0.000000,0.000000,0.000000}%
\pgfsetstrokecolor{textcolor}%
\pgfsetfillcolor{textcolor}%
\pgftext[x=0.743731in, y=3.492079in, left, base]{\color{textcolor}\fontsize{20.000000}{24.000000}\selectfont \(\displaystyle {80000}\)}%
\end{pgfscope}%
\begin{pgfscope}%
\pgfsetbuttcap%
\pgfsetroundjoin%
\definecolor{currentfill}{rgb}{0.000000,0.000000,0.000000}%
\pgfsetfillcolor{currentfill}%
\pgfsetlinewidth{0.803000pt}%
\definecolor{currentstroke}{rgb}{0.000000,0.000000,0.000000}%
\pgfsetstrokecolor{currentstroke}%
\pgfsetdash{}{0pt}%
\pgfsys@defobject{currentmarker}{\pgfqpoint{-0.048611in}{0.000000in}}{\pgfqpoint{-0.000000in}{0.000000in}}{%
\pgfpathmoveto{\pgfqpoint{-0.000000in}{0.000000in}}%
\pgfpathlineto{\pgfqpoint{-0.048611in}{0.000000in}}%
\pgfusepath{stroke,fill}%
}%
\begin{pgfscope}%
\pgfsys@transformshift{1.501490in}{4.287151in}%
\pgfsys@useobject{currentmarker}{}%
\end{pgfscope}%
\end{pgfscope}%
\begin{pgfscope}%
\definecolor{textcolor}{rgb}{0.000000,0.000000,0.000000}%
\pgfsetstrokecolor{textcolor}%
\pgfsetfillcolor{textcolor}%
\pgftext[x=0.611623in, y=4.187131in, left, base]{\color{textcolor}\fontsize{20.000000}{24.000000}\selectfont \(\displaystyle {100000}\)}%
\end{pgfscope}%
\begin{pgfscope}%
\definecolor{textcolor}{rgb}{0.000000,0.000000,0.000000}%
\pgfsetstrokecolor{textcolor}%
\pgfsetfillcolor{textcolor}%
\pgftext[x=0.556068in,y=2.732457in,,bottom,rotate=90.000000]{\color{textcolor}\fontsize{20.000000}{24.000000}\selectfont Number of operations}%
\end{pgfscope}%
\begin{pgfscope}%
\pgfpathrectangle{\pgfqpoint{1.501490in}{0.964913in}}{\pgfqpoint{4.598510in}{3.535087in}}%
\pgfusepath{clip}%
\pgfsetrectcap%
\pgfsetroundjoin%
\pgfsetlinewidth{1.505625pt}%
\definecolor{currentstroke}{rgb}{0.121569,0.466667,0.705882}%
\pgfsetstrokecolor{currentstroke}%
\pgfsetdash{}{0pt}%
\pgfpathmoveto{\pgfqpoint{1.710513in}{1.125599in}}%
\pgfpathlineto{\pgfqpoint{1.752740in}{1.156112in}}%
\pgfpathlineto{\pgfqpoint{1.794967in}{1.178145in}}%
\pgfpathlineto{\pgfqpoint{1.837194in}{1.200178in}}%
\pgfpathlineto{\pgfqpoint{1.879421in}{1.222211in}}%
\pgfpathlineto{\pgfqpoint{1.921648in}{1.244244in}}%
\pgfpathlineto{\pgfqpoint{1.963874in}{1.266278in}}%
\pgfpathlineto{\pgfqpoint{2.006101in}{1.288311in}}%
\pgfpathlineto{\pgfqpoint{2.048328in}{1.310344in}}%
\pgfpathlineto{\pgfqpoint{2.090555in}{1.332377in}}%
\pgfpathlineto{\pgfqpoint{2.132782in}{1.354410in}}%
\pgfpathlineto{\pgfqpoint{2.175009in}{1.376443in}}%
\pgfpathlineto{\pgfqpoint{2.217236in}{1.398477in}}%
\pgfpathlineto{\pgfqpoint{2.259463in}{1.420510in}}%
\pgfpathlineto{\pgfqpoint{2.301690in}{1.442543in}}%
\pgfpathlineto{\pgfqpoint{2.343917in}{1.464576in}}%
\pgfpathlineto{\pgfqpoint{2.386144in}{1.486609in}}%
\pgfpathlineto{\pgfqpoint{2.428370in}{1.508642in}}%
\pgfpathlineto{\pgfqpoint{2.470597in}{1.530676in}}%
\pgfpathlineto{\pgfqpoint{2.512824in}{1.552709in}}%
\pgfpathlineto{\pgfqpoint{2.555051in}{1.574742in}}%
\pgfpathlineto{\pgfqpoint{2.597278in}{1.596775in}}%
\pgfpathlineto{\pgfqpoint{2.639505in}{1.618808in}}%
\pgfpathlineto{\pgfqpoint{2.681732in}{1.640841in}}%
\pgfpathlineto{\pgfqpoint{2.723959in}{1.662875in}}%
\pgfpathlineto{\pgfqpoint{2.766186in}{1.684908in}}%
\pgfpathlineto{\pgfqpoint{2.808413in}{1.706941in}}%
\pgfpathlineto{\pgfqpoint{2.850640in}{1.728974in}}%
\pgfpathlineto{\pgfqpoint{2.892866in}{1.751007in}}%
\pgfpathlineto{\pgfqpoint{2.935093in}{1.773040in}}%
\pgfpathlineto{\pgfqpoint{2.977320in}{1.795074in}}%
\pgfpathlineto{\pgfqpoint{3.019547in}{1.817107in}}%
\pgfpathlineto{\pgfqpoint{3.061774in}{1.839140in}}%
\pgfpathlineto{\pgfqpoint{3.104001in}{1.861173in}}%
\pgfpathlineto{\pgfqpoint{3.146228in}{1.883206in}}%
\pgfpathlineto{\pgfqpoint{3.188455in}{1.905239in}}%
\pgfpathlineto{\pgfqpoint{3.230682in}{1.927273in}}%
\pgfpathlineto{\pgfqpoint{3.272909in}{1.949306in}}%
\pgfpathlineto{\pgfqpoint{3.315136in}{1.971339in}}%
\pgfpathlineto{\pgfqpoint{3.357362in}{1.993372in}}%
\pgfpathlineto{\pgfqpoint{3.399589in}{2.015405in}}%
\pgfpathlineto{\pgfqpoint{3.441816in}{2.037439in}}%
\pgfpathlineto{\pgfqpoint{3.484043in}{2.059472in}}%
\pgfpathlineto{\pgfqpoint{3.526270in}{2.081505in}}%
\pgfpathlineto{\pgfqpoint{3.568497in}{2.103538in}}%
\pgfpathlineto{\pgfqpoint{3.610724in}{2.125571in}}%
\pgfpathlineto{\pgfqpoint{3.652951in}{2.147604in}}%
\pgfpathlineto{\pgfqpoint{3.695178in}{2.169638in}}%
\pgfpathlineto{\pgfqpoint{3.737405in}{2.191671in}}%
\pgfpathlineto{\pgfqpoint{3.779631in}{2.213704in}}%
\pgfpathlineto{\pgfqpoint{3.821858in}{2.235737in}}%
\pgfpathlineto{\pgfqpoint{3.864085in}{2.257770in}}%
\pgfpathlineto{\pgfqpoint{3.906312in}{2.279803in}}%
\pgfpathlineto{\pgfqpoint{3.948539in}{2.301837in}}%
\pgfpathlineto{\pgfqpoint{3.990766in}{2.323870in}}%
\pgfpathlineto{\pgfqpoint{4.032993in}{2.345903in}}%
\pgfpathlineto{\pgfqpoint{4.075220in}{2.367936in}}%
\pgfpathlineto{\pgfqpoint{4.117447in}{2.389969in}}%
\pgfpathlineto{\pgfqpoint{4.159674in}{2.412002in}}%
\pgfpathlineto{\pgfqpoint{4.201901in}{2.434036in}}%
\pgfpathlineto{\pgfqpoint{4.244127in}{2.456069in}}%
\pgfpathlineto{\pgfqpoint{4.286354in}{2.478102in}}%
\pgfpathlineto{\pgfqpoint{4.328581in}{2.500135in}}%
\pgfpathlineto{\pgfqpoint{4.370808in}{2.522168in}}%
\pgfpathlineto{\pgfqpoint{4.413035in}{2.544201in}}%
\pgfpathlineto{\pgfqpoint{4.455262in}{2.566235in}}%
\pgfpathlineto{\pgfqpoint{4.497489in}{2.588268in}}%
\pgfpathlineto{\pgfqpoint{4.539716in}{2.610301in}}%
\pgfpathlineto{\pgfqpoint{4.581943in}{2.632334in}}%
\pgfpathlineto{\pgfqpoint{4.624170in}{2.654367in}}%
\pgfpathlineto{\pgfqpoint{4.666397in}{2.676401in}}%
\pgfpathlineto{\pgfqpoint{4.708623in}{2.698434in}}%
\pgfpathlineto{\pgfqpoint{4.750850in}{2.720467in}}%
\pgfpathlineto{\pgfqpoint{4.793077in}{2.742500in}}%
\pgfpathlineto{\pgfqpoint{4.835304in}{2.764533in}}%
\pgfpathlineto{\pgfqpoint{4.877531in}{2.786566in}}%
\pgfpathlineto{\pgfqpoint{4.919758in}{2.808600in}}%
\pgfpathlineto{\pgfqpoint{4.961985in}{2.830633in}}%
\pgfpathlineto{\pgfqpoint{5.004212in}{2.852666in}}%
\pgfpathlineto{\pgfqpoint{5.046439in}{2.874699in}}%
\pgfpathlineto{\pgfqpoint{5.088666in}{2.896732in}}%
\pgfpathlineto{\pgfqpoint{5.130892in}{2.918765in}}%
\pgfpathlineto{\pgfqpoint{5.173119in}{2.940799in}}%
\pgfpathlineto{\pgfqpoint{5.215346in}{2.962832in}}%
\pgfpathlineto{\pgfqpoint{5.257573in}{2.984865in}}%
\pgfpathlineto{\pgfqpoint{5.299800in}{3.006898in}}%
\pgfpathlineto{\pgfqpoint{5.342027in}{3.028931in}}%
\pgfpathlineto{\pgfqpoint{5.384254in}{3.050964in}}%
\pgfpathlineto{\pgfqpoint{5.426481in}{3.072998in}}%
\pgfpathlineto{\pgfqpoint{5.468708in}{3.095031in}}%
\pgfpathlineto{\pgfqpoint{5.510935in}{3.117064in}}%
\pgfpathlineto{\pgfqpoint{5.553162in}{3.139097in}}%
\pgfpathlineto{\pgfqpoint{5.595388in}{3.161130in}}%
\pgfpathlineto{\pgfqpoint{5.637615in}{3.183163in}}%
\pgfpathlineto{\pgfqpoint{5.679842in}{3.205197in}}%
\pgfpathlineto{\pgfqpoint{5.722069in}{3.227230in}}%
\pgfpathlineto{\pgfqpoint{5.764296in}{3.249263in}}%
\pgfpathlineto{\pgfqpoint{5.806523in}{3.271296in}}%
\pgfpathlineto{\pgfqpoint{5.848750in}{3.293329in}}%
\pgfpathlineto{\pgfqpoint{5.890977in}{3.315363in}}%
\pgfusepath{stroke}%
\end{pgfscope}%
\begin{pgfscope}%
\pgfpathrectangle{\pgfqpoint{1.501490in}{0.964913in}}{\pgfqpoint{4.598510in}{3.535087in}}%
\pgfusepath{clip}%
\pgfsetrectcap%
\pgfsetroundjoin%
\pgfsetlinewidth{1.505625pt}%
\definecolor{currentstroke}{rgb}{1.000000,0.498039,0.054902}%
\pgfsetstrokecolor{currentstroke}%
\pgfsetdash{}{0pt}%
\pgfpathmoveto{\pgfqpoint{1.710513in}{1.230969in}}%
\pgfpathlineto{\pgfqpoint{1.752740in}{1.267320in}}%
\pgfpathlineto{\pgfqpoint{1.794967in}{1.298667in}}%
\pgfpathlineto{\pgfqpoint{1.837194in}{1.330014in}}%
\pgfpathlineto{\pgfqpoint{1.879421in}{1.361361in}}%
\pgfpathlineto{\pgfqpoint{1.921648in}{1.392708in}}%
\pgfpathlineto{\pgfqpoint{1.963874in}{1.424054in}}%
\pgfpathlineto{\pgfqpoint{2.006101in}{1.455401in}}%
\pgfpathlineto{\pgfqpoint{2.048328in}{1.486748in}}%
\pgfpathlineto{\pgfqpoint{2.090555in}{1.518095in}}%
\pgfpathlineto{\pgfqpoint{2.132782in}{1.549442in}}%
\pgfpathlineto{\pgfqpoint{2.175009in}{1.580789in}}%
\pgfpathlineto{\pgfqpoint{2.217236in}{1.612136in}}%
\pgfpathlineto{\pgfqpoint{2.259463in}{1.643483in}}%
\pgfpathlineto{\pgfqpoint{2.301690in}{1.674830in}}%
\pgfpathlineto{\pgfqpoint{2.343917in}{1.706176in}}%
\pgfpathlineto{\pgfqpoint{2.386144in}{1.737523in}}%
\pgfpathlineto{\pgfqpoint{2.428370in}{1.768870in}}%
\pgfpathlineto{\pgfqpoint{2.470597in}{1.800217in}}%
\pgfpathlineto{\pgfqpoint{2.512824in}{1.831564in}}%
\pgfpathlineto{\pgfqpoint{2.555051in}{1.862911in}}%
\pgfpathlineto{\pgfqpoint{2.597278in}{1.894258in}}%
\pgfpathlineto{\pgfqpoint{2.639505in}{1.925605in}}%
\pgfpathlineto{\pgfqpoint{2.681732in}{1.956951in}}%
\pgfpathlineto{\pgfqpoint{2.723959in}{1.988298in}}%
\pgfpathlineto{\pgfqpoint{2.766186in}{2.019645in}}%
\pgfpathlineto{\pgfqpoint{2.808413in}{2.050992in}}%
\pgfpathlineto{\pgfqpoint{2.850640in}{2.082339in}}%
\pgfpathlineto{\pgfqpoint{2.892866in}{2.113686in}}%
\pgfpathlineto{\pgfqpoint{2.935093in}{2.145033in}}%
\pgfpathlineto{\pgfqpoint{2.977320in}{2.176380in}}%
\pgfpathlineto{\pgfqpoint{3.019547in}{2.207726in}}%
\pgfpathlineto{\pgfqpoint{3.061774in}{2.239073in}}%
\pgfpathlineto{\pgfqpoint{3.104001in}{2.270420in}}%
\pgfpathlineto{\pgfqpoint{3.146228in}{2.301767in}}%
\pgfpathlineto{\pgfqpoint{3.188455in}{2.333114in}}%
\pgfpathlineto{\pgfqpoint{3.230682in}{2.364461in}}%
\pgfpathlineto{\pgfqpoint{3.272909in}{2.395808in}}%
\pgfpathlineto{\pgfqpoint{3.315136in}{2.427155in}}%
\pgfpathlineto{\pgfqpoint{3.357362in}{2.458501in}}%
\pgfpathlineto{\pgfqpoint{3.399589in}{2.489848in}}%
\pgfpathlineto{\pgfqpoint{3.441816in}{2.521195in}}%
\pgfpathlineto{\pgfqpoint{3.484043in}{2.552542in}}%
\pgfpathlineto{\pgfqpoint{3.526270in}{2.583889in}}%
\pgfpathlineto{\pgfqpoint{3.568497in}{2.615236in}}%
\pgfpathlineto{\pgfqpoint{3.610724in}{2.646583in}}%
\pgfpathlineto{\pgfqpoint{3.652951in}{2.677930in}}%
\pgfpathlineto{\pgfqpoint{3.695178in}{2.709277in}}%
\pgfpathlineto{\pgfqpoint{3.737405in}{2.740623in}}%
\pgfpathlineto{\pgfqpoint{3.779631in}{2.771970in}}%
\pgfpathlineto{\pgfqpoint{3.821858in}{2.803317in}}%
\pgfpathlineto{\pgfqpoint{3.864085in}{2.834664in}}%
\pgfpathlineto{\pgfqpoint{3.906312in}{2.866011in}}%
\pgfpathlineto{\pgfqpoint{3.948539in}{2.897358in}}%
\pgfpathlineto{\pgfqpoint{3.990766in}{2.928705in}}%
\pgfpathlineto{\pgfqpoint{4.032993in}{2.960052in}}%
\pgfpathlineto{\pgfqpoint{4.075220in}{2.991398in}}%
\pgfpathlineto{\pgfqpoint{4.117447in}{3.022745in}}%
\pgfpathlineto{\pgfqpoint{4.159674in}{3.054092in}}%
\pgfpathlineto{\pgfqpoint{4.201901in}{3.085439in}}%
\pgfpathlineto{\pgfqpoint{4.244127in}{3.116786in}}%
\pgfpathlineto{\pgfqpoint{4.286354in}{3.148133in}}%
\pgfpathlineto{\pgfqpoint{4.328581in}{3.179480in}}%
\pgfpathlineto{\pgfqpoint{4.370808in}{3.210827in}}%
\pgfpathlineto{\pgfqpoint{4.413035in}{3.242173in}}%
\pgfpathlineto{\pgfqpoint{4.455262in}{3.273520in}}%
\pgfpathlineto{\pgfqpoint{4.497489in}{3.304867in}}%
\pgfpathlineto{\pgfqpoint{4.539716in}{3.336214in}}%
\pgfpathlineto{\pgfqpoint{4.581943in}{3.367561in}}%
\pgfpathlineto{\pgfqpoint{4.624170in}{3.398908in}}%
\pgfpathlineto{\pgfqpoint{4.666397in}{3.430255in}}%
\pgfpathlineto{\pgfqpoint{4.708623in}{3.461602in}}%
\pgfpathlineto{\pgfqpoint{4.750850in}{3.492948in}}%
\pgfpathlineto{\pgfqpoint{4.793077in}{3.524295in}}%
\pgfpathlineto{\pgfqpoint{4.835304in}{3.555642in}}%
\pgfpathlineto{\pgfqpoint{4.877531in}{3.586989in}}%
\pgfpathlineto{\pgfqpoint{4.919758in}{3.618336in}}%
\pgfpathlineto{\pgfqpoint{4.961985in}{3.649683in}}%
\pgfpathlineto{\pgfqpoint{5.004212in}{3.681030in}}%
\pgfpathlineto{\pgfqpoint{5.046439in}{3.712377in}}%
\pgfpathlineto{\pgfqpoint{5.088666in}{3.743724in}}%
\pgfpathlineto{\pgfqpoint{5.130892in}{3.775070in}}%
\pgfpathlineto{\pgfqpoint{5.173119in}{3.806417in}}%
\pgfpathlineto{\pgfqpoint{5.215346in}{3.837764in}}%
\pgfpathlineto{\pgfqpoint{5.257573in}{3.869111in}}%
\pgfpathlineto{\pgfqpoint{5.299800in}{3.900458in}}%
\pgfpathlineto{\pgfqpoint{5.342027in}{3.931805in}}%
\pgfpathlineto{\pgfqpoint{5.384254in}{3.963152in}}%
\pgfpathlineto{\pgfqpoint{5.426481in}{3.994499in}}%
\pgfpathlineto{\pgfqpoint{5.468708in}{4.025845in}}%
\pgfpathlineto{\pgfqpoint{5.510935in}{4.057192in}}%
\pgfpathlineto{\pgfqpoint{5.553162in}{4.088539in}}%
\pgfpathlineto{\pgfqpoint{5.595388in}{4.119886in}}%
\pgfpathlineto{\pgfqpoint{5.637615in}{4.151233in}}%
\pgfpathlineto{\pgfqpoint{5.679842in}{4.182580in}}%
\pgfpathlineto{\pgfqpoint{5.722069in}{4.213927in}}%
\pgfpathlineto{\pgfqpoint{5.764296in}{4.245274in}}%
\pgfpathlineto{\pgfqpoint{5.806523in}{4.276620in}}%
\pgfpathlineto{\pgfqpoint{5.848750in}{4.307967in}}%
\pgfpathlineto{\pgfqpoint{5.890977in}{4.339314in}}%
\pgfusepath{stroke}%
\end{pgfscope}%
\begin{pgfscope}%
\pgfpathrectangle{\pgfqpoint{1.501490in}{0.964913in}}{\pgfqpoint{4.598510in}{3.535087in}}%
\pgfusepath{clip}%
\pgfsetrectcap%
\pgfsetroundjoin%
\pgfsetlinewidth{1.505625pt}%
\definecolor{currentstroke}{rgb}{0.172549,0.627451,0.172549}%
\pgfsetstrokecolor{currentstroke}%
\pgfsetdash{}{0pt}%
\pgfpathmoveto{\pgfqpoint{1.710513in}{1.125599in}}%
\pgfpathlineto{\pgfqpoint{1.752740in}{1.156112in}}%
\pgfpathlineto{\pgfqpoint{1.794967in}{1.178145in}}%
\pgfpathlineto{\pgfqpoint{1.837194in}{1.200178in}}%
\pgfpathlineto{\pgfqpoint{1.879421in}{1.222211in}}%
\pgfpathlineto{\pgfqpoint{1.921648in}{1.244244in}}%
\pgfpathlineto{\pgfqpoint{1.963874in}{1.266278in}}%
\pgfpathlineto{\pgfqpoint{2.006101in}{1.288311in}}%
\pgfpathlineto{\pgfqpoint{2.048328in}{1.310344in}}%
\pgfpathlineto{\pgfqpoint{2.090555in}{1.332377in}}%
\pgfpathlineto{\pgfqpoint{2.132782in}{1.354410in}}%
\pgfpathlineto{\pgfqpoint{2.175009in}{1.376443in}}%
\pgfpathlineto{\pgfqpoint{2.217236in}{1.398477in}}%
\pgfpathlineto{\pgfqpoint{2.259463in}{1.420510in}}%
\pgfpathlineto{\pgfqpoint{2.301690in}{1.442543in}}%
\pgfpathlineto{\pgfqpoint{2.343917in}{1.464576in}}%
\pgfpathlineto{\pgfqpoint{2.386144in}{1.486609in}}%
\pgfpathlineto{\pgfqpoint{2.428370in}{1.508642in}}%
\pgfpathlineto{\pgfqpoint{2.470597in}{1.530676in}}%
\pgfpathlineto{\pgfqpoint{2.512824in}{1.552709in}}%
\pgfpathlineto{\pgfqpoint{2.555051in}{1.574742in}}%
\pgfpathlineto{\pgfqpoint{2.597278in}{1.596775in}}%
\pgfpathlineto{\pgfqpoint{2.639505in}{1.618808in}}%
\pgfpathlineto{\pgfqpoint{2.681732in}{1.640841in}}%
\pgfpathlineto{\pgfqpoint{2.723959in}{1.662875in}}%
\pgfpathlineto{\pgfqpoint{2.766186in}{1.684908in}}%
\pgfpathlineto{\pgfqpoint{2.808413in}{1.706941in}}%
\pgfpathlineto{\pgfqpoint{2.850640in}{1.728974in}}%
\pgfpathlineto{\pgfqpoint{2.892866in}{1.751007in}}%
\pgfpathlineto{\pgfqpoint{2.935093in}{1.773040in}}%
\pgfpathlineto{\pgfqpoint{2.977320in}{1.795074in}}%
\pgfpathlineto{\pgfqpoint{3.019547in}{1.817107in}}%
\pgfpathlineto{\pgfqpoint{3.061774in}{1.839140in}}%
\pgfpathlineto{\pgfqpoint{3.104001in}{1.861173in}}%
\pgfpathlineto{\pgfqpoint{3.146228in}{1.883206in}}%
\pgfpathlineto{\pgfqpoint{3.188455in}{1.905239in}}%
\pgfpathlineto{\pgfqpoint{3.230682in}{1.927273in}}%
\pgfpathlineto{\pgfqpoint{3.272909in}{1.949306in}}%
\pgfpathlineto{\pgfqpoint{3.315136in}{1.971339in}}%
\pgfpathlineto{\pgfqpoint{3.357362in}{1.993372in}}%
\pgfpathlineto{\pgfqpoint{3.399589in}{2.015405in}}%
\pgfpathlineto{\pgfqpoint{3.441816in}{2.037439in}}%
\pgfpathlineto{\pgfqpoint{3.484043in}{2.059472in}}%
\pgfpathlineto{\pgfqpoint{3.526270in}{2.081505in}}%
\pgfpathlineto{\pgfqpoint{3.568497in}{2.103538in}}%
\pgfpathlineto{\pgfqpoint{3.610724in}{2.125571in}}%
\pgfpathlineto{\pgfqpoint{3.652951in}{2.147604in}}%
\pgfpathlineto{\pgfqpoint{3.695178in}{2.169638in}}%
\pgfpathlineto{\pgfqpoint{3.737405in}{2.191671in}}%
\pgfpathlineto{\pgfqpoint{3.779631in}{2.213704in}}%
\pgfpathlineto{\pgfqpoint{3.821858in}{2.235737in}}%
\pgfpathlineto{\pgfqpoint{3.864085in}{2.257770in}}%
\pgfpathlineto{\pgfqpoint{3.906312in}{2.279803in}}%
\pgfpathlineto{\pgfqpoint{3.948539in}{2.301837in}}%
\pgfpathlineto{\pgfqpoint{3.990766in}{2.323870in}}%
\pgfpathlineto{\pgfqpoint{4.032993in}{2.345903in}}%
\pgfpathlineto{\pgfqpoint{4.075220in}{2.367936in}}%
\pgfpathlineto{\pgfqpoint{4.117447in}{2.389969in}}%
\pgfpathlineto{\pgfqpoint{4.159674in}{2.412002in}}%
\pgfpathlineto{\pgfqpoint{4.201901in}{2.434036in}}%
\pgfpathlineto{\pgfqpoint{4.244127in}{2.456069in}}%
\pgfpathlineto{\pgfqpoint{4.286354in}{2.478102in}}%
\pgfpathlineto{\pgfqpoint{4.328581in}{2.500135in}}%
\pgfpathlineto{\pgfqpoint{4.370808in}{2.522168in}}%
\pgfpathlineto{\pgfqpoint{4.413035in}{2.544201in}}%
\pgfpathlineto{\pgfqpoint{4.455262in}{2.566235in}}%
\pgfpathlineto{\pgfqpoint{4.497489in}{2.588268in}}%
\pgfpathlineto{\pgfqpoint{4.539716in}{2.610301in}}%
\pgfpathlineto{\pgfqpoint{4.581943in}{2.632334in}}%
\pgfpathlineto{\pgfqpoint{4.624170in}{2.654367in}}%
\pgfpathlineto{\pgfqpoint{4.666397in}{2.676401in}}%
\pgfpathlineto{\pgfqpoint{4.708623in}{2.698434in}}%
\pgfpathlineto{\pgfqpoint{4.750850in}{2.720467in}}%
\pgfpathlineto{\pgfqpoint{4.793077in}{2.742500in}}%
\pgfpathlineto{\pgfqpoint{4.835304in}{2.764533in}}%
\pgfpathlineto{\pgfqpoint{4.877531in}{2.786566in}}%
\pgfpathlineto{\pgfqpoint{4.919758in}{2.808600in}}%
\pgfpathlineto{\pgfqpoint{4.961985in}{2.830633in}}%
\pgfpathlineto{\pgfqpoint{5.004212in}{2.852666in}}%
\pgfpathlineto{\pgfqpoint{5.046439in}{2.874699in}}%
\pgfpathlineto{\pgfqpoint{5.088666in}{2.896732in}}%
\pgfpathlineto{\pgfqpoint{5.130892in}{2.918765in}}%
\pgfpathlineto{\pgfqpoint{5.173119in}{2.940799in}}%
\pgfpathlineto{\pgfqpoint{5.215346in}{2.962832in}}%
\pgfpathlineto{\pgfqpoint{5.257573in}{2.984865in}}%
\pgfpathlineto{\pgfqpoint{5.299800in}{3.006898in}}%
\pgfpathlineto{\pgfqpoint{5.342027in}{3.028931in}}%
\pgfpathlineto{\pgfqpoint{5.384254in}{3.050964in}}%
\pgfpathlineto{\pgfqpoint{5.426481in}{3.072998in}}%
\pgfpathlineto{\pgfqpoint{5.468708in}{3.095031in}}%
\pgfpathlineto{\pgfqpoint{5.510935in}{3.117064in}}%
\pgfpathlineto{\pgfqpoint{5.553162in}{3.139097in}}%
\pgfpathlineto{\pgfqpoint{5.595388in}{3.161130in}}%
\pgfpathlineto{\pgfqpoint{5.637615in}{3.183163in}}%
\pgfpathlineto{\pgfqpoint{5.679842in}{3.205197in}}%
\pgfpathlineto{\pgfqpoint{5.722069in}{3.227230in}}%
\pgfpathlineto{\pgfqpoint{5.764296in}{3.249263in}}%
\pgfpathlineto{\pgfqpoint{5.806523in}{3.271296in}}%
\pgfpathlineto{\pgfqpoint{5.848750in}{3.293329in}}%
\pgfpathlineto{\pgfqpoint{5.890977in}{3.315363in}}%
\pgfusepath{stroke}%
\end{pgfscope}%
\begin{pgfscope}%
\pgfsetrectcap%
\pgfsetmiterjoin%
\pgfsetlinewidth{0.803000pt}%
\definecolor{currentstroke}{rgb}{0.000000,0.000000,0.000000}%
\pgfsetstrokecolor{currentstroke}%
\pgfsetdash{}{0pt}%
\pgfpathmoveto{\pgfqpoint{1.501490in}{0.964913in}}%
\pgfpathlineto{\pgfqpoint{1.501490in}{4.500000in}}%
\pgfusepath{stroke}%
\end{pgfscope}%
\begin{pgfscope}%
\pgfsetrectcap%
\pgfsetmiterjoin%
\pgfsetlinewidth{0.803000pt}%
\definecolor{currentstroke}{rgb}{0.000000,0.000000,0.000000}%
\pgfsetstrokecolor{currentstroke}%
\pgfsetdash{}{0pt}%
\pgfpathmoveto{\pgfqpoint{6.100000in}{0.964913in}}%
\pgfpathlineto{\pgfqpoint{6.100000in}{4.500000in}}%
\pgfusepath{stroke}%
\end{pgfscope}%
\begin{pgfscope}%
\pgfsetrectcap%
\pgfsetmiterjoin%
\pgfsetlinewidth{0.803000pt}%
\definecolor{currentstroke}{rgb}{0.000000,0.000000,0.000000}%
\pgfsetstrokecolor{currentstroke}%
\pgfsetdash{}{0pt}%
\pgfpathmoveto{\pgfqpoint{1.501490in}{0.964913in}}%
\pgfpathlineto{\pgfqpoint{6.100000in}{0.964913in}}%
\pgfusepath{stroke}%
\end{pgfscope}%
\begin{pgfscope}%
\pgfsetrectcap%
\pgfsetmiterjoin%
\pgfsetlinewidth{0.803000pt}%
\definecolor{currentstroke}{rgb}{0.000000,0.000000,0.000000}%
\pgfsetstrokecolor{currentstroke}%
\pgfsetdash{}{0pt}%
\pgfpathmoveto{\pgfqpoint{1.501490in}{4.500000in}}%
\pgfpathlineto{\pgfqpoint{6.100000in}{4.500000in}}%
\pgfusepath{stroke}%
\end{pgfscope}%
\begin{pgfscope}%
\pgfsetbuttcap%
\pgfsetmiterjoin%
\definecolor{currentfill}{rgb}{1.000000,1.000000,1.000000}%
\pgfsetfillcolor{currentfill}%
\pgfsetfillopacity{0.800000}%
\pgfsetlinewidth{1.003750pt}%
\definecolor{currentstroke}{rgb}{0.800000,0.800000,0.800000}%
\pgfsetstrokecolor{currentstroke}%
\pgfsetstrokeopacity{0.800000}%
\pgfsetdash{}{0pt}%
\pgfpathmoveto{\pgfqpoint{1.695934in}{3.092908in}}%
\pgfpathlineto{\pgfqpoint{3.746195in}{3.092908in}}%
\pgfpathquadraticcurveto{\pgfqpoint{3.801750in}{3.092908in}}{\pgfqpoint{3.801750in}{3.148464in}}%
\pgfpathlineto{\pgfqpoint{3.801750in}{4.305556in}}%
\pgfpathquadraticcurveto{\pgfqpoint{3.801750in}{4.361111in}}{\pgfqpoint{3.746195in}{4.361111in}}%
\pgfpathlineto{\pgfqpoint{1.695934in}{4.361111in}}%
\pgfpathquadraticcurveto{\pgfqpoint{1.640379in}{4.361111in}}{\pgfqpoint{1.640379in}{4.305556in}}%
\pgfpathlineto{\pgfqpoint{1.640379in}{3.148464in}}%
\pgfpathquadraticcurveto{\pgfqpoint{1.640379in}{3.092908in}}{\pgfqpoint{1.695934in}{3.092908in}}%
\pgfpathclose%
\pgfusepath{stroke,fill}%
\end{pgfscope}%
\begin{pgfscope}%
\pgfsetrectcap%
\pgfsetroundjoin%
\pgfsetlinewidth{1.505625pt}%
\definecolor{currentstroke}{rgb}{0.121569,0.466667,0.705882}%
\pgfsetstrokecolor{currentstroke}%
\pgfsetdash{}{0pt}%
\pgfpathmoveto{\pgfqpoint{1.751490in}{4.147184in}}%
\pgfpathlineto{\pgfqpoint{2.307045in}{4.147184in}}%
\pgfusepath{stroke}%
\end{pgfscope}%
\begin{pgfscope}%
\definecolor{textcolor}{rgb}{0.000000,0.000000,0.000000}%
\pgfsetstrokecolor{textcolor}%
\pgfsetfillcolor{textcolor}%
\pgftext[x=2.529268in,y=4.049962in,left,base]{\color{textcolor}\fontsize{20.000000}{24.000000}\selectfont PV-OSIM}%
\end{pgfscope}%
\begin{pgfscope}%
\pgfsetrectcap%
\pgfsetroundjoin%
\pgfsetlinewidth{1.505625pt}%
\definecolor{currentstroke}{rgb}{1.000000,0.498039,0.054902}%
\pgfsetstrokecolor{currentstroke}%
\pgfsetdash{}{0pt}%
\pgfpathmoveto{\pgfqpoint{1.751490in}{3.752227in}}%
\pgfpathlineto{\pgfqpoint{2.307045in}{3.752227in}}%
\pgfusepath{stroke}%
\end{pgfscope}%
\begin{pgfscope}%
\definecolor{textcolor}{rgb}{0.000000,0.000000,0.000000}%
\pgfsetstrokecolor{textcolor}%
\pgfsetfillcolor{textcolor}%
\pgftext[x=2.529268in,y=3.655005in,left,base]{\color{textcolor}\fontsize{20.000000}{24.000000}\selectfont EFPA}%
\end{pgfscope}%
\begin{pgfscope}%
\pgfsetrectcap%
\pgfsetroundjoin%
\pgfsetlinewidth{1.505625pt}%
\definecolor{currentstroke}{rgb}{0.172549,0.627451,0.172549}%
\pgfsetstrokecolor{currentstroke}%
\pgfsetdash{}{0pt}%
\pgfpathmoveto{\pgfqpoint{1.751490in}{3.357271in}}%
\pgfpathlineto{\pgfqpoint{2.307045in}{3.357271in}}%
\pgfusepath{stroke}%
\end{pgfscope}%
\begin{pgfscope}%
\definecolor{textcolor}{rgb}{0.000000,0.000000,0.000000}%
\pgfsetstrokecolor{textcolor}%
\pgfsetfillcolor{textcolor}%
\pgftext[x=2.529268in,y=3.260048in,left,base]{\color{textcolor}\fontsize{20.000000}{24.000000}\selectfont PV-OSIMr}%
\end{pgfscope}%
\end{pgfpicture}%
\makeatother%
\endgroup%

%% file: graphics/tree_comparison2.pgf
%% Creator: Matplotlib, PGF backend
%%
%% To include the figure in your LaTeX document, write
%%   \input{<filename>.pgf}
%%
%% Make sure the required packages are loaded in your preamble
%%   \usepackage{pgf}
%%
%% Figures using additional raster images can only be included by \input if
%% they are in the same directory as the main LaTeX file. For loading figures
%% from other directories you can use the `import` package
%% and then include the figures with
%%   \import{<path to file>}{<filename>.pgf}
%%
%% Matplotlib used the following preamble
%%
\begingroup%
\makeatletter%
\begin{pgfpicture}%
\pgfpathrectangle{\pgfpointorigin}{\pgfqpoint{6.400000in}{4.800000in}}%
\pgfusepath{use as bounding box, clip}%
\begin{pgfscope}%
\pgfsetbuttcap%
\pgfsetmiterjoin%
\definecolor{currentfill}{rgb}{1.000000,1.000000,1.000000}%
\pgfsetfillcolor{currentfill}%
\pgfsetlinewidth{0.000000pt}%
\definecolor{currentstroke}{rgb}{1.000000,1.000000,1.000000}%
\pgfsetstrokecolor{currentstroke}%
\pgfsetdash{}{0pt}%
\pgfpathmoveto{\pgfqpoint{0.000000in}{0.000000in}}%
\pgfpathlineto{\pgfqpoint{6.400000in}{0.000000in}}%
\pgfpathlineto{\pgfqpoint{6.400000in}{4.800000in}}%
\pgfpathlineto{\pgfqpoint{0.000000in}{4.800000in}}%
\pgfpathclose%
\pgfusepath{fill}%
\end{pgfscope}%
\begin{pgfscope}%
\pgfsetbuttcap%
\pgfsetmiterjoin%
\definecolor{currentfill}{rgb}{1.000000,1.000000,1.000000}%
\pgfsetfillcolor{currentfill}%
\pgfsetlinewidth{0.000000pt}%
\definecolor{currentstroke}{rgb}{0.000000,0.000000,0.000000}%
\pgfsetstrokecolor{currentstroke}%
\pgfsetstrokeopacity{0.000000}%
\pgfsetdash{}{0pt}%
\pgfpathmoveto{\pgfqpoint{1.501490in}{0.964913in}}%
\pgfpathlineto{\pgfqpoint{6.100000in}{0.964913in}}%
\pgfpathlineto{\pgfqpoint{6.100000in}{4.500000in}}%
\pgfpathlineto{\pgfqpoint{1.501490in}{4.500000in}}%
\pgfpathclose%
\pgfusepath{fill}%
\end{pgfscope}%
\begin{pgfscope}%
\pgfsetbuttcap%
\pgfsetroundjoin%
\definecolor{currentfill}{rgb}{0.000000,0.000000,0.000000}%
\pgfsetfillcolor{currentfill}%
\pgfsetlinewidth{0.803000pt}%
\definecolor{currentstroke}{rgb}{0.000000,0.000000,0.000000}%
\pgfsetstrokecolor{currentstroke}%
\pgfsetdash{}{0pt}%
\pgfsys@defobject{currentmarker}{\pgfqpoint{0.000000in}{-0.048611in}}{\pgfqpoint{0.000000in}{0.000000in}}{%
\pgfpathmoveto{\pgfqpoint{0.000000in}{0.000000in}}%
\pgfpathlineto{\pgfqpoint{0.000000in}{-0.048611in}}%
\pgfusepath{stroke,fill}%
}%
\begin{pgfscope}%
\pgfsys@transformshift{1.668286in}{0.964913in}%
\pgfsys@useobject{currentmarker}{}%
\end{pgfscope}%
\end{pgfscope}%
\begin{pgfscope}%
\definecolor{textcolor}{rgb}{0.000000,0.000000,0.000000}%
\pgfsetstrokecolor{textcolor}%
\pgfsetfillcolor{textcolor}%
\pgftext[x=1.668286in,y=0.867691in,,top]{\color{textcolor}\fontsize{20.000000}{24.000000}\selectfont \(\displaystyle {0}\)}%
\end{pgfscope}%
\begin{pgfscope}%
\pgfsetbuttcap%
\pgfsetroundjoin%
\definecolor{currentfill}{rgb}{0.000000,0.000000,0.000000}%
\pgfsetfillcolor{currentfill}%
\pgfsetlinewidth{0.803000pt}%
\definecolor{currentstroke}{rgb}{0.000000,0.000000,0.000000}%
\pgfsetstrokecolor{currentstroke}%
\pgfsetdash{}{0pt}%
\pgfsys@defobject{currentmarker}{\pgfqpoint{0.000000in}{-0.048611in}}{\pgfqpoint{0.000000in}{0.000000in}}{%
\pgfpathmoveto{\pgfqpoint{0.000000in}{0.000000in}}%
\pgfpathlineto{\pgfqpoint{0.000000in}{-0.048611in}}%
\pgfusepath{stroke,fill}%
}%
\begin{pgfscope}%
\pgfsys@transformshift{2.723959in}{0.964913in}%
\pgfsys@useobject{currentmarker}{}%
\end{pgfscope}%
\end{pgfscope}%
\begin{pgfscope}%
\definecolor{textcolor}{rgb}{0.000000,0.000000,0.000000}%
\pgfsetstrokecolor{textcolor}%
\pgfsetfillcolor{textcolor}%
\pgftext[x=2.723959in,y=0.867691in,,top]{\color{textcolor}\fontsize{20.000000}{24.000000}\selectfont \(\displaystyle {25}\)}%
\end{pgfscope}%
\begin{pgfscope}%
\pgfsetbuttcap%
\pgfsetroundjoin%
\definecolor{currentfill}{rgb}{0.000000,0.000000,0.000000}%
\pgfsetfillcolor{currentfill}%
\pgfsetlinewidth{0.803000pt}%
\definecolor{currentstroke}{rgb}{0.000000,0.000000,0.000000}%
\pgfsetstrokecolor{currentstroke}%
\pgfsetdash{}{0pt}%
\pgfsys@defobject{currentmarker}{\pgfqpoint{0.000000in}{-0.048611in}}{\pgfqpoint{0.000000in}{0.000000in}}{%
\pgfpathmoveto{\pgfqpoint{0.000000in}{0.000000in}}%
\pgfpathlineto{\pgfqpoint{0.000000in}{-0.048611in}}%
\pgfusepath{stroke,fill}%
}%
\begin{pgfscope}%
\pgfsys@transformshift{3.779631in}{0.964913in}%
\pgfsys@useobject{currentmarker}{}%
\end{pgfscope}%
\end{pgfscope}%
\begin{pgfscope}%
\definecolor{textcolor}{rgb}{0.000000,0.000000,0.000000}%
\pgfsetstrokecolor{textcolor}%
\pgfsetfillcolor{textcolor}%
\pgftext[x=3.779631in,y=0.867691in,,top]{\color{textcolor}\fontsize{20.000000}{24.000000}\selectfont \(\displaystyle {50}\)}%
\end{pgfscope}%
\begin{pgfscope}%
\pgfsetbuttcap%
\pgfsetroundjoin%
\definecolor{currentfill}{rgb}{0.000000,0.000000,0.000000}%
\pgfsetfillcolor{currentfill}%
\pgfsetlinewidth{0.803000pt}%
\definecolor{currentstroke}{rgb}{0.000000,0.000000,0.000000}%
\pgfsetstrokecolor{currentstroke}%
\pgfsetdash{}{0pt}%
\pgfsys@defobject{currentmarker}{\pgfqpoint{0.000000in}{-0.048611in}}{\pgfqpoint{0.000000in}{0.000000in}}{%
\pgfpathmoveto{\pgfqpoint{0.000000in}{0.000000in}}%
\pgfpathlineto{\pgfqpoint{0.000000in}{-0.048611in}}%
\pgfusepath{stroke,fill}%
}%
\begin{pgfscope}%
\pgfsys@transformshift{4.835304in}{0.964913in}%
\pgfsys@useobject{currentmarker}{}%
\end{pgfscope}%
\end{pgfscope}%
\begin{pgfscope}%
\definecolor{textcolor}{rgb}{0.000000,0.000000,0.000000}%
\pgfsetstrokecolor{textcolor}%
\pgfsetfillcolor{textcolor}%
\pgftext[x=4.835304in,y=0.867691in,,top]{\color{textcolor}\fontsize{20.000000}{24.000000}\selectfont \(\displaystyle {75}\)}%
\end{pgfscope}%
\begin{pgfscope}%
\pgfsetbuttcap%
\pgfsetroundjoin%
\definecolor{currentfill}{rgb}{0.000000,0.000000,0.000000}%
\pgfsetfillcolor{currentfill}%
\pgfsetlinewidth{0.803000pt}%
\definecolor{currentstroke}{rgb}{0.000000,0.000000,0.000000}%
\pgfsetstrokecolor{currentstroke}%
\pgfsetdash{}{0pt}%
\pgfsys@defobject{currentmarker}{\pgfqpoint{0.000000in}{-0.048611in}}{\pgfqpoint{0.000000in}{0.000000in}}{%
\pgfpathmoveto{\pgfqpoint{0.000000in}{0.000000in}}%
\pgfpathlineto{\pgfqpoint{0.000000in}{-0.048611in}}%
\pgfusepath{stroke,fill}%
}%
\begin{pgfscope}%
\pgfsys@transformshift{5.890977in}{0.964913in}%
\pgfsys@useobject{currentmarker}{}%
\end{pgfscope}%
\end{pgfscope}%
\begin{pgfscope}%
\definecolor{textcolor}{rgb}{0.000000,0.000000,0.000000}%
\pgfsetstrokecolor{textcolor}%
\pgfsetfillcolor{textcolor}%
\pgftext[x=5.890977in,y=0.867691in,,top]{\color{textcolor}\fontsize{20.000000}{24.000000}\selectfont \(\displaystyle {100}\)}%
\end{pgfscope}%
\begin{pgfscope}%
\definecolor{textcolor}{rgb}{0.000000,0.000000,0.000000}%
\pgfsetstrokecolor{textcolor}%
\pgfsetfillcolor{textcolor}%
\pgftext[x=3.800745in,y=0.556068in,,top]{\color{textcolor}\fontsize{20.000000}{24.000000}\selectfont Number links in the stem}%
\end{pgfscope}%
\begin{pgfscope}%
\pgfsetbuttcap%
\pgfsetroundjoin%
\definecolor{currentfill}{rgb}{0.000000,0.000000,0.000000}%
\pgfsetfillcolor{currentfill}%
\pgfsetlinewidth{0.803000pt}%
\definecolor{currentstroke}{rgb}{0.000000,0.000000,0.000000}%
\pgfsetstrokecolor{currentstroke}%
\pgfsetdash{}{0pt}%
\pgfsys@defobject{currentmarker}{\pgfqpoint{-0.048611in}{0.000000in}}{\pgfqpoint{-0.000000in}{0.000000in}}{%
\pgfpathmoveto{\pgfqpoint{-0.000000in}{0.000000in}}%
\pgfpathlineto{\pgfqpoint{-0.048611in}{0.000000in}}%
\pgfusepath{stroke,fill}%
}%
\begin{pgfscope}%
\pgfsys@transformshift{1.501490in}{1.778129in}%
\pgfsys@useobject{currentmarker}{}%
\end{pgfscope}%
\end{pgfscope}%
\begin{pgfscope}%
\definecolor{textcolor}{rgb}{0.000000,0.000000,0.000000}%
\pgfsetstrokecolor{textcolor}%
\pgfsetfillcolor{textcolor}%
\pgftext[x=0.743731in, y=1.678109in, left, base]{\color{textcolor}\fontsize{20.000000}{24.000000}\selectfont \(\displaystyle {50000}\)}%
\end{pgfscope}%
\begin{pgfscope}%
\pgfsetbuttcap%
\pgfsetroundjoin%
\definecolor{currentfill}{rgb}{0.000000,0.000000,0.000000}%
\pgfsetfillcolor{currentfill}%
\pgfsetlinewidth{0.803000pt}%
\definecolor{currentstroke}{rgb}{0.000000,0.000000,0.000000}%
\pgfsetstrokecolor{currentstroke}%
\pgfsetdash{}{0pt}%
\pgfsys@defobject{currentmarker}{\pgfqpoint{-0.048611in}{0.000000in}}{\pgfqpoint{-0.000000in}{0.000000in}}{%
\pgfpathmoveto{\pgfqpoint{-0.000000in}{0.000000in}}%
\pgfpathlineto{\pgfqpoint{-0.048611in}{0.000000in}}%
\pgfusepath{stroke,fill}%
}%
\begin{pgfscope}%
\pgfsys@transformshift{1.501490in}{2.823280in}%
\pgfsys@useobject{currentmarker}{}%
\end{pgfscope}%
\end{pgfscope}%
\begin{pgfscope}%
\definecolor{textcolor}{rgb}{0.000000,0.000000,0.000000}%
\pgfsetstrokecolor{textcolor}%
\pgfsetfillcolor{textcolor}%
\pgftext[x=0.611623in, y=2.723261in, left, base]{\color{textcolor}\fontsize{20.000000}{24.000000}\selectfont \(\displaystyle {100000}\)}%
\end{pgfscope}%
\begin{pgfscope}%
\pgfsetbuttcap%
\pgfsetroundjoin%
\definecolor{currentfill}{rgb}{0.000000,0.000000,0.000000}%
\pgfsetfillcolor{currentfill}%
\pgfsetlinewidth{0.803000pt}%
\definecolor{currentstroke}{rgb}{0.000000,0.000000,0.000000}%
\pgfsetstrokecolor{currentstroke}%
\pgfsetdash{}{0pt}%
\pgfsys@defobject{currentmarker}{\pgfqpoint{-0.048611in}{0.000000in}}{\pgfqpoint{-0.000000in}{0.000000in}}{%
\pgfpathmoveto{\pgfqpoint{-0.000000in}{0.000000in}}%
\pgfpathlineto{\pgfqpoint{-0.048611in}{0.000000in}}%
\pgfusepath{stroke,fill}%
}%
\begin{pgfscope}%
\pgfsys@transformshift{1.501490in}{3.868432in}%
\pgfsys@useobject{currentmarker}{}%
\end{pgfscope}%
\end{pgfscope}%
\begin{pgfscope}%
\definecolor{textcolor}{rgb}{0.000000,0.000000,0.000000}%
\pgfsetstrokecolor{textcolor}%
\pgfsetfillcolor{textcolor}%
\pgftext[x=0.611623in, y=3.768412in, left, base]{\color{textcolor}\fontsize{20.000000}{24.000000}\selectfont \(\displaystyle {150000}\)}%
\end{pgfscope}%
\begin{pgfscope}%
\definecolor{textcolor}{rgb}{0.000000,0.000000,0.000000}%
\pgfsetstrokecolor{textcolor}%
\pgfsetfillcolor{textcolor}%
\pgftext[x=0.556068in,y=2.732457in,,bottom,rotate=90.000000]{\color{textcolor}\fontsize{20.000000}{24.000000}\selectfont Number of operations}%
\end{pgfscope}%
\begin{pgfscope}%
\pgfpathrectangle{\pgfqpoint{1.501490in}{0.964913in}}{\pgfqpoint{4.598510in}{3.535087in}}%
\pgfusepath{clip}%
\pgfsetrectcap%
\pgfsetroundjoin%
\pgfsetlinewidth{1.505625pt}%
\definecolor{currentstroke}{rgb}{0.121569,0.466667,0.705882}%
\pgfsetstrokecolor{currentstroke}%
\pgfsetdash{}{0pt}%
\pgfpathmoveto{\pgfqpoint{1.710513in}{1.125599in}}%
\pgfpathlineto{\pgfqpoint{1.752740in}{1.158876in}}%
\pgfpathlineto{\pgfqpoint{1.794967in}{1.180281in}}%
\pgfpathlineto{\pgfqpoint{1.837194in}{1.201686in}}%
\pgfpathlineto{\pgfqpoint{1.879421in}{1.223091in}}%
\pgfpathlineto{\pgfqpoint{1.921648in}{1.244495in}}%
\pgfpathlineto{\pgfqpoint{1.963874in}{1.265900in}}%
\pgfpathlineto{\pgfqpoint{2.006101in}{1.287305in}}%
\pgfpathlineto{\pgfqpoint{2.048328in}{1.308709in}}%
\pgfpathlineto{\pgfqpoint{2.090555in}{1.330114in}}%
\pgfpathlineto{\pgfqpoint{2.132782in}{1.351519in}}%
\pgfpathlineto{\pgfqpoint{2.175009in}{1.372923in}}%
\pgfpathlineto{\pgfqpoint{2.217236in}{1.394328in}}%
\pgfpathlineto{\pgfqpoint{2.259463in}{1.415733in}}%
\pgfpathlineto{\pgfqpoint{2.301690in}{1.437138in}}%
\pgfpathlineto{\pgfqpoint{2.343917in}{1.458542in}}%
\pgfpathlineto{\pgfqpoint{2.386144in}{1.479947in}}%
\pgfpathlineto{\pgfqpoint{2.428370in}{1.501352in}}%
\pgfpathlineto{\pgfqpoint{2.470597in}{1.522756in}}%
\pgfpathlineto{\pgfqpoint{2.512824in}{1.544161in}}%
\pgfpathlineto{\pgfqpoint{2.555051in}{1.565566in}}%
\pgfpathlineto{\pgfqpoint{2.597278in}{1.586970in}}%
\pgfpathlineto{\pgfqpoint{2.639505in}{1.608375in}}%
\pgfpathlineto{\pgfqpoint{2.681732in}{1.629780in}}%
\pgfpathlineto{\pgfqpoint{2.723959in}{1.651185in}}%
\pgfpathlineto{\pgfqpoint{2.766186in}{1.672589in}}%
\pgfpathlineto{\pgfqpoint{2.808413in}{1.693994in}}%
\pgfpathlineto{\pgfqpoint{2.850640in}{1.715399in}}%
\pgfpathlineto{\pgfqpoint{2.892866in}{1.736803in}}%
\pgfpathlineto{\pgfqpoint{2.935093in}{1.758208in}}%
\pgfpathlineto{\pgfqpoint{2.977320in}{1.779613in}}%
\pgfpathlineto{\pgfqpoint{3.019547in}{1.801018in}}%
\pgfpathlineto{\pgfqpoint{3.061774in}{1.822422in}}%
\pgfpathlineto{\pgfqpoint{3.104001in}{1.843827in}}%
\pgfpathlineto{\pgfqpoint{3.146228in}{1.865232in}}%
\pgfpathlineto{\pgfqpoint{3.188455in}{1.886636in}}%
\pgfpathlineto{\pgfqpoint{3.230682in}{1.908041in}}%
\pgfpathlineto{\pgfqpoint{3.272909in}{1.929446in}}%
\pgfpathlineto{\pgfqpoint{3.315136in}{1.950850in}}%
\pgfpathlineto{\pgfqpoint{3.357362in}{1.972255in}}%
\pgfpathlineto{\pgfqpoint{3.399589in}{1.993660in}}%
\pgfpathlineto{\pgfqpoint{3.441816in}{2.015065in}}%
\pgfpathlineto{\pgfqpoint{3.484043in}{2.036469in}}%
\pgfpathlineto{\pgfqpoint{3.526270in}{2.057874in}}%
\pgfpathlineto{\pgfqpoint{3.568497in}{2.079279in}}%
\pgfpathlineto{\pgfqpoint{3.610724in}{2.100683in}}%
\pgfpathlineto{\pgfqpoint{3.652951in}{2.122088in}}%
\pgfpathlineto{\pgfqpoint{3.695178in}{2.143493in}}%
\pgfpathlineto{\pgfqpoint{3.737405in}{2.164897in}}%
\pgfpathlineto{\pgfqpoint{3.779631in}{2.186302in}}%
\pgfpathlineto{\pgfqpoint{3.821858in}{2.207707in}}%
\pgfpathlineto{\pgfqpoint{3.864085in}{2.229112in}}%
\pgfpathlineto{\pgfqpoint{3.906312in}{2.250516in}}%
\pgfpathlineto{\pgfqpoint{3.948539in}{2.271921in}}%
\pgfpathlineto{\pgfqpoint{3.990766in}{2.293326in}}%
\pgfpathlineto{\pgfqpoint{4.032993in}{2.314730in}}%
\pgfpathlineto{\pgfqpoint{4.075220in}{2.336135in}}%
\pgfpathlineto{\pgfqpoint{4.117447in}{2.357540in}}%
\pgfpathlineto{\pgfqpoint{4.159674in}{2.378944in}}%
\pgfpathlineto{\pgfqpoint{4.201901in}{2.400349in}}%
\pgfpathlineto{\pgfqpoint{4.244127in}{2.421754in}}%
\pgfpathlineto{\pgfqpoint{4.286354in}{2.443159in}}%
\pgfpathlineto{\pgfqpoint{4.328581in}{2.464563in}}%
\pgfpathlineto{\pgfqpoint{4.370808in}{2.485968in}}%
\pgfpathlineto{\pgfqpoint{4.413035in}{2.507373in}}%
\pgfpathlineto{\pgfqpoint{4.455262in}{2.528777in}}%
\pgfpathlineto{\pgfqpoint{4.497489in}{2.550182in}}%
\pgfpathlineto{\pgfqpoint{4.539716in}{2.571587in}}%
\pgfpathlineto{\pgfqpoint{4.581943in}{2.592992in}}%
\pgfpathlineto{\pgfqpoint{4.624170in}{2.614396in}}%
\pgfpathlineto{\pgfqpoint{4.666397in}{2.635801in}}%
\pgfpathlineto{\pgfqpoint{4.708623in}{2.657206in}}%
\pgfpathlineto{\pgfqpoint{4.750850in}{2.678610in}}%
\pgfpathlineto{\pgfqpoint{4.793077in}{2.700015in}}%
\pgfpathlineto{\pgfqpoint{4.835304in}{2.721420in}}%
\pgfpathlineto{\pgfqpoint{4.877531in}{2.742824in}}%
\pgfpathlineto{\pgfqpoint{4.919758in}{2.764229in}}%
\pgfpathlineto{\pgfqpoint{4.961985in}{2.785634in}}%
\pgfpathlineto{\pgfqpoint{5.004212in}{2.807039in}}%
\pgfpathlineto{\pgfqpoint{5.046439in}{2.828443in}}%
\pgfpathlineto{\pgfqpoint{5.088666in}{2.849848in}}%
\pgfpathlineto{\pgfqpoint{5.130892in}{2.871253in}}%
\pgfpathlineto{\pgfqpoint{5.173119in}{2.892657in}}%
\pgfpathlineto{\pgfqpoint{5.215346in}{2.914062in}}%
\pgfpathlineto{\pgfqpoint{5.257573in}{2.935467in}}%
\pgfpathlineto{\pgfqpoint{5.299800in}{2.956871in}}%
\pgfpathlineto{\pgfqpoint{5.342027in}{2.978276in}}%
\pgfpathlineto{\pgfqpoint{5.384254in}{2.999681in}}%
\pgfpathlineto{\pgfqpoint{5.426481in}{3.021086in}}%
\pgfpathlineto{\pgfqpoint{5.468708in}{3.042490in}}%
\pgfpathlineto{\pgfqpoint{5.510935in}{3.063895in}}%
\pgfpathlineto{\pgfqpoint{5.553162in}{3.085300in}}%
\pgfpathlineto{\pgfqpoint{5.595388in}{3.106704in}}%
\pgfpathlineto{\pgfqpoint{5.637615in}{3.128109in}}%
\pgfpathlineto{\pgfqpoint{5.679842in}{3.149514in}}%
\pgfpathlineto{\pgfqpoint{5.722069in}{3.170918in}}%
\pgfpathlineto{\pgfqpoint{5.764296in}{3.192323in}}%
\pgfpathlineto{\pgfqpoint{5.806523in}{3.213728in}}%
\pgfpathlineto{\pgfqpoint{5.848750in}{3.235133in}}%
\pgfpathlineto{\pgfqpoint{5.890977in}{3.256537in}}%
\pgfusepath{stroke}%
\end{pgfscope}%
\begin{pgfscope}%
\pgfpathrectangle{\pgfqpoint{1.501490in}{0.964913in}}{\pgfqpoint{4.598510in}{3.535087in}}%
\pgfusepath{clip}%
\pgfsetrectcap%
\pgfsetroundjoin%
\pgfsetlinewidth{1.505625pt}%
\definecolor{currentstroke}{rgb}{1.000000,0.498039,0.054902}%
\pgfsetstrokecolor{currentstroke}%
\pgfsetdash{}{0pt}%
\pgfpathmoveto{\pgfqpoint{1.710513in}{1.252355in}}%
\pgfpathlineto{\pgfqpoint{1.752740in}{1.291151in}}%
\pgfpathlineto{\pgfqpoint{1.794967in}{1.322255in}}%
\pgfpathlineto{\pgfqpoint{1.837194in}{1.353358in}}%
\pgfpathlineto{\pgfqpoint{1.879421in}{1.384462in}}%
\pgfpathlineto{\pgfqpoint{1.921648in}{1.415566in}}%
\pgfpathlineto{\pgfqpoint{1.963874in}{1.446669in}}%
\pgfpathlineto{\pgfqpoint{2.006101in}{1.477773in}}%
\pgfpathlineto{\pgfqpoint{2.048328in}{1.508877in}}%
\pgfpathlineto{\pgfqpoint{2.090555in}{1.539980in}}%
\pgfpathlineto{\pgfqpoint{2.132782in}{1.571084in}}%
\pgfpathlineto{\pgfqpoint{2.175009in}{1.602188in}}%
\pgfpathlineto{\pgfqpoint{2.217236in}{1.633292in}}%
\pgfpathlineto{\pgfqpoint{2.259463in}{1.664395in}}%
\pgfpathlineto{\pgfqpoint{2.301690in}{1.695499in}}%
\pgfpathlineto{\pgfqpoint{2.343917in}{1.726603in}}%
\pgfpathlineto{\pgfqpoint{2.386144in}{1.757706in}}%
\pgfpathlineto{\pgfqpoint{2.428370in}{1.788810in}}%
\pgfpathlineto{\pgfqpoint{2.470597in}{1.819914in}}%
\pgfpathlineto{\pgfqpoint{2.512824in}{1.851018in}}%
\pgfpathlineto{\pgfqpoint{2.555051in}{1.882121in}}%
\pgfpathlineto{\pgfqpoint{2.597278in}{1.913225in}}%
\pgfpathlineto{\pgfqpoint{2.639505in}{1.944329in}}%
\pgfpathlineto{\pgfqpoint{2.681732in}{1.975432in}}%
\pgfpathlineto{\pgfqpoint{2.723959in}{2.006536in}}%
\pgfpathlineto{\pgfqpoint{2.766186in}{2.037640in}}%
\pgfpathlineto{\pgfqpoint{2.808413in}{2.068744in}}%
\pgfpathlineto{\pgfqpoint{2.850640in}{2.099847in}}%
\pgfpathlineto{\pgfqpoint{2.892866in}{2.130951in}}%
\pgfpathlineto{\pgfqpoint{2.935093in}{2.162055in}}%
\pgfpathlineto{\pgfqpoint{2.977320in}{2.193158in}}%
\pgfpathlineto{\pgfqpoint{3.019547in}{2.224262in}}%
\pgfpathlineto{\pgfqpoint{3.061774in}{2.255366in}}%
\pgfpathlineto{\pgfqpoint{3.104001in}{2.286469in}}%
\pgfpathlineto{\pgfqpoint{3.146228in}{2.317573in}}%
\pgfpathlineto{\pgfqpoint{3.188455in}{2.348677in}}%
\pgfpathlineto{\pgfqpoint{3.230682in}{2.379781in}}%
\pgfpathlineto{\pgfqpoint{3.272909in}{2.410884in}}%
\pgfpathlineto{\pgfqpoint{3.315136in}{2.441988in}}%
\pgfpathlineto{\pgfqpoint{3.357362in}{2.473092in}}%
\pgfpathlineto{\pgfqpoint{3.399589in}{2.504195in}}%
\pgfpathlineto{\pgfqpoint{3.441816in}{2.535299in}}%
\pgfpathlineto{\pgfqpoint{3.484043in}{2.566403in}}%
\pgfpathlineto{\pgfqpoint{3.526270in}{2.597507in}}%
\pgfpathlineto{\pgfqpoint{3.568497in}{2.628610in}}%
\pgfpathlineto{\pgfqpoint{3.610724in}{2.659714in}}%
\pgfpathlineto{\pgfqpoint{3.652951in}{2.690818in}}%
\pgfpathlineto{\pgfqpoint{3.695178in}{2.721921in}}%
\pgfpathlineto{\pgfqpoint{3.737405in}{2.753025in}}%
\pgfpathlineto{\pgfqpoint{3.779631in}{2.784129in}}%
\pgfpathlineto{\pgfqpoint{3.821858in}{2.815233in}}%
\pgfpathlineto{\pgfqpoint{3.864085in}{2.846336in}}%
\pgfpathlineto{\pgfqpoint{3.906312in}{2.877440in}}%
\pgfpathlineto{\pgfqpoint{3.948539in}{2.908544in}}%
\pgfpathlineto{\pgfqpoint{3.990766in}{2.939647in}}%
\pgfpathlineto{\pgfqpoint{4.032993in}{2.970751in}}%
\pgfpathlineto{\pgfqpoint{4.075220in}{3.001855in}}%
\pgfpathlineto{\pgfqpoint{4.117447in}{3.032958in}}%
\pgfpathlineto{\pgfqpoint{4.159674in}{3.064062in}}%
\pgfpathlineto{\pgfqpoint{4.201901in}{3.095166in}}%
\pgfpathlineto{\pgfqpoint{4.244127in}{3.126270in}}%
\pgfpathlineto{\pgfqpoint{4.286354in}{3.157373in}}%
\pgfpathlineto{\pgfqpoint{4.328581in}{3.188477in}}%
\pgfpathlineto{\pgfqpoint{4.370808in}{3.219581in}}%
\pgfpathlineto{\pgfqpoint{4.413035in}{3.250684in}}%
\pgfpathlineto{\pgfqpoint{4.455262in}{3.281788in}}%
\pgfpathlineto{\pgfqpoint{4.497489in}{3.312892in}}%
\pgfpathlineto{\pgfqpoint{4.539716in}{3.343996in}}%
\pgfpathlineto{\pgfqpoint{4.581943in}{3.375099in}}%
\pgfpathlineto{\pgfqpoint{4.624170in}{3.406203in}}%
\pgfpathlineto{\pgfqpoint{4.666397in}{3.437307in}}%
\pgfpathlineto{\pgfqpoint{4.708623in}{3.468410in}}%
\pgfpathlineto{\pgfqpoint{4.750850in}{3.499514in}}%
\pgfpathlineto{\pgfqpoint{4.793077in}{3.530618in}}%
\pgfpathlineto{\pgfqpoint{4.835304in}{3.561722in}}%
\pgfpathlineto{\pgfqpoint{4.877531in}{3.592825in}}%
\pgfpathlineto{\pgfqpoint{4.919758in}{3.623929in}}%
\pgfpathlineto{\pgfqpoint{4.961985in}{3.655033in}}%
\pgfpathlineto{\pgfqpoint{5.004212in}{3.686136in}}%
\pgfpathlineto{\pgfqpoint{5.046439in}{3.717240in}}%
\pgfpathlineto{\pgfqpoint{5.088666in}{3.748344in}}%
\pgfpathlineto{\pgfqpoint{5.130892in}{3.779447in}}%
\pgfpathlineto{\pgfqpoint{5.173119in}{3.810551in}}%
\pgfpathlineto{\pgfqpoint{5.215346in}{3.841655in}}%
\pgfpathlineto{\pgfqpoint{5.257573in}{3.872759in}}%
\pgfpathlineto{\pgfqpoint{5.299800in}{3.903862in}}%
\pgfpathlineto{\pgfqpoint{5.342027in}{3.934966in}}%
\pgfpathlineto{\pgfqpoint{5.384254in}{3.966070in}}%
\pgfpathlineto{\pgfqpoint{5.426481in}{3.997173in}}%
\pgfpathlineto{\pgfqpoint{5.468708in}{4.028277in}}%
\pgfpathlineto{\pgfqpoint{5.510935in}{4.059381in}}%
\pgfpathlineto{\pgfqpoint{5.553162in}{4.090485in}}%
\pgfpathlineto{\pgfqpoint{5.595388in}{4.121588in}}%
\pgfpathlineto{\pgfqpoint{5.637615in}{4.152692in}}%
\pgfpathlineto{\pgfqpoint{5.679842in}{4.183796in}}%
\pgfpathlineto{\pgfqpoint{5.722069in}{4.214899in}}%
\pgfpathlineto{\pgfqpoint{5.764296in}{4.246003in}}%
\pgfpathlineto{\pgfqpoint{5.806523in}{4.277107in}}%
\pgfpathlineto{\pgfqpoint{5.848750in}{4.308211in}}%
\pgfpathlineto{\pgfqpoint{5.890977in}{4.339314in}}%
\pgfusepath{stroke}%
\end{pgfscope}%
\begin{pgfscope}%
\pgfpathrectangle{\pgfqpoint{1.501490in}{0.964913in}}{\pgfqpoint{4.598510in}{3.535087in}}%
\pgfusepath{clip}%
\pgfsetrectcap%
\pgfsetroundjoin%
\pgfsetlinewidth{1.505625pt}%
\definecolor{currentstroke}{rgb}{0.172549,0.627451,0.172549}%
\pgfsetstrokecolor{currentstroke}%
\pgfsetdash{}{0pt}%
\pgfpathmoveto{\pgfqpoint{1.710513in}{1.125599in}}%
\pgfpathlineto{\pgfqpoint{1.752740in}{1.156431in}}%
\pgfpathlineto{\pgfqpoint{1.794967in}{1.177543in}}%
\pgfpathlineto{\pgfqpoint{1.837194in}{1.189583in}}%
\pgfpathlineto{\pgfqpoint{1.879421in}{1.201623in}}%
\pgfpathlineto{\pgfqpoint{1.921648in}{1.213663in}}%
\pgfpathlineto{\pgfqpoint{1.963874in}{1.225703in}}%
\pgfpathlineto{\pgfqpoint{2.006101in}{1.237744in}}%
\pgfpathlineto{\pgfqpoint{2.048328in}{1.249784in}}%
\pgfpathlineto{\pgfqpoint{2.090555in}{1.261824in}}%
\pgfpathlineto{\pgfqpoint{2.132782in}{1.273864in}}%
\pgfpathlineto{\pgfqpoint{2.175009in}{1.285904in}}%
\pgfpathlineto{\pgfqpoint{2.217236in}{1.297944in}}%
\pgfpathlineto{\pgfqpoint{2.259463in}{1.309984in}}%
\pgfpathlineto{\pgfqpoint{2.301690in}{1.322025in}}%
\pgfpathlineto{\pgfqpoint{2.343917in}{1.334065in}}%
\pgfpathlineto{\pgfqpoint{2.386144in}{1.346105in}}%
\pgfpathlineto{\pgfqpoint{2.428370in}{1.358145in}}%
\pgfpathlineto{\pgfqpoint{2.470597in}{1.370185in}}%
\pgfpathlineto{\pgfqpoint{2.512824in}{1.382225in}}%
\pgfpathlineto{\pgfqpoint{2.555051in}{1.394265in}}%
\pgfpathlineto{\pgfqpoint{2.597278in}{1.406306in}}%
\pgfpathlineto{\pgfqpoint{2.639505in}{1.418346in}}%
\pgfpathlineto{\pgfqpoint{2.681732in}{1.430386in}}%
\pgfpathlineto{\pgfqpoint{2.723959in}{1.442426in}}%
\pgfpathlineto{\pgfqpoint{2.766186in}{1.454466in}}%
\pgfpathlineto{\pgfqpoint{2.808413in}{1.466506in}}%
\pgfpathlineto{\pgfqpoint{2.850640in}{1.478546in}}%
\pgfpathlineto{\pgfqpoint{2.892866in}{1.490587in}}%
\pgfpathlineto{\pgfqpoint{2.935093in}{1.502627in}}%
\pgfpathlineto{\pgfqpoint{2.977320in}{1.514667in}}%
\pgfpathlineto{\pgfqpoint{3.019547in}{1.526707in}}%
\pgfpathlineto{\pgfqpoint{3.061774in}{1.538747in}}%
\pgfpathlineto{\pgfqpoint{3.104001in}{1.550787in}}%
\pgfpathlineto{\pgfqpoint{3.146228in}{1.562827in}}%
\pgfpathlineto{\pgfqpoint{3.188455in}{1.574868in}}%
\pgfpathlineto{\pgfqpoint{3.230682in}{1.586908in}}%
\pgfpathlineto{\pgfqpoint{3.272909in}{1.598948in}}%
\pgfpathlineto{\pgfqpoint{3.315136in}{1.610988in}}%
\pgfpathlineto{\pgfqpoint{3.357362in}{1.623028in}}%
\pgfpathlineto{\pgfqpoint{3.399589in}{1.635068in}}%
\pgfpathlineto{\pgfqpoint{3.441816in}{1.647109in}}%
\pgfpathlineto{\pgfqpoint{3.484043in}{1.659149in}}%
\pgfpathlineto{\pgfqpoint{3.526270in}{1.671189in}}%
\pgfpathlineto{\pgfqpoint{3.568497in}{1.683229in}}%
\pgfpathlineto{\pgfqpoint{3.610724in}{1.695269in}}%
\pgfpathlineto{\pgfqpoint{3.652951in}{1.707309in}}%
\pgfpathlineto{\pgfqpoint{3.695178in}{1.719349in}}%
\pgfpathlineto{\pgfqpoint{3.737405in}{1.731390in}}%
\pgfpathlineto{\pgfqpoint{3.779631in}{1.743430in}}%
\pgfpathlineto{\pgfqpoint{3.821858in}{1.755470in}}%
\pgfpathlineto{\pgfqpoint{3.864085in}{1.767510in}}%
\pgfpathlineto{\pgfqpoint{3.906312in}{1.779550in}}%
\pgfpathlineto{\pgfqpoint{3.948539in}{1.791590in}}%
\pgfpathlineto{\pgfqpoint{3.990766in}{1.803630in}}%
\pgfpathlineto{\pgfqpoint{4.032993in}{1.815671in}}%
\pgfpathlineto{\pgfqpoint{4.075220in}{1.827711in}}%
\pgfpathlineto{\pgfqpoint{4.117447in}{1.839751in}}%
\pgfpathlineto{\pgfqpoint{4.159674in}{1.851791in}}%
\pgfpathlineto{\pgfqpoint{4.201901in}{1.863831in}}%
\pgfpathlineto{\pgfqpoint{4.244127in}{1.875871in}}%
\pgfpathlineto{\pgfqpoint{4.286354in}{1.887911in}}%
\pgfpathlineto{\pgfqpoint{4.328581in}{1.899952in}}%
\pgfpathlineto{\pgfqpoint{4.370808in}{1.911992in}}%
\pgfpathlineto{\pgfqpoint{4.413035in}{1.924032in}}%
\pgfpathlineto{\pgfqpoint{4.455262in}{1.936072in}}%
\pgfpathlineto{\pgfqpoint{4.497489in}{1.948112in}}%
\pgfpathlineto{\pgfqpoint{4.539716in}{1.960152in}}%
\pgfpathlineto{\pgfqpoint{4.581943in}{1.972192in}}%
\pgfpathlineto{\pgfqpoint{4.624170in}{1.984233in}}%
\pgfpathlineto{\pgfqpoint{4.666397in}{1.996273in}}%
\pgfpathlineto{\pgfqpoint{4.708623in}{2.008313in}}%
\pgfpathlineto{\pgfqpoint{4.750850in}{2.020353in}}%
\pgfpathlineto{\pgfqpoint{4.793077in}{2.032393in}}%
\pgfpathlineto{\pgfqpoint{4.835304in}{2.044433in}}%
\pgfpathlineto{\pgfqpoint{4.877531in}{2.056473in}}%
\pgfpathlineto{\pgfqpoint{4.919758in}{2.068514in}}%
\pgfpathlineto{\pgfqpoint{4.961985in}{2.080554in}}%
\pgfpathlineto{\pgfqpoint{5.004212in}{2.092594in}}%
\pgfpathlineto{\pgfqpoint{5.046439in}{2.104634in}}%
\pgfpathlineto{\pgfqpoint{5.088666in}{2.116674in}}%
\pgfpathlineto{\pgfqpoint{5.130892in}{2.128714in}}%
\pgfpathlineto{\pgfqpoint{5.173119in}{2.140754in}}%
\pgfpathlineto{\pgfqpoint{5.215346in}{2.152795in}}%
\pgfpathlineto{\pgfqpoint{5.257573in}{2.164835in}}%
\pgfpathlineto{\pgfqpoint{5.299800in}{2.176875in}}%
\pgfpathlineto{\pgfqpoint{5.342027in}{2.188915in}}%
\pgfpathlineto{\pgfqpoint{5.384254in}{2.200955in}}%
\pgfpathlineto{\pgfqpoint{5.426481in}{2.212995in}}%
\pgfpathlineto{\pgfqpoint{5.468708in}{2.225035in}}%
\pgfpathlineto{\pgfqpoint{5.510935in}{2.237076in}}%
\pgfpathlineto{\pgfqpoint{5.553162in}{2.249116in}}%
\pgfpathlineto{\pgfqpoint{5.595388in}{2.261156in}}%
\pgfpathlineto{\pgfqpoint{5.637615in}{2.273196in}}%
\pgfpathlineto{\pgfqpoint{5.679842in}{2.285236in}}%
\pgfpathlineto{\pgfqpoint{5.722069in}{2.297276in}}%
\pgfpathlineto{\pgfqpoint{5.764296in}{2.309316in}}%
\pgfpathlineto{\pgfqpoint{5.806523in}{2.321357in}}%
\pgfpathlineto{\pgfqpoint{5.848750in}{2.333397in}}%
\pgfpathlineto{\pgfqpoint{5.890977in}{2.345437in}}%
\pgfusepath{stroke}%
\end{pgfscope}%
\begin{pgfscope}%
\pgfsetrectcap%
\pgfsetmiterjoin%
\pgfsetlinewidth{0.803000pt}%
\definecolor{currentstroke}{rgb}{0.000000,0.000000,0.000000}%
\pgfsetstrokecolor{currentstroke}%
\pgfsetdash{}{0pt}%
\pgfpathmoveto{\pgfqpoint{1.501490in}{0.964913in}}%
\pgfpathlineto{\pgfqpoint{1.501490in}{4.500000in}}%
\pgfusepath{stroke}%
\end{pgfscope}%
\begin{pgfscope}%
\pgfsetrectcap%
\pgfsetmiterjoin%
\pgfsetlinewidth{0.803000pt}%
\definecolor{currentstroke}{rgb}{0.000000,0.000000,0.000000}%
\pgfsetstrokecolor{currentstroke}%
\pgfsetdash{}{0pt}%
\pgfpathmoveto{\pgfqpoint{6.100000in}{0.964913in}}%
\pgfpathlineto{\pgfqpoint{6.100000in}{4.500000in}}%
\pgfusepath{stroke}%
\end{pgfscope}%
\begin{pgfscope}%
\pgfsetrectcap%
\pgfsetmiterjoin%
\pgfsetlinewidth{0.803000pt}%
\definecolor{currentstroke}{rgb}{0.000000,0.000000,0.000000}%
\pgfsetstrokecolor{currentstroke}%
\pgfsetdash{}{0pt}%
\pgfpathmoveto{\pgfqpoint{1.501490in}{0.964913in}}%
\pgfpathlineto{\pgfqpoint{6.100000in}{0.964913in}}%
\pgfusepath{stroke}%
\end{pgfscope}%
\begin{pgfscope}%
\pgfsetrectcap%
\pgfsetmiterjoin%
\pgfsetlinewidth{0.803000pt}%
\definecolor{currentstroke}{rgb}{0.000000,0.000000,0.000000}%
\pgfsetstrokecolor{currentstroke}%
\pgfsetdash{}{0pt}%
\pgfpathmoveto{\pgfqpoint{1.501490in}{4.500000in}}%
\pgfpathlineto{\pgfqpoint{6.100000in}{4.500000in}}%
\pgfusepath{stroke}%
\end{pgfscope}%
\begin{pgfscope}%
\pgfsetbuttcap%
\pgfsetmiterjoin%
\definecolor{currentfill}{rgb}{1.000000,1.000000,1.000000}%
\pgfsetfillcolor{currentfill}%
\pgfsetfillopacity{0.800000}%
\pgfsetlinewidth{1.003750pt}%
\definecolor{currentstroke}{rgb}{0.800000,0.800000,0.800000}%
\pgfsetstrokecolor{currentstroke}%
\pgfsetstrokeopacity{0.800000}%
\pgfsetdash{}{0pt}%
\pgfpathmoveto{\pgfqpoint{1.695934in}{3.092908in}}%
\pgfpathlineto{\pgfqpoint{3.746195in}{3.092908in}}%
\pgfpathquadraticcurveto{\pgfqpoint{3.801750in}{3.092908in}}{\pgfqpoint{3.801750in}{3.148464in}}%
\pgfpathlineto{\pgfqpoint{3.801750in}{4.305556in}}%
\pgfpathquadraticcurveto{\pgfqpoint{3.801750in}{4.361111in}}{\pgfqpoint{3.746195in}{4.361111in}}%
\pgfpathlineto{\pgfqpoint{1.695934in}{4.361111in}}%
\pgfpathquadraticcurveto{\pgfqpoint{1.640379in}{4.361111in}}{\pgfqpoint{1.640379in}{4.305556in}}%
\pgfpathlineto{\pgfqpoint{1.640379in}{3.148464in}}%
\pgfpathquadraticcurveto{\pgfqpoint{1.640379in}{3.092908in}}{\pgfqpoint{1.695934in}{3.092908in}}%
\pgfpathclose%
\pgfusepath{stroke,fill}%
\end{pgfscope}%
\begin{pgfscope}%
\pgfsetrectcap%
\pgfsetroundjoin%
\pgfsetlinewidth{1.505625pt}%
\definecolor{currentstroke}{rgb}{0.121569,0.466667,0.705882}%
\pgfsetstrokecolor{currentstroke}%
\pgfsetdash{}{0pt}%
\pgfpathmoveto{\pgfqpoint{1.751490in}{4.147184in}}%
\pgfpathlineto{\pgfqpoint{2.307045in}{4.147184in}}%
\pgfusepath{stroke}%
\end{pgfscope}%
\begin{pgfscope}%
\definecolor{textcolor}{rgb}{0.000000,0.000000,0.000000}%
\pgfsetstrokecolor{textcolor}%
\pgfsetfillcolor{textcolor}%
\pgftext[x=2.529268in,y=4.049962in,left,base]{\color{textcolor}\fontsize{20.000000}{24.000000}\selectfont PV-OSIM}%
\end{pgfscope}%
\begin{pgfscope}%
\pgfsetrectcap%
\pgfsetroundjoin%
\pgfsetlinewidth{1.505625pt}%
\definecolor{currentstroke}{rgb}{1.000000,0.498039,0.054902}%
\pgfsetstrokecolor{currentstroke}%
\pgfsetdash{}{0pt}%
\pgfpathmoveto{\pgfqpoint{1.751490in}{3.752227in}}%
\pgfpathlineto{\pgfqpoint{2.307045in}{3.752227in}}%
\pgfusepath{stroke}%
\end{pgfscope}%
\begin{pgfscope}%
\definecolor{textcolor}{rgb}{0.000000,0.000000,0.000000}%
\pgfsetstrokecolor{textcolor}%
\pgfsetfillcolor{textcolor}%
\pgftext[x=2.529268in,y=3.655005in,left,base]{\color{textcolor}\fontsize{20.000000}{24.000000}\selectfont EFPA}%
\end{pgfscope}%
\begin{pgfscope}%
\pgfsetrectcap%
\pgfsetroundjoin%
\pgfsetlinewidth{1.505625pt}%
\definecolor{currentstroke}{rgb}{0.172549,0.627451,0.172549}%
\pgfsetstrokecolor{currentstroke}%
\pgfsetdash{}{0pt}%
\pgfpathmoveto{\pgfqpoint{1.751490in}{3.357271in}}%
\pgfpathlineto{\pgfqpoint{2.307045in}{3.357271in}}%
\pgfusepath{stroke}%
\end{pgfscope}%
\begin{pgfscope}%
\definecolor{textcolor}{rgb}{0.000000,0.000000,0.000000}%
\pgfsetstrokecolor{textcolor}%
\pgfsetfillcolor{textcolor}%
\pgftext[x=2.529268in,y=3.260048in,left,base]{\color{textcolor}\fontsize{20.000000}{24.000000}\selectfont PV-OSIMr}%
\end{pgfscope}%
\end{pgfpicture}%
\makeatother%
\endgroup%

%% file: graphics/tree_comparison4.pgf
%% Creator: Matplotlib, PGF backend
%%
%% To include the figure in your LaTeX document, write
%%   \input{<filename>.pgf}
%%
%% Make sure the required packages are loaded in your preamble
%%   \usepackage{pgf}
%%
%% Figures using additional raster images can only be included by \input if
%% they are in the same directory as the main LaTeX file. For loading figures
%% from other directories you can use the `import` package
%% and then include the figures with
%%   \import{<path to file>}{<filename>.pgf}
%%
%% Matplotlib used the following preamble
%%
\begingroup%
\makeatletter%
\begin{pgfpicture}%
\pgfpathrectangle{\pgfpointorigin}{\pgfqpoint{6.400000in}{4.800000in}}%
\pgfusepath{use as bounding box, clip}%
\begin{pgfscope}%
\pgfsetbuttcap%
\pgfsetmiterjoin%
\definecolor{currentfill}{rgb}{1.000000,1.000000,1.000000}%
\pgfsetfillcolor{currentfill}%
\pgfsetlinewidth{0.000000pt}%
\definecolor{currentstroke}{rgb}{1.000000,1.000000,1.000000}%
\pgfsetstrokecolor{currentstroke}%
\pgfsetdash{}{0pt}%
\pgfpathmoveto{\pgfqpoint{0.000000in}{0.000000in}}%
\pgfpathlineto{\pgfqpoint{6.400000in}{0.000000in}}%
\pgfpathlineto{\pgfqpoint{6.400000in}{4.800000in}}%
\pgfpathlineto{\pgfqpoint{0.000000in}{4.800000in}}%
\pgfpathclose%
\pgfusepath{fill}%
\end{pgfscope}%
\begin{pgfscope}%
\pgfsetbuttcap%
\pgfsetmiterjoin%
\definecolor{currentfill}{rgb}{1.000000,1.000000,1.000000}%
\pgfsetfillcolor{currentfill}%
\pgfsetlinewidth{0.000000pt}%
\definecolor{currentstroke}{rgb}{0.000000,0.000000,0.000000}%
\pgfsetstrokecolor{currentstroke}%
\pgfsetstrokeopacity{0.000000}%
\pgfsetdash{}{0pt}%
\pgfpathmoveto{\pgfqpoint{1.501490in}{0.964913in}}%
\pgfpathlineto{\pgfqpoint{6.100000in}{0.964913in}}%
\pgfpathlineto{\pgfqpoint{6.100000in}{4.500000in}}%
\pgfpathlineto{\pgfqpoint{1.501490in}{4.500000in}}%
\pgfpathclose%
\pgfusepath{fill}%
\end{pgfscope}%
\begin{pgfscope}%
\pgfsetbuttcap%
\pgfsetroundjoin%
\definecolor{currentfill}{rgb}{0.000000,0.000000,0.000000}%
\pgfsetfillcolor{currentfill}%
\pgfsetlinewidth{0.803000pt}%
\definecolor{currentstroke}{rgb}{0.000000,0.000000,0.000000}%
\pgfsetstrokecolor{currentstroke}%
\pgfsetdash{}{0pt}%
\pgfsys@defobject{currentmarker}{\pgfqpoint{0.000000in}{-0.048611in}}{\pgfqpoint{0.000000in}{0.000000in}}{%
\pgfpathmoveto{\pgfqpoint{0.000000in}{0.000000in}}%
\pgfpathlineto{\pgfqpoint{0.000000in}{-0.048611in}}%
\pgfusepath{stroke,fill}%
}%
\begin{pgfscope}%
\pgfsys@transformshift{1.668286in}{0.964913in}%
\pgfsys@useobject{currentmarker}{}%
\end{pgfscope}%
\end{pgfscope}%
\begin{pgfscope}%
\definecolor{textcolor}{rgb}{0.000000,0.000000,0.000000}%
\pgfsetstrokecolor{textcolor}%
\pgfsetfillcolor{textcolor}%
\pgftext[x=1.668286in,y=0.867691in,,top]{\color{textcolor}\fontsize{20.000000}{24.000000}\selectfont \(\displaystyle {0}\)}%
\end{pgfscope}%
\begin{pgfscope}%
\pgfsetbuttcap%
\pgfsetroundjoin%
\definecolor{currentfill}{rgb}{0.000000,0.000000,0.000000}%
\pgfsetfillcolor{currentfill}%
\pgfsetlinewidth{0.803000pt}%
\definecolor{currentstroke}{rgb}{0.000000,0.000000,0.000000}%
\pgfsetstrokecolor{currentstroke}%
\pgfsetdash{}{0pt}%
\pgfsys@defobject{currentmarker}{\pgfqpoint{0.000000in}{-0.048611in}}{\pgfqpoint{0.000000in}{0.000000in}}{%
\pgfpathmoveto{\pgfqpoint{0.000000in}{0.000000in}}%
\pgfpathlineto{\pgfqpoint{0.000000in}{-0.048611in}}%
\pgfusepath{stroke,fill}%
}%
\begin{pgfscope}%
\pgfsys@transformshift{2.723959in}{0.964913in}%
\pgfsys@useobject{currentmarker}{}%
\end{pgfscope}%
\end{pgfscope}%
\begin{pgfscope}%
\definecolor{textcolor}{rgb}{0.000000,0.000000,0.000000}%
\pgfsetstrokecolor{textcolor}%
\pgfsetfillcolor{textcolor}%
\pgftext[x=2.723959in,y=0.867691in,,top]{\color{textcolor}\fontsize{20.000000}{24.000000}\selectfont \(\displaystyle {25}\)}%
\end{pgfscope}%
\begin{pgfscope}%
\pgfsetbuttcap%
\pgfsetroundjoin%
\definecolor{currentfill}{rgb}{0.000000,0.000000,0.000000}%
\pgfsetfillcolor{currentfill}%
\pgfsetlinewidth{0.803000pt}%
\definecolor{currentstroke}{rgb}{0.000000,0.000000,0.000000}%
\pgfsetstrokecolor{currentstroke}%
\pgfsetdash{}{0pt}%
\pgfsys@defobject{currentmarker}{\pgfqpoint{0.000000in}{-0.048611in}}{\pgfqpoint{0.000000in}{0.000000in}}{%
\pgfpathmoveto{\pgfqpoint{0.000000in}{0.000000in}}%
\pgfpathlineto{\pgfqpoint{0.000000in}{-0.048611in}}%
\pgfusepath{stroke,fill}%
}%
\begin{pgfscope}%
\pgfsys@transformshift{3.779631in}{0.964913in}%
\pgfsys@useobject{currentmarker}{}%
\end{pgfscope}%
\end{pgfscope}%
\begin{pgfscope}%
\definecolor{textcolor}{rgb}{0.000000,0.000000,0.000000}%
\pgfsetstrokecolor{textcolor}%
\pgfsetfillcolor{textcolor}%
\pgftext[x=3.779631in,y=0.867691in,,top]{\color{textcolor}\fontsize{20.000000}{24.000000}\selectfont \(\displaystyle {50}\)}%
\end{pgfscope}%
\begin{pgfscope}%
\pgfsetbuttcap%
\pgfsetroundjoin%
\definecolor{currentfill}{rgb}{0.000000,0.000000,0.000000}%
\pgfsetfillcolor{currentfill}%
\pgfsetlinewidth{0.803000pt}%
\definecolor{currentstroke}{rgb}{0.000000,0.000000,0.000000}%
\pgfsetstrokecolor{currentstroke}%
\pgfsetdash{}{0pt}%
\pgfsys@defobject{currentmarker}{\pgfqpoint{0.000000in}{-0.048611in}}{\pgfqpoint{0.000000in}{0.000000in}}{%
\pgfpathmoveto{\pgfqpoint{0.000000in}{0.000000in}}%
\pgfpathlineto{\pgfqpoint{0.000000in}{-0.048611in}}%
\pgfusepath{stroke,fill}%
}%
\begin{pgfscope}%
\pgfsys@transformshift{4.835304in}{0.964913in}%
\pgfsys@useobject{currentmarker}{}%
\end{pgfscope}%
\end{pgfscope}%
\begin{pgfscope}%
\definecolor{textcolor}{rgb}{0.000000,0.000000,0.000000}%
\pgfsetstrokecolor{textcolor}%
\pgfsetfillcolor{textcolor}%
\pgftext[x=4.835304in,y=0.867691in,,top]{\color{textcolor}\fontsize{20.000000}{24.000000}\selectfont \(\displaystyle {75}\)}%
\end{pgfscope}%
\begin{pgfscope}%
\pgfsetbuttcap%
\pgfsetroundjoin%
\definecolor{currentfill}{rgb}{0.000000,0.000000,0.000000}%
\pgfsetfillcolor{currentfill}%
\pgfsetlinewidth{0.803000pt}%
\definecolor{currentstroke}{rgb}{0.000000,0.000000,0.000000}%
\pgfsetstrokecolor{currentstroke}%
\pgfsetdash{}{0pt}%
\pgfsys@defobject{currentmarker}{\pgfqpoint{0.000000in}{-0.048611in}}{\pgfqpoint{0.000000in}{0.000000in}}{%
\pgfpathmoveto{\pgfqpoint{0.000000in}{0.000000in}}%
\pgfpathlineto{\pgfqpoint{0.000000in}{-0.048611in}}%
\pgfusepath{stroke,fill}%
}%
\begin{pgfscope}%
\pgfsys@transformshift{5.890977in}{0.964913in}%
\pgfsys@useobject{currentmarker}{}%
\end{pgfscope}%
\end{pgfscope}%
\begin{pgfscope}%
\definecolor{textcolor}{rgb}{0.000000,0.000000,0.000000}%
\pgfsetstrokecolor{textcolor}%
\pgfsetfillcolor{textcolor}%
\pgftext[x=5.890977in,y=0.867691in,,top]{\color{textcolor}\fontsize{20.000000}{24.000000}\selectfont \(\displaystyle {100}\)}%
\end{pgfscope}%
\begin{pgfscope}%
\definecolor{textcolor}{rgb}{0.000000,0.000000,0.000000}%
\pgfsetstrokecolor{textcolor}%
\pgfsetfillcolor{textcolor}%
\pgftext[x=3.800745in,y=0.556068in,,top]{\color{textcolor}\fontsize{20.000000}{24.000000}\selectfont Number links in the stem}%
\end{pgfscope}%
\begin{pgfscope}%
\pgfsetbuttcap%
\pgfsetroundjoin%
\definecolor{currentfill}{rgb}{0.000000,0.000000,0.000000}%
\pgfsetfillcolor{currentfill}%
\pgfsetlinewidth{0.803000pt}%
\definecolor{currentstroke}{rgb}{0.000000,0.000000,0.000000}%
\pgfsetstrokecolor{currentstroke}%
\pgfsetdash{}{0pt}%
\pgfsys@defobject{currentmarker}{\pgfqpoint{-0.048611in}{0.000000in}}{\pgfqpoint{-0.000000in}{0.000000in}}{%
\pgfpathmoveto{\pgfqpoint{-0.000000in}{0.000000in}}%
\pgfpathlineto{\pgfqpoint{-0.048611in}{0.000000in}}%
\pgfusepath{stroke,fill}%
}%
\begin{pgfscope}%
\pgfsys@transformshift{1.501490in}{1.797072in}%
\pgfsys@useobject{currentmarker}{}%
\end{pgfscope}%
\end{pgfscope}%
\begin{pgfscope}%
\definecolor{textcolor}{rgb}{0.000000,0.000000,0.000000}%
\pgfsetstrokecolor{textcolor}%
\pgfsetfillcolor{textcolor}%
\pgftext[x=0.611623in, y=1.697053in, left, base]{\color{textcolor}\fontsize{20.000000}{24.000000}\selectfont \(\displaystyle {100000}\)}%
\end{pgfscope}%
\begin{pgfscope}%
\pgfsetbuttcap%
\pgfsetroundjoin%
\definecolor{currentfill}{rgb}{0.000000,0.000000,0.000000}%
\pgfsetfillcolor{currentfill}%
\pgfsetlinewidth{0.803000pt}%
\definecolor{currentstroke}{rgb}{0.000000,0.000000,0.000000}%
\pgfsetstrokecolor{currentstroke}%
\pgfsetdash{}{0pt}%
\pgfsys@defobject{currentmarker}{\pgfqpoint{-0.048611in}{0.000000in}}{\pgfqpoint{-0.000000in}{0.000000in}}{%
\pgfpathmoveto{\pgfqpoint{-0.000000in}{0.000000in}}%
\pgfpathlineto{\pgfqpoint{-0.048611in}{0.000000in}}%
\pgfusepath{stroke,fill}%
}%
\begin{pgfscope}%
\pgfsys@transformshift{1.501490in}{2.958449in}%
\pgfsys@useobject{currentmarker}{}%
\end{pgfscope}%
\end{pgfscope}%
\begin{pgfscope}%
\definecolor{textcolor}{rgb}{0.000000,0.000000,0.000000}%
\pgfsetstrokecolor{textcolor}%
\pgfsetfillcolor{textcolor}%
\pgftext[x=0.611623in, y=2.858430in, left, base]{\color{textcolor}\fontsize{20.000000}{24.000000}\selectfont \(\displaystyle {200000}\)}%
\end{pgfscope}%
\begin{pgfscope}%
\pgfsetbuttcap%
\pgfsetroundjoin%
\definecolor{currentfill}{rgb}{0.000000,0.000000,0.000000}%
\pgfsetfillcolor{currentfill}%
\pgfsetlinewidth{0.803000pt}%
\definecolor{currentstroke}{rgb}{0.000000,0.000000,0.000000}%
\pgfsetstrokecolor{currentstroke}%
\pgfsetdash{}{0pt}%
\pgfsys@defobject{currentmarker}{\pgfqpoint{-0.048611in}{0.000000in}}{\pgfqpoint{-0.000000in}{0.000000in}}{%
\pgfpathmoveto{\pgfqpoint{-0.000000in}{0.000000in}}%
\pgfpathlineto{\pgfqpoint{-0.048611in}{0.000000in}}%
\pgfusepath{stroke,fill}%
}%
\begin{pgfscope}%
\pgfsys@transformshift{1.501490in}{4.119826in}%
\pgfsys@useobject{currentmarker}{}%
\end{pgfscope}%
\end{pgfscope}%
\begin{pgfscope}%
\definecolor{textcolor}{rgb}{0.000000,0.000000,0.000000}%
\pgfsetstrokecolor{textcolor}%
\pgfsetfillcolor{textcolor}%
\pgftext[x=0.611623in, y=4.019806in, left, base]{\color{textcolor}\fontsize{20.000000}{24.000000}\selectfont \(\displaystyle {300000}\)}%
\end{pgfscope}%
\begin{pgfscope}%
\definecolor{textcolor}{rgb}{0.000000,0.000000,0.000000}%
\pgfsetstrokecolor{textcolor}%
\pgfsetfillcolor{textcolor}%
\pgftext[x=0.556068in,y=2.732457in,,bottom,rotate=90.000000]{\color{textcolor}\fontsize{20.000000}{24.000000}\selectfont Number of operations}%
\end{pgfscope}%
\begin{pgfscope}%
\pgfpathrectangle{\pgfqpoint{1.501490in}{0.964913in}}{\pgfqpoint{4.598510in}{3.535087in}}%
\pgfusepath{clip}%
\pgfsetrectcap%
\pgfsetroundjoin%
\pgfsetlinewidth{1.505625pt}%
\definecolor{currentstroke}{rgb}{0.121569,0.466667,0.705882}%
\pgfsetstrokecolor{currentstroke}%
\pgfsetdash{}{0pt}%
\pgfpathmoveto{\pgfqpoint{1.710513in}{1.125599in}}%
\pgfpathlineto{\pgfqpoint{1.752740in}{1.168198in}}%
\pgfpathlineto{\pgfqpoint{1.794967in}{1.191658in}}%
\pgfpathlineto{\pgfqpoint{1.837194in}{1.215118in}}%
\pgfpathlineto{\pgfqpoint{1.879421in}{1.238578in}}%
\pgfpathlineto{\pgfqpoint{1.921648in}{1.262037in}}%
\pgfpathlineto{\pgfqpoint{1.963874in}{1.285497in}}%
\pgfpathlineto{\pgfqpoint{2.006101in}{1.308957in}}%
\pgfpathlineto{\pgfqpoint{2.048328in}{1.332417in}}%
\pgfpathlineto{\pgfqpoint{2.090555in}{1.355877in}}%
\pgfpathlineto{\pgfqpoint{2.132782in}{1.379336in}}%
\pgfpathlineto{\pgfqpoint{2.175009in}{1.402796in}}%
\pgfpathlineto{\pgfqpoint{2.217236in}{1.426256in}}%
\pgfpathlineto{\pgfqpoint{2.259463in}{1.449716in}}%
\pgfpathlineto{\pgfqpoint{2.301690in}{1.473176in}}%
\pgfpathlineto{\pgfqpoint{2.343917in}{1.496635in}}%
\pgfpathlineto{\pgfqpoint{2.386144in}{1.520095in}}%
\pgfpathlineto{\pgfqpoint{2.428370in}{1.543555in}}%
\pgfpathlineto{\pgfqpoint{2.470597in}{1.567015in}}%
\pgfpathlineto{\pgfqpoint{2.512824in}{1.590475in}}%
\pgfpathlineto{\pgfqpoint{2.555051in}{1.613935in}}%
\pgfpathlineto{\pgfqpoint{2.597278in}{1.637394in}}%
\pgfpathlineto{\pgfqpoint{2.639505in}{1.660854in}}%
\pgfpathlineto{\pgfqpoint{2.681732in}{1.684314in}}%
\pgfpathlineto{\pgfqpoint{2.723959in}{1.707774in}}%
\pgfpathlineto{\pgfqpoint{2.766186in}{1.731234in}}%
\pgfpathlineto{\pgfqpoint{2.808413in}{1.754693in}}%
\pgfpathlineto{\pgfqpoint{2.850640in}{1.778153in}}%
\pgfpathlineto{\pgfqpoint{2.892866in}{1.801613in}}%
\pgfpathlineto{\pgfqpoint{2.935093in}{1.825073in}}%
\pgfpathlineto{\pgfqpoint{2.977320in}{1.848533in}}%
\pgfpathlineto{\pgfqpoint{3.019547in}{1.871992in}}%
\pgfpathlineto{\pgfqpoint{3.061774in}{1.895452in}}%
\pgfpathlineto{\pgfqpoint{3.104001in}{1.918912in}}%
\pgfpathlineto{\pgfqpoint{3.146228in}{1.942372in}}%
\pgfpathlineto{\pgfqpoint{3.188455in}{1.965832in}}%
\pgfpathlineto{\pgfqpoint{3.230682in}{1.989292in}}%
\pgfpathlineto{\pgfqpoint{3.272909in}{2.012751in}}%
\pgfpathlineto{\pgfqpoint{3.315136in}{2.036211in}}%
\pgfpathlineto{\pgfqpoint{3.357362in}{2.059671in}}%
\pgfpathlineto{\pgfqpoint{3.399589in}{2.083131in}}%
\pgfpathlineto{\pgfqpoint{3.441816in}{2.106591in}}%
\pgfpathlineto{\pgfqpoint{3.484043in}{2.130050in}}%
\pgfpathlineto{\pgfqpoint{3.526270in}{2.153510in}}%
\pgfpathlineto{\pgfqpoint{3.568497in}{2.176970in}}%
\pgfpathlineto{\pgfqpoint{3.610724in}{2.200430in}}%
\pgfpathlineto{\pgfqpoint{3.652951in}{2.223890in}}%
\pgfpathlineto{\pgfqpoint{3.695178in}{2.247349in}}%
\pgfpathlineto{\pgfqpoint{3.737405in}{2.270809in}}%
\pgfpathlineto{\pgfqpoint{3.779631in}{2.294269in}}%
\pgfpathlineto{\pgfqpoint{3.821858in}{2.317729in}}%
\pgfpathlineto{\pgfqpoint{3.864085in}{2.341189in}}%
\pgfpathlineto{\pgfqpoint{3.906312in}{2.364648in}}%
\pgfpathlineto{\pgfqpoint{3.948539in}{2.388108in}}%
\pgfpathlineto{\pgfqpoint{3.990766in}{2.411568in}}%
\pgfpathlineto{\pgfqpoint{4.032993in}{2.435028in}}%
\pgfpathlineto{\pgfqpoint{4.075220in}{2.458488in}}%
\pgfpathlineto{\pgfqpoint{4.117447in}{2.481948in}}%
\pgfpathlineto{\pgfqpoint{4.159674in}{2.505407in}}%
\pgfpathlineto{\pgfqpoint{4.201901in}{2.528867in}}%
\pgfpathlineto{\pgfqpoint{4.244127in}{2.552327in}}%
\pgfpathlineto{\pgfqpoint{4.286354in}{2.575787in}}%
\pgfpathlineto{\pgfqpoint{4.328581in}{2.599247in}}%
\pgfpathlineto{\pgfqpoint{4.370808in}{2.622706in}}%
\pgfpathlineto{\pgfqpoint{4.413035in}{2.646166in}}%
\pgfpathlineto{\pgfqpoint{4.455262in}{2.669626in}}%
\pgfpathlineto{\pgfqpoint{4.497489in}{2.693086in}}%
\pgfpathlineto{\pgfqpoint{4.539716in}{2.716546in}}%
\pgfpathlineto{\pgfqpoint{4.581943in}{2.740005in}}%
\pgfpathlineto{\pgfqpoint{4.624170in}{2.763465in}}%
\pgfpathlineto{\pgfqpoint{4.666397in}{2.786925in}}%
\pgfpathlineto{\pgfqpoint{4.708623in}{2.810385in}}%
\pgfpathlineto{\pgfqpoint{4.750850in}{2.833845in}}%
\pgfpathlineto{\pgfqpoint{4.793077in}{2.857305in}}%
\pgfpathlineto{\pgfqpoint{4.835304in}{2.880764in}}%
\pgfpathlineto{\pgfqpoint{4.877531in}{2.904224in}}%
\pgfpathlineto{\pgfqpoint{4.919758in}{2.927684in}}%
\pgfpathlineto{\pgfqpoint{4.961985in}{2.951144in}}%
\pgfpathlineto{\pgfqpoint{5.004212in}{2.974604in}}%
\pgfpathlineto{\pgfqpoint{5.046439in}{2.998063in}}%
\pgfpathlineto{\pgfqpoint{5.088666in}{3.021523in}}%
\pgfpathlineto{\pgfqpoint{5.130892in}{3.044983in}}%
\pgfpathlineto{\pgfqpoint{5.173119in}{3.068443in}}%
\pgfpathlineto{\pgfqpoint{5.215346in}{3.091903in}}%
\pgfpathlineto{\pgfqpoint{5.257573in}{3.115362in}}%
\pgfpathlineto{\pgfqpoint{5.299800in}{3.138822in}}%
\pgfpathlineto{\pgfqpoint{5.342027in}{3.162282in}}%
\pgfpathlineto{\pgfqpoint{5.384254in}{3.185742in}}%
\pgfpathlineto{\pgfqpoint{5.426481in}{3.209202in}}%
\pgfpathlineto{\pgfqpoint{5.468708in}{3.232662in}}%
\pgfpathlineto{\pgfqpoint{5.510935in}{3.256121in}}%
\pgfpathlineto{\pgfqpoint{5.553162in}{3.279581in}}%
\pgfpathlineto{\pgfqpoint{5.595388in}{3.303041in}}%
\pgfpathlineto{\pgfqpoint{5.637615in}{3.326501in}}%
\pgfpathlineto{\pgfqpoint{5.679842in}{3.349961in}}%
\pgfpathlineto{\pgfqpoint{5.722069in}{3.373420in}}%
\pgfpathlineto{\pgfqpoint{5.764296in}{3.396880in}}%
\pgfpathlineto{\pgfqpoint{5.806523in}{3.420340in}}%
\pgfpathlineto{\pgfqpoint{5.848750in}{3.443800in}}%
\pgfpathlineto{\pgfqpoint{5.890977in}{3.467260in}}%
\pgfusepath{stroke}%
\end{pgfscope}%
\begin{pgfscope}%
\pgfpathrectangle{\pgfqpoint{1.501490in}{0.964913in}}{\pgfqpoint{4.598510in}{3.535087in}}%
\pgfusepath{clip}%
\pgfsetrectcap%
\pgfsetroundjoin%
\pgfsetlinewidth{1.505625pt}%
\definecolor{currentstroke}{rgb}{1.000000,0.498039,0.054902}%
\pgfsetstrokecolor{currentstroke}%
\pgfsetdash{}{0pt}%
\pgfpathmoveto{\pgfqpoint{1.710513in}{1.266451in}}%
\pgfpathlineto{\pgfqpoint{1.752740in}{1.311837in}}%
\pgfpathlineto{\pgfqpoint{1.794967in}{1.342730in}}%
\pgfpathlineto{\pgfqpoint{1.837194in}{1.373622in}}%
\pgfpathlineto{\pgfqpoint{1.879421in}{1.404515in}}%
\pgfpathlineto{\pgfqpoint{1.921648in}{1.435408in}}%
\pgfpathlineto{\pgfqpoint{1.963874in}{1.466300in}}%
\pgfpathlineto{\pgfqpoint{2.006101in}{1.497193in}}%
\pgfpathlineto{\pgfqpoint{2.048328in}{1.528086in}}%
\pgfpathlineto{\pgfqpoint{2.090555in}{1.558978in}}%
\pgfpathlineto{\pgfqpoint{2.132782in}{1.589871in}}%
\pgfpathlineto{\pgfqpoint{2.175009in}{1.620763in}}%
\pgfpathlineto{\pgfqpoint{2.217236in}{1.651656in}}%
\pgfpathlineto{\pgfqpoint{2.259463in}{1.682549in}}%
\pgfpathlineto{\pgfqpoint{2.301690in}{1.713441in}}%
\pgfpathlineto{\pgfqpoint{2.343917in}{1.744334in}}%
\pgfpathlineto{\pgfqpoint{2.386144in}{1.775227in}}%
\pgfpathlineto{\pgfqpoint{2.428370in}{1.806119in}}%
\pgfpathlineto{\pgfqpoint{2.470597in}{1.837012in}}%
\pgfpathlineto{\pgfqpoint{2.512824in}{1.867904in}}%
\pgfpathlineto{\pgfqpoint{2.555051in}{1.898797in}}%
\pgfpathlineto{\pgfqpoint{2.597278in}{1.929690in}}%
\pgfpathlineto{\pgfqpoint{2.639505in}{1.960582in}}%
\pgfpathlineto{\pgfqpoint{2.681732in}{1.991475in}}%
\pgfpathlineto{\pgfqpoint{2.723959in}{2.022368in}}%
\pgfpathlineto{\pgfqpoint{2.766186in}{2.053260in}}%
\pgfpathlineto{\pgfqpoint{2.808413in}{2.084153in}}%
\pgfpathlineto{\pgfqpoint{2.850640in}{2.115045in}}%
\pgfpathlineto{\pgfqpoint{2.892866in}{2.145938in}}%
\pgfpathlineto{\pgfqpoint{2.935093in}{2.176831in}}%
\pgfpathlineto{\pgfqpoint{2.977320in}{2.207723in}}%
\pgfpathlineto{\pgfqpoint{3.019547in}{2.238616in}}%
\pgfpathlineto{\pgfqpoint{3.061774in}{2.269509in}}%
\pgfpathlineto{\pgfqpoint{3.104001in}{2.300401in}}%
\pgfpathlineto{\pgfqpoint{3.146228in}{2.331294in}}%
\pgfpathlineto{\pgfqpoint{3.188455in}{2.362186in}}%
\pgfpathlineto{\pgfqpoint{3.230682in}{2.393079in}}%
\pgfpathlineto{\pgfqpoint{3.272909in}{2.423972in}}%
\pgfpathlineto{\pgfqpoint{3.315136in}{2.454864in}}%
\pgfpathlineto{\pgfqpoint{3.357362in}{2.485757in}}%
\pgfpathlineto{\pgfqpoint{3.399589in}{2.516649in}}%
\pgfpathlineto{\pgfqpoint{3.441816in}{2.547542in}}%
\pgfpathlineto{\pgfqpoint{3.484043in}{2.578435in}}%
\pgfpathlineto{\pgfqpoint{3.526270in}{2.609327in}}%
\pgfpathlineto{\pgfqpoint{3.568497in}{2.640220in}}%
\pgfpathlineto{\pgfqpoint{3.610724in}{2.671113in}}%
\pgfpathlineto{\pgfqpoint{3.652951in}{2.702005in}}%
\pgfpathlineto{\pgfqpoint{3.695178in}{2.732898in}}%
\pgfpathlineto{\pgfqpoint{3.737405in}{2.763790in}}%
\pgfpathlineto{\pgfqpoint{3.779631in}{2.794683in}}%
\pgfpathlineto{\pgfqpoint{3.821858in}{2.825576in}}%
\pgfpathlineto{\pgfqpoint{3.864085in}{2.856468in}}%
\pgfpathlineto{\pgfqpoint{3.906312in}{2.887361in}}%
\pgfpathlineto{\pgfqpoint{3.948539in}{2.918254in}}%
\pgfpathlineto{\pgfqpoint{3.990766in}{2.949146in}}%
\pgfpathlineto{\pgfqpoint{4.032993in}{2.980039in}}%
\pgfpathlineto{\pgfqpoint{4.075220in}{3.010931in}}%
\pgfpathlineto{\pgfqpoint{4.117447in}{3.041824in}}%
\pgfpathlineto{\pgfqpoint{4.159674in}{3.072717in}}%
\pgfpathlineto{\pgfqpoint{4.201901in}{3.103609in}}%
\pgfpathlineto{\pgfqpoint{4.244127in}{3.134502in}}%
\pgfpathlineto{\pgfqpoint{4.286354in}{3.165395in}}%
\pgfpathlineto{\pgfqpoint{4.328581in}{3.196287in}}%
\pgfpathlineto{\pgfqpoint{4.370808in}{3.227180in}}%
\pgfpathlineto{\pgfqpoint{4.413035in}{3.258072in}}%
\pgfpathlineto{\pgfqpoint{4.455262in}{3.288965in}}%
\pgfpathlineto{\pgfqpoint{4.497489in}{3.319858in}}%
\pgfpathlineto{\pgfqpoint{4.539716in}{3.350750in}}%
\pgfpathlineto{\pgfqpoint{4.581943in}{3.381643in}}%
\pgfpathlineto{\pgfqpoint{4.624170in}{3.412536in}}%
\pgfpathlineto{\pgfqpoint{4.666397in}{3.443428in}}%
\pgfpathlineto{\pgfqpoint{4.708623in}{3.474321in}}%
\pgfpathlineto{\pgfqpoint{4.750850in}{3.505213in}}%
\pgfpathlineto{\pgfqpoint{4.793077in}{3.536106in}}%
\pgfpathlineto{\pgfqpoint{4.835304in}{3.566999in}}%
\pgfpathlineto{\pgfqpoint{4.877531in}{3.597891in}}%
\pgfpathlineto{\pgfqpoint{4.919758in}{3.628784in}}%
\pgfpathlineto{\pgfqpoint{4.961985in}{3.659677in}}%
\pgfpathlineto{\pgfqpoint{5.004212in}{3.690569in}}%
\pgfpathlineto{\pgfqpoint{5.046439in}{3.721462in}}%
\pgfpathlineto{\pgfqpoint{5.088666in}{3.752354in}}%
\pgfpathlineto{\pgfqpoint{5.130892in}{3.783247in}}%
\pgfpathlineto{\pgfqpoint{5.173119in}{3.814140in}}%
\pgfpathlineto{\pgfqpoint{5.215346in}{3.845032in}}%
\pgfpathlineto{\pgfqpoint{5.257573in}{3.875925in}}%
\pgfpathlineto{\pgfqpoint{5.299800in}{3.906818in}}%
\pgfpathlineto{\pgfqpoint{5.342027in}{3.937710in}}%
\pgfpathlineto{\pgfqpoint{5.384254in}{3.968603in}}%
\pgfpathlineto{\pgfqpoint{5.426481in}{3.999495in}}%
\pgfpathlineto{\pgfqpoint{5.468708in}{4.030388in}}%
\pgfpathlineto{\pgfqpoint{5.510935in}{4.061281in}}%
\pgfpathlineto{\pgfqpoint{5.553162in}{4.092173in}}%
\pgfpathlineto{\pgfqpoint{5.595388in}{4.123066in}}%
\pgfpathlineto{\pgfqpoint{5.637615in}{4.153958in}}%
\pgfpathlineto{\pgfqpoint{5.679842in}{4.184851in}}%
\pgfpathlineto{\pgfqpoint{5.722069in}{4.215744in}}%
\pgfpathlineto{\pgfqpoint{5.764296in}{4.246636in}}%
\pgfpathlineto{\pgfqpoint{5.806523in}{4.277529in}}%
\pgfpathlineto{\pgfqpoint{5.848750in}{4.308422in}}%
\pgfpathlineto{\pgfqpoint{5.890977in}{4.339314in}}%
\pgfusepath{stroke}%
\end{pgfscope}%
\begin{pgfscope}%
\pgfpathrectangle{\pgfqpoint{1.501490in}{0.964913in}}{\pgfqpoint{4.598510in}{3.535087in}}%
\pgfusepath{clip}%
\pgfsetrectcap%
\pgfsetroundjoin%
\pgfsetlinewidth{1.505625pt}%
\definecolor{currentstroke}{rgb}{0.172549,0.627451,0.172549}%
\pgfsetstrokecolor{currentstroke}%
\pgfsetdash{}{0pt}%
\pgfpathmoveto{\pgfqpoint{1.710513in}{1.125599in}}%
\pgfpathlineto{\pgfqpoint{1.752740in}{1.158756in}}%
\pgfpathlineto{\pgfqpoint{1.794967in}{1.173273in}}%
\pgfpathlineto{\pgfqpoint{1.837194in}{1.179963in}}%
\pgfpathlineto{\pgfqpoint{1.879421in}{1.186652in}}%
\pgfpathlineto{\pgfqpoint{1.921648in}{1.193342in}}%
\pgfpathlineto{\pgfqpoint{1.963874in}{1.200031in}}%
\pgfpathlineto{\pgfqpoint{2.006101in}{1.206721in}}%
\pgfpathlineto{\pgfqpoint{2.048328in}{1.213411in}}%
\pgfpathlineto{\pgfqpoint{2.090555in}{1.220100in}}%
\pgfpathlineto{\pgfqpoint{2.132782in}{1.226790in}}%
\pgfpathlineto{\pgfqpoint{2.175009in}{1.233479in}}%
\pgfpathlineto{\pgfqpoint{2.217236in}{1.240169in}}%
\pgfpathlineto{\pgfqpoint{2.259463in}{1.246858in}}%
\pgfpathlineto{\pgfqpoint{2.301690in}{1.253548in}}%
\pgfpathlineto{\pgfqpoint{2.343917in}{1.260237in}}%
\pgfpathlineto{\pgfqpoint{2.386144in}{1.266927in}}%
\pgfpathlineto{\pgfqpoint{2.428370in}{1.273616in}}%
\pgfpathlineto{\pgfqpoint{2.470597in}{1.280306in}}%
\pgfpathlineto{\pgfqpoint{2.512824in}{1.286995in}}%
\pgfpathlineto{\pgfqpoint{2.555051in}{1.293685in}}%
\pgfpathlineto{\pgfqpoint{2.597278in}{1.300374in}}%
\pgfpathlineto{\pgfqpoint{2.639505in}{1.307064in}}%
\pgfpathlineto{\pgfqpoint{2.681732in}{1.313753in}}%
\pgfpathlineto{\pgfqpoint{2.723959in}{1.320443in}}%
\pgfpathlineto{\pgfqpoint{2.766186in}{1.327133in}}%
\pgfpathlineto{\pgfqpoint{2.808413in}{1.333822in}}%
\pgfpathlineto{\pgfqpoint{2.850640in}{1.340512in}}%
\pgfpathlineto{\pgfqpoint{2.892866in}{1.347201in}}%
\pgfpathlineto{\pgfqpoint{2.935093in}{1.353891in}}%
\pgfpathlineto{\pgfqpoint{2.977320in}{1.360580in}}%
\pgfpathlineto{\pgfqpoint{3.019547in}{1.367270in}}%
\pgfpathlineto{\pgfqpoint{3.061774in}{1.373959in}}%
\pgfpathlineto{\pgfqpoint{3.104001in}{1.380649in}}%
\pgfpathlineto{\pgfqpoint{3.146228in}{1.387338in}}%
\pgfpathlineto{\pgfqpoint{3.188455in}{1.394028in}}%
\pgfpathlineto{\pgfqpoint{3.230682in}{1.400717in}}%
\pgfpathlineto{\pgfqpoint{3.272909in}{1.407407in}}%
\pgfpathlineto{\pgfqpoint{3.315136in}{1.414096in}}%
\pgfpathlineto{\pgfqpoint{3.357362in}{1.420786in}}%
\pgfpathlineto{\pgfqpoint{3.399589in}{1.427475in}}%
\pgfpathlineto{\pgfqpoint{3.441816in}{1.434165in}}%
\pgfpathlineto{\pgfqpoint{3.484043in}{1.440855in}}%
\pgfpathlineto{\pgfqpoint{3.526270in}{1.447544in}}%
\pgfpathlineto{\pgfqpoint{3.568497in}{1.454234in}}%
\pgfpathlineto{\pgfqpoint{3.610724in}{1.460923in}}%
\pgfpathlineto{\pgfqpoint{3.652951in}{1.467613in}}%
\pgfpathlineto{\pgfqpoint{3.695178in}{1.474302in}}%
\pgfpathlineto{\pgfqpoint{3.737405in}{1.480992in}}%
\pgfpathlineto{\pgfqpoint{3.779631in}{1.487681in}}%
\pgfpathlineto{\pgfqpoint{3.821858in}{1.494371in}}%
\pgfpathlineto{\pgfqpoint{3.864085in}{1.501060in}}%
\pgfpathlineto{\pgfqpoint{3.906312in}{1.507750in}}%
\pgfpathlineto{\pgfqpoint{3.948539in}{1.514439in}}%
\pgfpathlineto{\pgfqpoint{3.990766in}{1.521129in}}%
\pgfpathlineto{\pgfqpoint{4.032993in}{1.527818in}}%
\pgfpathlineto{\pgfqpoint{4.075220in}{1.534508in}}%
\pgfpathlineto{\pgfqpoint{4.117447in}{1.541198in}}%
\pgfpathlineto{\pgfqpoint{4.159674in}{1.547887in}}%
\pgfpathlineto{\pgfqpoint{4.201901in}{1.554577in}}%
\pgfpathlineto{\pgfqpoint{4.244127in}{1.561266in}}%
\pgfpathlineto{\pgfqpoint{4.286354in}{1.567956in}}%
\pgfpathlineto{\pgfqpoint{4.328581in}{1.574645in}}%
\pgfpathlineto{\pgfqpoint{4.370808in}{1.581335in}}%
\pgfpathlineto{\pgfqpoint{4.413035in}{1.588024in}}%
\pgfpathlineto{\pgfqpoint{4.455262in}{1.594714in}}%
\pgfpathlineto{\pgfqpoint{4.497489in}{1.601403in}}%
\pgfpathlineto{\pgfqpoint{4.539716in}{1.608093in}}%
\pgfpathlineto{\pgfqpoint{4.581943in}{1.614782in}}%
\pgfpathlineto{\pgfqpoint{4.624170in}{1.621472in}}%
\pgfpathlineto{\pgfqpoint{4.666397in}{1.628161in}}%
\pgfpathlineto{\pgfqpoint{4.708623in}{1.634851in}}%
\pgfpathlineto{\pgfqpoint{4.750850in}{1.641540in}}%
\pgfpathlineto{\pgfqpoint{4.793077in}{1.648230in}}%
\pgfpathlineto{\pgfqpoint{4.835304in}{1.654920in}}%
\pgfpathlineto{\pgfqpoint{4.877531in}{1.661609in}}%
\pgfpathlineto{\pgfqpoint{4.919758in}{1.668299in}}%
\pgfpathlineto{\pgfqpoint{4.961985in}{1.674988in}}%
\pgfpathlineto{\pgfqpoint{5.004212in}{1.681678in}}%
\pgfpathlineto{\pgfqpoint{5.046439in}{1.688367in}}%
\pgfpathlineto{\pgfqpoint{5.088666in}{1.695057in}}%
\pgfpathlineto{\pgfqpoint{5.130892in}{1.701746in}}%
\pgfpathlineto{\pgfqpoint{5.173119in}{1.708436in}}%
\pgfpathlineto{\pgfqpoint{5.215346in}{1.715125in}}%
\pgfpathlineto{\pgfqpoint{5.257573in}{1.721815in}}%
\pgfpathlineto{\pgfqpoint{5.299800in}{1.728504in}}%
\pgfpathlineto{\pgfqpoint{5.342027in}{1.735194in}}%
\pgfpathlineto{\pgfqpoint{5.384254in}{1.741883in}}%
\pgfpathlineto{\pgfqpoint{5.426481in}{1.748573in}}%
\pgfpathlineto{\pgfqpoint{5.468708in}{1.755262in}}%
\pgfpathlineto{\pgfqpoint{5.510935in}{1.761952in}}%
\pgfpathlineto{\pgfqpoint{5.553162in}{1.768642in}}%
\pgfpathlineto{\pgfqpoint{5.595388in}{1.775331in}}%
\pgfpathlineto{\pgfqpoint{5.637615in}{1.782021in}}%
\pgfpathlineto{\pgfqpoint{5.679842in}{1.788710in}}%
\pgfpathlineto{\pgfqpoint{5.722069in}{1.795400in}}%
\pgfpathlineto{\pgfqpoint{5.764296in}{1.802089in}}%
\pgfpathlineto{\pgfqpoint{5.806523in}{1.808779in}}%
\pgfpathlineto{\pgfqpoint{5.848750in}{1.815468in}}%
\pgfpathlineto{\pgfqpoint{5.890977in}{1.822158in}}%
\pgfusepath{stroke}%
\end{pgfscope}%
\begin{pgfscope}%
\pgfsetrectcap%
\pgfsetmiterjoin%
\pgfsetlinewidth{0.803000pt}%
\definecolor{currentstroke}{rgb}{0.000000,0.000000,0.000000}%
\pgfsetstrokecolor{currentstroke}%
\pgfsetdash{}{0pt}%
\pgfpathmoveto{\pgfqpoint{1.501490in}{0.964913in}}%
\pgfpathlineto{\pgfqpoint{1.501490in}{4.500000in}}%
\pgfusepath{stroke}%
\end{pgfscope}%
\begin{pgfscope}%
\pgfsetrectcap%
\pgfsetmiterjoin%
\pgfsetlinewidth{0.803000pt}%
\definecolor{currentstroke}{rgb}{0.000000,0.000000,0.000000}%
\pgfsetstrokecolor{currentstroke}%
\pgfsetdash{}{0pt}%
\pgfpathmoveto{\pgfqpoint{6.100000in}{0.964913in}}%
\pgfpathlineto{\pgfqpoint{6.100000in}{4.500000in}}%
\pgfusepath{stroke}%
\end{pgfscope}%
\begin{pgfscope}%
\pgfsetrectcap%
\pgfsetmiterjoin%
\pgfsetlinewidth{0.803000pt}%
\definecolor{currentstroke}{rgb}{0.000000,0.000000,0.000000}%
\pgfsetstrokecolor{currentstroke}%
\pgfsetdash{}{0pt}%
\pgfpathmoveto{\pgfqpoint{1.501490in}{0.964913in}}%
\pgfpathlineto{\pgfqpoint{6.100000in}{0.964913in}}%
\pgfusepath{stroke}%
\end{pgfscope}%
\begin{pgfscope}%
\pgfsetrectcap%
\pgfsetmiterjoin%
\pgfsetlinewidth{0.803000pt}%
\definecolor{currentstroke}{rgb}{0.000000,0.000000,0.000000}%
\pgfsetstrokecolor{currentstroke}%
\pgfsetdash{}{0pt}%
\pgfpathmoveto{\pgfqpoint{1.501490in}{4.500000in}}%
\pgfpathlineto{\pgfqpoint{6.100000in}{4.500000in}}%
\pgfusepath{stroke}%
\end{pgfscope}%
\begin{pgfscope}%
\pgfsetbuttcap%
\pgfsetmiterjoin%
\definecolor{currentfill}{rgb}{1.000000,1.000000,1.000000}%
\pgfsetfillcolor{currentfill}%
\pgfsetfillopacity{0.800000}%
\pgfsetlinewidth{1.003750pt}%
\definecolor{currentstroke}{rgb}{0.800000,0.800000,0.800000}%
\pgfsetstrokecolor{currentstroke}%
\pgfsetstrokeopacity{0.800000}%
\pgfsetdash{}{0pt}%
\pgfpathmoveto{\pgfqpoint{1.695934in}{3.092908in}}%
\pgfpathlineto{\pgfqpoint{3.746195in}{3.092908in}}%
\pgfpathquadraticcurveto{\pgfqpoint{3.801750in}{3.092908in}}{\pgfqpoint{3.801750in}{3.148464in}}%
\pgfpathlineto{\pgfqpoint{3.801750in}{4.305556in}}%
\pgfpathquadraticcurveto{\pgfqpoint{3.801750in}{4.361111in}}{\pgfqpoint{3.746195in}{4.361111in}}%
\pgfpathlineto{\pgfqpoint{1.695934in}{4.361111in}}%
\pgfpathquadraticcurveto{\pgfqpoint{1.640379in}{4.361111in}}{\pgfqpoint{1.640379in}{4.305556in}}%
\pgfpathlineto{\pgfqpoint{1.640379in}{3.148464in}}%
\pgfpathquadraticcurveto{\pgfqpoint{1.640379in}{3.092908in}}{\pgfqpoint{1.695934in}{3.092908in}}%
\pgfpathclose%
\pgfusepath{stroke,fill}%
\end{pgfscope}%
\begin{pgfscope}%
\pgfsetrectcap%
\pgfsetroundjoin%
\pgfsetlinewidth{1.505625pt}%
\definecolor{currentstroke}{rgb}{0.121569,0.466667,0.705882}%
\pgfsetstrokecolor{currentstroke}%
\pgfsetdash{}{0pt}%
\pgfpathmoveto{\pgfqpoint{1.751490in}{4.147184in}}%
\pgfpathlineto{\pgfqpoint{2.307045in}{4.147184in}}%
\pgfusepath{stroke}%
\end{pgfscope}%
\begin{pgfscope}%
\definecolor{textcolor}{rgb}{0.000000,0.000000,0.000000}%
\pgfsetstrokecolor{textcolor}%
\pgfsetfillcolor{textcolor}%
\pgftext[x=2.529268in,y=4.049962in,left,base]{\color{textcolor}\fontsize{20.000000}{24.000000}\selectfont PV-OSIM}%
\end{pgfscope}%
\begin{pgfscope}%
\pgfsetrectcap%
\pgfsetroundjoin%
\pgfsetlinewidth{1.505625pt}%
\definecolor{currentstroke}{rgb}{1.000000,0.498039,0.054902}%
\pgfsetstrokecolor{currentstroke}%
\pgfsetdash{}{0pt}%
\pgfpathmoveto{\pgfqpoint{1.751490in}{3.752227in}}%
\pgfpathlineto{\pgfqpoint{2.307045in}{3.752227in}}%
\pgfusepath{stroke}%
\end{pgfscope}%
\begin{pgfscope}%
\definecolor{textcolor}{rgb}{0.000000,0.000000,0.000000}%
\pgfsetstrokecolor{textcolor}%
\pgfsetfillcolor{textcolor}%
\pgftext[x=2.529268in,y=3.655005in,left,base]{\color{textcolor}\fontsize{20.000000}{24.000000}\selectfont EFPA}%
\end{pgfscope}%
\begin{pgfscope}%
\pgfsetrectcap%
\pgfsetroundjoin%
\pgfsetlinewidth{1.505625pt}%
\definecolor{currentstroke}{rgb}{0.172549,0.627451,0.172549}%
\pgfsetstrokecolor{currentstroke}%
\pgfsetdash{}{0pt}%
\pgfpathmoveto{\pgfqpoint{1.751490in}{3.357271in}}%
\pgfpathlineto{\pgfqpoint{2.307045in}{3.357271in}}%
\pgfusepath{stroke}%
\end{pgfscope}%
\begin{pgfscope}%
\definecolor{textcolor}{rgb}{0.000000,0.000000,0.000000}%
\pgfsetstrokecolor{textcolor}%
\pgfsetfillcolor{textcolor}%
\pgftext[x=2.529268in,y=3.260048in,left,base]{\color{textcolor}\fontsize{20.000000}{24.000000}\selectfont PV-OSIMr}%
\end{pgfscope}%
\end{pgfpicture}%
\makeatother%
\endgroup%

%% file: graphics/tree_comparison7.pgf
%% Creator: Matplotlib, PGF backend
%%
%% To include the figure in your LaTeX document, write
%%   \input{<filename>.pgf}
%%
%% Make sure the required packages are loaded in your preamble
%%   \usepackage{pgf}
%%
%% Figures using additional raster images can only be included by \input if
%% they are in the same directory as the main LaTeX file. For loading figures
%% from other directories you can use the `import` package
%% and then include the figures with
%%   \import{<path to file>}{<filename>.pgf}
%%
%% Matplotlib used the following preamble
%%
\begingroup%
\makeatletter%
\begin{pgfpicture}%
\pgfpathrectangle{\pgfpointorigin}{\pgfqpoint{6.400000in}{4.800000in}}%
\pgfusepath{use as bounding box, clip}%
\begin{pgfscope}%
\pgfsetbuttcap%
\pgfsetmiterjoin%
\definecolor{currentfill}{rgb}{1.000000,1.000000,1.000000}%
\pgfsetfillcolor{currentfill}%
\pgfsetlinewidth{0.000000pt}%
\definecolor{currentstroke}{rgb}{1.000000,1.000000,1.000000}%
\pgfsetstrokecolor{currentstroke}%
\pgfsetdash{}{0pt}%
\pgfpathmoveto{\pgfqpoint{0.000000in}{0.000000in}}%
\pgfpathlineto{\pgfqpoint{6.400000in}{0.000000in}}%
\pgfpathlineto{\pgfqpoint{6.400000in}{4.800000in}}%
\pgfpathlineto{\pgfqpoint{0.000000in}{4.800000in}}%
\pgfpathclose%
\pgfusepath{fill}%
\end{pgfscope}%
\begin{pgfscope}%
\pgfsetbuttcap%
\pgfsetmiterjoin%
\definecolor{currentfill}{rgb}{1.000000,1.000000,1.000000}%
\pgfsetfillcolor{currentfill}%
\pgfsetlinewidth{0.000000pt}%
\definecolor{currentstroke}{rgb}{0.000000,0.000000,0.000000}%
\pgfsetstrokecolor{currentstroke}%
\pgfsetstrokeopacity{0.000000}%
\pgfsetdash{}{0pt}%
\pgfpathmoveto{\pgfqpoint{1.501490in}{0.964913in}}%
\pgfpathlineto{\pgfqpoint{6.100000in}{0.964913in}}%
\pgfpathlineto{\pgfqpoint{6.100000in}{4.500000in}}%
\pgfpathlineto{\pgfqpoint{1.501490in}{4.500000in}}%
\pgfpathclose%
\pgfusepath{fill}%
\end{pgfscope}%
\begin{pgfscope}%
\pgfsetbuttcap%
\pgfsetroundjoin%
\definecolor{currentfill}{rgb}{0.000000,0.000000,0.000000}%
\pgfsetfillcolor{currentfill}%
\pgfsetlinewidth{0.803000pt}%
\definecolor{currentstroke}{rgb}{0.000000,0.000000,0.000000}%
\pgfsetstrokecolor{currentstroke}%
\pgfsetdash{}{0pt}%
\pgfsys@defobject{currentmarker}{\pgfqpoint{0.000000in}{-0.048611in}}{\pgfqpoint{0.000000in}{0.000000in}}{%
\pgfpathmoveto{\pgfqpoint{0.000000in}{0.000000in}}%
\pgfpathlineto{\pgfqpoint{0.000000in}{-0.048611in}}%
\pgfusepath{stroke,fill}%
}%
\begin{pgfscope}%
\pgfsys@transformshift{1.668286in}{0.964913in}%
\pgfsys@useobject{currentmarker}{}%
\end{pgfscope}%
\end{pgfscope}%
\begin{pgfscope}%
\definecolor{textcolor}{rgb}{0.000000,0.000000,0.000000}%
\pgfsetstrokecolor{textcolor}%
\pgfsetfillcolor{textcolor}%
\pgftext[x=1.668286in,y=0.867691in,,top]{\color{textcolor}\fontsize{20.000000}{24.000000}\selectfont \(\displaystyle {0}\)}%
\end{pgfscope}%
\begin{pgfscope}%
\pgfsetbuttcap%
\pgfsetroundjoin%
\definecolor{currentfill}{rgb}{0.000000,0.000000,0.000000}%
\pgfsetfillcolor{currentfill}%
\pgfsetlinewidth{0.803000pt}%
\definecolor{currentstroke}{rgb}{0.000000,0.000000,0.000000}%
\pgfsetstrokecolor{currentstroke}%
\pgfsetdash{}{0pt}%
\pgfsys@defobject{currentmarker}{\pgfqpoint{0.000000in}{-0.048611in}}{\pgfqpoint{0.000000in}{0.000000in}}{%
\pgfpathmoveto{\pgfqpoint{0.000000in}{0.000000in}}%
\pgfpathlineto{\pgfqpoint{0.000000in}{-0.048611in}}%
\pgfusepath{stroke,fill}%
}%
\begin{pgfscope}%
\pgfsys@transformshift{2.723959in}{0.964913in}%
\pgfsys@useobject{currentmarker}{}%
\end{pgfscope}%
\end{pgfscope}%
\begin{pgfscope}%
\definecolor{textcolor}{rgb}{0.000000,0.000000,0.000000}%
\pgfsetstrokecolor{textcolor}%
\pgfsetfillcolor{textcolor}%
\pgftext[x=2.723959in,y=0.867691in,,top]{\color{textcolor}\fontsize{20.000000}{24.000000}\selectfont \(\displaystyle {25}\)}%
\end{pgfscope}%
\begin{pgfscope}%
\pgfsetbuttcap%
\pgfsetroundjoin%
\definecolor{currentfill}{rgb}{0.000000,0.000000,0.000000}%
\pgfsetfillcolor{currentfill}%
\pgfsetlinewidth{0.803000pt}%
\definecolor{currentstroke}{rgb}{0.000000,0.000000,0.000000}%
\pgfsetstrokecolor{currentstroke}%
\pgfsetdash{}{0pt}%
\pgfsys@defobject{currentmarker}{\pgfqpoint{0.000000in}{-0.048611in}}{\pgfqpoint{0.000000in}{0.000000in}}{%
\pgfpathmoveto{\pgfqpoint{0.000000in}{0.000000in}}%
\pgfpathlineto{\pgfqpoint{0.000000in}{-0.048611in}}%
\pgfusepath{stroke,fill}%
}%
\begin{pgfscope}%
\pgfsys@transformshift{3.779631in}{0.964913in}%
\pgfsys@useobject{currentmarker}{}%
\end{pgfscope}%
\end{pgfscope}%
\begin{pgfscope}%
\definecolor{textcolor}{rgb}{0.000000,0.000000,0.000000}%
\pgfsetstrokecolor{textcolor}%
\pgfsetfillcolor{textcolor}%
\pgftext[x=3.779631in,y=0.867691in,,top]{\color{textcolor}\fontsize{20.000000}{24.000000}\selectfont \(\displaystyle {50}\)}%
\end{pgfscope}%
\begin{pgfscope}%
\pgfsetbuttcap%
\pgfsetroundjoin%
\definecolor{currentfill}{rgb}{0.000000,0.000000,0.000000}%
\pgfsetfillcolor{currentfill}%
\pgfsetlinewidth{0.803000pt}%
\definecolor{currentstroke}{rgb}{0.000000,0.000000,0.000000}%
\pgfsetstrokecolor{currentstroke}%
\pgfsetdash{}{0pt}%
\pgfsys@defobject{currentmarker}{\pgfqpoint{0.000000in}{-0.048611in}}{\pgfqpoint{0.000000in}{0.000000in}}{%
\pgfpathmoveto{\pgfqpoint{0.000000in}{0.000000in}}%
\pgfpathlineto{\pgfqpoint{0.000000in}{-0.048611in}}%
\pgfusepath{stroke,fill}%
}%
\begin{pgfscope}%
\pgfsys@transformshift{4.835304in}{0.964913in}%
\pgfsys@useobject{currentmarker}{}%
\end{pgfscope}%
\end{pgfscope}%
\begin{pgfscope}%
\definecolor{textcolor}{rgb}{0.000000,0.000000,0.000000}%
\pgfsetstrokecolor{textcolor}%
\pgfsetfillcolor{textcolor}%
\pgftext[x=4.835304in,y=0.867691in,,top]{\color{textcolor}\fontsize{20.000000}{24.000000}\selectfont \(\displaystyle {75}\)}%
\end{pgfscope}%
\begin{pgfscope}%
\pgfsetbuttcap%
\pgfsetroundjoin%
\definecolor{currentfill}{rgb}{0.000000,0.000000,0.000000}%
\pgfsetfillcolor{currentfill}%
\pgfsetlinewidth{0.803000pt}%
\definecolor{currentstroke}{rgb}{0.000000,0.000000,0.000000}%
\pgfsetstrokecolor{currentstroke}%
\pgfsetdash{}{0pt}%
\pgfsys@defobject{currentmarker}{\pgfqpoint{0.000000in}{-0.048611in}}{\pgfqpoint{0.000000in}{0.000000in}}{%
\pgfpathmoveto{\pgfqpoint{0.000000in}{0.000000in}}%
\pgfpathlineto{\pgfqpoint{0.000000in}{-0.048611in}}%
\pgfusepath{stroke,fill}%
}%
\begin{pgfscope}%
\pgfsys@transformshift{5.890977in}{0.964913in}%
\pgfsys@useobject{currentmarker}{}%
\end{pgfscope}%
\end{pgfscope}%
\begin{pgfscope}%
\definecolor{textcolor}{rgb}{0.000000,0.000000,0.000000}%
\pgfsetstrokecolor{textcolor}%
\pgfsetfillcolor{textcolor}%
\pgftext[x=5.890977in,y=0.867691in,,top]{\color{textcolor}\fontsize{20.000000}{24.000000}\selectfont \(\displaystyle {100}\)}%
\end{pgfscope}%
\begin{pgfscope}%
\definecolor{textcolor}{rgb}{0.000000,0.000000,0.000000}%
\pgfsetstrokecolor{textcolor}%
\pgfsetfillcolor{textcolor}%
\pgftext[x=3.800745in,y=0.556068in,,top]{\color{textcolor}\fontsize{20.000000}{24.000000}\selectfont Number links in the stem}%
\end{pgfscope}%
\begin{pgfscope}%
\pgfsetbuttcap%
\pgfsetroundjoin%
\definecolor{currentfill}{rgb}{0.000000,0.000000,0.000000}%
\pgfsetfillcolor{currentfill}%
\pgfsetlinewidth{0.803000pt}%
\definecolor{currentstroke}{rgb}{0.000000,0.000000,0.000000}%
\pgfsetstrokecolor{currentstroke}%
\pgfsetdash{}{0pt}%
\pgfsys@defobject{currentmarker}{\pgfqpoint{-0.048611in}{0.000000in}}{\pgfqpoint{-0.000000in}{0.000000in}}{%
\pgfpathmoveto{\pgfqpoint{-0.000000in}{0.000000in}}%
\pgfpathlineto{\pgfqpoint{-0.048611in}{0.000000in}}%
\pgfusepath{stroke,fill}%
}%
\begin{pgfscope}%
\pgfsys@transformshift{1.501490in}{1.215958in}%
\pgfsys@useobject{currentmarker}{}%
\end{pgfscope}%
\end{pgfscope}%
\begin{pgfscope}%
\definecolor{textcolor}{rgb}{0.000000,0.000000,0.000000}%
\pgfsetstrokecolor{textcolor}%
\pgfsetfillcolor{textcolor}%
\pgftext[x=0.611623in, y=1.115938in, left, base]{\color{textcolor}\fontsize{20.000000}{24.000000}\selectfont \(\displaystyle {100000}\)}%
\end{pgfscope}%
\begin{pgfscope}%
\pgfsetbuttcap%
\pgfsetroundjoin%
\definecolor{currentfill}{rgb}{0.000000,0.000000,0.000000}%
\pgfsetfillcolor{currentfill}%
\pgfsetlinewidth{0.803000pt}%
\definecolor{currentstroke}{rgb}{0.000000,0.000000,0.000000}%
\pgfsetstrokecolor{currentstroke}%
\pgfsetdash{}{0pt}%
\pgfsys@defobject{currentmarker}{\pgfqpoint{-0.048611in}{0.000000in}}{\pgfqpoint{-0.000000in}{0.000000in}}{%
\pgfpathmoveto{\pgfqpoint{-0.000000in}{0.000000in}}%
\pgfpathlineto{\pgfqpoint{-0.048611in}{0.000000in}}%
\pgfusepath{stroke,fill}%
}%
\begin{pgfscope}%
\pgfsys@transformshift{1.501490in}{1.911185in}%
\pgfsys@useobject{currentmarker}{}%
\end{pgfscope}%
\end{pgfscope}%
\begin{pgfscope}%
\definecolor{textcolor}{rgb}{0.000000,0.000000,0.000000}%
\pgfsetstrokecolor{textcolor}%
\pgfsetfillcolor{textcolor}%
\pgftext[x=0.611623in, y=1.811165in, left, base]{\color{textcolor}\fontsize{20.000000}{24.000000}\selectfont \(\displaystyle {200000}\)}%
\end{pgfscope}%
\begin{pgfscope}%
\pgfsetbuttcap%
\pgfsetroundjoin%
\definecolor{currentfill}{rgb}{0.000000,0.000000,0.000000}%
\pgfsetfillcolor{currentfill}%
\pgfsetlinewidth{0.803000pt}%
\definecolor{currentstroke}{rgb}{0.000000,0.000000,0.000000}%
\pgfsetstrokecolor{currentstroke}%
\pgfsetdash{}{0pt}%
\pgfsys@defobject{currentmarker}{\pgfqpoint{-0.048611in}{0.000000in}}{\pgfqpoint{-0.000000in}{0.000000in}}{%
\pgfpathmoveto{\pgfqpoint{-0.000000in}{0.000000in}}%
\pgfpathlineto{\pgfqpoint{-0.048611in}{0.000000in}}%
\pgfusepath{stroke,fill}%
}%
\begin{pgfscope}%
\pgfsys@transformshift{1.501490in}{2.606412in}%
\pgfsys@useobject{currentmarker}{}%
\end{pgfscope}%
\end{pgfscope}%
\begin{pgfscope}%
\definecolor{textcolor}{rgb}{0.000000,0.000000,0.000000}%
\pgfsetstrokecolor{textcolor}%
\pgfsetfillcolor{textcolor}%
\pgftext[x=0.611623in, y=2.506393in, left, base]{\color{textcolor}\fontsize{20.000000}{24.000000}\selectfont \(\displaystyle {300000}\)}%
\end{pgfscope}%
\begin{pgfscope}%
\pgfsetbuttcap%
\pgfsetroundjoin%
\definecolor{currentfill}{rgb}{0.000000,0.000000,0.000000}%
\pgfsetfillcolor{currentfill}%
\pgfsetlinewidth{0.803000pt}%
\definecolor{currentstroke}{rgb}{0.000000,0.000000,0.000000}%
\pgfsetstrokecolor{currentstroke}%
\pgfsetdash{}{0pt}%
\pgfsys@defobject{currentmarker}{\pgfqpoint{-0.048611in}{0.000000in}}{\pgfqpoint{-0.000000in}{0.000000in}}{%
\pgfpathmoveto{\pgfqpoint{-0.000000in}{0.000000in}}%
\pgfpathlineto{\pgfqpoint{-0.048611in}{0.000000in}}%
\pgfusepath{stroke,fill}%
}%
\begin{pgfscope}%
\pgfsys@transformshift{1.501490in}{3.301639in}%
\pgfsys@useobject{currentmarker}{}%
\end{pgfscope}%
\end{pgfscope}%
\begin{pgfscope}%
\definecolor{textcolor}{rgb}{0.000000,0.000000,0.000000}%
\pgfsetstrokecolor{textcolor}%
\pgfsetfillcolor{textcolor}%
\pgftext[x=0.611623in, y=3.201620in, left, base]{\color{textcolor}\fontsize{20.000000}{24.000000}\selectfont \(\displaystyle {400000}\)}%
\end{pgfscope}%
\begin{pgfscope}%
\pgfsetbuttcap%
\pgfsetroundjoin%
\definecolor{currentfill}{rgb}{0.000000,0.000000,0.000000}%
\pgfsetfillcolor{currentfill}%
\pgfsetlinewidth{0.803000pt}%
\definecolor{currentstroke}{rgb}{0.000000,0.000000,0.000000}%
\pgfsetstrokecolor{currentstroke}%
\pgfsetdash{}{0pt}%
\pgfsys@defobject{currentmarker}{\pgfqpoint{-0.048611in}{0.000000in}}{\pgfqpoint{-0.000000in}{0.000000in}}{%
\pgfpathmoveto{\pgfqpoint{-0.000000in}{0.000000in}}%
\pgfpathlineto{\pgfqpoint{-0.048611in}{0.000000in}}%
\pgfusepath{stroke,fill}%
}%
\begin{pgfscope}%
\pgfsys@transformshift{1.501490in}{3.996866in}%
\pgfsys@useobject{currentmarker}{}%
\end{pgfscope}%
\end{pgfscope}%
\begin{pgfscope}%
\definecolor{textcolor}{rgb}{0.000000,0.000000,0.000000}%
\pgfsetstrokecolor{textcolor}%
\pgfsetfillcolor{textcolor}%
\pgftext[x=0.611623in, y=3.896847in, left, base]{\color{textcolor}\fontsize{20.000000}{24.000000}\selectfont \(\displaystyle {500000}\)}%
\end{pgfscope}%
\begin{pgfscope}%
\definecolor{textcolor}{rgb}{0.000000,0.000000,0.000000}%
\pgfsetstrokecolor{textcolor}%
\pgfsetfillcolor{textcolor}%
\pgftext[x=0.556068in,y=2.732457in,,bottom,rotate=90.000000]{\color{textcolor}\fontsize{20.000000}{24.000000}\selectfont Number of operations}%
\end{pgfscope}%
\begin{pgfscope}%
\pgfpathrectangle{\pgfqpoint{1.501490in}{0.964913in}}{\pgfqpoint{4.598510in}{3.535087in}}%
\pgfusepath{clip}%
\pgfsetrectcap%
\pgfsetroundjoin%
\pgfsetlinewidth{1.505625pt}%
\definecolor{currentstroke}{rgb}{0.121569,0.466667,0.705882}%
\pgfsetstrokecolor{currentstroke}%
\pgfsetdash{}{0pt}%
\pgfpathmoveto{\pgfqpoint{1.710513in}{1.125599in}}%
\pgfpathlineto{\pgfqpoint{1.752740in}{1.184012in}}%
\pgfpathlineto{\pgfqpoint{1.794967in}{1.212196in}}%
\pgfpathlineto{\pgfqpoint{1.837194in}{1.240381in}}%
\pgfpathlineto{\pgfqpoint{1.879421in}{1.268565in}}%
\pgfpathlineto{\pgfqpoint{1.921648in}{1.296750in}}%
\pgfpathlineto{\pgfqpoint{1.963874in}{1.324934in}}%
\pgfpathlineto{\pgfqpoint{2.006101in}{1.353119in}}%
\pgfpathlineto{\pgfqpoint{2.048328in}{1.381303in}}%
\pgfpathlineto{\pgfqpoint{2.090555in}{1.409488in}}%
\pgfpathlineto{\pgfqpoint{2.132782in}{1.437672in}}%
\pgfpathlineto{\pgfqpoint{2.175009in}{1.465857in}}%
\pgfpathlineto{\pgfqpoint{2.217236in}{1.494041in}}%
\pgfpathlineto{\pgfqpoint{2.259463in}{1.522226in}}%
\pgfpathlineto{\pgfqpoint{2.301690in}{1.550410in}}%
\pgfpathlineto{\pgfqpoint{2.343917in}{1.578595in}}%
\pgfpathlineto{\pgfqpoint{2.386144in}{1.606779in}}%
\pgfpathlineto{\pgfqpoint{2.428370in}{1.634964in}}%
\pgfpathlineto{\pgfqpoint{2.470597in}{1.663148in}}%
\pgfpathlineto{\pgfqpoint{2.512824in}{1.691333in}}%
\pgfpathlineto{\pgfqpoint{2.555051in}{1.719517in}}%
\pgfpathlineto{\pgfqpoint{2.597278in}{1.747702in}}%
\pgfpathlineto{\pgfqpoint{2.639505in}{1.775887in}}%
\pgfpathlineto{\pgfqpoint{2.681732in}{1.804071in}}%
\pgfpathlineto{\pgfqpoint{2.723959in}{1.832256in}}%
\pgfpathlineto{\pgfqpoint{2.766186in}{1.860440in}}%
\pgfpathlineto{\pgfqpoint{2.808413in}{1.888625in}}%
\pgfpathlineto{\pgfqpoint{2.850640in}{1.916809in}}%
\pgfpathlineto{\pgfqpoint{2.892866in}{1.944994in}}%
\pgfpathlineto{\pgfqpoint{2.935093in}{1.973178in}}%
\pgfpathlineto{\pgfqpoint{2.977320in}{2.001363in}}%
\pgfpathlineto{\pgfqpoint{3.019547in}{2.029547in}}%
\pgfpathlineto{\pgfqpoint{3.061774in}{2.057732in}}%
\pgfpathlineto{\pgfqpoint{3.104001in}{2.085916in}}%
\pgfpathlineto{\pgfqpoint{3.146228in}{2.114101in}}%
\pgfpathlineto{\pgfqpoint{3.188455in}{2.142285in}}%
\pgfpathlineto{\pgfqpoint{3.230682in}{2.170470in}}%
\pgfpathlineto{\pgfqpoint{3.272909in}{2.198654in}}%
\pgfpathlineto{\pgfqpoint{3.315136in}{2.226839in}}%
\pgfpathlineto{\pgfqpoint{3.357362in}{2.255023in}}%
\pgfpathlineto{\pgfqpoint{3.399589in}{2.283208in}}%
\pgfpathlineto{\pgfqpoint{3.441816in}{2.311392in}}%
\pgfpathlineto{\pgfqpoint{3.484043in}{2.339577in}}%
\pgfpathlineto{\pgfqpoint{3.526270in}{2.367761in}}%
\pgfpathlineto{\pgfqpoint{3.568497in}{2.395946in}}%
\pgfpathlineto{\pgfqpoint{3.610724in}{2.424130in}}%
\pgfpathlineto{\pgfqpoint{3.652951in}{2.452315in}}%
\pgfpathlineto{\pgfqpoint{3.695178in}{2.480499in}}%
\pgfpathlineto{\pgfqpoint{3.737405in}{2.508684in}}%
\pgfpathlineto{\pgfqpoint{3.779631in}{2.536868in}}%
\pgfpathlineto{\pgfqpoint{3.821858in}{2.565053in}}%
\pgfpathlineto{\pgfqpoint{3.864085in}{2.593237in}}%
\pgfpathlineto{\pgfqpoint{3.906312in}{2.621422in}}%
\pgfpathlineto{\pgfqpoint{3.948539in}{2.649606in}}%
\pgfpathlineto{\pgfqpoint{3.990766in}{2.677791in}}%
\pgfpathlineto{\pgfqpoint{4.032993in}{2.705975in}}%
\pgfpathlineto{\pgfqpoint{4.075220in}{2.734160in}}%
\pgfpathlineto{\pgfqpoint{4.117447in}{2.762344in}}%
\pgfpathlineto{\pgfqpoint{4.159674in}{2.790529in}}%
\pgfpathlineto{\pgfqpoint{4.201901in}{2.818713in}}%
\pgfpathlineto{\pgfqpoint{4.244127in}{2.846898in}}%
\pgfpathlineto{\pgfqpoint{4.286354in}{2.875082in}}%
\pgfpathlineto{\pgfqpoint{4.328581in}{2.903267in}}%
\pgfpathlineto{\pgfqpoint{4.370808in}{2.931451in}}%
\pgfpathlineto{\pgfqpoint{4.413035in}{2.959636in}}%
\pgfpathlineto{\pgfqpoint{4.455262in}{2.987820in}}%
\pgfpathlineto{\pgfqpoint{4.497489in}{3.016005in}}%
\pgfpathlineto{\pgfqpoint{4.539716in}{3.044189in}}%
\pgfpathlineto{\pgfqpoint{4.581943in}{3.072374in}}%
\pgfpathlineto{\pgfqpoint{4.624170in}{3.100558in}}%
\pgfpathlineto{\pgfqpoint{4.666397in}{3.128743in}}%
\pgfpathlineto{\pgfqpoint{4.708623in}{3.156927in}}%
\pgfpathlineto{\pgfqpoint{4.750850in}{3.185112in}}%
\pgfpathlineto{\pgfqpoint{4.793077in}{3.213296in}}%
\pgfpathlineto{\pgfqpoint{4.835304in}{3.241481in}}%
\pgfpathlineto{\pgfqpoint{4.877531in}{3.269666in}}%
\pgfpathlineto{\pgfqpoint{4.919758in}{3.297850in}}%
\pgfpathlineto{\pgfqpoint{4.961985in}{3.326035in}}%
\pgfpathlineto{\pgfqpoint{5.004212in}{3.354219in}}%
\pgfpathlineto{\pgfqpoint{5.046439in}{3.382404in}}%
\pgfpathlineto{\pgfqpoint{5.088666in}{3.410588in}}%
\pgfpathlineto{\pgfqpoint{5.130892in}{3.438773in}}%
\pgfpathlineto{\pgfqpoint{5.173119in}{3.466957in}}%
\pgfpathlineto{\pgfqpoint{5.215346in}{3.495142in}}%
\pgfpathlineto{\pgfqpoint{5.257573in}{3.523326in}}%
\pgfpathlineto{\pgfqpoint{5.299800in}{3.551511in}}%
\pgfpathlineto{\pgfqpoint{5.342027in}{3.579695in}}%
\pgfpathlineto{\pgfqpoint{5.384254in}{3.607880in}}%
\pgfpathlineto{\pgfqpoint{5.426481in}{3.636064in}}%
\pgfpathlineto{\pgfqpoint{5.468708in}{3.664249in}}%
\pgfpathlineto{\pgfqpoint{5.510935in}{3.692433in}}%
\pgfpathlineto{\pgfqpoint{5.553162in}{3.720618in}}%
\pgfpathlineto{\pgfqpoint{5.595388in}{3.748802in}}%
\pgfpathlineto{\pgfqpoint{5.637615in}{3.776987in}}%
\pgfpathlineto{\pgfqpoint{5.679842in}{3.805171in}}%
\pgfpathlineto{\pgfqpoint{5.722069in}{3.833356in}}%
\pgfpathlineto{\pgfqpoint{5.764296in}{3.861540in}}%
\pgfpathlineto{\pgfqpoint{5.806523in}{3.889725in}}%
\pgfpathlineto{\pgfqpoint{5.848750in}{3.917909in}}%
\pgfpathlineto{\pgfqpoint{5.890977in}{3.946094in}}%
\pgfusepath{stroke}%
\end{pgfscope}%
\begin{pgfscope}%
\pgfpathrectangle{\pgfqpoint{1.501490in}{0.964913in}}{\pgfqpoint{4.598510in}{3.535087in}}%
\pgfusepath{clip}%
\pgfsetrectcap%
\pgfsetroundjoin%
\pgfsetlinewidth{1.505625pt}%
\definecolor{currentstroke}{rgb}{1.000000,0.498039,0.054902}%
\pgfsetstrokecolor{currentstroke}%
\pgfsetdash{}{0pt}%
\pgfpathmoveto{\pgfqpoint{1.710513in}{1.273154in}}%
\pgfpathlineto{\pgfqpoint{1.752740in}{1.329231in}}%
\pgfpathlineto{\pgfqpoint{1.794967in}{1.359946in}}%
\pgfpathlineto{\pgfqpoint{1.837194in}{1.390661in}}%
\pgfpathlineto{\pgfqpoint{1.879421in}{1.421376in}}%
\pgfpathlineto{\pgfqpoint{1.921648in}{1.452091in}}%
\pgfpathlineto{\pgfqpoint{1.963874in}{1.482807in}}%
\pgfpathlineto{\pgfqpoint{2.006101in}{1.513522in}}%
\pgfpathlineto{\pgfqpoint{2.048328in}{1.544237in}}%
\pgfpathlineto{\pgfqpoint{2.090555in}{1.574952in}}%
\pgfpathlineto{\pgfqpoint{2.132782in}{1.605667in}}%
\pgfpathlineto{\pgfqpoint{2.175009in}{1.636382in}}%
\pgfpathlineto{\pgfqpoint{2.217236in}{1.667097in}}%
\pgfpathlineto{\pgfqpoint{2.259463in}{1.697813in}}%
\pgfpathlineto{\pgfqpoint{2.301690in}{1.728528in}}%
\pgfpathlineto{\pgfqpoint{2.343917in}{1.759243in}}%
\pgfpathlineto{\pgfqpoint{2.386144in}{1.789958in}}%
\pgfpathlineto{\pgfqpoint{2.428370in}{1.820673in}}%
\pgfpathlineto{\pgfqpoint{2.470597in}{1.851388in}}%
\pgfpathlineto{\pgfqpoint{2.512824in}{1.882103in}}%
\pgfpathlineto{\pgfqpoint{2.555051in}{1.912818in}}%
\pgfpathlineto{\pgfqpoint{2.597278in}{1.943534in}}%
\pgfpathlineto{\pgfqpoint{2.639505in}{1.974249in}}%
\pgfpathlineto{\pgfqpoint{2.681732in}{2.004964in}}%
\pgfpathlineto{\pgfqpoint{2.723959in}{2.035679in}}%
\pgfpathlineto{\pgfqpoint{2.766186in}{2.066394in}}%
\pgfpathlineto{\pgfqpoint{2.808413in}{2.097109in}}%
\pgfpathlineto{\pgfqpoint{2.850640in}{2.127824in}}%
\pgfpathlineto{\pgfqpoint{2.892866in}{2.158540in}}%
\pgfpathlineto{\pgfqpoint{2.935093in}{2.189255in}}%
\pgfpathlineto{\pgfqpoint{2.977320in}{2.219970in}}%
\pgfpathlineto{\pgfqpoint{3.019547in}{2.250685in}}%
\pgfpathlineto{\pgfqpoint{3.061774in}{2.281400in}}%
\pgfpathlineto{\pgfqpoint{3.104001in}{2.312115in}}%
\pgfpathlineto{\pgfqpoint{3.146228in}{2.342830in}}%
\pgfpathlineto{\pgfqpoint{3.188455in}{2.373546in}}%
\pgfpathlineto{\pgfqpoint{3.230682in}{2.404261in}}%
\pgfpathlineto{\pgfqpoint{3.272909in}{2.434976in}}%
\pgfpathlineto{\pgfqpoint{3.315136in}{2.465691in}}%
\pgfpathlineto{\pgfqpoint{3.357362in}{2.496406in}}%
\pgfpathlineto{\pgfqpoint{3.399589in}{2.527121in}}%
\pgfpathlineto{\pgfqpoint{3.441816in}{2.557836in}}%
\pgfpathlineto{\pgfqpoint{3.484043in}{2.588551in}}%
\pgfpathlineto{\pgfqpoint{3.526270in}{2.619267in}}%
\pgfpathlineto{\pgfqpoint{3.568497in}{2.649982in}}%
\pgfpathlineto{\pgfqpoint{3.610724in}{2.680697in}}%
\pgfpathlineto{\pgfqpoint{3.652951in}{2.711412in}}%
\pgfpathlineto{\pgfqpoint{3.695178in}{2.742127in}}%
\pgfpathlineto{\pgfqpoint{3.737405in}{2.772842in}}%
\pgfpathlineto{\pgfqpoint{3.779631in}{2.803557in}}%
\pgfpathlineto{\pgfqpoint{3.821858in}{2.834273in}}%
\pgfpathlineto{\pgfqpoint{3.864085in}{2.864988in}}%
\pgfpathlineto{\pgfqpoint{3.906312in}{2.895703in}}%
\pgfpathlineto{\pgfqpoint{3.948539in}{2.926418in}}%
\pgfpathlineto{\pgfqpoint{3.990766in}{2.957133in}}%
\pgfpathlineto{\pgfqpoint{4.032993in}{2.987848in}}%
\pgfpathlineto{\pgfqpoint{4.075220in}{3.018563in}}%
\pgfpathlineto{\pgfqpoint{4.117447in}{3.049279in}}%
\pgfpathlineto{\pgfqpoint{4.159674in}{3.079994in}}%
\pgfpathlineto{\pgfqpoint{4.201901in}{3.110709in}}%
\pgfpathlineto{\pgfqpoint{4.244127in}{3.141424in}}%
\pgfpathlineto{\pgfqpoint{4.286354in}{3.172139in}}%
\pgfpathlineto{\pgfqpoint{4.328581in}{3.202854in}}%
\pgfpathlineto{\pgfqpoint{4.370808in}{3.233569in}}%
\pgfpathlineto{\pgfqpoint{4.413035in}{3.264284in}}%
\pgfpathlineto{\pgfqpoint{4.455262in}{3.295000in}}%
\pgfpathlineto{\pgfqpoint{4.497489in}{3.325715in}}%
\pgfpathlineto{\pgfqpoint{4.539716in}{3.356430in}}%
\pgfpathlineto{\pgfqpoint{4.581943in}{3.387145in}}%
\pgfpathlineto{\pgfqpoint{4.624170in}{3.417860in}}%
\pgfpathlineto{\pgfqpoint{4.666397in}{3.448575in}}%
\pgfpathlineto{\pgfqpoint{4.708623in}{3.479290in}}%
\pgfpathlineto{\pgfqpoint{4.750850in}{3.510006in}}%
\pgfpathlineto{\pgfqpoint{4.793077in}{3.540721in}}%
\pgfpathlineto{\pgfqpoint{4.835304in}{3.571436in}}%
\pgfpathlineto{\pgfqpoint{4.877531in}{3.602151in}}%
\pgfpathlineto{\pgfqpoint{4.919758in}{3.632866in}}%
\pgfpathlineto{\pgfqpoint{4.961985in}{3.663581in}}%
\pgfpathlineto{\pgfqpoint{5.004212in}{3.694296in}}%
\pgfpathlineto{\pgfqpoint{5.046439in}{3.725012in}}%
\pgfpathlineto{\pgfqpoint{5.088666in}{3.755727in}}%
\pgfpathlineto{\pgfqpoint{5.130892in}{3.786442in}}%
\pgfpathlineto{\pgfqpoint{5.173119in}{3.817157in}}%
\pgfpathlineto{\pgfqpoint{5.215346in}{3.847872in}}%
\pgfpathlineto{\pgfqpoint{5.257573in}{3.878587in}}%
\pgfpathlineto{\pgfqpoint{5.299800in}{3.909302in}}%
\pgfpathlineto{\pgfqpoint{5.342027in}{3.940017in}}%
\pgfpathlineto{\pgfqpoint{5.384254in}{3.970733in}}%
\pgfpathlineto{\pgfqpoint{5.426481in}{4.001448in}}%
\pgfpathlineto{\pgfqpoint{5.468708in}{4.032163in}}%
\pgfpathlineto{\pgfqpoint{5.510935in}{4.062878in}}%
\pgfpathlineto{\pgfqpoint{5.553162in}{4.093593in}}%
\pgfpathlineto{\pgfqpoint{5.595388in}{4.124308in}}%
\pgfpathlineto{\pgfqpoint{5.637615in}{4.155023in}}%
\pgfpathlineto{\pgfqpoint{5.679842in}{4.185739in}}%
\pgfpathlineto{\pgfqpoint{5.722069in}{4.216454in}}%
\pgfpathlineto{\pgfqpoint{5.764296in}{4.247169in}}%
\pgfpathlineto{\pgfqpoint{5.806523in}{4.277884in}}%
\pgfpathlineto{\pgfqpoint{5.848750in}{4.308599in}}%
\pgfpathlineto{\pgfqpoint{5.890977in}{4.339314in}}%
\pgfusepath{stroke}%
\end{pgfscope}%
\begin{pgfscope}%
\pgfpathrectangle{\pgfqpoint{1.501490in}{0.964913in}}{\pgfqpoint{4.598510in}{3.535087in}}%
\pgfusepath{clip}%
\pgfsetrectcap%
\pgfsetroundjoin%
\pgfsetlinewidth{1.505625pt}%
\definecolor{currentstroke}{rgb}{0.172549,0.627451,0.172549}%
\pgfsetstrokecolor{currentstroke}%
\pgfsetdash{}{0pt}%
\pgfpathmoveto{\pgfqpoint{1.710513in}{1.125599in}}%
\pgfpathlineto{\pgfqpoint{1.752740in}{1.167347in}}%
\pgfpathlineto{\pgfqpoint{1.794967in}{1.178540in}}%
\pgfpathlineto{\pgfqpoint{1.837194in}{1.182545in}}%
\pgfpathlineto{\pgfqpoint{1.879421in}{1.186549in}}%
\pgfpathlineto{\pgfqpoint{1.921648in}{1.190554in}}%
\pgfpathlineto{\pgfqpoint{1.963874in}{1.194558in}}%
\pgfpathlineto{\pgfqpoint{2.006101in}{1.198563in}}%
\pgfpathlineto{\pgfqpoint{2.048328in}{1.202567in}}%
\pgfpathlineto{\pgfqpoint{2.090555in}{1.206572in}}%
\pgfpathlineto{\pgfqpoint{2.132782in}{1.210576in}}%
\pgfpathlineto{\pgfqpoint{2.175009in}{1.214581in}}%
\pgfpathlineto{\pgfqpoint{2.217236in}{1.218585in}}%
\pgfpathlineto{\pgfqpoint{2.259463in}{1.222590in}}%
\pgfpathlineto{\pgfqpoint{2.301690in}{1.226594in}}%
\pgfpathlineto{\pgfqpoint{2.343917in}{1.230599in}}%
\pgfpathlineto{\pgfqpoint{2.386144in}{1.234603in}}%
\pgfpathlineto{\pgfqpoint{2.428370in}{1.238608in}}%
\pgfpathlineto{\pgfqpoint{2.470597in}{1.242613in}}%
\pgfpathlineto{\pgfqpoint{2.512824in}{1.246617in}}%
\pgfpathlineto{\pgfqpoint{2.555051in}{1.250622in}}%
\pgfpathlineto{\pgfqpoint{2.597278in}{1.254626in}}%
\pgfpathlineto{\pgfqpoint{2.639505in}{1.258631in}}%
\pgfpathlineto{\pgfqpoint{2.681732in}{1.262635in}}%
\pgfpathlineto{\pgfqpoint{2.723959in}{1.266640in}}%
\pgfpathlineto{\pgfqpoint{2.766186in}{1.270644in}}%
\pgfpathlineto{\pgfqpoint{2.808413in}{1.274649in}}%
\pgfpathlineto{\pgfqpoint{2.850640in}{1.278653in}}%
\pgfpathlineto{\pgfqpoint{2.892866in}{1.282658in}}%
\pgfpathlineto{\pgfqpoint{2.935093in}{1.286662in}}%
\pgfpathlineto{\pgfqpoint{2.977320in}{1.290667in}}%
\pgfpathlineto{\pgfqpoint{3.019547in}{1.294671in}}%
\pgfpathlineto{\pgfqpoint{3.061774in}{1.298676in}}%
\pgfpathlineto{\pgfqpoint{3.104001in}{1.302680in}}%
\pgfpathlineto{\pgfqpoint{3.146228in}{1.306685in}}%
\pgfpathlineto{\pgfqpoint{3.188455in}{1.310689in}}%
\pgfpathlineto{\pgfqpoint{3.230682in}{1.314694in}}%
\pgfpathlineto{\pgfqpoint{3.272909in}{1.318698in}}%
\pgfpathlineto{\pgfqpoint{3.315136in}{1.322703in}}%
\pgfpathlineto{\pgfqpoint{3.357362in}{1.326707in}}%
\pgfpathlineto{\pgfqpoint{3.399589in}{1.330712in}}%
\pgfpathlineto{\pgfqpoint{3.441816in}{1.334716in}}%
\pgfpathlineto{\pgfqpoint{3.484043in}{1.338721in}}%
\pgfpathlineto{\pgfqpoint{3.526270in}{1.342725in}}%
\pgfpathlineto{\pgfqpoint{3.568497in}{1.346730in}}%
\pgfpathlineto{\pgfqpoint{3.610724in}{1.350734in}}%
\pgfpathlineto{\pgfqpoint{3.652951in}{1.354739in}}%
\pgfpathlineto{\pgfqpoint{3.695178in}{1.358743in}}%
\pgfpathlineto{\pgfqpoint{3.737405in}{1.362748in}}%
\pgfpathlineto{\pgfqpoint{3.779631in}{1.366752in}}%
\pgfpathlineto{\pgfqpoint{3.821858in}{1.370757in}}%
\pgfpathlineto{\pgfqpoint{3.864085in}{1.374761in}}%
\pgfpathlineto{\pgfqpoint{3.906312in}{1.378766in}}%
\pgfpathlineto{\pgfqpoint{3.948539in}{1.382770in}}%
\pgfpathlineto{\pgfqpoint{3.990766in}{1.386775in}}%
\pgfpathlineto{\pgfqpoint{4.032993in}{1.390779in}}%
\pgfpathlineto{\pgfqpoint{4.075220in}{1.394784in}}%
\pgfpathlineto{\pgfqpoint{4.117447in}{1.398788in}}%
\pgfpathlineto{\pgfqpoint{4.159674in}{1.402793in}}%
\pgfpathlineto{\pgfqpoint{4.201901in}{1.406797in}}%
\pgfpathlineto{\pgfqpoint{4.244127in}{1.410802in}}%
\pgfpathlineto{\pgfqpoint{4.286354in}{1.414806in}}%
\pgfpathlineto{\pgfqpoint{4.328581in}{1.418811in}}%
\pgfpathlineto{\pgfqpoint{4.370808in}{1.422815in}}%
\pgfpathlineto{\pgfqpoint{4.413035in}{1.426820in}}%
\pgfpathlineto{\pgfqpoint{4.455262in}{1.430824in}}%
\pgfpathlineto{\pgfqpoint{4.497489in}{1.434829in}}%
\pgfpathlineto{\pgfqpoint{4.539716in}{1.438833in}}%
\pgfpathlineto{\pgfqpoint{4.581943in}{1.442838in}}%
\pgfpathlineto{\pgfqpoint{4.624170in}{1.446842in}}%
\pgfpathlineto{\pgfqpoint{4.666397in}{1.450847in}}%
\pgfpathlineto{\pgfqpoint{4.708623in}{1.454851in}}%
\pgfpathlineto{\pgfqpoint{4.750850in}{1.458856in}}%
\pgfpathlineto{\pgfqpoint{4.793077in}{1.462860in}}%
\pgfpathlineto{\pgfqpoint{4.835304in}{1.466865in}}%
\pgfpathlineto{\pgfqpoint{4.877531in}{1.470869in}}%
\pgfpathlineto{\pgfqpoint{4.919758in}{1.474874in}}%
\pgfpathlineto{\pgfqpoint{4.961985in}{1.478879in}}%
\pgfpathlineto{\pgfqpoint{5.004212in}{1.482883in}}%
\pgfpathlineto{\pgfqpoint{5.046439in}{1.486888in}}%
\pgfpathlineto{\pgfqpoint{5.088666in}{1.490892in}}%
\pgfpathlineto{\pgfqpoint{5.130892in}{1.494897in}}%
\pgfpathlineto{\pgfqpoint{5.173119in}{1.498901in}}%
\pgfpathlineto{\pgfqpoint{5.215346in}{1.502906in}}%
\pgfpathlineto{\pgfqpoint{5.257573in}{1.506910in}}%
\pgfpathlineto{\pgfqpoint{5.299800in}{1.510915in}}%
\pgfpathlineto{\pgfqpoint{5.342027in}{1.514919in}}%
\pgfpathlineto{\pgfqpoint{5.384254in}{1.518924in}}%
\pgfpathlineto{\pgfqpoint{5.426481in}{1.522928in}}%
\pgfpathlineto{\pgfqpoint{5.468708in}{1.526933in}}%
\pgfpathlineto{\pgfqpoint{5.510935in}{1.530937in}}%
\pgfpathlineto{\pgfqpoint{5.553162in}{1.534942in}}%
\pgfpathlineto{\pgfqpoint{5.595388in}{1.538946in}}%
\pgfpathlineto{\pgfqpoint{5.637615in}{1.542951in}}%
\pgfpathlineto{\pgfqpoint{5.679842in}{1.546955in}}%
\pgfpathlineto{\pgfqpoint{5.722069in}{1.550960in}}%
\pgfpathlineto{\pgfqpoint{5.764296in}{1.554964in}}%
\pgfpathlineto{\pgfqpoint{5.806523in}{1.558969in}}%
\pgfpathlineto{\pgfqpoint{5.848750in}{1.562973in}}%
\pgfpathlineto{\pgfqpoint{5.890977in}{1.566978in}}%
\pgfusepath{stroke}%
\end{pgfscope}%
\begin{pgfscope}%
\pgfsetrectcap%
\pgfsetmiterjoin%
\pgfsetlinewidth{0.803000pt}%
\definecolor{currentstroke}{rgb}{0.000000,0.000000,0.000000}%
\pgfsetstrokecolor{currentstroke}%
\pgfsetdash{}{0pt}%
\pgfpathmoveto{\pgfqpoint{1.501490in}{0.964913in}}%
\pgfpathlineto{\pgfqpoint{1.501490in}{4.500000in}}%
\pgfusepath{stroke}%
\end{pgfscope}%
\begin{pgfscope}%
\pgfsetrectcap%
\pgfsetmiterjoin%
\pgfsetlinewidth{0.803000pt}%
\definecolor{currentstroke}{rgb}{0.000000,0.000000,0.000000}%
\pgfsetstrokecolor{currentstroke}%
\pgfsetdash{}{0pt}%
\pgfpathmoveto{\pgfqpoint{6.100000in}{0.964913in}}%
\pgfpathlineto{\pgfqpoint{6.100000in}{4.500000in}}%
\pgfusepath{stroke}%
\end{pgfscope}%
\begin{pgfscope}%
\pgfsetrectcap%
\pgfsetmiterjoin%
\pgfsetlinewidth{0.803000pt}%
\definecolor{currentstroke}{rgb}{0.000000,0.000000,0.000000}%
\pgfsetstrokecolor{currentstroke}%
\pgfsetdash{}{0pt}%
\pgfpathmoveto{\pgfqpoint{1.501490in}{0.964913in}}%
\pgfpathlineto{\pgfqpoint{6.100000in}{0.964913in}}%
\pgfusepath{stroke}%
\end{pgfscope}%
\begin{pgfscope}%
\pgfsetrectcap%
\pgfsetmiterjoin%
\pgfsetlinewidth{0.803000pt}%
\definecolor{currentstroke}{rgb}{0.000000,0.000000,0.000000}%
\pgfsetstrokecolor{currentstroke}%
\pgfsetdash{}{0pt}%
\pgfpathmoveto{\pgfqpoint{1.501490in}{4.500000in}}%
\pgfpathlineto{\pgfqpoint{6.100000in}{4.500000in}}%
\pgfusepath{stroke}%
\end{pgfscope}%
\begin{pgfscope}%
\pgfsetbuttcap%
\pgfsetmiterjoin%
\definecolor{currentfill}{rgb}{1.000000,1.000000,1.000000}%
\pgfsetfillcolor{currentfill}%
\pgfsetfillopacity{0.800000}%
\pgfsetlinewidth{1.003750pt}%
\definecolor{currentstroke}{rgb}{0.800000,0.800000,0.800000}%
\pgfsetstrokecolor{currentstroke}%
\pgfsetstrokeopacity{0.800000}%
\pgfsetdash{}{0pt}%
\pgfpathmoveto{\pgfqpoint{1.695934in}{3.092908in}}%
\pgfpathlineto{\pgfqpoint{3.746195in}{3.092908in}}%
\pgfpathquadraticcurveto{\pgfqpoint{3.801750in}{3.092908in}}{\pgfqpoint{3.801750in}{3.148464in}}%
\pgfpathlineto{\pgfqpoint{3.801750in}{4.305556in}}%
\pgfpathquadraticcurveto{\pgfqpoint{3.801750in}{4.361111in}}{\pgfqpoint{3.746195in}{4.361111in}}%
\pgfpathlineto{\pgfqpoint{1.695934in}{4.361111in}}%
\pgfpathquadraticcurveto{\pgfqpoint{1.640379in}{4.361111in}}{\pgfqpoint{1.640379in}{4.305556in}}%
\pgfpathlineto{\pgfqpoint{1.640379in}{3.148464in}}%
\pgfpathquadraticcurveto{\pgfqpoint{1.640379in}{3.092908in}}{\pgfqpoint{1.695934in}{3.092908in}}%
\pgfpathclose%
\pgfusepath{stroke,fill}%
\end{pgfscope}%
\begin{pgfscope}%
\pgfsetrectcap%
\pgfsetroundjoin%
\pgfsetlinewidth{1.505625pt}%
\definecolor{currentstroke}{rgb}{0.121569,0.466667,0.705882}%
\pgfsetstrokecolor{currentstroke}%
\pgfsetdash{}{0pt}%
\pgfpathmoveto{\pgfqpoint{1.751490in}{4.147184in}}%
\pgfpathlineto{\pgfqpoint{2.307045in}{4.147184in}}%
\pgfusepath{stroke}%
\end{pgfscope}%
\begin{pgfscope}%
\definecolor{textcolor}{rgb}{0.000000,0.000000,0.000000}%
\pgfsetstrokecolor{textcolor}%
\pgfsetfillcolor{textcolor}%
\pgftext[x=2.529268in,y=4.049962in,left,base]{\color{textcolor}\fontsize{20.000000}{24.000000}\selectfont PV-OSIM}%
\end{pgfscope}%
\begin{pgfscope}%
\pgfsetrectcap%
\pgfsetroundjoin%
\pgfsetlinewidth{1.505625pt}%
\definecolor{currentstroke}{rgb}{1.000000,0.498039,0.054902}%
\pgfsetstrokecolor{currentstroke}%
\pgfsetdash{}{0pt}%
\pgfpathmoveto{\pgfqpoint{1.751490in}{3.752227in}}%
\pgfpathlineto{\pgfqpoint{2.307045in}{3.752227in}}%
\pgfusepath{stroke}%
\end{pgfscope}%
\begin{pgfscope}%
\definecolor{textcolor}{rgb}{0.000000,0.000000,0.000000}%
\pgfsetstrokecolor{textcolor}%
\pgfsetfillcolor{textcolor}%
\pgftext[x=2.529268in,y=3.655005in,left,base]{\color{textcolor}\fontsize{20.000000}{24.000000}\selectfont EFPA}%
\end{pgfscope}%
\begin{pgfscope}%
\pgfsetrectcap%
\pgfsetroundjoin%
\pgfsetlinewidth{1.505625pt}%
\definecolor{currentstroke}{rgb}{0.172549,0.627451,0.172549}%
\pgfsetstrokecolor{currentstroke}%
\pgfsetdash{}{0pt}%
\pgfpathmoveto{\pgfqpoint{1.751490in}{3.357271in}}%
\pgfpathlineto{\pgfqpoint{2.307045in}{3.357271in}}%
\pgfusepath{stroke}%
\end{pgfscope}%
\begin{pgfscope}%
\definecolor{textcolor}{rgb}{0.000000,0.000000,0.000000}%
\pgfsetstrokecolor{textcolor}%
\pgfsetfillcolor{textcolor}%
\pgftext[x=2.529268in,y=3.260048in,left,base]{\color{textcolor}\fontsize{20.000000}{24.000000}\selectfont PV-OSIMr}%
\end{pgfscope}%
\end{pgfpicture}%
\makeatother%
\endgroup%

%% file: graphics/tree_comparison10.pgf
%% Creator: Matplotlib, PGF backend
%%
%% To include the figure in your LaTeX document, write
%%   \input{<filename>.pgf}
%%
%% Make sure the required packages are loaded in your preamble
%%   \usepackage{pgf}
%%
%% Figures using additional raster images can only be included by \input if
%% they are in the same directory as the main LaTeX file. For loading figures
%% from other directories you can use the `import` package
%% and then include the figures with
%%   \import{<path to file>}{<filename>.pgf}
%%
%% Matplotlib used the following preamble
%%
\begingroup%
\makeatletter%
\begin{pgfpicture}%
\pgfpathrectangle{\pgfpointorigin}{\pgfqpoint{6.400000in}{4.800000in}}%
\pgfusepath{use as bounding box, clip}%
\begin{pgfscope}%
\pgfsetbuttcap%
\pgfsetmiterjoin%
\definecolor{currentfill}{rgb}{1.000000,1.000000,1.000000}%
\pgfsetfillcolor{currentfill}%
\pgfsetlinewidth{0.000000pt}%
\definecolor{currentstroke}{rgb}{1.000000,1.000000,1.000000}%
\pgfsetstrokecolor{currentstroke}%
\pgfsetdash{}{0pt}%
\pgfpathmoveto{\pgfqpoint{0.000000in}{0.000000in}}%
\pgfpathlineto{\pgfqpoint{6.400000in}{0.000000in}}%
\pgfpathlineto{\pgfqpoint{6.400000in}{4.800000in}}%
\pgfpathlineto{\pgfqpoint{0.000000in}{4.800000in}}%
\pgfpathclose%
\pgfusepath{fill}%
\end{pgfscope}%
\begin{pgfscope}%
\pgfsetbuttcap%
\pgfsetmiterjoin%
\definecolor{currentfill}{rgb}{1.000000,1.000000,1.000000}%
\pgfsetfillcolor{currentfill}%
\pgfsetlinewidth{0.000000pt}%
\definecolor{currentstroke}{rgb}{0.000000,0.000000,0.000000}%
\pgfsetstrokecolor{currentstroke}%
\pgfsetstrokeopacity{0.000000}%
\pgfsetdash{}{0pt}%
\pgfpathmoveto{\pgfqpoint{1.501490in}{0.964913in}}%
\pgfpathlineto{\pgfqpoint{6.100000in}{0.964913in}}%
\pgfpathlineto{\pgfqpoint{6.100000in}{4.500000in}}%
\pgfpathlineto{\pgfqpoint{1.501490in}{4.500000in}}%
\pgfpathclose%
\pgfusepath{fill}%
\end{pgfscope}%
\begin{pgfscope}%
\pgfsetbuttcap%
\pgfsetroundjoin%
\definecolor{currentfill}{rgb}{0.000000,0.000000,0.000000}%
\pgfsetfillcolor{currentfill}%
\pgfsetlinewidth{0.803000pt}%
\definecolor{currentstroke}{rgb}{0.000000,0.000000,0.000000}%
\pgfsetstrokecolor{currentstroke}%
\pgfsetdash{}{0pt}%
\pgfsys@defobject{currentmarker}{\pgfqpoint{0.000000in}{-0.048611in}}{\pgfqpoint{0.000000in}{0.000000in}}{%
\pgfpathmoveto{\pgfqpoint{0.000000in}{0.000000in}}%
\pgfpathlineto{\pgfqpoint{0.000000in}{-0.048611in}}%
\pgfusepath{stroke,fill}%
}%
\begin{pgfscope}%
\pgfsys@transformshift{1.668286in}{0.964913in}%
\pgfsys@useobject{currentmarker}{}%
\end{pgfscope}%
\end{pgfscope}%
\begin{pgfscope}%
\definecolor{textcolor}{rgb}{0.000000,0.000000,0.000000}%
\pgfsetstrokecolor{textcolor}%
\pgfsetfillcolor{textcolor}%
\pgftext[x=1.668286in,y=0.867691in,,top]{\color{textcolor}\fontsize{20.000000}{24.000000}\selectfont \(\displaystyle {0}\)}%
\end{pgfscope}%
\begin{pgfscope}%
\pgfsetbuttcap%
\pgfsetroundjoin%
\definecolor{currentfill}{rgb}{0.000000,0.000000,0.000000}%
\pgfsetfillcolor{currentfill}%
\pgfsetlinewidth{0.803000pt}%
\definecolor{currentstroke}{rgb}{0.000000,0.000000,0.000000}%
\pgfsetstrokecolor{currentstroke}%
\pgfsetdash{}{0pt}%
\pgfsys@defobject{currentmarker}{\pgfqpoint{0.000000in}{-0.048611in}}{\pgfqpoint{0.000000in}{0.000000in}}{%
\pgfpathmoveto{\pgfqpoint{0.000000in}{0.000000in}}%
\pgfpathlineto{\pgfqpoint{0.000000in}{-0.048611in}}%
\pgfusepath{stroke,fill}%
}%
\begin{pgfscope}%
\pgfsys@transformshift{2.723959in}{0.964913in}%
\pgfsys@useobject{currentmarker}{}%
\end{pgfscope}%
\end{pgfscope}%
\begin{pgfscope}%
\definecolor{textcolor}{rgb}{0.000000,0.000000,0.000000}%
\pgfsetstrokecolor{textcolor}%
\pgfsetfillcolor{textcolor}%
\pgftext[x=2.723959in,y=0.867691in,,top]{\color{textcolor}\fontsize{20.000000}{24.000000}\selectfont \(\displaystyle {25}\)}%
\end{pgfscope}%
\begin{pgfscope}%
\pgfsetbuttcap%
\pgfsetroundjoin%
\definecolor{currentfill}{rgb}{0.000000,0.000000,0.000000}%
\pgfsetfillcolor{currentfill}%
\pgfsetlinewidth{0.803000pt}%
\definecolor{currentstroke}{rgb}{0.000000,0.000000,0.000000}%
\pgfsetstrokecolor{currentstroke}%
\pgfsetdash{}{0pt}%
\pgfsys@defobject{currentmarker}{\pgfqpoint{0.000000in}{-0.048611in}}{\pgfqpoint{0.000000in}{0.000000in}}{%
\pgfpathmoveto{\pgfqpoint{0.000000in}{0.000000in}}%
\pgfpathlineto{\pgfqpoint{0.000000in}{-0.048611in}}%
\pgfusepath{stroke,fill}%
}%
\begin{pgfscope}%
\pgfsys@transformshift{3.779631in}{0.964913in}%
\pgfsys@useobject{currentmarker}{}%
\end{pgfscope}%
\end{pgfscope}%
\begin{pgfscope}%
\definecolor{textcolor}{rgb}{0.000000,0.000000,0.000000}%
\pgfsetstrokecolor{textcolor}%
\pgfsetfillcolor{textcolor}%
\pgftext[x=3.779631in,y=0.867691in,,top]{\color{textcolor}\fontsize{20.000000}{24.000000}\selectfont \(\displaystyle {50}\)}%
\end{pgfscope}%
\begin{pgfscope}%
\pgfsetbuttcap%
\pgfsetroundjoin%
\definecolor{currentfill}{rgb}{0.000000,0.000000,0.000000}%
\pgfsetfillcolor{currentfill}%
\pgfsetlinewidth{0.803000pt}%
\definecolor{currentstroke}{rgb}{0.000000,0.000000,0.000000}%
\pgfsetstrokecolor{currentstroke}%
\pgfsetdash{}{0pt}%
\pgfsys@defobject{currentmarker}{\pgfqpoint{0.000000in}{-0.048611in}}{\pgfqpoint{0.000000in}{0.000000in}}{%
\pgfpathmoveto{\pgfqpoint{0.000000in}{0.000000in}}%
\pgfpathlineto{\pgfqpoint{0.000000in}{-0.048611in}}%
\pgfusepath{stroke,fill}%
}%
\begin{pgfscope}%
\pgfsys@transformshift{4.835304in}{0.964913in}%
\pgfsys@useobject{currentmarker}{}%
\end{pgfscope}%
\end{pgfscope}%
\begin{pgfscope}%
\definecolor{textcolor}{rgb}{0.000000,0.000000,0.000000}%
\pgfsetstrokecolor{textcolor}%
\pgfsetfillcolor{textcolor}%
\pgftext[x=4.835304in,y=0.867691in,,top]{\color{textcolor}\fontsize{20.000000}{24.000000}\selectfont \(\displaystyle {75}\)}%
\end{pgfscope}%
\begin{pgfscope}%
\pgfsetbuttcap%
\pgfsetroundjoin%
\definecolor{currentfill}{rgb}{0.000000,0.000000,0.000000}%
\pgfsetfillcolor{currentfill}%
\pgfsetlinewidth{0.803000pt}%
\definecolor{currentstroke}{rgb}{0.000000,0.000000,0.000000}%
\pgfsetstrokecolor{currentstroke}%
\pgfsetdash{}{0pt}%
\pgfsys@defobject{currentmarker}{\pgfqpoint{0.000000in}{-0.048611in}}{\pgfqpoint{0.000000in}{0.000000in}}{%
\pgfpathmoveto{\pgfqpoint{0.000000in}{0.000000in}}%
\pgfpathlineto{\pgfqpoint{0.000000in}{-0.048611in}}%
\pgfusepath{stroke,fill}%
}%
\begin{pgfscope}%
\pgfsys@transformshift{5.890977in}{0.964913in}%
\pgfsys@useobject{currentmarker}{}%
\end{pgfscope}%
\end{pgfscope}%
\begin{pgfscope}%
\definecolor{textcolor}{rgb}{0.000000,0.000000,0.000000}%
\pgfsetstrokecolor{textcolor}%
\pgfsetfillcolor{textcolor}%
\pgftext[x=5.890977in,y=0.867691in,,top]{\color{textcolor}\fontsize{20.000000}{24.000000}\selectfont \(\displaystyle {100}\)}%
\end{pgfscope}%
\begin{pgfscope}%
\definecolor{textcolor}{rgb}{0.000000,0.000000,0.000000}%
\pgfsetstrokecolor{textcolor}%
\pgfsetfillcolor{textcolor}%
\pgftext[x=3.800745in,y=0.556068in,,top]{\color{textcolor}\fontsize{20.000000}{24.000000}\selectfont Number links in the stem}%
\end{pgfscope}%
\begin{pgfscope}%
\pgfsetbuttcap%
\pgfsetroundjoin%
\definecolor{currentfill}{rgb}{0.000000,0.000000,0.000000}%
\pgfsetfillcolor{currentfill}%
\pgfsetlinewidth{0.803000pt}%
\definecolor{currentstroke}{rgb}{0.000000,0.000000,0.000000}%
\pgfsetstrokecolor{currentstroke}%
\pgfsetdash{}{0pt}%
\pgfsys@defobject{currentmarker}{\pgfqpoint{-0.048611in}{0.000000in}}{\pgfqpoint{-0.000000in}{0.000000in}}{%
\pgfpathmoveto{\pgfqpoint{-0.000000in}{0.000000in}}%
\pgfpathlineto{\pgfqpoint{-0.048611in}{0.000000in}}%
\pgfusepath{stroke,fill}%
}%
\begin{pgfscope}%
\pgfsys@transformshift{1.501490in}{1.394578in}%
\pgfsys@useobject{currentmarker}{}%
\end{pgfscope}%
\end{pgfscope}%
\begin{pgfscope}%
\definecolor{textcolor}{rgb}{0.000000,0.000000,0.000000}%
\pgfsetstrokecolor{textcolor}%
\pgfsetfillcolor{textcolor}%
\pgftext[x=0.611623in, y=1.294558in, left, base]{\color{textcolor}\fontsize{20.000000}{24.000000}\selectfont \(\displaystyle {200000}\)}%
\end{pgfscope}%
\begin{pgfscope}%
\pgfsetbuttcap%
\pgfsetroundjoin%
\definecolor{currentfill}{rgb}{0.000000,0.000000,0.000000}%
\pgfsetfillcolor{currentfill}%
\pgfsetlinewidth{0.803000pt}%
\definecolor{currentstroke}{rgb}{0.000000,0.000000,0.000000}%
\pgfsetstrokecolor{currentstroke}%
\pgfsetdash{}{0pt}%
\pgfsys@defobject{currentmarker}{\pgfqpoint{-0.048611in}{0.000000in}}{\pgfqpoint{-0.000000in}{0.000000in}}{%
\pgfpathmoveto{\pgfqpoint{-0.000000in}{0.000000in}}%
\pgfpathlineto{\pgfqpoint{-0.048611in}{0.000000in}}%
\pgfusepath{stroke,fill}%
}%
\begin{pgfscope}%
\pgfsys@transformshift{1.501490in}{2.346496in}%
\pgfsys@useobject{currentmarker}{}%
\end{pgfscope}%
\end{pgfscope}%
\begin{pgfscope}%
\definecolor{textcolor}{rgb}{0.000000,0.000000,0.000000}%
\pgfsetstrokecolor{textcolor}%
\pgfsetfillcolor{textcolor}%
\pgftext[x=0.611623in, y=2.246477in, left, base]{\color{textcolor}\fontsize{20.000000}{24.000000}\selectfont \(\displaystyle {400000}\)}%
\end{pgfscope}%
\begin{pgfscope}%
\pgfsetbuttcap%
\pgfsetroundjoin%
\definecolor{currentfill}{rgb}{0.000000,0.000000,0.000000}%
\pgfsetfillcolor{currentfill}%
\pgfsetlinewidth{0.803000pt}%
\definecolor{currentstroke}{rgb}{0.000000,0.000000,0.000000}%
\pgfsetstrokecolor{currentstroke}%
\pgfsetdash{}{0pt}%
\pgfsys@defobject{currentmarker}{\pgfqpoint{-0.048611in}{0.000000in}}{\pgfqpoint{-0.000000in}{0.000000in}}{%
\pgfpathmoveto{\pgfqpoint{-0.000000in}{0.000000in}}%
\pgfpathlineto{\pgfqpoint{-0.048611in}{0.000000in}}%
\pgfusepath{stroke,fill}%
}%
\begin{pgfscope}%
\pgfsys@transformshift{1.501490in}{3.298415in}%
\pgfsys@useobject{currentmarker}{}%
\end{pgfscope}%
\end{pgfscope}%
\begin{pgfscope}%
\definecolor{textcolor}{rgb}{0.000000,0.000000,0.000000}%
\pgfsetstrokecolor{textcolor}%
\pgfsetfillcolor{textcolor}%
\pgftext[x=0.611623in, y=3.198396in, left, base]{\color{textcolor}\fontsize{20.000000}{24.000000}\selectfont \(\displaystyle {600000}\)}%
\end{pgfscope}%
\begin{pgfscope}%
\pgfsetbuttcap%
\pgfsetroundjoin%
\definecolor{currentfill}{rgb}{0.000000,0.000000,0.000000}%
\pgfsetfillcolor{currentfill}%
\pgfsetlinewidth{0.803000pt}%
\definecolor{currentstroke}{rgb}{0.000000,0.000000,0.000000}%
\pgfsetstrokecolor{currentstroke}%
\pgfsetdash{}{0pt}%
\pgfsys@defobject{currentmarker}{\pgfqpoint{-0.048611in}{0.000000in}}{\pgfqpoint{-0.000000in}{0.000000in}}{%
\pgfpathmoveto{\pgfqpoint{-0.000000in}{0.000000in}}%
\pgfpathlineto{\pgfqpoint{-0.048611in}{0.000000in}}%
\pgfusepath{stroke,fill}%
}%
\begin{pgfscope}%
\pgfsys@transformshift{1.501490in}{4.250334in}%
\pgfsys@useobject{currentmarker}{}%
\end{pgfscope}%
\end{pgfscope}%
\begin{pgfscope}%
\definecolor{textcolor}{rgb}{0.000000,0.000000,0.000000}%
\pgfsetstrokecolor{textcolor}%
\pgfsetfillcolor{textcolor}%
\pgftext[x=0.611623in, y=4.150314in, left, base]{\color{textcolor}\fontsize{20.000000}{24.000000}\selectfont \(\displaystyle {800000}\)}%
\end{pgfscope}%
\begin{pgfscope}%
\definecolor{textcolor}{rgb}{0.000000,0.000000,0.000000}%
\pgfsetstrokecolor{textcolor}%
\pgfsetfillcolor{textcolor}%
\pgftext[x=0.556068in,y=2.732457in,,bottom,rotate=90.000000]{\color{textcolor}\fontsize{20.000000}{24.000000}\selectfont Number of operations}%
\end{pgfscope}%
\begin{pgfscope}%
\pgfpathrectangle{\pgfqpoint{1.501490in}{0.964913in}}{\pgfqpoint{4.598510in}{3.535087in}}%
\pgfusepath{clip}%
\pgfsetrectcap%
\pgfsetroundjoin%
\pgfsetlinewidth{1.505625pt}%
\definecolor{currentstroke}{rgb}{0.121569,0.466667,0.705882}%
\pgfsetstrokecolor{currentstroke}%
\pgfsetdash{}{0pt}%
\pgfpathmoveto{\pgfqpoint{1.710513in}{1.125599in}}%
\pgfpathlineto{\pgfqpoint{1.752740in}{1.197373in}}%
\pgfpathlineto{\pgfqpoint{1.794967in}{1.229434in}}%
\pgfpathlineto{\pgfqpoint{1.837194in}{1.261495in}}%
\pgfpathlineto{\pgfqpoint{1.879421in}{1.293555in}}%
\pgfpathlineto{\pgfqpoint{1.921648in}{1.325616in}}%
\pgfpathlineto{\pgfqpoint{1.963874in}{1.357677in}}%
\pgfpathlineto{\pgfqpoint{2.006101in}{1.389737in}}%
\pgfpathlineto{\pgfqpoint{2.048328in}{1.421798in}}%
\pgfpathlineto{\pgfqpoint{2.090555in}{1.453858in}}%
\pgfpathlineto{\pgfqpoint{2.132782in}{1.485919in}}%
\pgfpathlineto{\pgfqpoint{2.175009in}{1.517980in}}%
\pgfpathlineto{\pgfqpoint{2.217236in}{1.550040in}}%
\pgfpathlineto{\pgfqpoint{2.259463in}{1.582101in}}%
\pgfpathlineto{\pgfqpoint{2.301690in}{1.614162in}}%
\pgfpathlineto{\pgfqpoint{2.343917in}{1.646222in}}%
\pgfpathlineto{\pgfqpoint{2.386144in}{1.678283in}}%
\pgfpathlineto{\pgfqpoint{2.428370in}{1.710343in}}%
\pgfpathlineto{\pgfqpoint{2.470597in}{1.742404in}}%
\pgfpathlineto{\pgfqpoint{2.512824in}{1.774465in}}%
\pgfpathlineto{\pgfqpoint{2.555051in}{1.806525in}}%
\pgfpathlineto{\pgfqpoint{2.597278in}{1.838586in}}%
\pgfpathlineto{\pgfqpoint{2.639505in}{1.870647in}}%
\pgfpathlineto{\pgfqpoint{2.681732in}{1.902707in}}%
\pgfpathlineto{\pgfqpoint{2.723959in}{1.934768in}}%
\pgfpathlineto{\pgfqpoint{2.766186in}{1.966828in}}%
\pgfpathlineto{\pgfqpoint{2.808413in}{1.998889in}}%
\pgfpathlineto{\pgfqpoint{2.850640in}{2.030950in}}%
\pgfpathlineto{\pgfqpoint{2.892866in}{2.063010in}}%
\pgfpathlineto{\pgfqpoint{2.935093in}{2.095071in}}%
\pgfpathlineto{\pgfqpoint{2.977320in}{2.127131in}}%
\pgfpathlineto{\pgfqpoint{3.019547in}{2.159192in}}%
\pgfpathlineto{\pgfqpoint{3.061774in}{2.191253in}}%
\pgfpathlineto{\pgfqpoint{3.104001in}{2.223313in}}%
\pgfpathlineto{\pgfqpoint{3.146228in}{2.255374in}}%
\pgfpathlineto{\pgfqpoint{3.188455in}{2.287435in}}%
\pgfpathlineto{\pgfqpoint{3.230682in}{2.319495in}}%
\pgfpathlineto{\pgfqpoint{3.272909in}{2.351556in}}%
\pgfpathlineto{\pgfqpoint{3.315136in}{2.383616in}}%
\pgfpathlineto{\pgfqpoint{3.357362in}{2.415677in}}%
\pgfpathlineto{\pgfqpoint{3.399589in}{2.447738in}}%
\pgfpathlineto{\pgfqpoint{3.441816in}{2.479798in}}%
\pgfpathlineto{\pgfqpoint{3.484043in}{2.511859in}}%
\pgfpathlineto{\pgfqpoint{3.526270in}{2.543920in}}%
\pgfpathlineto{\pgfqpoint{3.568497in}{2.575980in}}%
\pgfpathlineto{\pgfqpoint{3.610724in}{2.608041in}}%
\pgfpathlineto{\pgfqpoint{3.652951in}{2.640101in}}%
\pgfpathlineto{\pgfqpoint{3.695178in}{2.672162in}}%
\pgfpathlineto{\pgfqpoint{3.737405in}{2.704223in}}%
\pgfpathlineto{\pgfqpoint{3.779631in}{2.736283in}}%
\pgfpathlineto{\pgfqpoint{3.821858in}{2.768344in}}%
\pgfpathlineto{\pgfqpoint{3.864085in}{2.800404in}}%
\pgfpathlineto{\pgfqpoint{3.906312in}{2.832465in}}%
\pgfpathlineto{\pgfqpoint{3.948539in}{2.864526in}}%
\pgfpathlineto{\pgfqpoint{3.990766in}{2.896586in}}%
\pgfpathlineto{\pgfqpoint{4.032993in}{2.928647in}}%
\pgfpathlineto{\pgfqpoint{4.075220in}{2.960708in}}%
\pgfpathlineto{\pgfqpoint{4.117447in}{2.992768in}}%
\pgfpathlineto{\pgfqpoint{4.159674in}{3.024829in}}%
\pgfpathlineto{\pgfqpoint{4.201901in}{3.056889in}}%
\pgfpathlineto{\pgfqpoint{4.244127in}{3.088950in}}%
\pgfpathlineto{\pgfqpoint{4.286354in}{3.121011in}}%
\pgfpathlineto{\pgfqpoint{4.328581in}{3.153071in}}%
\pgfpathlineto{\pgfqpoint{4.370808in}{3.185132in}}%
\pgfpathlineto{\pgfqpoint{4.413035in}{3.217193in}}%
\pgfpathlineto{\pgfqpoint{4.455262in}{3.249253in}}%
\pgfpathlineto{\pgfqpoint{4.497489in}{3.281314in}}%
\pgfpathlineto{\pgfqpoint{4.539716in}{3.313374in}}%
\pgfpathlineto{\pgfqpoint{4.581943in}{3.345435in}}%
\pgfpathlineto{\pgfqpoint{4.624170in}{3.377496in}}%
\pgfpathlineto{\pgfqpoint{4.666397in}{3.409556in}}%
\pgfpathlineto{\pgfqpoint{4.708623in}{3.441617in}}%
\pgfpathlineto{\pgfqpoint{4.750850in}{3.473677in}}%
\pgfpathlineto{\pgfqpoint{4.793077in}{3.505738in}}%
\pgfpathlineto{\pgfqpoint{4.835304in}{3.537799in}}%
\pgfpathlineto{\pgfqpoint{4.877531in}{3.569859in}}%
\pgfpathlineto{\pgfqpoint{4.919758in}{3.601920in}}%
\pgfpathlineto{\pgfqpoint{4.961985in}{3.633981in}}%
\pgfpathlineto{\pgfqpoint{5.004212in}{3.666041in}}%
\pgfpathlineto{\pgfqpoint{5.046439in}{3.698102in}}%
\pgfpathlineto{\pgfqpoint{5.088666in}{3.730162in}}%
\pgfpathlineto{\pgfqpoint{5.130892in}{3.762223in}}%
\pgfpathlineto{\pgfqpoint{5.173119in}{3.794284in}}%
\pgfpathlineto{\pgfqpoint{5.215346in}{3.826344in}}%
\pgfpathlineto{\pgfqpoint{5.257573in}{3.858405in}}%
\pgfpathlineto{\pgfqpoint{5.299800in}{3.890466in}}%
\pgfpathlineto{\pgfqpoint{5.342027in}{3.922526in}}%
\pgfpathlineto{\pgfqpoint{5.384254in}{3.954587in}}%
\pgfpathlineto{\pgfqpoint{5.426481in}{3.986647in}}%
\pgfpathlineto{\pgfqpoint{5.468708in}{4.018708in}}%
\pgfpathlineto{\pgfqpoint{5.510935in}{4.050769in}}%
\pgfpathlineto{\pgfqpoint{5.553162in}{4.082829in}}%
\pgfpathlineto{\pgfqpoint{5.595388in}{4.114890in}}%
\pgfpathlineto{\pgfqpoint{5.637615in}{4.146951in}}%
\pgfpathlineto{\pgfqpoint{5.679842in}{4.179011in}}%
\pgfpathlineto{\pgfqpoint{5.722069in}{4.211072in}}%
\pgfpathlineto{\pgfqpoint{5.764296in}{4.243132in}}%
\pgfpathlineto{\pgfqpoint{5.806523in}{4.275193in}}%
\pgfpathlineto{\pgfqpoint{5.848750in}{4.307254in}}%
\pgfpathlineto{\pgfqpoint{5.890977in}{4.339314in}}%
\pgfusepath{stroke}%
\end{pgfscope}%
\begin{pgfscope}%
\pgfpathrectangle{\pgfqpoint{1.501490in}{0.964913in}}{\pgfqpoint{4.598510in}{3.535087in}}%
\pgfusepath{clip}%
\pgfsetrectcap%
\pgfsetroundjoin%
\pgfsetlinewidth{1.505625pt}%
\definecolor{currentstroke}{rgb}{1.000000,0.498039,0.054902}%
\pgfsetstrokecolor{currentstroke}%
\pgfsetdash{}{0pt}%
\pgfpathmoveto{\pgfqpoint{1.710513in}{1.269910in}}%
\pgfpathlineto{\pgfqpoint{1.752740in}{1.334259in}}%
\pgfpathlineto{\pgfqpoint{1.794967in}{1.363655in}}%
\pgfpathlineto{\pgfqpoint{1.837194in}{1.393050in}}%
\pgfpathlineto{\pgfqpoint{1.879421in}{1.422445in}}%
\pgfpathlineto{\pgfqpoint{1.921648in}{1.451840in}}%
\pgfpathlineto{\pgfqpoint{1.963874in}{1.481236in}}%
\pgfpathlineto{\pgfqpoint{2.006101in}{1.510631in}}%
\pgfpathlineto{\pgfqpoint{2.048328in}{1.540026in}}%
\pgfpathlineto{\pgfqpoint{2.090555in}{1.569421in}}%
\pgfpathlineto{\pgfqpoint{2.132782in}{1.598817in}}%
\pgfpathlineto{\pgfqpoint{2.175009in}{1.628212in}}%
\pgfpathlineto{\pgfqpoint{2.217236in}{1.657607in}}%
\pgfpathlineto{\pgfqpoint{2.259463in}{1.687002in}}%
\pgfpathlineto{\pgfqpoint{2.301690in}{1.716398in}}%
\pgfpathlineto{\pgfqpoint{2.343917in}{1.745793in}}%
\pgfpathlineto{\pgfqpoint{2.386144in}{1.775188in}}%
\pgfpathlineto{\pgfqpoint{2.428370in}{1.804583in}}%
\pgfpathlineto{\pgfqpoint{2.470597in}{1.833979in}}%
\pgfpathlineto{\pgfqpoint{2.512824in}{1.863374in}}%
\pgfpathlineto{\pgfqpoint{2.555051in}{1.892769in}}%
\pgfpathlineto{\pgfqpoint{2.597278in}{1.922164in}}%
\pgfpathlineto{\pgfqpoint{2.639505in}{1.951560in}}%
\pgfpathlineto{\pgfqpoint{2.681732in}{1.980955in}}%
\pgfpathlineto{\pgfqpoint{2.723959in}{2.010350in}}%
\pgfpathlineto{\pgfqpoint{2.766186in}{2.039745in}}%
\pgfpathlineto{\pgfqpoint{2.808413in}{2.069141in}}%
\pgfpathlineto{\pgfqpoint{2.850640in}{2.098536in}}%
\pgfpathlineto{\pgfqpoint{2.892866in}{2.127931in}}%
\pgfpathlineto{\pgfqpoint{2.935093in}{2.157326in}}%
\pgfpathlineto{\pgfqpoint{2.977320in}{2.186722in}}%
\pgfpathlineto{\pgfqpoint{3.019547in}{2.216117in}}%
\pgfpathlineto{\pgfqpoint{3.061774in}{2.245512in}}%
\pgfpathlineto{\pgfqpoint{3.104001in}{2.274907in}}%
\pgfpathlineto{\pgfqpoint{3.146228in}{2.304303in}}%
\pgfpathlineto{\pgfqpoint{3.188455in}{2.333698in}}%
\pgfpathlineto{\pgfqpoint{3.230682in}{2.363093in}}%
\pgfpathlineto{\pgfqpoint{3.272909in}{2.392488in}}%
\pgfpathlineto{\pgfqpoint{3.315136in}{2.421884in}}%
\pgfpathlineto{\pgfqpoint{3.357362in}{2.451279in}}%
\pgfpathlineto{\pgfqpoint{3.399589in}{2.480674in}}%
\pgfpathlineto{\pgfqpoint{3.441816in}{2.510069in}}%
\pgfpathlineto{\pgfqpoint{3.484043in}{2.539465in}}%
\pgfpathlineto{\pgfqpoint{3.526270in}{2.568860in}}%
\pgfpathlineto{\pgfqpoint{3.568497in}{2.598255in}}%
\pgfpathlineto{\pgfqpoint{3.610724in}{2.627650in}}%
\pgfpathlineto{\pgfqpoint{3.652951in}{2.657046in}}%
\pgfpathlineto{\pgfqpoint{3.695178in}{2.686441in}}%
\pgfpathlineto{\pgfqpoint{3.737405in}{2.715836in}}%
\pgfpathlineto{\pgfqpoint{3.779631in}{2.745231in}}%
\pgfpathlineto{\pgfqpoint{3.821858in}{2.774627in}}%
\pgfpathlineto{\pgfqpoint{3.864085in}{2.804022in}}%
\pgfpathlineto{\pgfqpoint{3.906312in}{2.833417in}}%
\pgfpathlineto{\pgfqpoint{3.948539in}{2.862812in}}%
\pgfpathlineto{\pgfqpoint{3.990766in}{2.892208in}}%
\pgfpathlineto{\pgfqpoint{4.032993in}{2.921603in}}%
\pgfpathlineto{\pgfqpoint{4.075220in}{2.950998in}}%
\pgfpathlineto{\pgfqpoint{4.117447in}{2.980393in}}%
\pgfpathlineto{\pgfqpoint{4.159674in}{3.009789in}}%
\pgfpathlineto{\pgfqpoint{4.201901in}{3.039184in}}%
\pgfpathlineto{\pgfqpoint{4.244127in}{3.068579in}}%
\pgfpathlineto{\pgfqpoint{4.286354in}{3.097974in}}%
\pgfpathlineto{\pgfqpoint{4.328581in}{3.127369in}}%
\pgfpathlineto{\pgfqpoint{4.370808in}{3.156765in}}%
\pgfpathlineto{\pgfqpoint{4.413035in}{3.186160in}}%
\pgfpathlineto{\pgfqpoint{4.455262in}{3.215555in}}%
\pgfpathlineto{\pgfqpoint{4.497489in}{3.244950in}}%
\pgfpathlineto{\pgfqpoint{4.539716in}{3.274346in}}%
\pgfpathlineto{\pgfqpoint{4.581943in}{3.303741in}}%
\pgfpathlineto{\pgfqpoint{4.624170in}{3.333136in}}%
\pgfpathlineto{\pgfqpoint{4.666397in}{3.362531in}}%
\pgfpathlineto{\pgfqpoint{4.708623in}{3.391927in}}%
\pgfpathlineto{\pgfqpoint{4.750850in}{3.421322in}}%
\pgfpathlineto{\pgfqpoint{4.793077in}{3.450717in}}%
\pgfpathlineto{\pgfqpoint{4.835304in}{3.480112in}}%
\pgfpathlineto{\pgfqpoint{4.877531in}{3.509508in}}%
\pgfpathlineto{\pgfqpoint{4.919758in}{3.538903in}}%
\pgfpathlineto{\pgfqpoint{4.961985in}{3.568298in}}%
\pgfpathlineto{\pgfqpoint{5.004212in}{3.597693in}}%
\pgfpathlineto{\pgfqpoint{5.046439in}{3.627089in}}%
\pgfpathlineto{\pgfqpoint{5.088666in}{3.656484in}}%
\pgfpathlineto{\pgfqpoint{5.130892in}{3.685879in}}%
\pgfpathlineto{\pgfqpoint{5.173119in}{3.715274in}}%
\pgfpathlineto{\pgfqpoint{5.215346in}{3.744670in}}%
\pgfpathlineto{\pgfqpoint{5.257573in}{3.774065in}}%
\pgfpathlineto{\pgfqpoint{5.299800in}{3.803460in}}%
\pgfpathlineto{\pgfqpoint{5.342027in}{3.832855in}}%
\pgfpathlineto{\pgfqpoint{5.384254in}{3.862251in}}%
\pgfpathlineto{\pgfqpoint{5.426481in}{3.891646in}}%
\pgfpathlineto{\pgfqpoint{5.468708in}{3.921041in}}%
\pgfpathlineto{\pgfqpoint{5.510935in}{3.950436in}}%
\pgfpathlineto{\pgfqpoint{5.553162in}{3.979832in}}%
\pgfpathlineto{\pgfqpoint{5.595388in}{4.009227in}}%
\pgfpathlineto{\pgfqpoint{5.637615in}{4.038622in}}%
\pgfpathlineto{\pgfqpoint{5.679842in}{4.068017in}}%
\pgfpathlineto{\pgfqpoint{5.722069in}{4.097413in}}%
\pgfpathlineto{\pgfqpoint{5.764296in}{4.126808in}}%
\pgfpathlineto{\pgfqpoint{5.806523in}{4.156203in}}%
\pgfpathlineto{\pgfqpoint{5.848750in}{4.185598in}}%
\pgfpathlineto{\pgfqpoint{5.890977in}{4.214994in}}%
\pgfusepath{stroke}%
\end{pgfscope}%
\begin{pgfscope}%
\pgfpathrectangle{\pgfqpoint{1.501490in}{0.964913in}}{\pgfqpoint{4.598510in}{3.535087in}}%
\pgfusepath{clip}%
\pgfsetrectcap%
\pgfsetroundjoin%
\pgfsetlinewidth{1.505625pt}%
\definecolor{currentstroke}{rgb}{0.172549,0.627451,0.172549}%
\pgfsetstrokecolor{currentstroke}%
\pgfsetdash{}{0pt}%
\pgfpathmoveto{\pgfqpoint{1.710513in}{1.125599in}}%
\pgfpathlineto{\pgfqpoint{1.752740in}{1.175341in}}%
\pgfpathlineto{\pgfqpoint{1.794967in}{1.184718in}}%
\pgfpathlineto{\pgfqpoint{1.837194in}{1.187459in}}%
\pgfpathlineto{\pgfqpoint{1.879421in}{1.190201in}}%
\pgfpathlineto{\pgfqpoint{1.921648in}{1.192942in}}%
\pgfpathlineto{\pgfqpoint{1.963874in}{1.195684in}}%
\pgfpathlineto{\pgfqpoint{2.006101in}{1.198425in}}%
\pgfpathlineto{\pgfqpoint{2.048328in}{1.201167in}}%
\pgfpathlineto{\pgfqpoint{2.090555in}{1.203908in}}%
\pgfpathlineto{\pgfqpoint{2.132782in}{1.206650in}}%
\pgfpathlineto{\pgfqpoint{2.175009in}{1.209391in}}%
\pgfpathlineto{\pgfqpoint{2.217236in}{1.212133in}}%
\pgfpathlineto{\pgfqpoint{2.259463in}{1.214875in}}%
\pgfpathlineto{\pgfqpoint{2.301690in}{1.217616in}}%
\pgfpathlineto{\pgfqpoint{2.343917in}{1.220358in}}%
\pgfpathlineto{\pgfqpoint{2.386144in}{1.223099in}}%
\pgfpathlineto{\pgfqpoint{2.428370in}{1.225841in}}%
\pgfpathlineto{\pgfqpoint{2.470597in}{1.228582in}}%
\pgfpathlineto{\pgfqpoint{2.512824in}{1.231324in}}%
\pgfpathlineto{\pgfqpoint{2.555051in}{1.234065in}}%
\pgfpathlineto{\pgfqpoint{2.597278in}{1.236807in}}%
\pgfpathlineto{\pgfqpoint{2.639505in}{1.239548in}}%
\pgfpathlineto{\pgfqpoint{2.681732in}{1.242290in}}%
\pgfpathlineto{\pgfqpoint{2.723959in}{1.245031in}}%
\pgfpathlineto{\pgfqpoint{2.766186in}{1.247773in}}%
\pgfpathlineto{\pgfqpoint{2.808413in}{1.250514in}}%
\pgfpathlineto{\pgfqpoint{2.850640in}{1.253256in}}%
\pgfpathlineto{\pgfqpoint{2.892866in}{1.255997in}}%
\pgfpathlineto{\pgfqpoint{2.935093in}{1.258739in}}%
\pgfpathlineto{\pgfqpoint{2.977320in}{1.261480in}}%
\pgfpathlineto{\pgfqpoint{3.019547in}{1.264222in}}%
\pgfpathlineto{\pgfqpoint{3.061774in}{1.266964in}}%
\pgfpathlineto{\pgfqpoint{3.104001in}{1.269705in}}%
\pgfpathlineto{\pgfqpoint{3.146228in}{1.272447in}}%
\pgfpathlineto{\pgfqpoint{3.188455in}{1.275188in}}%
\pgfpathlineto{\pgfqpoint{3.230682in}{1.277930in}}%
\pgfpathlineto{\pgfqpoint{3.272909in}{1.280671in}}%
\pgfpathlineto{\pgfqpoint{3.315136in}{1.283413in}}%
\pgfpathlineto{\pgfqpoint{3.357362in}{1.286154in}}%
\pgfpathlineto{\pgfqpoint{3.399589in}{1.288896in}}%
\pgfpathlineto{\pgfqpoint{3.441816in}{1.291637in}}%
\pgfpathlineto{\pgfqpoint{3.484043in}{1.294379in}}%
\pgfpathlineto{\pgfqpoint{3.526270in}{1.297120in}}%
\pgfpathlineto{\pgfqpoint{3.568497in}{1.299862in}}%
\pgfpathlineto{\pgfqpoint{3.610724in}{1.302603in}}%
\pgfpathlineto{\pgfqpoint{3.652951in}{1.305345in}}%
\pgfpathlineto{\pgfqpoint{3.695178in}{1.308086in}}%
\pgfpathlineto{\pgfqpoint{3.737405in}{1.310828in}}%
\pgfpathlineto{\pgfqpoint{3.779631in}{1.313569in}}%
\pgfpathlineto{\pgfqpoint{3.821858in}{1.316311in}}%
\pgfpathlineto{\pgfqpoint{3.864085in}{1.319052in}}%
\pgfpathlineto{\pgfqpoint{3.906312in}{1.321794in}}%
\pgfpathlineto{\pgfqpoint{3.948539in}{1.324536in}}%
\pgfpathlineto{\pgfqpoint{3.990766in}{1.327277in}}%
\pgfpathlineto{\pgfqpoint{4.032993in}{1.330019in}}%
\pgfpathlineto{\pgfqpoint{4.075220in}{1.332760in}}%
\pgfpathlineto{\pgfqpoint{4.117447in}{1.335502in}}%
\pgfpathlineto{\pgfqpoint{4.159674in}{1.338243in}}%
\pgfpathlineto{\pgfqpoint{4.201901in}{1.340985in}}%
\pgfpathlineto{\pgfqpoint{4.244127in}{1.343726in}}%
\pgfpathlineto{\pgfqpoint{4.286354in}{1.346468in}}%
\pgfpathlineto{\pgfqpoint{4.328581in}{1.349209in}}%
\pgfpathlineto{\pgfqpoint{4.370808in}{1.351951in}}%
\pgfpathlineto{\pgfqpoint{4.413035in}{1.354692in}}%
\pgfpathlineto{\pgfqpoint{4.455262in}{1.357434in}}%
\pgfpathlineto{\pgfqpoint{4.497489in}{1.360175in}}%
\pgfpathlineto{\pgfqpoint{4.539716in}{1.362917in}}%
\pgfpathlineto{\pgfqpoint{4.581943in}{1.365658in}}%
\pgfpathlineto{\pgfqpoint{4.624170in}{1.368400in}}%
\pgfpathlineto{\pgfqpoint{4.666397in}{1.371141in}}%
\pgfpathlineto{\pgfqpoint{4.708623in}{1.373883in}}%
\pgfpathlineto{\pgfqpoint{4.750850in}{1.376625in}}%
\pgfpathlineto{\pgfqpoint{4.793077in}{1.379366in}}%
\pgfpathlineto{\pgfqpoint{4.835304in}{1.382108in}}%
\pgfpathlineto{\pgfqpoint{4.877531in}{1.384849in}}%
\pgfpathlineto{\pgfqpoint{4.919758in}{1.387591in}}%
\pgfpathlineto{\pgfqpoint{4.961985in}{1.390332in}}%
\pgfpathlineto{\pgfqpoint{5.004212in}{1.393074in}}%
\pgfpathlineto{\pgfqpoint{5.046439in}{1.395815in}}%
\pgfpathlineto{\pgfqpoint{5.088666in}{1.398557in}}%
\pgfpathlineto{\pgfqpoint{5.130892in}{1.401298in}}%
\pgfpathlineto{\pgfqpoint{5.173119in}{1.404040in}}%
\pgfpathlineto{\pgfqpoint{5.215346in}{1.406781in}}%
\pgfpathlineto{\pgfqpoint{5.257573in}{1.409523in}}%
\pgfpathlineto{\pgfqpoint{5.299800in}{1.412264in}}%
\pgfpathlineto{\pgfqpoint{5.342027in}{1.415006in}}%
\pgfpathlineto{\pgfqpoint{5.384254in}{1.417747in}}%
\pgfpathlineto{\pgfqpoint{5.426481in}{1.420489in}}%
\pgfpathlineto{\pgfqpoint{5.468708in}{1.423230in}}%
\pgfpathlineto{\pgfqpoint{5.510935in}{1.425972in}}%
\pgfpathlineto{\pgfqpoint{5.553162in}{1.428714in}}%
\pgfpathlineto{\pgfqpoint{5.595388in}{1.431455in}}%
\pgfpathlineto{\pgfqpoint{5.637615in}{1.434197in}}%
\pgfpathlineto{\pgfqpoint{5.679842in}{1.436938in}}%
\pgfpathlineto{\pgfqpoint{5.722069in}{1.439680in}}%
\pgfpathlineto{\pgfqpoint{5.764296in}{1.442421in}}%
\pgfpathlineto{\pgfqpoint{5.806523in}{1.445163in}}%
\pgfpathlineto{\pgfqpoint{5.848750in}{1.447904in}}%
\pgfpathlineto{\pgfqpoint{5.890977in}{1.450646in}}%
\pgfusepath{stroke}%
\end{pgfscope}%
\begin{pgfscope}%
\pgfsetrectcap%
\pgfsetmiterjoin%
\pgfsetlinewidth{0.803000pt}%
\definecolor{currentstroke}{rgb}{0.000000,0.000000,0.000000}%
\pgfsetstrokecolor{currentstroke}%
\pgfsetdash{}{0pt}%
\pgfpathmoveto{\pgfqpoint{1.501490in}{0.964913in}}%
\pgfpathlineto{\pgfqpoint{1.501490in}{4.500000in}}%
\pgfusepath{stroke}%
\end{pgfscope}%
\begin{pgfscope}%
\pgfsetrectcap%
\pgfsetmiterjoin%
\pgfsetlinewidth{0.803000pt}%
\definecolor{currentstroke}{rgb}{0.000000,0.000000,0.000000}%
\pgfsetstrokecolor{currentstroke}%
\pgfsetdash{}{0pt}%
\pgfpathmoveto{\pgfqpoint{6.100000in}{0.964913in}}%
\pgfpathlineto{\pgfqpoint{6.100000in}{4.500000in}}%
\pgfusepath{stroke}%
\end{pgfscope}%
\begin{pgfscope}%
\pgfsetrectcap%
\pgfsetmiterjoin%
\pgfsetlinewidth{0.803000pt}%
\definecolor{currentstroke}{rgb}{0.000000,0.000000,0.000000}%
\pgfsetstrokecolor{currentstroke}%
\pgfsetdash{}{0pt}%
\pgfpathmoveto{\pgfqpoint{1.501490in}{0.964913in}}%
\pgfpathlineto{\pgfqpoint{6.100000in}{0.964913in}}%
\pgfusepath{stroke}%
\end{pgfscope}%
\begin{pgfscope}%
\pgfsetrectcap%
\pgfsetmiterjoin%
\pgfsetlinewidth{0.803000pt}%
\definecolor{currentstroke}{rgb}{0.000000,0.000000,0.000000}%
\pgfsetstrokecolor{currentstroke}%
\pgfsetdash{}{0pt}%
\pgfpathmoveto{\pgfqpoint{1.501490in}{4.500000in}}%
\pgfpathlineto{\pgfqpoint{6.100000in}{4.500000in}}%
\pgfusepath{stroke}%
\end{pgfscope}%
\begin{pgfscope}%
\pgfsetbuttcap%
\pgfsetmiterjoin%
\definecolor{currentfill}{rgb}{1.000000,1.000000,1.000000}%
\pgfsetfillcolor{currentfill}%
\pgfsetfillopacity{0.800000}%
\pgfsetlinewidth{1.003750pt}%
\definecolor{currentstroke}{rgb}{0.800000,0.800000,0.800000}%
\pgfsetstrokecolor{currentstroke}%
\pgfsetstrokeopacity{0.800000}%
\pgfsetdash{}{0pt}%
\pgfpathmoveto{\pgfqpoint{1.695934in}{3.092908in}}%
\pgfpathlineto{\pgfqpoint{3.746195in}{3.092908in}}%
\pgfpathquadraticcurveto{\pgfqpoint{3.801750in}{3.092908in}}{\pgfqpoint{3.801750in}{3.148464in}}%
\pgfpathlineto{\pgfqpoint{3.801750in}{4.305556in}}%
\pgfpathquadraticcurveto{\pgfqpoint{3.801750in}{4.361111in}}{\pgfqpoint{3.746195in}{4.361111in}}%
\pgfpathlineto{\pgfqpoint{1.695934in}{4.361111in}}%
\pgfpathquadraticcurveto{\pgfqpoint{1.640379in}{4.361111in}}{\pgfqpoint{1.640379in}{4.305556in}}%
\pgfpathlineto{\pgfqpoint{1.640379in}{3.148464in}}%
\pgfpathquadraticcurveto{\pgfqpoint{1.640379in}{3.092908in}}{\pgfqpoint{1.695934in}{3.092908in}}%
\pgfpathclose%
\pgfusepath{stroke,fill}%
\end{pgfscope}%
\begin{pgfscope}%
\pgfsetrectcap%
\pgfsetroundjoin%
\pgfsetlinewidth{1.505625pt}%
\definecolor{currentstroke}{rgb}{0.121569,0.466667,0.705882}%
\pgfsetstrokecolor{currentstroke}%
\pgfsetdash{}{0pt}%
\pgfpathmoveto{\pgfqpoint{1.751490in}{4.147184in}}%
\pgfpathlineto{\pgfqpoint{2.307045in}{4.147184in}}%
\pgfusepath{stroke}%
\end{pgfscope}%
\begin{pgfscope}%
\definecolor{textcolor}{rgb}{0.000000,0.000000,0.000000}%
\pgfsetstrokecolor{textcolor}%
\pgfsetfillcolor{textcolor}%
\pgftext[x=2.529268in,y=4.049962in,left,base]{\color{textcolor}\fontsize{20.000000}{24.000000}\selectfont PV-OSIM}%
\end{pgfscope}%
\begin{pgfscope}%
\pgfsetrectcap%
\pgfsetroundjoin%
\pgfsetlinewidth{1.505625pt}%
\definecolor{currentstroke}{rgb}{1.000000,0.498039,0.054902}%
\pgfsetstrokecolor{currentstroke}%
\pgfsetdash{}{0pt}%
\pgfpathmoveto{\pgfqpoint{1.751490in}{3.752227in}}%
\pgfpathlineto{\pgfqpoint{2.307045in}{3.752227in}}%
\pgfusepath{stroke}%
\end{pgfscope}%
\begin{pgfscope}%
\definecolor{textcolor}{rgb}{0.000000,0.000000,0.000000}%
\pgfsetstrokecolor{textcolor}%
\pgfsetfillcolor{textcolor}%
\pgftext[x=2.529268in,y=3.655005in,left,base]{\color{textcolor}\fontsize{20.000000}{24.000000}\selectfont EFPA}%
\end{pgfscope}%
\begin{pgfscope}%
\pgfsetrectcap%
\pgfsetroundjoin%
\pgfsetlinewidth{1.505625pt}%
\definecolor{currentstroke}{rgb}{0.172549,0.627451,0.172549}%
\pgfsetstrokecolor{currentstroke}%
\pgfsetdash{}{0pt}%
\pgfpathmoveto{\pgfqpoint{1.751490in}{3.357271in}}%
\pgfpathlineto{\pgfqpoint{2.307045in}{3.357271in}}%
\pgfusepath{stroke}%
\end{pgfscope}%
\begin{pgfscope}%
\definecolor{textcolor}{rgb}{0.000000,0.000000,0.000000}%
\pgfsetstrokecolor{textcolor}%
\pgfsetfillcolor{textcolor}%
\pgftext[x=2.529268in,y=3.260048in,left,base]{\color{textcolor}\fontsize{20.000000}{24.000000}\selectfont PV-OSIMr}%
\end{pgfscope}%
\end{pgfpicture}%
\makeatother%
\endgroup%

%% file: graphics/chain_md.pgf
%% Creator: Matplotlib, PGF backend
%%
%% To include the figure in your LaTeX document, write
%%   \input{<filename>.pgf}
%%
%% Make sure the required packages are loaded in your preamble
%%   \usepackage{pgf}
%%
%% Figures using additional raster images can only be included by \input if
%% they are in the same directory as the main LaTeX file. For loading figures
%% from other directories you can use the `import` package
%% and then include the figures with
%%   \import{<path to file>}{<filename>.pgf}
%%
%% Matplotlib used the following preamble
%%
\begingroup%
\makeatletter%
\begin{pgfpicture}%
\pgfpathrectangle{\pgfpointorigin}{\pgfqpoint{6.400000in}{4.800000in}}%
\pgfusepath{use as bounding box, clip}%
\begin{pgfscope}%
\pgfsetbuttcap%
\pgfsetmiterjoin%
\definecolor{currentfill}{rgb}{1.000000,1.000000,1.000000}%
\pgfsetfillcolor{currentfill}%
\pgfsetlinewidth{0.000000pt}%
\definecolor{currentstroke}{rgb}{1.000000,1.000000,1.000000}%
\pgfsetstrokecolor{currentstroke}%
\pgfsetdash{}{0pt}%
\pgfpathmoveto{\pgfqpoint{0.000000in}{0.000000in}}%
\pgfpathlineto{\pgfqpoint{6.400000in}{0.000000in}}%
\pgfpathlineto{\pgfqpoint{6.400000in}{4.800000in}}%
\pgfpathlineto{\pgfqpoint{0.000000in}{4.800000in}}%
\pgfpathclose%
\pgfusepath{fill}%
\end{pgfscope}%
\begin{pgfscope}%
\pgfsetbuttcap%
\pgfsetmiterjoin%
\definecolor{currentfill}{rgb}{1.000000,1.000000,1.000000}%
\pgfsetfillcolor{currentfill}%
\pgfsetlinewidth{0.000000pt}%
\definecolor{currentstroke}{rgb}{0.000000,0.000000,0.000000}%
\pgfsetstrokecolor{currentstroke}%
\pgfsetstrokeopacity{0.000000}%
\pgfsetdash{}{0pt}%
\pgfpathmoveto{\pgfqpoint{1.047462in}{0.944795in}}%
\pgfpathlineto{\pgfqpoint{6.100000in}{0.944795in}}%
\pgfpathlineto{\pgfqpoint{6.100000in}{4.500000in}}%
\pgfpathlineto{\pgfqpoint{1.047462in}{4.500000in}}%
\pgfpathclose%
\pgfusepath{fill}%
\end{pgfscope}%
\begin{pgfscope}%
\pgfsetbuttcap%
\pgfsetroundjoin%
\definecolor{currentfill}{rgb}{0.000000,0.000000,0.000000}%
\pgfsetfillcolor{currentfill}%
\pgfsetlinewidth{0.803000pt}%
\definecolor{currentstroke}{rgb}{0.000000,0.000000,0.000000}%
\pgfsetstrokecolor{currentstroke}%
\pgfsetdash{}{0pt}%
\pgfsys@defobject{currentmarker}{\pgfqpoint{0.000000in}{-0.048611in}}{\pgfqpoint{0.000000in}{0.000000in}}{%
\pgfpathmoveto{\pgfqpoint{0.000000in}{0.000000in}}%
\pgfpathlineto{\pgfqpoint{0.000000in}{-0.048611in}}%
\pgfusepath{stroke,fill}%
}%
\begin{pgfscope}%
\pgfsys@transformshift{4.487643in}{0.944795in}%
\pgfsys@useobject{currentmarker}{}%
\end{pgfscope}%
\end{pgfscope}%
\begin{pgfscope}%
\definecolor{textcolor}{rgb}{0.000000,0.000000,0.000000}%
\pgfsetstrokecolor{textcolor}%
\pgfsetfillcolor{textcolor}%
\pgftext[x=4.487643in,y=0.847573in,,top]{\color{textcolor}\fontsize{50.000000}{24.000000}\selectfont \(\displaystyle {10^{1}}\)}%
\end{pgfscope}%
\begin{pgfscope}%
\pgfsetbuttcap%
\pgfsetroundjoin%
\definecolor{currentfill}{rgb}{0.000000,0.000000,0.000000}%
\pgfsetfillcolor{currentfill}%
\pgfsetlinewidth{0.602250pt}%
\definecolor{currentstroke}{rgb}{0.000000,0.000000,0.000000}%
\pgfsetstrokecolor{currentstroke}%
\pgfsetdash{}{0pt}%
\pgfsys@defobject{currentmarker}{\pgfqpoint{0.000000in}{-0.027778in}}{\pgfqpoint{0.000000in}{0.000000in}}{%
\pgfpathmoveto{\pgfqpoint{0.000000in}{0.000000in}}%
\pgfpathlineto{\pgfqpoint{0.000000in}{-0.027778in}}%
\pgfusepath{stroke,fill}%
}%
\begin{pgfscope}%
\pgfsys@transformshift{1.277123in}{0.944795in}%
\pgfsys@useobject{currentmarker}{}%
\end{pgfscope}%
\end{pgfscope}%
\begin{pgfscope}%
\pgfsetbuttcap%
\pgfsetroundjoin%
\definecolor{currentfill}{rgb}{0.000000,0.000000,0.000000}%
\pgfsetfillcolor{currentfill}%
\pgfsetlinewidth{0.602250pt}%
\definecolor{currentstroke}{rgb}{0.000000,0.000000,0.000000}%
\pgfsetstrokecolor{currentstroke}%
\pgfsetdash{}{0pt}%
\pgfsys@defobject{currentmarker}{\pgfqpoint{0.000000in}{-0.027778in}}{\pgfqpoint{0.000000in}{0.000000in}}{%
\pgfpathmoveto{\pgfqpoint{0.000000in}{0.000000in}}%
\pgfpathlineto{\pgfqpoint{0.000000in}{-0.027778in}}%
\pgfusepath{stroke,fill}%
}%
\begin{pgfscope}%
\pgfsys@transformshift{2.085948in}{0.944795in}%
\pgfsys@useobject{currentmarker}{}%
\end{pgfscope}%
\end{pgfscope}%
\begin{pgfscope}%
\pgfsetbuttcap%
\pgfsetroundjoin%
\definecolor{currentfill}{rgb}{0.000000,0.000000,0.000000}%
\pgfsetfillcolor{currentfill}%
\pgfsetlinewidth{0.602250pt}%
\definecolor{currentstroke}{rgb}{0.000000,0.000000,0.000000}%
\pgfsetstrokecolor{currentstroke}%
\pgfsetdash{}{0pt}%
\pgfsys@defobject{currentmarker}{\pgfqpoint{0.000000in}{-0.027778in}}{\pgfqpoint{0.000000in}{0.000000in}}{%
\pgfpathmoveto{\pgfqpoint{0.000000in}{0.000000in}}%
\pgfpathlineto{\pgfqpoint{0.000000in}{-0.027778in}}%
\pgfusepath{stroke,fill}%
}%
\begin{pgfscope}%
\pgfsys@transformshift{2.659819in}{0.944795in}%
\pgfsys@useobject{currentmarker}{}%
\end{pgfscope}%
\end{pgfscope}%
\begin{pgfscope}%
\pgfsetbuttcap%
\pgfsetroundjoin%
\definecolor{currentfill}{rgb}{0.000000,0.000000,0.000000}%
\pgfsetfillcolor{currentfill}%
\pgfsetlinewidth{0.602250pt}%
\definecolor{currentstroke}{rgb}{0.000000,0.000000,0.000000}%
\pgfsetstrokecolor{currentstroke}%
\pgfsetdash{}{0pt}%
\pgfsys@defobject{currentmarker}{\pgfqpoint{0.000000in}{-0.027778in}}{\pgfqpoint{0.000000in}{0.000000in}}{%
\pgfpathmoveto{\pgfqpoint{0.000000in}{0.000000in}}%
\pgfpathlineto{\pgfqpoint{0.000000in}{-0.027778in}}%
\pgfusepath{stroke,fill}%
}%
\begin{pgfscope}%
\pgfsys@transformshift{3.104948in}{0.944795in}%
\pgfsys@useobject{currentmarker}{}%
\end{pgfscope}%
\end{pgfscope}%
\begin{pgfscope}%
\pgfsetbuttcap%
\pgfsetroundjoin%
\definecolor{currentfill}{rgb}{0.000000,0.000000,0.000000}%
\pgfsetfillcolor{currentfill}%
\pgfsetlinewidth{0.602250pt}%
\definecolor{currentstroke}{rgb}{0.000000,0.000000,0.000000}%
\pgfsetstrokecolor{currentstroke}%
\pgfsetdash{}{0pt}%
\pgfsys@defobject{currentmarker}{\pgfqpoint{0.000000in}{-0.027778in}}{\pgfqpoint{0.000000in}{0.000000in}}{%
\pgfpathmoveto{\pgfqpoint{0.000000in}{0.000000in}}%
\pgfpathlineto{\pgfqpoint{0.000000in}{-0.027778in}}%
\pgfusepath{stroke,fill}%
}%
\begin{pgfscope}%
\pgfsys@transformshift{3.468644in}{0.944795in}%
\pgfsys@useobject{currentmarker}{}%
\end{pgfscope}%
\end{pgfscope}%
\begin{pgfscope}%
\pgfsetbuttcap%
\pgfsetroundjoin%
\definecolor{currentfill}{rgb}{0.000000,0.000000,0.000000}%
\pgfsetfillcolor{currentfill}%
\pgfsetlinewidth{0.602250pt}%
\definecolor{currentstroke}{rgb}{0.000000,0.000000,0.000000}%
\pgfsetstrokecolor{currentstroke}%
\pgfsetdash{}{0pt}%
\pgfsys@defobject{currentmarker}{\pgfqpoint{0.000000in}{-0.027778in}}{\pgfqpoint{0.000000in}{0.000000in}}{%
\pgfpathmoveto{\pgfqpoint{0.000000in}{0.000000in}}%
\pgfpathlineto{\pgfqpoint{0.000000in}{-0.027778in}}%
\pgfusepath{stroke,fill}%
}%
\begin{pgfscope}%
\pgfsys@transformshift{3.776145in}{0.944795in}%
\pgfsys@useobject{currentmarker}{}%
\end{pgfscope}%
\end{pgfscope}%
\begin{pgfscope}%
\pgfsetbuttcap%
\pgfsetroundjoin%
\definecolor{currentfill}{rgb}{0.000000,0.000000,0.000000}%
\pgfsetfillcolor{currentfill}%
\pgfsetlinewidth{0.602250pt}%
\definecolor{currentstroke}{rgb}{0.000000,0.000000,0.000000}%
\pgfsetstrokecolor{currentstroke}%
\pgfsetdash{}{0pt}%
\pgfsys@defobject{currentmarker}{\pgfqpoint{0.000000in}{-0.027778in}}{\pgfqpoint{0.000000in}{0.000000in}}{%
\pgfpathmoveto{\pgfqpoint{0.000000in}{0.000000in}}%
\pgfpathlineto{\pgfqpoint{0.000000in}{-0.027778in}}%
\pgfusepath{stroke,fill}%
}%
\begin{pgfscope}%
\pgfsys@transformshift{4.042515in}{0.944795in}%
\pgfsys@useobject{currentmarker}{}%
\end{pgfscope}%
\end{pgfscope}%
\begin{pgfscope}%
\pgfsetbuttcap%
\pgfsetroundjoin%
\definecolor{currentfill}{rgb}{0.000000,0.000000,0.000000}%
\pgfsetfillcolor{currentfill}%
\pgfsetlinewidth{0.602250pt}%
\definecolor{currentstroke}{rgb}{0.000000,0.000000,0.000000}%
\pgfsetstrokecolor{currentstroke}%
\pgfsetdash{}{0pt}%
\pgfsys@defobject{currentmarker}{\pgfqpoint{0.000000in}{-0.027778in}}{\pgfqpoint{0.000000in}{0.000000in}}{%
\pgfpathmoveto{\pgfqpoint{0.000000in}{0.000000in}}%
\pgfpathlineto{\pgfqpoint{0.000000in}{-0.027778in}}%
\pgfusepath{stroke,fill}%
}%
\begin{pgfscope}%
\pgfsys@transformshift{4.277469in}{0.944795in}%
\pgfsys@useobject{currentmarker}{}%
\end{pgfscope}%
\end{pgfscope}%
\begin{pgfscope}%
\pgfsetbuttcap%
\pgfsetroundjoin%
\definecolor{currentfill}{rgb}{0.000000,0.000000,0.000000}%
\pgfsetfillcolor{currentfill}%
\pgfsetlinewidth{0.602250pt}%
\definecolor{currentstroke}{rgb}{0.000000,0.000000,0.000000}%
\pgfsetstrokecolor{currentstroke}%
\pgfsetdash{}{0pt}%
\pgfsys@defobject{currentmarker}{\pgfqpoint{0.000000in}{-0.027778in}}{\pgfqpoint{0.000000in}{0.000000in}}{%
\pgfpathmoveto{\pgfqpoint{0.000000in}{0.000000in}}%
\pgfpathlineto{\pgfqpoint{0.000000in}{-0.027778in}}%
\pgfusepath{stroke,fill}%
}%
\begin{pgfscope}%
\pgfsys@transformshift{5.870339in}{0.944795in}%
\pgfsys@useobject{currentmarker}{}%
\end{pgfscope}%
\end{pgfscope}%
\begin{pgfscope}%
\definecolor{textcolor}{rgb}{0.000000,0.000000,0.000000}%
\pgfsetstrokecolor{textcolor}%
\pgfsetfillcolor{textcolor}%
\pgftext[x=3.573731in,y=0.535950in,,top]{\color{textcolor}\fontsize{30.000000}{24.000000}\selectfont k}%
\end{pgfscope}%
\begin{pgfscope}%
\pgfsetbuttcap%
\pgfsetroundjoin%
\definecolor{currentfill}{rgb}{0.000000,0.000000,0.000000}%
\pgfsetfillcolor{currentfill}%
\pgfsetlinewidth{0.803000pt}%
\definecolor{currentstroke}{rgb}{0.000000,0.000000,0.000000}%
\pgfsetstrokecolor{currentstroke}%
\pgfsetdash{}{0pt}%
\pgfsys@defobject{currentmarker}{\pgfqpoint{-0.048611in}{0.000000in}}{\pgfqpoint{-0.000000in}{0.000000in}}{%
\pgfpathmoveto{\pgfqpoint{-0.000000in}{0.000000in}}%
\pgfpathlineto{\pgfqpoint{-0.048611in}{0.000000in}}%
\pgfusepath{stroke,fill}%
}%
\begin{pgfscope}%
\pgfsys@transformshift{1.047462in}{1.948508in}%
\pgfsys@useobject{currentmarker}{}%
\end{pgfscope}%
\end{pgfscope}%
\begin{pgfscope}%
\definecolor{textcolor}{rgb}{0.000000,0.000000,0.000000}%
\pgfsetstrokecolor{textcolor}%
\pgfsetfillcolor{textcolor}%
\pgftext[x=0.581165in, y=1.848489in, left, base]{\color{textcolor}\fontsize{30.000000}{24.000000}\selectfont \(\displaystyle {10^{4}}\)}%
\end{pgfscope}%
\begin{pgfscope}%
\pgfsetbuttcap%
\pgfsetroundjoin%
\definecolor{currentfill}{rgb}{0.000000,0.000000,0.000000}%
\pgfsetfillcolor{currentfill}%
\pgfsetlinewidth{0.803000pt}%
\definecolor{currentstroke}{rgb}{0.000000,0.000000,0.000000}%
\pgfsetstrokecolor{currentstroke}%
\pgfsetdash{}{0pt}%
\pgfsys@defobject{currentmarker}{\pgfqpoint{-0.048611in}{0.000000in}}{\pgfqpoint{-0.000000in}{0.000000in}}{%
\pgfpathmoveto{\pgfqpoint{-0.000000in}{0.000000in}}%
\pgfpathlineto{\pgfqpoint{-0.048611in}{0.000000in}}%
\pgfusepath{stroke,fill}%
}%
\begin{pgfscope}%
\pgfsys@transformshift{1.047462in}{3.006597in}%
\pgfsys@useobject{currentmarker}{}%
\end{pgfscope}%
\end{pgfscope}%
\begin{pgfscope}%
\definecolor{textcolor}{rgb}{0.000000,0.000000,0.000000}%
\pgfsetstrokecolor{textcolor}%
\pgfsetfillcolor{textcolor}%
\pgftext[x=0.581165in, y=2.906578in, left, base]{\color{textcolor}\fontsize{30.000000}{24.000000}\selectfont \(\displaystyle {10^{5}}\)}%
\end{pgfscope}%
\begin{pgfscope}%
\pgfsetbuttcap%
\pgfsetroundjoin%
\definecolor{currentfill}{rgb}{0.000000,0.000000,0.000000}%
\pgfsetfillcolor{currentfill}%
\pgfsetlinewidth{0.803000pt}%
\definecolor{currentstroke}{rgb}{0.000000,0.000000,0.000000}%
\pgfsetstrokecolor{currentstroke}%
\pgfsetdash{}{0pt}%
\pgfsys@defobject{currentmarker}{\pgfqpoint{-0.048611in}{0.000000in}}{\pgfqpoint{-0.000000in}{0.000000in}}{%
\pgfpathmoveto{\pgfqpoint{-0.000000in}{0.000000in}}%
\pgfpathlineto{\pgfqpoint{-0.048611in}{0.000000in}}%
\pgfusepath{stroke,fill}%
}%
\begin{pgfscope}%
\pgfsys@transformshift{1.047462in}{4.064687in}%
\pgfsys@useobject{currentmarker}{}%
\end{pgfscope}%
\end{pgfscope}%
\begin{pgfscope}%
\definecolor{textcolor}{rgb}{0.000000,0.000000,0.000000}%
\pgfsetstrokecolor{textcolor}%
\pgfsetfillcolor{textcolor}%
\pgftext[x=0.581165in, y=3.964667in, left, base]{\color{textcolor}\fontsize{30.000000}{24.000000}\selectfont \(\displaystyle {10^{6}}\)}%
\end{pgfscope}%
\begin{pgfscope}%
\pgfsetbuttcap%
\pgfsetroundjoin%
\definecolor{currentfill}{rgb}{0.000000,0.000000,0.000000}%
\pgfsetfillcolor{currentfill}%
\pgfsetlinewidth{0.602250pt}%
\definecolor{currentstroke}{rgb}{0.000000,0.000000,0.000000}%
\pgfsetstrokecolor{currentstroke}%
\pgfsetdash{}{0pt}%
\pgfsys@defobject{currentmarker}{\pgfqpoint{-0.027778in}{0.000000in}}{\pgfqpoint{-0.000000in}{0.000000in}}{%
\pgfpathmoveto{\pgfqpoint{-0.000000in}{0.000000in}}%
\pgfpathlineto{\pgfqpoint{-0.027778in}{0.000000in}}%
\pgfusepath{stroke,fill}%
}%
\begin{pgfscope}%
\pgfsys@transformshift{1.047462in}{1.208935in}%
\pgfsys@useobject{currentmarker}{}%
\end{pgfscope}%
\end{pgfscope}%
\begin{pgfscope}%
\pgfsetbuttcap%
\pgfsetroundjoin%
\definecolor{currentfill}{rgb}{0.000000,0.000000,0.000000}%
\pgfsetfillcolor{currentfill}%
\pgfsetlinewidth{0.602250pt}%
\definecolor{currentstroke}{rgb}{0.000000,0.000000,0.000000}%
\pgfsetstrokecolor{currentstroke}%
\pgfsetdash{}{0pt}%
\pgfsys@defobject{currentmarker}{\pgfqpoint{-0.027778in}{0.000000in}}{\pgfqpoint{-0.000000in}{0.000000in}}{%
\pgfpathmoveto{\pgfqpoint{-0.000000in}{0.000000in}}%
\pgfpathlineto{\pgfqpoint{-0.027778in}{0.000000in}}%
\pgfusepath{stroke,fill}%
}%
\begin{pgfscope}%
\pgfsys@transformshift{1.047462in}{1.395255in}%
\pgfsys@useobject{currentmarker}{}%
\end{pgfscope}%
\end{pgfscope}%
\begin{pgfscope}%
\pgfsetbuttcap%
\pgfsetroundjoin%
\definecolor{currentfill}{rgb}{0.000000,0.000000,0.000000}%
\pgfsetfillcolor{currentfill}%
\pgfsetlinewidth{0.602250pt}%
\definecolor{currentstroke}{rgb}{0.000000,0.000000,0.000000}%
\pgfsetstrokecolor{currentstroke}%
\pgfsetdash{}{0pt}%
\pgfsys@defobject{currentmarker}{\pgfqpoint{-0.027778in}{0.000000in}}{\pgfqpoint{-0.000000in}{0.000000in}}{%
\pgfpathmoveto{\pgfqpoint{-0.000000in}{0.000000in}}%
\pgfpathlineto{\pgfqpoint{-0.027778in}{0.000000in}}%
\pgfusepath{stroke,fill}%
}%
\begin{pgfscope}%
\pgfsys@transformshift{1.047462in}{1.527452in}%
\pgfsys@useobject{currentmarker}{}%
\end{pgfscope}%
\end{pgfscope}%
\begin{pgfscope}%
\pgfsetbuttcap%
\pgfsetroundjoin%
\definecolor{currentfill}{rgb}{0.000000,0.000000,0.000000}%
\pgfsetfillcolor{currentfill}%
\pgfsetlinewidth{0.602250pt}%
\definecolor{currentstroke}{rgb}{0.000000,0.000000,0.000000}%
\pgfsetstrokecolor{currentstroke}%
\pgfsetdash{}{0pt}%
\pgfsys@defobject{currentmarker}{\pgfqpoint{-0.027778in}{0.000000in}}{\pgfqpoint{-0.000000in}{0.000000in}}{%
\pgfpathmoveto{\pgfqpoint{-0.000000in}{0.000000in}}%
\pgfpathlineto{\pgfqpoint{-0.027778in}{0.000000in}}%
\pgfusepath{stroke,fill}%
}%
\begin{pgfscope}%
\pgfsys@transformshift{1.047462in}{1.629991in}%
\pgfsys@useobject{currentmarker}{}%
\end{pgfscope}%
\end{pgfscope}%
\begin{pgfscope}%
\pgfsetbuttcap%
\pgfsetroundjoin%
\definecolor{currentfill}{rgb}{0.000000,0.000000,0.000000}%
\pgfsetfillcolor{currentfill}%
\pgfsetlinewidth{0.602250pt}%
\definecolor{currentstroke}{rgb}{0.000000,0.000000,0.000000}%
\pgfsetstrokecolor{currentstroke}%
\pgfsetdash{}{0pt}%
\pgfsys@defobject{currentmarker}{\pgfqpoint{-0.027778in}{0.000000in}}{\pgfqpoint{-0.000000in}{0.000000in}}{%
\pgfpathmoveto{\pgfqpoint{-0.000000in}{0.000000in}}%
\pgfpathlineto{\pgfqpoint{-0.027778in}{0.000000in}}%
\pgfusepath{stroke,fill}%
}%
\begin{pgfscope}%
\pgfsys@transformshift{1.047462in}{1.713772in}%
\pgfsys@useobject{currentmarker}{}%
\end{pgfscope}%
\end{pgfscope}%
\begin{pgfscope}%
\pgfsetbuttcap%
\pgfsetroundjoin%
\definecolor{currentfill}{rgb}{0.000000,0.000000,0.000000}%
\pgfsetfillcolor{currentfill}%
\pgfsetlinewidth{0.602250pt}%
\definecolor{currentstroke}{rgb}{0.000000,0.000000,0.000000}%
\pgfsetstrokecolor{currentstroke}%
\pgfsetdash{}{0pt}%
\pgfsys@defobject{currentmarker}{\pgfqpoint{-0.027778in}{0.000000in}}{\pgfqpoint{-0.000000in}{0.000000in}}{%
\pgfpathmoveto{\pgfqpoint{-0.000000in}{0.000000in}}%
\pgfpathlineto{\pgfqpoint{-0.027778in}{0.000000in}}%
\pgfusepath{stroke,fill}%
}%
\begin{pgfscope}%
\pgfsys@transformshift{1.047462in}{1.784608in}%
\pgfsys@useobject{currentmarker}{}%
\end{pgfscope}%
\end{pgfscope}%
\begin{pgfscope}%
\pgfsetbuttcap%
\pgfsetroundjoin%
\definecolor{currentfill}{rgb}{0.000000,0.000000,0.000000}%
\pgfsetfillcolor{currentfill}%
\pgfsetlinewidth{0.602250pt}%
\definecolor{currentstroke}{rgb}{0.000000,0.000000,0.000000}%
\pgfsetstrokecolor{currentstroke}%
\pgfsetdash{}{0pt}%
\pgfsys@defobject{currentmarker}{\pgfqpoint{-0.027778in}{0.000000in}}{\pgfqpoint{-0.000000in}{0.000000in}}{%
\pgfpathmoveto{\pgfqpoint{-0.000000in}{0.000000in}}%
\pgfpathlineto{\pgfqpoint{-0.027778in}{0.000000in}}%
\pgfusepath{stroke,fill}%
}%
\begin{pgfscope}%
\pgfsys@transformshift{1.047462in}{1.845968in}%
\pgfsys@useobject{currentmarker}{}%
\end{pgfscope}%
\end{pgfscope}%
\begin{pgfscope}%
\pgfsetbuttcap%
\pgfsetroundjoin%
\definecolor{currentfill}{rgb}{0.000000,0.000000,0.000000}%
\pgfsetfillcolor{currentfill}%
\pgfsetlinewidth{0.602250pt}%
\definecolor{currentstroke}{rgb}{0.000000,0.000000,0.000000}%
\pgfsetstrokecolor{currentstroke}%
\pgfsetdash{}{0pt}%
\pgfsys@defobject{currentmarker}{\pgfqpoint{-0.027778in}{0.000000in}}{\pgfqpoint{-0.000000in}{0.000000in}}{%
\pgfpathmoveto{\pgfqpoint{-0.000000in}{0.000000in}}%
\pgfpathlineto{\pgfqpoint{-0.027778in}{0.000000in}}%
\pgfusepath{stroke,fill}%
}%
\begin{pgfscope}%
\pgfsys@transformshift{1.047462in}{1.900092in}%
\pgfsys@useobject{currentmarker}{}%
\end{pgfscope}%
\end{pgfscope}%
\begin{pgfscope}%
\pgfsetbuttcap%
\pgfsetroundjoin%
\definecolor{currentfill}{rgb}{0.000000,0.000000,0.000000}%
\pgfsetfillcolor{currentfill}%
\pgfsetlinewidth{0.602250pt}%
\definecolor{currentstroke}{rgb}{0.000000,0.000000,0.000000}%
\pgfsetstrokecolor{currentstroke}%
\pgfsetdash{}{0pt}%
\pgfsys@defobject{currentmarker}{\pgfqpoint{-0.027778in}{0.000000in}}{\pgfqpoint{-0.000000in}{0.000000in}}{%
\pgfpathmoveto{\pgfqpoint{-0.000000in}{0.000000in}}%
\pgfpathlineto{\pgfqpoint{-0.027778in}{0.000000in}}%
\pgfusepath{stroke,fill}%
}%
\begin{pgfscope}%
\pgfsys@transformshift{1.047462in}{2.267024in}%
\pgfsys@useobject{currentmarker}{}%
\end{pgfscope}%
\end{pgfscope}%
\begin{pgfscope}%
\pgfsetbuttcap%
\pgfsetroundjoin%
\definecolor{currentfill}{rgb}{0.000000,0.000000,0.000000}%
\pgfsetfillcolor{currentfill}%
\pgfsetlinewidth{0.602250pt}%
\definecolor{currentstroke}{rgb}{0.000000,0.000000,0.000000}%
\pgfsetstrokecolor{currentstroke}%
\pgfsetdash{}{0pt}%
\pgfsys@defobject{currentmarker}{\pgfqpoint{-0.027778in}{0.000000in}}{\pgfqpoint{-0.000000in}{0.000000in}}{%
\pgfpathmoveto{\pgfqpoint{-0.000000in}{0.000000in}}%
\pgfpathlineto{\pgfqpoint{-0.027778in}{0.000000in}}%
\pgfusepath{stroke,fill}%
}%
\begin{pgfscope}%
\pgfsys@transformshift{1.047462in}{2.453345in}%
\pgfsys@useobject{currentmarker}{}%
\end{pgfscope}%
\end{pgfscope}%
\begin{pgfscope}%
\pgfsetbuttcap%
\pgfsetroundjoin%
\definecolor{currentfill}{rgb}{0.000000,0.000000,0.000000}%
\pgfsetfillcolor{currentfill}%
\pgfsetlinewidth{0.602250pt}%
\definecolor{currentstroke}{rgb}{0.000000,0.000000,0.000000}%
\pgfsetstrokecolor{currentstroke}%
\pgfsetdash{}{0pt}%
\pgfsys@defobject{currentmarker}{\pgfqpoint{-0.027778in}{0.000000in}}{\pgfqpoint{-0.000000in}{0.000000in}}{%
\pgfpathmoveto{\pgfqpoint{-0.000000in}{0.000000in}}%
\pgfpathlineto{\pgfqpoint{-0.027778in}{0.000000in}}%
\pgfusepath{stroke,fill}%
}%
\begin{pgfscope}%
\pgfsys@transformshift{1.047462in}{2.585541in}%
\pgfsys@useobject{currentmarker}{}%
\end{pgfscope}%
\end{pgfscope}%
\begin{pgfscope}%
\pgfsetbuttcap%
\pgfsetroundjoin%
\definecolor{currentfill}{rgb}{0.000000,0.000000,0.000000}%
\pgfsetfillcolor{currentfill}%
\pgfsetlinewidth{0.602250pt}%
\definecolor{currentstroke}{rgb}{0.000000,0.000000,0.000000}%
\pgfsetstrokecolor{currentstroke}%
\pgfsetdash{}{0pt}%
\pgfsys@defobject{currentmarker}{\pgfqpoint{-0.027778in}{0.000000in}}{\pgfqpoint{-0.000000in}{0.000000in}}{%
\pgfpathmoveto{\pgfqpoint{-0.000000in}{0.000000in}}%
\pgfpathlineto{\pgfqpoint{-0.027778in}{0.000000in}}%
\pgfusepath{stroke,fill}%
}%
\begin{pgfscope}%
\pgfsys@transformshift{1.047462in}{2.688081in}%
\pgfsys@useobject{currentmarker}{}%
\end{pgfscope}%
\end{pgfscope}%
\begin{pgfscope}%
\pgfsetbuttcap%
\pgfsetroundjoin%
\definecolor{currentfill}{rgb}{0.000000,0.000000,0.000000}%
\pgfsetfillcolor{currentfill}%
\pgfsetlinewidth{0.602250pt}%
\definecolor{currentstroke}{rgb}{0.000000,0.000000,0.000000}%
\pgfsetstrokecolor{currentstroke}%
\pgfsetdash{}{0pt}%
\pgfsys@defobject{currentmarker}{\pgfqpoint{-0.027778in}{0.000000in}}{\pgfqpoint{-0.000000in}{0.000000in}}{%
\pgfpathmoveto{\pgfqpoint{-0.000000in}{0.000000in}}%
\pgfpathlineto{\pgfqpoint{-0.027778in}{0.000000in}}%
\pgfusepath{stroke,fill}%
}%
\begin{pgfscope}%
\pgfsys@transformshift{1.047462in}{2.771861in}%
\pgfsys@useobject{currentmarker}{}%
\end{pgfscope}%
\end{pgfscope}%
\begin{pgfscope}%
\pgfsetbuttcap%
\pgfsetroundjoin%
\definecolor{currentfill}{rgb}{0.000000,0.000000,0.000000}%
\pgfsetfillcolor{currentfill}%
\pgfsetlinewidth{0.602250pt}%
\definecolor{currentstroke}{rgb}{0.000000,0.000000,0.000000}%
\pgfsetstrokecolor{currentstroke}%
\pgfsetdash{}{0pt}%
\pgfsys@defobject{currentmarker}{\pgfqpoint{-0.027778in}{0.000000in}}{\pgfqpoint{-0.000000in}{0.000000in}}{%
\pgfpathmoveto{\pgfqpoint{-0.000000in}{0.000000in}}%
\pgfpathlineto{\pgfqpoint{-0.027778in}{0.000000in}}%
\pgfusepath{stroke,fill}%
}%
\begin{pgfscope}%
\pgfsys@transformshift{1.047462in}{2.842697in}%
\pgfsys@useobject{currentmarker}{}%
\end{pgfscope}%
\end{pgfscope}%
\begin{pgfscope}%
\pgfsetbuttcap%
\pgfsetroundjoin%
\definecolor{currentfill}{rgb}{0.000000,0.000000,0.000000}%
\pgfsetfillcolor{currentfill}%
\pgfsetlinewidth{0.602250pt}%
\definecolor{currentstroke}{rgb}{0.000000,0.000000,0.000000}%
\pgfsetstrokecolor{currentstroke}%
\pgfsetdash{}{0pt}%
\pgfsys@defobject{currentmarker}{\pgfqpoint{-0.027778in}{0.000000in}}{\pgfqpoint{-0.000000in}{0.000000in}}{%
\pgfpathmoveto{\pgfqpoint{-0.000000in}{0.000000in}}%
\pgfpathlineto{\pgfqpoint{-0.027778in}{0.000000in}}%
\pgfusepath{stroke,fill}%
}%
\begin{pgfscope}%
\pgfsys@transformshift{1.047462in}{2.904058in}%
\pgfsys@useobject{currentmarker}{}%
\end{pgfscope}%
\end{pgfscope}%
\begin{pgfscope}%
\pgfsetbuttcap%
\pgfsetroundjoin%
\definecolor{currentfill}{rgb}{0.000000,0.000000,0.000000}%
\pgfsetfillcolor{currentfill}%
\pgfsetlinewidth{0.602250pt}%
\definecolor{currentstroke}{rgb}{0.000000,0.000000,0.000000}%
\pgfsetstrokecolor{currentstroke}%
\pgfsetdash{}{0pt}%
\pgfsys@defobject{currentmarker}{\pgfqpoint{-0.027778in}{0.000000in}}{\pgfqpoint{-0.000000in}{0.000000in}}{%
\pgfpathmoveto{\pgfqpoint{-0.000000in}{0.000000in}}%
\pgfpathlineto{\pgfqpoint{-0.027778in}{0.000000in}}%
\pgfusepath{stroke,fill}%
}%
\begin{pgfscope}%
\pgfsys@transformshift{1.047462in}{2.958182in}%
\pgfsys@useobject{currentmarker}{}%
\end{pgfscope}%
\end{pgfscope}%
\begin{pgfscope}%
\pgfsetbuttcap%
\pgfsetroundjoin%
\definecolor{currentfill}{rgb}{0.000000,0.000000,0.000000}%
\pgfsetfillcolor{currentfill}%
\pgfsetlinewidth{0.602250pt}%
\definecolor{currentstroke}{rgb}{0.000000,0.000000,0.000000}%
\pgfsetstrokecolor{currentstroke}%
\pgfsetdash{}{0pt}%
\pgfsys@defobject{currentmarker}{\pgfqpoint{-0.027778in}{0.000000in}}{\pgfqpoint{-0.000000in}{0.000000in}}{%
\pgfpathmoveto{\pgfqpoint{-0.000000in}{0.000000in}}%
\pgfpathlineto{\pgfqpoint{-0.027778in}{0.000000in}}%
\pgfusepath{stroke,fill}%
}%
\begin{pgfscope}%
\pgfsys@transformshift{1.047462in}{3.325114in}%
\pgfsys@useobject{currentmarker}{}%
\end{pgfscope}%
\end{pgfscope}%
\begin{pgfscope}%
\pgfsetbuttcap%
\pgfsetroundjoin%
\definecolor{currentfill}{rgb}{0.000000,0.000000,0.000000}%
\pgfsetfillcolor{currentfill}%
\pgfsetlinewidth{0.602250pt}%
\definecolor{currentstroke}{rgb}{0.000000,0.000000,0.000000}%
\pgfsetstrokecolor{currentstroke}%
\pgfsetdash{}{0pt}%
\pgfsys@defobject{currentmarker}{\pgfqpoint{-0.027778in}{0.000000in}}{\pgfqpoint{-0.000000in}{0.000000in}}{%
\pgfpathmoveto{\pgfqpoint{-0.000000in}{0.000000in}}%
\pgfpathlineto{\pgfqpoint{-0.027778in}{0.000000in}}%
\pgfusepath{stroke,fill}%
}%
\begin{pgfscope}%
\pgfsys@transformshift{1.047462in}{3.511434in}%
\pgfsys@useobject{currentmarker}{}%
\end{pgfscope}%
\end{pgfscope}%
\begin{pgfscope}%
\pgfsetbuttcap%
\pgfsetroundjoin%
\definecolor{currentfill}{rgb}{0.000000,0.000000,0.000000}%
\pgfsetfillcolor{currentfill}%
\pgfsetlinewidth{0.602250pt}%
\definecolor{currentstroke}{rgb}{0.000000,0.000000,0.000000}%
\pgfsetstrokecolor{currentstroke}%
\pgfsetdash{}{0pt}%
\pgfsys@defobject{currentmarker}{\pgfqpoint{-0.027778in}{0.000000in}}{\pgfqpoint{-0.000000in}{0.000000in}}{%
\pgfpathmoveto{\pgfqpoint{-0.000000in}{0.000000in}}%
\pgfpathlineto{\pgfqpoint{-0.027778in}{0.000000in}}%
\pgfusepath{stroke,fill}%
}%
\begin{pgfscope}%
\pgfsys@transformshift{1.047462in}{3.643631in}%
\pgfsys@useobject{currentmarker}{}%
\end{pgfscope}%
\end{pgfscope}%
\begin{pgfscope}%
\pgfsetbuttcap%
\pgfsetroundjoin%
\definecolor{currentfill}{rgb}{0.000000,0.000000,0.000000}%
\pgfsetfillcolor{currentfill}%
\pgfsetlinewidth{0.602250pt}%
\definecolor{currentstroke}{rgb}{0.000000,0.000000,0.000000}%
\pgfsetstrokecolor{currentstroke}%
\pgfsetdash{}{0pt}%
\pgfsys@defobject{currentmarker}{\pgfqpoint{-0.027778in}{0.000000in}}{\pgfqpoint{-0.000000in}{0.000000in}}{%
\pgfpathmoveto{\pgfqpoint{-0.000000in}{0.000000in}}%
\pgfpathlineto{\pgfqpoint{-0.027778in}{0.000000in}}%
\pgfusepath{stroke,fill}%
}%
\begin{pgfscope}%
\pgfsys@transformshift{1.047462in}{3.746170in}%
\pgfsys@useobject{currentmarker}{}%
\end{pgfscope}%
\end{pgfscope}%
\begin{pgfscope}%
\pgfsetbuttcap%
\pgfsetroundjoin%
\definecolor{currentfill}{rgb}{0.000000,0.000000,0.000000}%
\pgfsetfillcolor{currentfill}%
\pgfsetlinewidth{0.602250pt}%
\definecolor{currentstroke}{rgb}{0.000000,0.000000,0.000000}%
\pgfsetstrokecolor{currentstroke}%
\pgfsetdash{}{0pt}%
\pgfsys@defobject{currentmarker}{\pgfqpoint{-0.027778in}{0.000000in}}{\pgfqpoint{-0.000000in}{0.000000in}}{%
\pgfpathmoveto{\pgfqpoint{-0.000000in}{0.000000in}}%
\pgfpathlineto{\pgfqpoint{-0.027778in}{0.000000in}}%
\pgfusepath{stroke,fill}%
}%
\begin{pgfscope}%
\pgfsys@transformshift{1.047462in}{3.829951in}%
\pgfsys@useobject{currentmarker}{}%
\end{pgfscope}%
\end{pgfscope}%
\begin{pgfscope}%
\pgfsetbuttcap%
\pgfsetroundjoin%
\definecolor{currentfill}{rgb}{0.000000,0.000000,0.000000}%
\pgfsetfillcolor{currentfill}%
\pgfsetlinewidth{0.602250pt}%
\definecolor{currentstroke}{rgb}{0.000000,0.000000,0.000000}%
\pgfsetstrokecolor{currentstroke}%
\pgfsetdash{}{0pt}%
\pgfsys@defobject{currentmarker}{\pgfqpoint{-0.027778in}{0.000000in}}{\pgfqpoint{-0.000000in}{0.000000in}}{%
\pgfpathmoveto{\pgfqpoint{-0.000000in}{0.000000in}}%
\pgfpathlineto{\pgfqpoint{-0.027778in}{0.000000in}}%
\pgfusepath{stroke,fill}%
}%
\begin{pgfscope}%
\pgfsys@transformshift{1.047462in}{3.900787in}%
\pgfsys@useobject{currentmarker}{}%
\end{pgfscope}%
\end{pgfscope}%
\begin{pgfscope}%
\pgfsetbuttcap%
\pgfsetroundjoin%
\definecolor{currentfill}{rgb}{0.000000,0.000000,0.000000}%
\pgfsetfillcolor{currentfill}%
\pgfsetlinewidth{0.602250pt}%
\definecolor{currentstroke}{rgb}{0.000000,0.000000,0.000000}%
\pgfsetstrokecolor{currentstroke}%
\pgfsetdash{}{0pt}%
\pgfsys@defobject{currentmarker}{\pgfqpoint{-0.027778in}{0.000000in}}{\pgfqpoint{-0.000000in}{0.000000in}}{%
\pgfpathmoveto{\pgfqpoint{-0.000000in}{0.000000in}}%
\pgfpathlineto{\pgfqpoint{-0.027778in}{0.000000in}}%
\pgfusepath{stroke,fill}%
}%
\begin{pgfscope}%
\pgfsys@transformshift{1.047462in}{3.962147in}%
\pgfsys@useobject{currentmarker}{}%
\end{pgfscope}%
\end{pgfscope}%
\begin{pgfscope}%
\pgfsetbuttcap%
\pgfsetroundjoin%
\definecolor{currentfill}{rgb}{0.000000,0.000000,0.000000}%
\pgfsetfillcolor{currentfill}%
\pgfsetlinewidth{0.602250pt}%
\definecolor{currentstroke}{rgb}{0.000000,0.000000,0.000000}%
\pgfsetstrokecolor{currentstroke}%
\pgfsetdash{}{0pt}%
\pgfsys@defobject{currentmarker}{\pgfqpoint{-0.027778in}{0.000000in}}{\pgfqpoint{-0.000000in}{0.000000in}}{%
\pgfpathmoveto{\pgfqpoint{-0.000000in}{0.000000in}}%
\pgfpathlineto{\pgfqpoint{-0.027778in}{0.000000in}}%
\pgfusepath{stroke,fill}%
}%
\begin{pgfscope}%
\pgfsys@transformshift{1.047462in}{4.016271in}%
\pgfsys@useobject{currentmarker}{}%
\end{pgfscope}%
\end{pgfscope}%
\begin{pgfscope}%
\pgfsetbuttcap%
\pgfsetroundjoin%
\definecolor{currentfill}{rgb}{0.000000,0.000000,0.000000}%
\pgfsetfillcolor{currentfill}%
\pgfsetlinewidth{0.602250pt}%
\definecolor{currentstroke}{rgb}{0.000000,0.000000,0.000000}%
\pgfsetstrokecolor{currentstroke}%
\pgfsetdash{}{0pt}%
\pgfsys@defobject{currentmarker}{\pgfqpoint{-0.027778in}{0.000000in}}{\pgfqpoint{-0.000000in}{0.000000in}}{%
\pgfpathmoveto{\pgfqpoint{-0.000000in}{0.000000in}}%
\pgfpathlineto{\pgfqpoint{-0.027778in}{0.000000in}}%
\pgfusepath{stroke,fill}%
}%
\begin{pgfscope}%
\pgfsys@transformshift{1.047462in}{4.383203in}%
\pgfsys@useobject{currentmarker}{}%
\end{pgfscope}%
\end{pgfscope}%
\begin{pgfscope}%
\definecolor{textcolor}{rgb}{0.000000,0.000000,0.000000}%
\pgfsetstrokecolor{textcolor}%
\pgfsetfillcolor{textcolor}%
\pgftext[x=0.525610in,y=2.722398in,,bottom,rotate=90.000000]{\color{textcolor}\fontsize{30.000000}{24.000000}\selectfont Number of operations}%
\end{pgfscope}%
\begin{pgfscope}%
\pgfpathrectangle{\pgfqpoint{1.047462in}{0.944795in}}{\pgfqpoint{5.052538in}{3.555205in}}%
\pgfusepath{clip}%
\pgfsetrectcap%
\pgfsetroundjoin%
\pgfsetlinewidth{1.505625pt}%
\definecolor{currentstroke}{rgb}{0.121569,0.466667,0.705882}%
\pgfsetstrokecolor{currentstroke}%
\pgfsetdash{}{0pt}%
\pgfpathmoveto{\pgfqpoint{1.277123in}{1.342221in}}%
\pgfpathlineto{\pgfqpoint{2.085948in}{1.779126in}}%
\pgfpathlineto{\pgfqpoint{2.659819in}{2.101480in}}%
\pgfpathlineto{\pgfqpoint{3.104948in}{2.367667in}}%
\pgfpathlineto{\pgfqpoint{3.468644in}{2.595204in}}%
\pgfpathlineto{\pgfqpoint{3.776145in}{2.794493in}}%
\pgfpathlineto{\pgfqpoint{4.042515in}{2.972171in}}%
\pgfpathlineto{\pgfqpoint{4.277469in}{3.132728in}}%
\pgfpathlineto{\pgfqpoint{4.487643in}{3.279351in}}%
\pgfpathlineto{\pgfqpoint{4.677769in}{3.414386in}}%
\pgfpathlineto{\pgfqpoint{4.851340in}{3.539615in}}%
\pgfpathlineto{\pgfqpoint{5.011010in}{3.656429in}}%
\pgfpathlineto{\pgfqpoint{5.158841in}{3.765930in}}%
\pgfpathlineto{\pgfqpoint{5.296469in}{3.869015in}}%
\pgfpathlineto{\pgfqpoint{5.425211in}{3.966418in}}%
\pgfpathlineto{\pgfqpoint{5.546145in}{4.058752in}}%
\pgfpathlineto{\pgfqpoint{5.660165in}{4.146534in}}%
\pgfpathlineto{\pgfqpoint{5.768019in}{4.230203in}}%
\pgfpathlineto{\pgfqpoint{5.870339in}{4.310136in}}%
\pgfusepath{stroke}%
\end{pgfscope}%
\begin{pgfscope}%
\pgfpathrectangle{\pgfqpoint{1.047462in}{0.944795in}}{\pgfqpoint{5.052538in}{3.555205in}}%
\pgfusepath{clip}%
\pgfsetrectcap%
\pgfsetroundjoin%
\pgfsetlinewidth{1.505625pt}%
\definecolor{currentstroke}{rgb}{1.000000,0.498039,0.054902}%
\pgfsetstrokecolor{currentstroke}%
\pgfsetdash{}{0pt}%
\pgfpathmoveto{\pgfqpoint{1.277123in}{1.466222in}}%
\pgfpathlineto{\pgfqpoint{2.085948in}{1.960571in}}%
\pgfpathlineto{\pgfqpoint{2.659819in}{2.307383in}}%
\pgfpathlineto{\pgfqpoint{3.104948in}{2.579167in}}%
\pgfpathlineto{\pgfqpoint{3.468644in}{2.803381in}}%
\pgfpathlineto{\pgfqpoint{3.776145in}{2.994605in}}%
\pgfpathlineto{\pgfqpoint{4.042515in}{3.161529in}}%
\pgfpathlineto{\pgfqpoint{4.277469in}{3.309767in}}%
\pgfpathlineto{\pgfqpoint{4.487643in}{3.443167in}}%
\pgfpathlineto{\pgfqpoint{4.677769in}{3.564481in}}%
\pgfpathlineto{\pgfqpoint{4.851340in}{3.675755in}}%
\pgfpathlineto{\pgfqpoint{5.011010in}{3.778548in}}%
\pgfpathlineto{\pgfqpoint{5.158841in}{3.874079in}}%
\pgfpathlineto{\pgfqpoint{5.296469in}{3.963320in}}%
\pgfpathlineto{\pgfqpoint{5.425211in}{4.047056in}}%
\pgfpathlineto{\pgfqpoint{5.546145in}{4.125936in}}%
\pgfpathlineto{\pgfqpoint{5.660165in}{4.200495in}}%
\pgfpathlineto{\pgfqpoint{5.768019in}{4.271189in}}%
\pgfpathlineto{\pgfqpoint{5.870339in}{4.338400in}}%
\pgfusepath{stroke}%
\end{pgfscope}%
\begin{pgfscope}%
\pgfpathrectangle{\pgfqpoint{1.047462in}{0.944795in}}{\pgfqpoint{5.052538in}{3.555205in}}%
\pgfusepath{clip}%
\pgfsetrectcap%
\pgfsetroundjoin%
\pgfsetlinewidth{1.505625pt}%
\definecolor{currentstroke}{rgb}{0.172549,0.627451,0.172549}%
\pgfsetstrokecolor{currentstroke}%
\pgfsetdash{}{0pt}%
\pgfpathmoveto{\pgfqpoint{1.277123in}{1.245787in}}%
\pgfpathlineto{\pgfqpoint{2.085948in}{1.639901in}}%
\pgfpathlineto{\pgfqpoint{2.659819in}{1.906682in}}%
\pgfpathlineto{\pgfqpoint{3.104948in}{2.116612in}}%
\pgfpathlineto{\pgfqpoint{3.468644in}{2.288788in}}%
\pgfpathlineto{\pgfqpoint{3.776145in}{2.434411in}}%
\pgfpathlineto{\pgfqpoint{4.042515in}{2.560444in}}%
\pgfpathlineto{\pgfqpoint{4.277469in}{2.671471in}}%
\pgfpathlineto{\pgfqpoint{4.487643in}{2.770650in}}%
\pgfpathlineto{\pgfqpoint{4.677769in}{2.860246in}}%
\pgfpathlineto{\pgfqpoint{4.851340in}{2.941937in}}%
\pgfpathlineto{\pgfqpoint{5.011010in}{3.016997in}}%
\pgfpathlineto{\pgfqpoint{5.158841in}{3.086417in}}%
\pgfpathlineto{\pgfqpoint{5.296469in}{3.150981in}}%
\pgfpathlineto{\pgfqpoint{5.425211in}{3.211323in}}%
\pgfpathlineto{\pgfqpoint{5.546145in}{3.267959in}}%
\pgfpathlineto{\pgfqpoint{5.660165in}{3.321316in}}%
\pgfpathlineto{\pgfqpoint{5.768019in}{3.371753in}}%
\pgfpathlineto{\pgfqpoint{5.870339in}{3.419572in}}%
\pgfusepath{stroke}%
\end{pgfscope}%
\begin{pgfscope}%
\pgfpathrectangle{\pgfqpoint{1.047462in}{0.944795in}}{\pgfqpoint{5.052538in}{3.555205in}}%
\pgfusepath{clip}%
\pgfsetrectcap%
\pgfsetroundjoin%
\pgfsetlinewidth{1.505625pt}%
\definecolor{currentstroke}{rgb}{0.839216,0.152941,0.156863}%
\pgfsetstrokecolor{currentstroke}%
\pgfsetdash{}{0pt}%
\pgfpathmoveto{\pgfqpoint{1.277123in}{1.106396in}}%
\pgfpathlineto{\pgfqpoint{2.085948in}{1.479036in}}%
\pgfpathlineto{\pgfqpoint{2.659819in}{1.743429in}}%
\pgfpathlineto{\pgfqpoint{3.104948in}{1.948508in}}%
\pgfpathlineto{\pgfqpoint{3.468644in}{2.116069in}}%
\pgfpathlineto{\pgfqpoint{3.776145in}{2.257741in}}%
\pgfpathlineto{\pgfqpoint{4.042515in}{2.380462in}}%
\pgfpathlineto{\pgfqpoint{4.277469in}{2.488710in}}%
\pgfpathlineto{\pgfqpoint{4.487643in}{2.585541in}}%
\pgfpathlineto{\pgfqpoint{4.677769in}{2.673135in}}%
\pgfpathlineto{\pgfqpoint{4.851340in}{2.753103in}}%
\pgfpathlineto{\pgfqpoint{5.011010in}{2.826666in}}%
\pgfpathlineto{\pgfqpoint{5.158841in}{2.894774in}}%
\pgfpathlineto{\pgfqpoint{5.296469in}{2.958182in}}%
\pgfpathlineto{\pgfqpoint{5.425211in}{3.017496in}}%
\pgfpathlineto{\pgfqpoint{5.546145in}{3.073212in}}%
\pgfpathlineto{\pgfqpoint{5.660165in}{3.125743in}}%
\pgfpathlineto{\pgfqpoint{5.768019in}{3.175434in}}%
\pgfpathlineto{\pgfqpoint{5.870339in}{3.222574in}}%
\pgfusepath{stroke}%
\end{pgfscope}%
\begin{pgfscope}%
\pgfsetrectcap%
\pgfsetmiterjoin%
\pgfsetlinewidth{0.803000pt}%
\definecolor{currentstroke}{rgb}{0.000000,0.000000,0.000000}%
\pgfsetstrokecolor{currentstroke}%
\pgfsetdash{}{0pt}%
\pgfpathmoveto{\pgfqpoint{1.047462in}{0.944795in}}%
\pgfpathlineto{\pgfqpoint{1.047462in}{4.500000in}}%
\pgfusepath{stroke}%
\end{pgfscope}%
\begin{pgfscope}%
\pgfsetrectcap%
\pgfsetmiterjoin%
\pgfsetlinewidth{0.803000pt}%
\definecolor{currentstroke}{rgb}{0.000000,0.000000,0.000000}%
\pgfsetstrokecolor{currentstroke}%
\pgfsetdash{}{0pt}%
\pgfpathmoveto{\pgfqpoint{6.100000in}{0.944795in}}%
\pgfpathlineto{\pgfqpoint{6.100000in}{4.500000in}}%
\pgfusepath{stroke}%
\end{pgfscope}%
\begin{pgfscope}%
\pgfsetrectcap%
\pgfsetmiterjoin%
\pgfsetlinewidth{0.803000pt}%
\definecolor{currentstroke}{rgb}{0.000000,0.000000,0.000000}%
\pgfsetstrokecolor{currentstroke}%
\pgfsetdash{}{0pt}%
\pgfpathmoveto{\pgfqpoint{1.047462in}{0.944795in}}%
\pgfpathlineto{\pgfqpoint{6.100000in}{0.944795in}}%
\pgfusepath{stroke}%
\end{pgfscope}%
\begin{pgfscope}%
\pgfsetrectcap%
\pgfsetmiterjoin%
\pgfsetlinewidth{0.803000pt}%
\definecolor{currentstroke}{rgb}{0.000000,0.000000,0.000000}%
\pgfsetstrokecolor{currentstroke}%
\pgfsetdash{}{0pt}%
\pgfpathmoveto{\pgfqpoint{1.047462in}{4.500000in}}%
\pgfpathlineto{\pgfqpoint{6.100000in}{4.500000in}}%
\pgfusepath{stroke}%
\end{pgfscope}%
\begin{pgfscope}%
\pgfsetbuttcap%
\pgfsetmiterjoin%
\definecolor{currentfill}{rgb}{1.000000,1.000000,1.000000}%
\pgfsetfillcolor{currentfill}%
\pgfsetfillopacity{0.800000}%
\pgfsetlinewidth{1.003750pt}%
\definecolor{currentstroke}{rgb}{0.800000,0.800000,0.800000}%
\pgfsetstrokecolor{currentstroke}%
\pgfsetstrokeopacity{0.800000}%
\pgfsetdash{}{0pt}%
\pgfpathmoveto{\pgfqpoint{1.241907in}{2.697952in}}%
\pgfpathlineto{\pgfqpoint{3.64176in}{2.697952in}}%
\pgfpathquadraticcurveto{\pgfqpoint{3.609732in}{2.697952in}}{\pgfqpoint{3.609732in}{2.753507in}}%
\pgfpathlineto{\pgfqpoint{3.609732in}{4.305556in}}%
\pgfpathquadraticcurveto{\pgfqpoint{3.609732in}{4.361111in}}{\pgfqpoint{3.60176in}{4.361111in}}%
\pgfpathlineto{\pgfqpoint{1.241907in}{4.361111in}}%
\pgfpathquadraticcurveto{\pgfqpoint{1.186351in}{4.361111in}}{\pgfqpoint{1.186351in}{4.305556in}}%
\pgfpathlineto{\pgfqpoint{1.186351in}{2.753507in}}%
\pgfpathquadraticcurveto{\pgfqpoint{1.186351in}{2.697952in}}{\pgfqpoint{1.241907in}{2.697952in}}%
\pgfpathclose%
\pgfusepath{stroke,fill}%
\end{pgfscope}%
\begin{pgfscope}%
\pgfsetrectcap%
\pgfsetroundjoin%
\pgfsetlinewidth{1.505625pt}%
\definecolor{currentstroke}{rgb}{0.121569,0.466667,0.705882}%
\pgfsetstrokecolor{currentstroke}%
\pgfsetdash{}{0pt}%
\pgfpathmoveto{\pgfqpoint{1.297462in}{4.147184in}}%
\pgfpathlineto{\pgfqpoint{1.853018in}{4.147184in}}%
\pgfusepath{stroke}%
\end{pgfscope}%
\begin{pgfscope}%
\definecolor{textcolor}{rgb}{0.000000,0.000000,0.000000}%
\pgfsetstrokecolor{textcolor}%
\pgfsetfillcolor{textcolor}%
\pgftext[x=2.075240in,y=4.049962in,left,base]{\color{textcolor}\fontsize{25.000000}{24.000000}\selectfont PV-OSIM}%
\end{pgfscope}%
\begin{pgfscope}%
\pgfsetrectcap%
\pgfsetroundjoin%
\pgfsetlinewidth{1.505625pt}%
\definecolor{currentstroke}{rgb}{1.000000,0.498039,0.054902}%
\pgfsetstrokecolor{currentstroke}%
\pgfsetdash{}{0pt}%
\pgfpathmoveto{\pgfqpoint{1.297462in}{3.752227in}}%
\pgfpathlineto{\pgfqpoint{1.853018in}{3.752227in}}%
\pgfusepath{stroke}%
\end{pgfscope}%
\begin{pgfscope}%
\definecolor{textcolor}{rgb}{0.000000,0.000000,0.000000}%
\pgfsetstrokecolor{textcolor}%
\pgfsetfillcolor{textcolor}%
\pgftext[x=2.075240in,y=3.655005in,left,base]{\color{textcolor}\fontsize{25.000000}{24.000000}\selectfont EFPA}%
\end{pgfscope}%
\begin{pgfscope}%
\pgfsetrectcap%
\pgfsetroundjoin%
\pgfsetlinewidth{1.505625pt}%
\definecolor{currentstroke}{rgb}{0.172549,0.627451,0.172549}%
\pgfsetstrokecolor{currentstroke}%
\pgfsetdash{}{0pt}%
\pgfpathmoveto{\pgfqpoint{1.297462in}{3.357271in}}%
\pgfpathlineto{\pgfqpoint{1.853018in}{3.357271in}}%
\pgfusepath{stroke}%
\end{pgfscope}%
\begin{pgfscope}%
\definecolor{textcolor}{rgb}{0.000000,0.000000,0.000000}%
\pgfsetstrokecolor{textcolor}%
\pgfsetfillcolor{textcolor}%
\pgftext[x=2.075240in,y=3.260048in,left,base]{\color{textcolor}\fontsize{24.000000}{24.000000}\selectfont PV-OSIMr}%
\end{pgfscope}%
\begin{pgfscope}%
\pgfsetrectcap%
\pgfsetroundjoin%
\pgfsetlinewidth{1.505625pt}%
\definecolor{currentstroke}{rgb}{0.839216,0.152941,0.156863}%
\pgfsetstrokecolor{currentstroke}%
\pgfsetdash{}{0pt}%
\pgfpathmoveto{\pgfqpoint{1.297462in}{2.962314in}}%
\pgfpathlineto{\pgfqpoint{1.853018in}{2.962314in}}%
\pgfusepath{stroke}%
\end{pgfscope}%
\begin{pgfscope}%
\definecolor{textcolor}{rgb}{0.000000,0.000000,0.000000}%
\pgfsetstrokecolor{textcolor}%
\pgfsetfillcolor{textcolor}%
\pgftext[x=2.075240in,y=2.865092in,left,base]{\color{textcolor}\fontsize{24.000000}{24.000000}\selectfont 400\(\displaystyle k^2\)}%
\end{pgfscope}%
\end{pgfpicture}%
\makeatother%
\endgroup%

%% file: graphics/chain_all_con.pgf
%% Creator: Matplotlib, PGF backend
%%
%% To include the figure in your LaTeX document, write
%%   \input{<filename>.pgf}
%%
%% Make sure the required packages are loaded in your preamble
%%   \usepackage{pgf}
%%
%% Figures using additional raster images can only be included by \input if
%% they are in the same directory as the main LaTeX file. For loading figures
%% from other directories you can use the `import` package
%% and then include the figures with
%%   \import{<path to file>}{<filename>.pgf}
%%
%% Matplotlib used the following preamble
%%
\begingroup%
\makeatletter%
\begin{pgfpicture}%
\pgfpathrectangle{\pgfpointorigin}{\pgfqpoint{6.400000in}{4.800000in}}%
\pgfusepath{use as bounding box, clip}%
\begin{pgfscope}%
\pgfsetbuttcap%
\pgfsetmiterjoin%
\definecolor{currentfill}{rgb}{1.000000,1.000000,1.000000}%
\pgfsetfillcolor{currentfill}%
\pgfsetlinewidth{0.000000pt}%
\definecolor{currentstroke}{rgb}{1.000000,1.000000,1.000000}%
\pgfsetstrokecolor{currentstroke}%
\pgfsetdash{}{0pt}%
\pgfpathmoveto{\pgfqpoint{0.000000in}{0.000000in}}%
\pgfpathlineto{\pgfqpoint{6.400000in}{0.000000in}}%
\pgfpathlineto{\pgfqpoint{6.400000in}{4.800000in}}%
\pgfpathlineto{\pgfqpoint{0.000000in}{4.800000in}}%
\pgfpathclose%
\pgfusepath{fill}%
\end{pgfscope}%
\begin{pgfscope}%
\pgfsetbuttcap%
\pgfsetmiterjoin%
\definecolor{currentfill}{rgb}{1.000000,1.000000,1.000000}%
\pgfsetfillcolor{currentfill}%
\pgfsetlinewidth{0.000000pt}%
\definecolor{currentstroke}{rgb}{0.000000,0.000000,0.000000}%
\pgfsetstrokecolor{currentstroke}%
\pgfsetstrokeopacity{0.000000}%
\pgfsetdash{}{0pt}%
\pgfpathmoveto{\pgfqpoint{1.047462in}{0.944795in}}%
\pgfpathlineto{\pgfqpoint{6.100000in}{0.944795in}}%
\pgfpathlineto{\pgfqpoint{6.100000in}{4.500000in}}%
\pgfpathlineto{\pgfqpoint{1.047462in}{4.500000in}}%
\pgfpathclose%
\pgfusepath{fill}%
\end{pgfscope}%
\begin{pgfscope}%
\pgfsetbuttcap%
\pgfsetroundjoin%
\definecolor{currentfill}{rgb}{0.000000,0.000000,0.000000}%
\pgfsetfillcolor{currentfill}%
\pgfsetlinewidth{0.803000pt}%
\definecolor{currentstroke}{rgb}{0.000000,0.000000,0.000000}%
\pgfsetstrokecolor{currentstroke}%
\pgfsetdash{}{0pt}%
\pgfsys@defobject{currentmarker}{\pgfqpoint{0.000000in}{-0.048611in}}{\pgfqpoint{0.000000in}{0.000000in}}{%
\pgfpathmoveto{\pgfqpoint{0.000000in}{0.000000in}}%
\pgfpathlineto{\pgfqpoint{0.000000in}{-0.048611in}}%
\pgfusepath{stroke,fill}%
}%
\begin{pgfscope}%
\pgfsys@transformshift{3.859924in}{0.944795in}%
\pgfsys@useobject{currentmarker}{}%
\end{pgfscope}%
\end{pgfscope}%
\begin{pgfscope}%
\definecolor{textcolor}{rgb}{0.000000,0.000000,0.000000}%
\pgfsetstrokecolor{textcolor}%
\pgfsetfillcolor{textcolor}%
\pgftext[x=3.859924in,y=0.847573in,,top]{\color{textcolor}\fontsize{25.000000}{24.000000}\selectfont \(\displaystyle {10^{1}}\)}%
\end{pgfscope}%
\begin{pgfscope}%
\pgfsetbuttcap%
\pgfsetroundjoin%
\definecolor{currentfill}{rgb}{0.000000,0.000000,0.000000}%
\pgfsetfillcolor{currentfill}%
\pgfsetlinewidth{0.602250pt}%
\definecolor{currentstroke}{rgb}{0.000000,0.000000,0.000000}%
\pgfsetstrokecolor{currentstroke}%
\pgfsetdash{}{0pt}%
\pgfsys@defobject{currentmarker}{\pgfqpoint{0.000000in}{-0.027778in}}{\pgfqpoint{0.000000in}{0.000000in}}{%
\pgfpathmoveto{\pgfqpoint{0.000000in}{0.000000in}}%
\pgfpathlineto{\pgfqpoint{0.000000in}{-0.027778in}}%
\pgfusepath{stroke,fill}%
}%
\begin{pgfscope}%
\pgfsys@transformshift{1.277123in}{0.944795in}%
\pgfsys@useobject{currentmarker}{}%
\end{pgfscope}%
\end{pgfscope}%
\begin{pgfscope}%
\pgfsetbuttcap%
\pgfsetroundjoin%
\definecolor{currentfill}{rgb}{0.000000,0.000000,0.000000}%
\pgfsetfillcolor{currentfill}%
\pgfsetlinewidth{0.602250pt}%
\definecolor{currentstroke}{rgb}{0.000000,0.000000,0.000000}%
\pgfsetstrokecolor{currentstroke}%
\pgfsetdash{}{0pt}%
\pgfsys@defobject{currentmarker}{\pgfqpoint{0.000000in}{-0.027778in}}{\pgfqpoint{0.000000in}{0.000000in}}{%
\pgfpathmoveto{\pgfqpoint{0.000000in}{0.000000in}}%
\pgfpathlineto{\pgfqpoint{0.000000in}{-0.027778in}}%
\pgfusepath{stroke,fill}%
}%
\begin{pgfscope}%
\pgfsys@transformshift{1.927807in}{0.944795in}%
\pgfsys@useobject{currentmarker}{}%
\end{pgfscope}%
\end{pgfscope}%
\begin{pgfscope}%
\pgfsetbuttcap%
\pgfsetroundjoin%
\definecolor{currentfill}{rgb}{0.000000,0.000000,0.000000}%
\pgfsetfillcolor{currentfill}%
\pgfsetlinewidth{0.602250pt}%
\definecolor{currentstroke}{rgb}{0.000000,0.000000,0.000000}%
\pgfsetstrokecolor{currentstroke}%
\pgfsetdash{}{0pt}%
\pgfsys@defobject{currentmarker}{\pgfqpoint{0.000000in}{-0.027778in}}{\pgfqpoint{0.000000in}{0.000000in}}{%
\pgfpathmoveto{\pgfqpoint{0.000000in}{0.000000in}}%
\pgfpathlineto{\pgfqpoint{0.000000in}{-0.027778in}}%
\pgfusepath{stroke,fill}%
}%
\begin{pgfscope}%
\pgfsys@transformshift{2.389475in}{0.944795in}%
\pgfsys@useobject{currentmarker}{}%
\end{pgfscope}%
\end{pgfscope}%
\begin{pgfscope}%
\pgfsetbuttcap%
\pgfsetroundjoin%
\definecolor{currentfill}{rgb}{0.000000,0.000000,0.000000}%
\pgfsetfillcolor{currentfill}%
\pgfsetlinewidth{0.602250pt}%
\definecolor{currentstroke}{rgb}{0.000000,0.000000,0.000000}%
\pgfsetstrokecolor{currentstroke}%
\pgfsetdash{}{0pt}%
\pgfsys@defobject{currentmarker}{\pgfqpoint{0.000000in}{-0.027778in}}{\pgfqpoint{0.000000in}{0.000000in}}{%
\pgfpathmoveto{\pgfqpoint{0.000000in}{0.000000in}}%
\pgfpathlineto{\pgfqpoint{0.000000in}{-0.027778in}}%
\pgfusepath{stroke,fill}%
}%
\begin{pgfscope}%
\pgfsys@transformshift{2.747572in}{0.944795in}%
\pgfsys@useobject{currentmarker}{}%
\end{pgfscope}%
\end{pgfscope}%
\begin{pgfscope}%
\pgfsetbuttcap%
\pgfsetroundjoin%
\definecolor{currentfill}{rgb}{0.000000,0.000000,0.000000}%
\pgfsetfillcolor{currentfill}%
\pgfsetlinewidth{0.602250pt}%
\definecolor{currentstroke}{rgb}{0.000000,0.000000,0.000000}%
\pgfsetstrokecolor{currentstroke}%
\pgfsetdash{}{0pt}%
\pgfsys@defobject{currentmarker}{\pgfqpoint{0.000000in}{-0.027778in}}{\pgfqpoint{0.000000in}{0.000000in}}{%
\pgfpathmoveto{\pgfqpoint{0.000000in}{0.000000in}}%
\pgfpathlineto{\pgfqpoint{0.000000in}{-0.027778in}}%
\pgfusepath{stroke,fill}%
}%
\begin{pgfscope}%
\pgfsys@transformshift{3.040159in}{0.944795in}%
\pgfsys@useobject{currentmarker}{}%
\end{pgfscope}%
\end{pgfscope}%
\begin{pgfscope}%
\pgfsetbuttcap%
\pgfsetroundjoin%
\definecolor{currentfill}{rgb}{0.000000,0.000000,0.000000}%
\pgfsetfillcolor{currentfill}%
\pgfsetlinewidth{0.602250pt}%
\definecolor{currentstroke}{rgb}{0.000000,0.000000,0.000000}%
\pgfsetstrokecolor{currentstroke}%
\pgfsetdash{}{0pt}%
\pgfsys@defobject{currentmarker}{\pgfqpoint{0.000000in}{-0.027778in}}{\pgfqpoint{0.000000in}{0.000000in}}{%
\pgfpathmoveto{\pgfqpoint{0.000000in}{0.000000in}}%
\pgfpathlineto{\pgfqpoint{0.000000in}{-0.027778in}}%
\pgfusepath{stroke,fill}%
}%
\begin{pgfscope}%
\pgfsys@transformshift{3.287538in}{0.944795in}%
\pgfsys@useobject{currentmarker}{}%
\end{pgfscope}%
\end{pgfscope}%
\begin{pgfscope}%
\pgfsetbuttcap%
\pgfsetroundjoin%
\definecolor{currentfill}{rgb}{0.000000,0.000000,0.000000}%
\pgfsetfillcolor{currentfill}%
\pgfsetlinewidth{0.602250pt}%
\definecolor{currentstroke}{rgb}{0.000000,0.000000,0.000000}%
\pgfsetstrokecolor{currentstroke}%
\pgfsetdash{}{0pt}%
\pgfsys@defobject{currentmarker}{\pgfqpoint{0.000000in}{-0.027778in}}{\pgfqpoint{0.000000in}{0.000000in}}{%
\pgfpathmoveto{\pgfqpoint{0.000000in}{0.000000in}}%
\pgfpathlineto{\pgfqpoint{0.000000in}{-0.027778in}}%
\pgfusepath{stroke,fill}%
}%
\begin{pgfscope}%
\pgfsys@transformshift{3.501827in}{0.944795in}%
\pgfsys@useobject{currentmarker}{}%
\end{pgfscope}%
\end{pgfscope}%
\begin{pgfscope}%
\pgfsetbuttcap%
\pgfsetroundjoin%
\definecolor{currentfill}{rgb}{0.000000,0.000000,0.000000}%
\pgfsetfillcolor{currentfill}%
\pgfsetlinewidth{0.602250pt}%
\definecolor{currentstroke}{rgb}{0.000000,0.000000,0.000000}%
\pgfsetstrokecolor{currentstroke}%
\pgfsetdash{}{0pt}%
\pgfsys@defobject{currentmarker}{\pgfqpoint{0.000000in}{-0.027778in}}{\pgfqpoint{0.000000in}{0.000000in}}{%
\pgfpathmoveto{\pgfqpoint{0.000000in}{0.000000in}}%
\pgfpathlineto{\pgfqpoint{0.000000in}{-0.027778in}}%
\pgfusepath{stroke,fill}%
}%
\begin{pgfscope}%
\pgfsys@transformshift{3.690843in}{0.944795in}%
\pgfsys@useobject{currentmarker}{}%
\end{pgfscope}%
\end{pgfscope}%
\begin{pgfscope}%
\pgfsetbuttcap%
\pgfsetroundjoin%
\definecolor{currentfill}{rgb}{0.000000,0.000000,0.000000}%
\pgfsetfillcolor{currentfill}%
\pgfsetlinewidth{0.602250pt}%
\definecolor{currentstroke}{rgb}{0.000000,0.000000,0.000000}%
\pgfsetstrokecolor{currentstroke}%
\pgfsetdash{}{0pt}%
\pgfsys@defobject{currentmarker}{\pgfqpoint{0.000000in}{-0.027778in}}{\pgfqpoint{0.000000in}{0.000000in}}{%
\pgfpathmoveto{\pgfqpoint{0.000000in}{0.000000in}}%
\pgfpathlineto{\pgfqpoint{0.000000in}{-0.027778in}}%
\pgfusepath{stroke,fill}%
}%
\begin{pgfscope}%
\pgfsys@transformshift{4.972276in}{0.944795in}%
\pgfsys@useobject{currentmarker}{}%
\end{pgfscope}%
\end{pgfscope}%
\begin{pgfscope}%
\pgfsetbuttcap%
\pgfsetroundjoin%
\definecolor{currentfill}{rgb}{0.000000,0.000000,0.000000}%
\pgfsetfillcolor{currentfill}%
\pgfsetlinewidth{0.602250pt}%
\definecolor{currentstroke}{rgb}{0.000000,0.000000,0.000000}%
\pgfsetstrokecolor{currentstroke}%
\pgfsetdash{}{0pt}%
\pgfsys@defobject{currentmarker}{\pgfqpoint{0.000000in}{-0.027778in}}{\pgfqpoint{0.000000in}{0.000000in}}{%
\pgfpathmoveto{\pgfqpoint{0.000000in}{0.000000in}}%
\pgfpathlineto{\pgfqpoint{0.000000in}{-0.027778in}}%
\pgfusepath{stroke,fill}%
}%
\begin{pgfscope}%
\pgfsys@transformshift{5.622961in}{0.944795in}%
\pgfsys@useobject{currentmarker}{}%
\end{pgfscope}%
\end{pgfscope}%
\begin{pgfscope}%
\pgfsetbuttcap%
\pgfsetroundjoin%
\definecolor{currentfill}{rgb}{0.000000,0.000000,0.000000}%
\pgfsetfillcolor{currentfill}%
\pgfsetlinewidth{0.602250pt}%
\definecolor{currentstroke}{rgb}{0.000000,0.000000,0.000000}%
\pgfsetstrokecolor{currentstroke}%
\pgfsetdash{}{0pt}%
\pgfsys@defobject{currentmarker}{\pgfqpoint{0.000000in}{-0.027778in}}{\pgfqpoint{0.000000in}{0.000000in}}{%
\pgfpathmoveto{\pgfqpoint{0.000000in}{0.000000in}}%
\pgfpathlineto{\pgfqpoint{0.000000in}{-0.027778in}}%
\pgfusepath{stroke,fill}%
}%
\begin{pgfscope}%
\pgfsys@transformshift{6.084628in}{0.944795in}%
\pgfsys@useobject{currentmarker}{}%
\end{pgfscope}%
\end{pgfscope}%
\begin{pgfscope}%
\definecolor{textcolor}{rgb}{0.000000,0.000000,0.000000}%
\pgfsetstrokecolor{textcolor}%
\pgfsetfillcolor{textcolor}%
\pgftext[x=3.573731in,y=0.405950in,,top]{\color{textcolor}\fontsize{30.000000}{24.000000}\selectfont n}%
\end{pgfscope}%
\begin{pgfscope}%
\pgfsetbuttcap%
\pgfsetroundjoin%
\definecolor{currentfill}{rgb}{0.000000,0.000000,0.000000}%
\pgfsetfillcolor{currentfill}%
\pgfsetlinewidth{0.803000pt}%
\definecolor{currentstroke}{rgb}{0.000000,0.000000,0.000000}%
\pgfsetstrokecolor{currentstroke}%
\pgfsetdash{}{0pt}%
\pgfsys@defobject{currentmarker}{\pgfqpoint{-0.048611in}{0.000000in}}{\pgfqpoint{-0.000000in}{0.000000in}}{%
\pgfpathmoveto{\pgfqpoint{-0.000000in}{0.000000in}}%
\pgfpathlineto{\pgfqpoint{-0.048611in}{0.000000in}}%
\pgfusepath{stroke,fill}%
}%
\begin{pgfscope}%
\pgfsys@transformshift{1.047462in}{1.131134in}%
\pgfsys@useobject{currentmarker}{}%
\end{pgfscope}%
\end{pgfscope}%
\begin{pgfscope}%
\definecolor{textcolor}{rgb}{0.000000,0.000000,0.000000}%
\pgfsetstrokecolor{textcolor}%
\pgfsetfillcolor{textcolor}%
\pgftext[x=0.581165in, y=1.031115in, left, base]{\color{textcolor}\fontsize{25.000000}{24.000000}\selectfont \(\displaystyle {10^{3}}\)}%
\end{pgfscope}%
\begin{pgfscope}%
\pgfsetbuttcap%
\pgfsetroundjoin%
\definecolor{currentfill}{rgb}{0.000000,0.000000,0.000000}%
\pgfsetfillcolor{currentfill}%
\pgfsetlinewidth{0.803000pt}%
\definecolor{currentstroke}{rgb}{0.000000,0.000000,0.000000}%
\pgfsetstrokecolor{currentstroke}%
\pgfsetdash{}{0pt}%
\pgfsys@defobject{currentmarker}{\pgfqpoint{-0.048611in}{0.000000in}}{\pgfqpoint{-0.000000in}{0.000000in}}{%
\pgfpathmoveto{\pgfqpoint{-0.000000in}{0.000000in}}%
\pgfpathlineto{\pgfqpoint{-0.048611in}{0.000000in}}%
\pgfusepath{stroke,fill}%
}%
\begin{pgfscope}%
\pgfsys@transformshift{1.047462in}{2.289130in}%
\pgfsys@useobject{currentmarker}{}%
\end{pgfscope}%
\end{pgfscope}%
\begin{pgfscope}%
\definecolor{textcolor}{rgb}{0.000000,0.000000,0.000000}%
\pgfsetstrokecolor{textcolor}%
\pgfsetfillcolor{textcolor}%
\pgftext[x=0.581165in, y=2.189111in, left, base]{\color{textcolor}\fontsize{25.000000}{24.000000}\selectfont \(\displaystyle {10^{4}}\)}%
\end{pgfscope}%
\begin{pgfscope}%
\pgfsetbuttcap%
\pgfsetroundjoin%
\definecolor{currentfill}{rgb}{0.000000,0.000000,0.000000}%
\pgfsetfillcolor{currentfill}%
\pgfsetlinewidth{0.803000pt}%
\definecolor{currentstroke}{rgb}{0.000000,0.000000,0.000000}%
\pgfsetstrokecolor{currentstroke}%
\pgfsetdash{}{0pt}%
\pgfsys@defobject{currentmarker}{\pgfqpoint{-0.048611in}{0.000000in}}{\pgfqpoint{-0.000000in}{0.000000in}}{%
\pgfpathmoveto{\pgfqpoint{-0.000000in}{0.000000in}}%
\pgfpathlineto{\pgfqpoint{-0.048611in}{0.000000in}}%
\pgfusepath{stroke,fill}%
}%
\begin{pgfscope}%
\pgfsys@transformshift{1.047462in}{3.447126in}%
\pgfsys@useobject{currentmarker}{}%
\end{pgfscope}%
\end{pgfscope}%
\begin{pgfscope}%
\definecolor{textcolor}{rgb}{0.000000,0.000000,0.000000}%
\pgfsetstrokecolor{textcolor}%
\pgfsetfillcolor{textcolor}%
\pgftext[x=0.581165in, y=3.347107in, left, base]{\color{textcolor}\fontsize{25.000000}{24.000000}\selectfont \(\displaystyle {10^{5}}\)}%
\end{pgfscope}%
\begin{pgfscope}%
\pgfsetbuttcap%
\pgfsetroundjoin%
\definecolor{currentfill}{rgb}{0.000000,0.000000,0.000000}%
\pgfsetfillcolor{currentfill}%
\pgfsetlinewidth{0.602250pt}%
\definecolor{currentstroke}{rgb}{0.000000,0.000000,0.000000}%
\pgfsetstrokecolor{currentstroke}%
\pgfsetdash{}{0pt}%
\pgfsys@defobject{currentmarker}{\pgfqpoint{-0.027778in}{0.000000in}}{\pgfqpoint{-0.000000in}{0.000000in}}{%
\pgfpathmoveto{\pgfqpoint{-0.000000in}{0.000000in}}%
\pgfpathlineto{\pgfqpoint{-0.027778in}{0.000000in}}%
\pgfusepath{stroke,fill}%
}%
\begin{pgfscope}%
\pgfsys@transformshift{1.047462in}{0.951758in}%
\pgfsys@useobject{currentmarker}{}%
\end{pgfscope}%
\end{pgfscope}%
\begin{pgfscope}%
\pgfsetbuttcap%
\pgfsetroundjoin%
\definecolor{currentfill}{rgb}{0.000000,0.000000,0.000000}%
\pgfsetfillcolor{currentfill}%
\pgfsetlinewidth{0.602250pt}%
\definecolor{currentstroke}{rgb}{0.000000,0.000000,0.000000}%
\pgfsetstrokecolor{currentstroke}%
\pgfsetdash{}{0pt}%
\pgfsys@defobject{currentmarker}{\pgfqpoint{-0.027778in}{0.000000in}}{\pgfqpoint{-0.000000in}{0.000000in}}{%
\pgfpathmoveto{\pgfqpoint{-0.000000in}{0.000000in}}%
\pgfpathlineto{\pgfqpoint{-0.027778in}{0.000000in}}%
\pgfusepath{stroke,fill}%
}%
\begin{pgfscope}%
\pgfsys@transformshift{1.047462in}{1.018912in}%
\pgfsys@useobject{currentmarker}{}%
\end{pgfscope}%
\end{pgfscope}%
\begin{pgfscope}%
\pgfsetbuttcap%
\pgfsetroundjoin%
\definecolor{currentfill}{rgb}{0.000000,0.000000,0.000000}%
\pgfsetfillcolor{currentfill}%
\pgfsetlinewidth{0.602250pt}%
\definecolor{currentstroke}{rgb}{0.000000,0.000000,0.000000}%
\pgfsetstrokecolor{currentstroke}%
\pgfsetdash{}{0pt}%
\pgfsys@defobject{currentmarker}{\pgfqpoint{-0.027778in}{0.000000in}}{\pgfqpoint{-0.000000in}{0.000000in}}{%
\pgfpathmoveto{\pgfqpoint{-0.000000in}{0.000000in}}%
\pgfpathlineto{\pgfqpoint{-0.027778in}{0.000000in}}%
\pgfusepath{stroke,fill}%
}%
\begin{pgfscope}%
\pgfsys@transformshift{1.047462in}{1.078147in}%
\pgfsys@useobject{currentmarker}{}%
\end{pgfscope}%
\end{pgfscope}%
\begin{pgfscope}%
\pgfsetbuttcap%
\pgfsetroundjoin%
\definecolor{currentfill}{rgb}{0.000000,0.000000,0.000000}%
\pgfsetfillcolor{currentfill}%
\pgfsetlinewidth{0.602250pt}%
\definecolor{currentstroke}{rgb}{0.000000,0.000000,0.000000}%
\pgfsetstrokecolor{currentstroke}%
\pgfsetdash{}{0pt}%
\pgfsys@defobject{currentmarker}{\pgfqpoint{-0.027778in}{0.000000in}}{\pgfqpoint{-0.000000in}{0.000000in}}{%
\pgfpathmoveto{\pgfqpoint{-0.000000in}{0.000000in}}%
\pgfpathlineto{\pgfqpoint{-0.027778in}{0.000000in}}%
\pgfusepath{stroke,fill}%
}%
\begin{pgfscope}%
\pgfsys@transformshift{1.047462in}{1.479725in}%
\pgfsys@useobject{currentmarker}{}%
\end{pgfscope}%
\end{pgfscope}%
\begin{pgfscope}%
\pgfsetbuttcap%
\pgfsetroundjoin%
\definecolor{currentfill}{rgb}{0.000000,0.000000,0.000000}%
\pgfsetfillcolor{currentfill}%
\pgfsetlinewidth{0.602250pt}%
\definecolor{currentstroke}{rgb}{0.000000,0.000000,0.000000}%
\pgfsetstrokecolor{currentstroke}%
\pgfsetdash{}{0pt}%
\pgfsys@defobject{currentmarker}{\pgfqpoint{-0.027778in}{0.000000in}}{\pgfqpoint{-0.000000in}{0.000000in}}{%
\pgfpathmoveto{\pgfqpoint{-0.000000in}{0.000000in}}%
\pgfpathlineto{\pgfqpoint{-0.027778in}{0.000000in}}%
\pgfusepath{stroke,fill}%
}%
\begin{pgfscope}%
\pgfsys@transformshift{1.047462in}{1.683638in}%
\pgfsys@useobject{currentmarker}{}%
\end{pgfscope}%
\end{pgfscope}%
\begin{pgfscope}%
\pgfsetbuttcap%
\pgfsetroundjoin%
\definecolor{currentfill}{rgb}{0.000000,0.000000,0.000000}%
\pgfsetfillcolor{currentfill}%
\pgfsetlinewidth{0.602250pt}%
\definecolor{currentstroke}{rgb}{0.000000,0.000000,0.000000}%
\pgfsetstrokecolor{currentstroke}%
\pgfsetdash{}{0pt}%
\pgfsys@defobject{currentmarker}{\pgfqpoint{-0.027778in}{0.000000in}}{\pgfqpoint{-0.000000in}{0.000000in}}{%
\pgfpathmoveto{\pgfqpoint{-0.000000in}{0.000000in}}%
\pgfpathlineto{\pgfqpoint{-0.027778in}{0.000000in}}%
\pgfusepath{stroke,fill}%
}%
\begin{pgfscope}%
\pgfsys@transformshift{1.047462in}{1.828317in}%
\pgfsys@useobject{currentmarker}{}%
\end{pgfscope}%
\end{pgfscope}%
\begin{pgfscope}%
\pgfsetbuttcap%
\pgfsetroundjoin%
\definecolor{currentfill}{rgb}{0.000000,0.000000,0.000000}%
\pgfsetfillcolor{currentfill}%
\pgfsetlinewidth{0.602250pt}%
\definecolor{currentstroke}{rgb}{0.000000,0.000000,0.000000}%
\pgfsetstrokecolor{currentstroke}%
\pgfsetdash{}{0pt}%
\pgfsys@defobject{currentmarker}{\pgfqpoint{-0.027778in}{0.000000in}}{\pgfqpoint{-0.000000in}{0.000000in}}{%
\pgfpathmoveto{\pgfqpoint{-0.000000in}{0.000000in}}%
\pgfpathlineto{\pgfqpoint{-0.027778in}{0.000000in}}%
\pgfusepath{stroke,fill}%
}%
\begin{pgfscope}%
\pgfsys@transformshift{1.047462in}{1.940538in}%
\pgfsys@useobject{currentmarker}{}%
\end{pgfscope}%
\end{pgfscope}%
\begin{pgfscope}%
\pgfsetbuttcap%
\pgfsetroundjoin%
\definecolor{currentfill}{rgb}{0.000000,0.000000,0.000000}%
\pgfsetfillcolor{currentfill}%
\pgfsetlinewidth{0.602250pt}%
\definecolor{currentstroke}{rgb}{0.000000,0.000000,0.000000}%
\pgfsetstrokecolor{currentstroke}%
\pgfsetdash{}{0pt}%
\pgfsys@defobject{currentmarker}{\pgfqpoint{-0.027778in}{0.000000in}}{\pgfqpoint{-0.000000in}{0.000000in}}{%
\pgfpathmoveto{\pgfqpoint{-0.000000in}{0.000000in}}%
\pgfpathlineto{\pgfqpoint{-0.027778in}{0.000000in}}%
\pgfusepath{stroke,fill}%
}%
\begin{pgfscope}%
\pgfsys@transformshift{1.047462in}{2.032230in}%
\pgfsys@useobject{currentmarker}{}%
\end{pgfscope}%
\end{pgfscope}%
\begin{pgfscope}%
\pgfsetbuttcap%
\pgfsetroundjoin%
\definecolor{currentfill}{rgb}{0.000000,0.000000,0.000000}%
\pgfsetfillcolor{currentfill}%
\pgfsetlinewidth{0.602250pt}%
\definecolor{currentstroke}{rgb}{0.000000,0.000000,0.000000}%
\pgfsetstrokecolor{currentstroke}%
\pgfsetdash{}{0pt}%
\pgfsys@defobject{currentmarker}{\pgfqpoint{-0.027778in}{0.000000in}}{\pgfqpoint{-0.000000in}{0.000000in}}{%
\pgfpathmoveto{\pgfqpoint{-0.000000in}{0.000000in}}%
\pgfpathlineto{\pgfqpoint{-0.027778in}{0.000000in}}%
\pgfusepath{stroke,fill}%
}%
\begin{pgfscope}%
\pgfsys@transformshift{1.047462in}{2.109754in}%
\pgfsys@useobject{currentmarker}{}%
\end{pgfscope}%
\end{pgfscope}%
\begin{pgfscope}%
\pgfsetbuttcap%
\pgfsetroundjoin%
\definecolor{currentfill}{rgb}{0.000000,0.000000,0.000000}%
\pgfsetfillcolor{currentfill}%
\pgfsetlinewidth{0.602250pt}%
\definecolor{currentstroke}{rgb}{0.000000,0.000000,0.000000}%
\pgfsetstrokecolor{currentstroke}%
\pgfsetdash{}{0pt}%
\pgfsys@defobject{currentmarker}{\pgfqpoint{-0.027778in}{0.000000in}}{\pgfqpoint{-0.000000in}{0.000000in}}{%
\pgfpathmoveto{\pgfqpoint{-0.000000in}{0.000000in}}%
\pgfpathlineto{\pgfqpoint{-0.027778in}{0.000000in}}%
\pgfusepath{stroke,fill}%
}%
\begin{pgfscope}%
\pgfsys@transformshift{1.047462in}{2.176909in}%
\pgfsys@useobject{currentmarker}{}%
\end{pgfscope}%
\end{pgfscope}%
\begin{pgfscope}%
\pgfsetbuttcap%
\pgfsetroundjoin%
\definecolor{currentfill}{rgb}{0.000000,0.000000,0.000000}%
\pgfsetfillcolor{currentfill}%
\pgfsetlinewidth{0.602250pt}%
\definecolor{currentstroke}{rgb}{0.000000,0.000000,0.000000}%
\pgfsetstrokecolor{currentstroke}%
\pgfsetdash{}{0pt}%
\pgfsys@defobject{currentmarker}{\pgfqpoint{-0.027778in}{0.000000in}}{\pgfqpoint{-0.000000in}{0.000000in}}{%
\pgfpathmoveto{\pgfqpoint{-0.000000in}{0.000000in}}%
\pgfpathlineto{\pgfqpoint{-0.027778in}{0.000000in}}%
\pgfusepath{stroke,fill}%
}%
\begin{pgfscope}%
\pgfsys@transformshift{1.047462in}{2.236143in}%
\pgfsys@useobject{currentmarker}{}%
\end{pgfscope}%
\end{pgfscope}%
\begin{pgfscope}%
\pgfsetbuttcap%
\pgfsetroundjoin%
\definecolor{currentfill}{rgb}{0.000000,0.000000,0.000000}%
\pgfsetfillcolor{currentfill}%
\pgfsetlinewidth{0.602250pt}%
\definecolor{currentstroke}{rgb}{0.000000,0.000000,0.000000}%
\pgfsetstrokecolor{currentstroke}%
\pgfsetdash{}{0pt}%
\pgfsys@defobject{currentmarker}{\pgfqpoint{-0.027778in}{0.000000in}}{\pgfqpoint{-0.000000in}{0.000000in}}{%
\pgfpathmoveto{\pgfqpoint{-0.000000in}{0.000000in}}%
\pgfpathlineto{\pgfqpoint{-0.027778in}{0.000000in}}%
\pgfusepath{stroke,fill}%
}%
\begin{pgfscope}%
\pgfsys@transformshift{1.047462in}{2.637722in}%
\pgfsys@useobject{currentmarker}{}%
\end{pgfscope}%
\end{pgfscope}%
\begin{pgfscope}%
\pgfsetbuttcap%
\pgfsetroundjoin%
\definecolor{currentfill}{rgb}{0.000000,0.000000,0.000000}%
\pgfsetfillcolor{currentfill}%
\pgfsetlinewidth{0.602250pt}%
\definecolor{currentstroke}{rgb}{0.000000,0.000000,0.000000}%
\pgfsetstrokecolor{currentstroke}%
\pgfsetdash{}{0pt}%
\pgfsys@defobject{currentmarker}{\pgfqpoint{-0.027778in}{0.000000in}}{\pgfqpoint{-0.000000in}{0.000000in}}{%
\pgfpathmoveto{\pgfqpoint{-0.000000in}{0.000000in}}%
\pgfpathlineto{\pgfqpoint{-0.027778in}{0.000000in}}%
\pgfusepath{stroke,fill}%
}%
\begin{pgfscope}%
\pgfsys@transformshift{1.047462in}{2.841635in}%
\pgfsys@useobject{currentmarker}{}%
\end{pgfscope}%
\end{pgfscope}%
\begin{pgfscope}%
\pgfsetbuttcap%
\pgfsetroundjoin%
\definecolor{currentfill}{rgb}{0.000000,0.000000,0.000000}%
\pgfsetfillcolor{currentfill}%
\pgfsetlinewidth{0.602250pt}%
\definecolor{currentstroke}{rgb}{0.000000,0.000000,0.000000}%
\pgfsetstrokecolor{currentstroke}%
\pgfsetdash{}{0pt}%
\pgfsys@defobject{currentmarker}{\pgfqpoint{-0.027778in}{0.000000in}}{\pgfqpoint{-0.000000in}{0.000000in}}{%
\pgfpathmoveto{\pgfqpoint{-0.000000in}{0.000000in}}%
\pgfpathlineto{\pgfqpoint{-0.027778in}{0.000000in}}%
\pgfusepath{stroke,fill}%
}%
\begin{pgfscope}%
\pgfsys@transformshift{1.047462in}{2.986313in}%
\pgfsys@useobject{currentmarker}{}%
\end{pgfscope}%
\end{pgfscope}%
\begin{pgfscope}%
\pgfsetbuttcap%
\pgfsetroundjoin%
\definecolor{currentfill}{rgb}{0.000000,0.000000,0.000000}%
\pgfsetfillcolor{currentfill}%
\pgfsetlinewidth{0.602250pt}%
\definecolor{currentstroke}{rgb}{0.000000,0.000000,0.000000}%
\pgfsetstrokecolor{currentstroke}%
\pgfsetdash{}{0pt}%
\pgfsys@defobject{currentmarker}{\pgfqpoint{-0.027778in}{0.000000in}}{\pgfqpoint{-0.000000in}{0.000000in}}{%
\pgfpathmoveto{\pgfqpoint{-0.000000in}{0.000000in}}%
\pgfpathlineto{\pgfqpoint{-0.027778in}{0.000000in}}%
\pgfusepath{stroke,fill}%
}%
\begin{pgfscope}%
\pgfsys@transformshift{1.047462in}{3.098535in}%
\pgfsys@useobject{currentmarker}{}%
\end{pgfscope}%
\end{pgfscope}%
\begin{pgfscope}%
\pgfsetbuttcap%
\pgfsetroundjoin%
\definecolor{currentfill}{rgb}{0.000000,0.000000,0.000000}%
\pgfsetfillcolor{currentfill}%
\pgfsetlinewidth{0.602250pt}%
\definecolor{currentstroke}{rgb}{0.000000,0.000000,0.000000}%
\pgfsetstrokecolor{currentstroke}%
\pgfsetdash{}{0pt}%
\pgfsys@defobject{currentmarker}{\pgfqpoint{-0.027778in}{0.000000in}}{\pgfqpoint{-0.000000in}{0.000000in}}{%
\pgfpathmoveto{\pgfqpoint{-0.000000in}{0.000000in}}%
\pgfpathlineto{\pgfqpoint{-0.027778in}{0.000000in}}%
\pgfusepath{stroke,fill}%
}%
\begin{pgfscope}%
\pgfsys@transformshift{1.047462in}{3.190226in}%
\pgfsys@useobject{currentmarker}{}%
\end{pgfscope}%
\end{pgfscope}%
\begin{pgfscope}%
\pgfsetbuttcap%
\pgfsetroundjoin%
\definecolor{currentfill}{rgb}{0.000000,0.000000,0.000000}%
\pgfsetfillcolor{currentfill}%
\pgfsetlinewidth{0.602250pt}%
\definecolor{currentstroke}{rgb}{0.000000,0.000000,0.000000}%
\pgfsetstrokecolor{currentstroke}%
\pgfsetdash{}{0pt}%
\pgfsys@defobject{currentmarker}{\pgfqpoint{-0.027778in}{0.000000in}}{\pgfqpoint{-0.000000in}{0.000000in}}{%
\pgfpathmoveto{\pgfqpoint{-0.000000in}{0.000000in}}%
\pgfpathlineto{\pgfqpoint{-0.027778in}{0.000000in}}%
\pgfusepath{stroke,fill}%
}%
\begin{pgfscope}%
\pgfsys@transformshift{1.047462in}{3.267750in}%
\pgfsys@useobject{currentmarker}{}%
\end{pgfscope}%
\end{pgfscope}%
\begin{pgfscope}%
\pgfsetbuttcap%
\pgfsetroundjoin%
\definecolor{currentfill}{rgb}{0.000000,0.000000,0.000000}%
\pgfsetfillcolor{currentfill}%
\pgfsetlinewidth{0.602250pt}%
\definecolor{currentstroke}{rgb}{0.000000,0.000000,0.000000}%
\pgfsetstrokecolor{currentstroke}%
\pgfsetdash{}{0pt}%
\pgfsys@defobject{currentmarker}{\pgfqpoint{-0.027778in}{0.000000in}}{\pgfqpoint{-0.000000in}{0.000000in}}{%
\pgfpathmoveto{\pgfqpoint{-0.000000in}{0.000000in}}%
\pgfpathlineto{\pgfqpoint{-0.027778in}{0.000000in}}%
\pgfusepath{stroke,fill}%
}%
\begin{pgfscope}%
\pgfsys@transformshift{1.047462in}{3.334905in}%
\pgfsys@useobject{currentmarker}{}%
\end{pgfscope}%
\end{pgfscope}%
\begin{pgfscope}%
\pgfsetbuttcap%
\pgfsetroundjoin%
\definecolor{currentfill}{rgb}{0.000000,0.000000,0.000000}%
\pgfsetfillcolor{currentfill}%
\pgfsetlinewidth{0.602250pt}%
\definecolor{currentstroke}{rgb}{0.000000,0.000000,0.000000}%
\pgfsetstrokecolor{currentstroke}%
\pgfsetdash{}{0pt}%
\pgfsys@defobject{currentmarker}{\pgfqpoint{-0.027778in}{0.000000in}}{\pgfqpoint{-0.000000in}{0.000000in}}{%
\pgfpathmoveto{\pgfqpoint{-0.000000in}{0.000000in}}%
\pgfpathlineto{\pgfqpoint{-0.027778in}{0.000000in}}%
\pgfusepath{stroke,fill}%
}%
\begin{pgfscope}%
\pgfsys@transformshift{1.047462in}{3.394139in}%
\pgfsys@useobject{currentmarker}{}%
\end{pgfscope}%
\end{pgfscope}%
\begin{pgfscope}%
\pgfsetbuttcap%
\pgfsetroundjoin%
\definecolor{currentfill}{rgb}{0.000000,0.000000,0.000000}%
\pgfsetfillcolor{currentfill}%
\pgfsetlinewidth{0.602250pt}%
\definecolor{currentstroke}{rgb}{0.000000,0.000000,0.000000}%
\pgfsetstrokecolor{currentstroke}%
\pgfsetdash{}{0pt}%
\pgfsys@defobject{currentmarker}{\pgfqpoint{-0.027778in}{0.000000in}}{\pgfqpoint{-0.000000in}{0.000000in}}{%
\pgfpathmoveto{\pgfqpoint{-0.000000in}{0.000000in}}%
\pgfpathlineto{\pgfqpoint{-0.027778in}{0.000000in}}%
\pgfusepath{stroke,fill}%
}%
\begin{pgfscope}%
\pgfsys@transformshift{1.047462in}{3.795718in}%
\pgfsys@useobject{currentmarker}{}%
\end{pgfscope}%
\end{pgfscope}%
\begin{pgfscope}%
\pgfsetbuttcap%
\pgfsetroundjoin%
\definecolor{currentfill}{rgb}{0.000000,0.000000,0.000000}%
\pgfsetfillcolor{currentfill}%
\pgfsetlinewidth{0.602250pt}%
\definecolor{currentstroke}{rgb}{0.000000,0.000000,0.000000}%
\pgfsetstrokecolor{currentstroke}%
\pgfsetdash{}{0pt}%
\pgfsys@defobject{currentmarker}{\pgfqpoint{-0.027778in}{0.000000in}}{\pgfqpoint{-0.000000in}{0.000000in}}{%
\pgfpathmoveto{\pgfqpoint{-0.000000in}{0.000000in}}%
\pgfpathlineto{\pgfqpoint{-0.027778in}{0.000000in}}%
\pgfusepath{stroke,fill}%
}%
\begin{pgfscope}%
\pgfsys@transformshift{1.047462in}{3.999631in}%
\pgfsys@useobject{currentmarker}{}%
\end{pgfscope}%
\end{pgfscope}%
\begin{pgfscope}%
\pgfsetbuttcap%
\pgfsetroundjoin%
\definecolor{currentfill}{rgb}{0.000000,0.000000,0.000000}%
\pgfsetfillcolor{currentfill}%
\pgfsetlinewidth{0.602250pt}%
\definecolor{currentstroke}{rgb}{0.000000,0.000000,0.000000}%
\pgfsetstrokecolor{currentstroke}%
\pgfsetdash{}{0pt}%
\pgfsys@defobject{currentmarker}{\pgfqpoint{-0.027778in}{0.000000in}}{\pgfqpoint{-0.000000in}{0.000000in}}{%
\pgfpathmoveto{\pgfqpoint{-0.000000in}{0.000000in}}%
\pgfpathlineto{\pgfqpoint{-0.027778in}{0.000000in}}%
\pgfusepath{stroke,fill}%
}%
\begin{pgfscope}%
\pgfsys@transformshift{1.047462in}{4.144309in}%
\pgfsys@useobject{currentmarker}{}%
\end{pgfscope}%
\end{pgfscope}%
\begin{pgfscope}%
\pgfsetbuttcap%
\pgfsetroundjoin%
\definecolor{currentfill}{rgb}{0.000000,0.000000,0.000000}%
\pgfsetfillcolor{currentfill}%
\pgfsetlinewidth{0.602250pt}%
\definecolor{currentstroke}{rgb}{0.000000,0.000000,0.000000}%
\pgfsetstrokecolor{currentstroke}%
\pgfsetdash{}{0pt}%
\pgfsys@defobject{currentmarker}{\pgfqpoint{-0.027778in}{0.000000in}}{\pgfqpoint{-0.000000in}{0.000000in}}{%
\pgfpathmoveto{\pgfqpoint{-0.000000in}{0.000000in}}%
\pgfpathlineto{\pgfqpoint{-0.027778in}{0.000000in}}%
\pgfusepath{stroke,fill}%
}%
\begin{pgfscope}%
\pgfsys@transformshift{1.047462in}{4.256531in}%
\pgfsys@useobject{currentmarker}{}%
\end{pgfscope}%
\end{pgfscope}%
\begin{pgfscope}%
\pgfsetbuttcap%
\pgfsetroundjoin%
\definecolor{currentfill}{rgb}{0.000000,0.000000,0.000000}%
\pgfsetfillcolor{currentfill}%
\pgfsetlinewidth{0.602250pt}%
\definecolor{currentstroke}{rgb}{0.000000,0.000000,0.000000}%
\pgfsetstrokecolor{currentstroke}%
\pgfsetdash{}{0pt}%
\pgfsys@defobject{currentmarker}{\pgfqpoint{-0.027778in}{0.000000in}}{\pgfqpoint{-0.000000in}{0.000000in}}{%
\pgfpathmoveto{\pgfqpoint{-0.000000in}{0.000000in}}%
\pgfpathlineto{\pgfqpoint{-0.027778in}{0.000000in}}%
\pgfusepath{stroke,fill}%
}%
\begin{pgfscope}%
\pgfsys@transformshift{1.047462in}{4.348222in}%
\pgfsys@useobject{currentmarker}{}%
\end{pgfscope}%
\end{pgfscope}%
\begin{pgfscope}%
\pgfsetbuttcap%
\pgfsetroundjoin%
\definecolor{currentfill}{rgb}{0.000000,0.000000,0.000000}%
\pgfsetfillcolor{currentfill}%
\pgfsetlinewidth{0.602250pt}%
\definecolor{currentstroke}{rgb}{0.000000,0.000000,0.000000}%
\pgfsetstrokecolor{currentstroke}%
\pgfsetdash{}{0pt}%
\pgfsys@defobject{currentmarker}{\pgfqpoint{-0.027778in}{0.000000in}}{\pgfqpoint{-0.000000in}{0.000000in}}{%
\pgfpathmoveto{\pgfqpoint{-0.000000in}{0.000000in}}%
\pgfpathlineto{\pgfqpoint{-0.027778in}{0.000000in}}%
\pgfusepath{stroke,fill}%
}%
\begin{pgfscope}%
\pgfsys@transformshift{1.047462in}{4.425746in}%
\pgfsys@useobject{currentmarker}{}%
\end{pgfscope}%
\end{pgfscope}%
\begin{pgfscope}%
\pgfsetbuttcap%
\pgfsetroundjoin%
\definecolor{currentfill}{rgb}{0.000000,0.000000,0.000000}%
\pgfsetfillcolor{currentfill}%
\pgfsetlinewidth{0.602250pt}%
\definecolor{currentstroke}{rgb}{0.000000,0.000000,0.000000}%
\pgfsetstrokecolor{currentstroke}%
\pgfsetdash{}{0pt}%
\pgfsys@defobject{currentmarker}{\pgfqpoint{-0.027778in}{0.000000in}}{\pgfqpoint{-0.000000in}{0.000000in}}{%
\pgfpathmoveto{\pgfqpoint{-0.000000in}{0.000000in}}%
\pgfpathlineto{\pgfqpoint{-0.027778in}{0.000000in}}%
\pgfusepath{stroke,fill}%
}%
\begin{pgfscope}%
\pgfsys@transformshift{1.047462in}{4.492901in}%
\pgfsys@useobject{currentmarker}{}%
\end{pgfscope}%
\end{pgfscope}%
\begin{pgfscope}%
\definecolor{textcolor}{rgb}{0.000000,0.000000,0.000000}%
\pgfsetstrokecolor{textcolor}%
\pgfsetfillcolor{textcolor}%
\pgftext[x=0.525610in,y=2.722398in,,bottom,rotate=90.000000]{\color{textcolor}\fontsize{30.000000}{24.000000}\selectfont Number of operations}%
\end{pgfscope}%
\begin{pgfscope}%
\pgfpathrectangle{\pgfqpoint{1.147462in}{0.944795in}}{\pgfqpoint{5.052538in}{3.555205in}}%
\pgfusepath{clip}%
\pgfsetrectcap%
\pgfsetroundjoin%
\pgfsetlinewidth{1.505625pt}%
\definecolor{currentstroke}{rgb}{0.121569,0.466667,0.705882}%
\pgfsetstrokecolor{currentstroke}%
\pgfsetdash{}{0pt}%
\pgfpathmoveto{\pgfqpoint{1.277123in}{1.106396in}}%
\pgfpathlineto{\pgfqpoint{1.927807in}{1.549138in}}%
\pgfpathlineto{\pgfqpoint{2.389475in}{1.831200in}}%
\pgfpathlineto{\pgfqpoint{2.747572in}{2.056288in}}%
\pgfpathlineto{\pgfqpoint{3.040159in}{2.243961in}}%
\pgfpathlineto{\pgfqpoint{3.287538in}{2.405199in}}%
\pgfpathlineto{\pgfqpoint{3.501827in}{2.546764in}}%
\pgfpathlineto{\pgfqpoint{3.690843in}{2.673109in}}%
\pgfpathlineto{\pgfqpoint{3.859924in}{2.787323in}}%
\pgfpathlineto{\pgfqpoint{4.012877in}{2.891635in}}%
\pgfpathlineto{\pgfqpoint{4.152511in}{2.987707in}}%
\pgfpathlineto{\pgfqpoint{4.280963in}{3.076809in}}%
\pgfpathlineto{\pgfqpoint{4.399890in}{3.159937in}}%
\pgfpathlineto{\pgfqpoint{4.510609in}{3.237884in}}%
\pgfpathlineto{\pgfqpoint{4.614179in}{3.311294in}}%
\pgfpathlineto{\pgfqpoint{4.711468in}{3.380695in}}%
\pgfpathlineto{\pgfqpoint{4.803195in}{3.446527in}}%
\pgfpathlineto{\pgfqpoint{4.889962in}{3.509160in}}%
\pgfpathlineto{\pgfqpoint{4.972276in}{3.568908in}}%
\pgfpathlineto{\pgfqpoint{5.050574in}{3.626040in}}%
\pgfpathlineto{\pgfqpoint{5.125229in}{3.680789in}}%
\pgfpathlineto{\pgfqpoint{5.196564in}{3.733357in}}%
\pgfpathlineto{\pgfqpoint{5.264863in}{3.783920in}}%
\pgfpathlineto{\pgfqpoint{5.330374in}{3.832636in}}%
\pgfpathlineto{\pgfqpoint{5.393314in}{3.879640in}}%
\pgfpathlineto{\pgfqpoint{5.453880in}{3.925057in}}%
\pgfpathlineto{\pgfqpoint{5.512242in}{3.968996in}}%
\pgfpathlineto{\pgfqpoint{5.568556in}{4.011555in}}%
\pgfpathlineto{\pgfqpoint{5.622961in}{4.052822in}}%
\pgfpathlineto{\pgfqpoint{5.675581in}{4.092879in}}%
\pgfpathlineto{\pgfqpoint{5.726531in}{4.131797in}}%
\pgfpathlineto{\pgfqpoint{5.775913in}{4.169643in}}%
\pgfpathlineto{\pgfqpoint{5.823820in}{4.206478in}}%
\pgfpathlineto{\pgfqpoint{5.870339in}{4.242356in}}%
\pgfusepath{stroke}%
\end{pgfscope}%
\begin{pgfscope}%
\pgfpathrectangle{\pgfqpoint{1.047462in}{0.944795in}}{\pgfqpoint{5.052538in}{3.555205in}}%
\pgfusepath{clip}%
\pgfsetrectcap%
\pgfsetroundjoin%
\pgfsetlinewidth{1.505625pt}%
\definecolor{currentstroke}{rgb}{1.000000,0.498039,0.054902}%
\pgfsetstrokecolor{currentstroke}%
\pgfsetdash{}{0pt}%
\pgfpathmoveto{\pgfqpoint{1.277123in}{1.159488in}}%
\pgfpathlineto{\pgfqpoint{1.927807in}{1.627475in}}%
\pgfpathlineto{\pgfqpoint{2.389475in}{1.968417in}}%
\pgfpathlineto{\pgfqpoint{2.747572in}{2.230524in}}%
\pgfpathlineto{\pgfqpoint{3.040159in}{2.441098in}}%
\pgfpathlineto{\pgfqpoint{3.287538in}{2.616379in}}%
\pgfpathlineto{\pgfqpoint{3.501827in}{2.766232in}}%
\pgfpathlineto{\pgfqpoint{3.690843in}{2.896980in}}%
\pgfpathlineto{\pgfqpoint{3.859924in}{3.012882in}}%
\pgfpathlineto{\pgfqpoint{4.012877in}{3.116932in}}%
\pgfpathlineto{\pgfqpoint{4.152511in}{3.211311in}}%
\pgfpathlineto{\pgfqpoint{4.280963in}{3.297650in}}%
\pgfpathlineto{\pgfqpoint{4.399890in}{3.377205in}}%
\pgfpathlineto{\pgfqpoint{4.510609in}{3.450959in}}%
\pgfpathlineto{\pgfqpoint{4.614179in}{3.519696in}}%
\pgfpathlineto{\pgfqpoint{4.711468in}{3.584053in}}%
\pgfpathlineto{\pgfqpoint{4.803195in}{3.644553in}}%
\pgfpathlineto{\pgfqpoint{4.889962in}{3.701632in}}%
\pgfpathlineto{\pgfqpoint{4.972276in}{3.755653in}}%
\pgfpathlineto{\pgfqpoint{5.050574in}{3.806927in}}%
\pgfpathlineto{\pgfqpoint{5.125229in}{3.855720in}}%
\pgfpathlineto{\pgfqpoint{5.196564in}{3.902260in}}%
\pgfpathlineto{\pgfqpoint{5.264863in}{3.946746in}}%
\pgfpathlineto{\pgfqpoint{5.330374in}{3.989351in}}%
\pgfpathlineto{\pgfqpoint{5.393314in}{4.030227in}}%
\pgfpathlineto{\pgfqpoint{5.453880in}{4.069509in}}%
\pgfpathlineto{\pgfqpoint{5.512242in}{4.107316in}}%
\pgfpathlineto{\pgfqpoint{5.568556in}{4.143756in}}%
\pgfpathlineto{\pgfqpoint{5.622961in}{4.178923in}}%
\pgfpathlineto{\pgfqpoint{5.675581in}{4.212903in}}%
\pgfpathlineto{\pgfqpoint{5.726531in}{4.245774in}}%
\pgfpathlineto{\pgfqpoint{5.775913in}{4.277605in}}%
\pgfpathlineto{\pgfqpoint{5.823820in}{4.308461in}}%
\pgfpathlineto{\pgfqpoint{5.870339in}{4.338400in}}%
\pgfusepath{stroke}%
\end{pgfscope}%
\begin{pgfscope}%
\pgfpathrectangle{\pgfqpoint{1.047462in}{0.944795in}}{\pgfqpoint{5.052538in}{3.555205in}}%
\pgfusepath{clip}%
\pgfsetrectcap%
\pgfsetroundjoin%
\pgfsetlinewidth{1.505625pt}%
\definecolor{currentstroke}{rgb}{0.172549,0.627451,0.172549}%
\pgfsetstrokecolor{currentstroke}%
\pgfsetdash{}{0pt}%
\pgfpathmoveto{\pgfqpoint{1.277123in}{1.106396in}}%
\pgfpathlineto{\pgfqpoint{1.927807in}{1.365930in}}%
\pgfpathlineto{\pgfqpoint{2.389475in}{1.606226in}}%
\pgfpathlineto{\pgfqpoint{2.747572in}{1.829573in}}%
\pgfpathlineto{\pgfqpoint{3.040159in}{2.023777in}}%
\pgfpathlineto{\pgfqpoint{3.287538in}{2.191896in}}%
\pgfpathlineto{\pgfqpoint{3.501827in}{2.338706in}}%
\pgfpathlineto{\pgfqpoint{3.690843in}{2.468375in}}%
\pgfpathlineto{\pgfqpoint{3.859924in}{2.584175in}}%
\pgfpathlineto{\pgfqpoint{4.012877in}{2.688617in}}%
\pgfpathlineto{\pgfqpoint{4.152511in}{2.783631in}}%
\pgfpathlineto{\pgfqpoint{4.280963in}{2.870717in}}%
\pgfpathlineto{\pgfqpoint{4.399890in}{2.951059in}}%
\pgfpathlineto{\pgfqpoint{4.510609in}{3.025599in}}%
\pgfpathlineto{\pgfqpoint{4.614179in}{3.095103in}}%
\pgfpathlineto{\pgfqpoint{4.711468in}{3.160195in}}%
\pgfpathlineto{\pgfqpoint{4.803195in}{3.221393in}}%
\pgfpathlineto{\pgfqpoint{4.889962in}{3.279130in}}%
\pgfpathlineto{\pgfqpoint{4.972276in}{3.333772in}}%
\pgfpathlineto{\pgfqpoint{5.050574in}{3.385630in}}%
\pgfpathlineto{\pgfqpoint{5.125229in}{3.434971in}}%
\pgfpathlineto{\pgfqpoint{5.196564in}{3.482026in}}%
\pgfpathlineto{\pgfqpoint{5.264863in}{3.526995in}}%
\pgfpathlineto{\pgfqpoint{5.330374in}{3.570055in}}%
\pgfpathlineto{\pgfqpoint{5.393314in}{3.611360in}}%
\pgfpathlineto{\pgfqpoint{5.453880in}{3.651046in}}%
\pgfpathlineto{\pgfqpoint{5.512242in}{3.689234in}}%
\pgfpathlineto{\pgfqpoint{5.568556in}{3.726034in}}%
\pgfpathlineto{\pgfqpoint{5.622961in}{3.761541in}}%
\pgfpathlineto{\pgfqpoint{5.675581in}{3.795843in}}%
\pgfpathlineto{\pgfqpoint{5.726531in}{3.829020in}}%
\pgfpathlineto{\pgfqpoint{5.775913in}{3.861141in}}%
\pgfpathlineto{\pgfqpoint{5.823820in}{3.892272in}}%
\pgfpathlineto{\pgfqpoint{5.870339in}{3.922472in}}%
\pgfusepath{stroke}%
\end{pgfscope}%
\begin{pgfscope}%
\pgfsetrectcap%
\pgfsetmiterjoin%
\pgfsetlinewidth{0.803000pt}%
\definecolor{currentstroke}{rgb}{0.000000,0.000000,0.000000}%
\pgfsetstrokecolor{currentstroke}%
\pgfsetdash{}{0pt}%
\pgfpathmoveto{\pgfqpoint{1.047462in}{0.944795in}}%
\pgfpathlineto{\pgfqpoint{1.047462in}{4.500000in}}%
\pgfusepath{stroke}%
\end{pgfscope}%
\begin{pgfscope}%
\pgfsetrectcap%
\pgfsetmiterjoin%
\pgfsetlinewidth{0.803000pt}%
\definecolor{currentstroke}{rgb}{0.000000,0.000000,0.000000}%
\pgfsetstrokecolor{currentstroke}%
\pgfsetdash{}{0pt}%
\pgfpathmoveto{\pgfqpoint{6.100000in}{0.944795in}}%
\pgfpathlineto{\pgfqpoint{6.100000in}{4.500000in}}%
\pgfusepath{stroke}%
\end{pgfscope}%
\begin{pgfscope}%
\pgfsetrectcap%
\pgfsetmiterjoin%
\pgfsetlinewidth{0.803000pt}%
\definecolor{currentstroke}{rgb}{0.000000,0.000000,0.000000}%
\pgfsetstrokecolor{currentstroke}%
\pgfsetdash{}{0pt}%
\pgfpathmoveto{\pgfqpoint{1.047462in}{0.944795in}}%
\pgfpathlineto{\pgfqpoint{6.100000in}{0.944795in}}%
\pgfusepath{stroke}%
\end{pgfscope}%
\begin{pgfscope}%
\pgfsetrectcap%
\pgfsetmiterjoin%
\pgfsetlinewidth{0.803000pt}%
\definecolor{currentstroke}{rgb}{0.000000,0.000000,0.000000}%
\pgfsetstrokecolor{currentstroke}%
\pgfsetdash{}{0pt}%
\pgfpathmoveto{\pgfqpoint{1.047462in}{4.500000in}}%
\pgfpathlineto{\pgfqpoint{6.100000in}{4.500000in}}%
\pgfusepath{stroke}%
\end{pgfscope}%
\begin{pgfscope}%
\pgfsetbuttcap%
\pgfsetmiterjoin%
\definecolor{currentfill}{rgb}{1.000000,1.000000,1.000000}%
\pgfsetfillcolor{currentfill}%
\pgfsetfillopacity{0.800000}%
\pgfsetlinewidth{1.003750pt}%
\definecolor{currentstroke}{rgb}{0.800000,0.800000,0.800000}%
\pgfsetstrokecolor{currentstroke}%
\pgfsetstrokeopacity{0.800000}%
\pgfsetdash{}{0pt}%
\pgfpathmoveto{\pgfqpoint{1.241907in}{3.092908in}}%
\pgfpathlineto{\pgfqpoint{3.64176in}{3.092908in}}%
\pgfpathquadraticcurveto{\pgfqpoint{3.649732in}{3.092908in}}{\pgfqpoint{3.649732in}{3.148464in}}%
\pgfpathlineto{\pgfqpoint{3.649732in}{4.305556in}}%
\pgfpathquadraticcurveto{\pgfqpoint{3.649732in}{4.361111in}}{\pgfqpoint{3.64176in}{4.361111in}}%
\pgfpathlineto{\pgfqpoint{1.241907in}{4.361111in}}%
\pgfpathquadraticcurveto{\pgfqpoint{1.186351in}{4.361111in}}{\pgfqpoint{1.186351in}{4.305556in}}%
\pgfpathlineto{\pgfqpoint{1.186351in}{3.148464in}}%
\pgfpathquadraticcurveto{\pgfqpoint{1.186351in}{3.092908in}}{\pgfqpoint{1.241907in}{3.092908in}}%
\pgfpathclose%
\pgfusepath{stroke,fill}%
\end{pgfscope}%
\begin{pgfscope}%
\pgfsetrectcap%
\pgfsetroundjoin%
\pgfsetlinewidth{1.505625pt}%
\definecolor{currentstroke}{rgb}{0.121569,0.466667,0.705882}%
\pgfsetstrokecolor{currentstroke}%
\pgfsetdash{}{0pt}%
\pgfpathmoveto{\pgfqpoint{1.297462in}{4.147184in}}%
\pgfpathlineto{\pgfqpoint{1.853018in}{4.147184in}}%
\pgfusepath{stroke}%
\end{pgfscope}%
\begin{pgfscope}%
\definecolor{textcolor}{rgb}{0.000000,0.000000,0.000000}%
\pgfsetstrokecolor{textcolor}%
\pgfsetfillcolor{textcolor}%
\pgftext[x=2.075240in,y=4.049962in,left,base]{\color{textcolor}\fontsize{25.000000}{24.000000}\selectfont PV-OSIM}%
\end{pgfscope}%
\begin{pgfscope}%
\pgfsetrectcap%
\pgfsetroundjoin%
\pgfsetlinewidth{1.505625pt}%
\definecolor{currentstroke}{rgb}{1.000000,0.498039,0.054902}%
\pgfsetstrokecolor{currentstroke}%
\pgfsetdash{}{0pt}%
\pgfpathmoveto{\pgfqpoint{1.297462in}{3.752227in}}%
\pgfpathlineto{\pgfqpoint{1.853018in}{3.752227in}}%
\pgfusepath{stroke}%
\end{pgfscope}%
\begin{pgfscope}%
\definecolor{textcolor}{rgb}{0.000000,0.000000,0.000000}%
\pgfsetstrokecolor{textcolor}%
\pgfsetfillcolor{textcolor}%
\pgftext[x=2.075240in,y=3.655005in,left,base]{\color{textcolor}\fontsize{25.000000}{24.000000}\selectfont EFPA}%
\end{pgfscope}%
\begin{pgfscope}%
\pgfsetrectcap%
\pgfsetroundjoin%
\pgfsetlinewidth{1.505625pt}%
\definecolor{currentstroke}{rgb}{0.172549,0.627451,0.172549}%
\pgfsetstrokecolor{currentstroke}%
\pgfsetdash{}{0pt}%
\pgfpathmoveto{\pgfqpoint{1.297462in}{3.357271in}}%
\pgfpathlineto{\pgfqpoint{1.853018in}{3.357271in}}%
\pgfusepath{stroke}%
\end{pgfscope}%
\begin{pgfscope}%
\definecolor{textcolor}{rgb}{0.000000,0.000000,0.000000}%
\pgfsetstrokecolor{textcolor}%
\pgfsetfillcolor{textcolor}%
\pgftext[x=2.075240in,y=3.260048in,left,base]{\color{textcolor}\fontsize{25.000000}{24.000000}\selectfont PV-OSIMr}%
\end{pgfscope}%
\end{pgfpicture}%
\makeatother%
\endgroup%

%% file: graphics/Atlas_6D.pgf
%% Creator: Matplotlib, PGF backend
%%
%% To include the figure in your LaTeX document, write
%%   \input{<filename>.pgf}
%%
%% Make sure the required packages are loaded in your preamble
%%   \usepackage{pgf}
%%
%% Figures using additional raster images can only be included by \input if
%% they are in the same directory as the main LaTeX file. For loading figures
%% from other directories you can use the `import` package
%% and then include the figures with
%%   \import{<path to file>}{<filename>.pgf}
%%
%% Matplotlib used the following preamble
%%
\begingroup%
\makeatletter%
\begin{pgfpicture}%
\pgfpathrectangle{\pgfpointorigin}{\pgfqpoint{2.000000in}{1.500000in}}%
\pgfusepath{use as bounding box, clip}%
\begin{pgfscope}%
\pgfsetbuttcap%
\pgfsetmiterjoin%
\definecolor{currentfill}{rgb}{1.000000,1.000000,1.000000}%
\pgfsetfillcolor{currentfill}%
\pgfsetlinewidth{0.000000pt}%
\definecolor{currentstroke}{rgb}{1.000000,1.000000,1.000000}%
\pgfsetstrokecolor{currentstroke}%
\pgfsetdash{}{0pt}%
\pgfpathmoveto{\pgfqpoint{0.000000in}{0.000000in}}%
\pgfpathlineto{\pgfqpoint{2.000000in}{0.000000in}}%
\pgfpathlineto{\pgfqpoint{2.000000in}{1.500000in}}%
\pgfpathlineto{\pgfqpoint{0.000000in}{1.500000in}}%
\pgfpathclose%
\pgfusepath{fill}%
\end{pgfscope}%
\begin{pgfscope}%
\pgfsetbuttcap%
\pgfsetmiterjoin%
\definecolor{currentfill}{rgb}{1.000000,1.000000,1.000000}%
\pgfsetfillcolor{currentfill}%
\pgfsetlinewidth{0.000000pt}%
\definecolor{currentstroke}{rgb}{0.000000,0.000000,0.000000}%
\pgfsetstrokecolor{currentstroke}%
\pgfsetstrokeopacity{0.000000}%
\pgfsetdash{}{0pt}%
\pgfpathmoveto{\pgfqpoint{0.621012in}{0.288642in}}%
\pgfpathlineto{\pgfqpoint{1.895000in}{0.288642in}}%
\pgfpathlineto{\pgfqpoint{1.895000in}{1.395000in}}%
\pgfpathlineto{\pgfqpoint{0.621012in}{1.395000in}}%
\pgfpathclose%
\pgfusepath{fill}%
\end{pgfscope}%
\begin{pgfscope}%
\pgfpathrectangle{\pgfqpoint{0.621012in}{0.288642in}}{\pgfqpoint{1.273988in}{1.106358in}}%
\pgfusepath{clip}%
\pgfsetbuttcap%
\pgfsetmiterjoin%
\definecolor{currentfill}{rgb}{0.000000,1.000000,1.000000}%
\pgfsetfillcolor{currentfill}%
\pgfsetlinewidth{0.000000pt}%
\definecolor{currentstroke}{rgb}{0.000000,0.000000,0.000000}%
\pgfsetstrokecolor{currentstroke}%
\pgfsetstrokeopacity{0.000000}%
\pgfsetdash{}{0pt}%
\pgfpathmoveto{\pgfqpoint{0.678920in}{0.288642in}}%
\pgfpathlineto{\pgfqpoint{0.922746in}{0.288642in}}%
\pgfpathlineto{\pgfqpoint{0.922746in}{0.952901in}}%
\pgfpathlineto{\pgfqpoint{0.678920in}{0.952901in}}%
\pgfpathclose%
\pgfusepath{fill}%
\end{pgfscope}%
\begin{pgfscope}%
\pgfpathrectangle{\pgfqpoint{0.621012in}{0.288642in}}{\pgfqpoint{1.273988in}{1.106358in}}%
\pgfusepath{clip}%
\pgfsetbuttcap%
\pgfsetmiterjoin%
\definecolor{currentfill}{rgb}{0.000000,0.000000,1.000000}%
\pgfsetfillcolor{currentfill}%
\pgfsetlinewidth{0.000000pt}%
\definecolor{currentstroke}{rgb}{0.000000,0.000000,0.000000}%
\pgfsetstrokecolor{currentstroke}%
\pgfsetstrokeopacity{0.000000}%
\pgfsetdash{}{0pt}%
\pgfpathmoveto{\pgfqpoint{0.983702in}{0.288642in}}%
\pgfpathlineto{\pgfqpoint{1.227528in}{0.288642in}}%
\pgfpathlineto{\pgfqpoint{1.227528in}{0.952901in}}%
\pgfpathlineto{\pgfqpoint{0.983702in}{0.952901in}}%
\pgfpathclose%
\pgfusepath{fill}%
\end{pgfscope}%
\begin{pgfscope}%
\pgfpathrectangle{\pgfqpoint{0.621012in}{0.288642in}}{\pgfqpoint{1.273988in}{1.106358in}}%
\pgfusepath{clip}%
\pgfsetbuttcap%
\pgfsetmiterjoin%
\definecolor{currentfill}{rgb}{1.000000,0.752941,0.796078}%
\pgfsetfillcolor{currentfill}%
\pgfsetlinewidth{0.000000pt}%
\definecolor{currentstroke}{rgb}{0.000000,0.000000,0.000000}%
\pgfsetstrokecolor{currentstroke}%
\pgfsetstrokeopacity{0.000000}%
\pgfsetdash{}{0pt}%
\pgfpathmoveto{\pgfqpoint{1.288484in}{0.288642in}}%
\pgfpathlineto{\pgfqpoint{1.532310in}{0.288642in}}%
\pgfpathlineto{\pgfqpoint{1.532310in}{1.089881in}}%
\pgfpathlineto{\pgfqpoint{1.288484in}{1.089881in}}%
\pgfpathclose%
\pgfusepath{fill}%
\end{pgfscope}%
\begin{pgfscope}%
\pgfpathrectangle{\pgfqpoint{0.621012in}{0.288642in}}{\pgfqpoint{1.273988in}{1.106358in}}%
\pgfusepath{clip}%
\pgfsetbuttcap%
\pgfsetmiterjoin%
\definecolor{currentfill}{rgb}{0.000000,0.501961,0.000000}%
\pgfsetfillcolor{currentfill}%
\pgfsetlinewidth{0.000000pt}%
\definecolor{currentstroke}{rgb}{0.000000,0.000000,0.000000}%
\pgfsetstrokecolor{currentstroke}%
\pgfsetstrokeopacity{0.000000}%
\pgfsetdash{}{0pt}%
\pgfpathmoveto{\pgfqpoint{1.593266in}{0.288642in}}%
\pgfpathlineto{\pgfqpoint{1.837091in}{0.288642in}}%
\pgfpathlineto{\pgfqpoint{1.837091in}{1.342316in}}%
\pgfpathlineto{\pgfqpoint{1.593266in}{1.342316in}}%
\pgfpathclose%
\pgfusepath{fill}%
\end{pgfscope}%
\begin{pgfscope}%
\pgfsetbuttcap%
\pgfsetroundjoin%
\definecolor{currentfill}{rgb}{0.000000,0.000000,0.000000}%
\pgfsetfillcolor{currentfill}%
\pgfsetlinewidth{0.803000pt}%
\definecolor{currentstroke}{rgb}{0.000000,0.000000,0.000000}%
\pgfsetstrokecolor{currentstroke}%
\pgfsetdash{}{0pt}%
\pgfsys@defobject{currentmarker}{\pgfqpoint{0.000000in}{-0.048611in}}{\pgfqpoint{0.000000in}{0.000000in}}{%
\pgfpathmoveto{\pgfqpoint{0.000000in}{0.000000in}}%
\pgfpathlineto{\pgfqpoint{0.000000in}{-0.048611in}}%
\pgfusepath{stroke,fill}%
}%
\begin{pgfscope}%
\pgfsys@transformshift{0.800833in}{0.288642in}%
\pgfsys@useobject{currentmarker}{}%
\end{pgfscope}%
\end{pgfscope}%
\begin{pgfscope}%
\definecolor{textcolor}{rgb}{0.000000,0.000000,0.000000}%
\pgfsetstrokecolor{textcolor}%
\pgfsetfillcolor{textcolor}%
\pgftext[x=0.800833in,y=0.191419in,,top]{\color{textcolor}\fontsize{7.000000}{8.400000}\selectfont PV-OSIMr}%
\end{pgfscope}%
\begin{pgfscope}%
\pgfsetbuttcap%
\pgfsetroundjoin%
\definecolor{currentfill}{rgb}{0.000000,0.000000,0.000000}%
\pgfsetfillcolor{currentfill}%
\pgfsetlinewidth{0.803000pt}%
\definecolor{currentstroke}{rgb}{0.000000,0.000000,0.000000}%
\pgfsetstrokecolor{currentstroke}%
\pgfsetdash{}{0pt}%
\pgfsys@defobject{currentmarker}{\pgfqpoint{0.000000in}{-0.048611in}}{\pgfqpoint{0.000000in}{0.000000in}}{%
\pgfpathmoveto{\pgfqpoint{0.000000in}{0.000000in}}%
\pgfpathlineto{\pgfqpoint{0.000000in}{-0.048611in}}%
\pgfusepath{stroke,fill}%
}%
\begin{pgfscope}%
\pgfsys@transformshift{1.105615in}{0.288642in}%
\pgfsys@useobject{currentmarker}{}%
\end{pgfscope}%
\end{pgfscope}%
\begin{pgfscope}%
\definecolor{textcolor}{rgb}{0.000000,0.000000,0.000000}%
\pgfsetstrokecolor{textcolor}%
\pgfsetfillcolor{textcolor}%
\pgftext[x=1.105615in,y=0.191419in,,top]{\color{textcolor}\fontsize{7.000000}{8.400000}\selectfont PV}%
\end{pgfscope}%
\begin{pgfscope}%
\pgfsetbuttcap%
\pgfsetroundjoin%
\definecolor{currentfill}{rgb}{0.000000,0.000000,0.000000}%
\pgfsetfillcolor{currentfill}%
\pgfsetlinewidth{0.803000pt}%
\definecolor{currentstroke}{rgb}{0.000000,0.000000,0.000000}%
\pgfsetstrokecolor{currentstroke}%
\pgfsetdash{}{0pt}%
\pgfsys@defobject{currentmarker}{\pgfqpoint{0.000000in}{-0.048611in}}{\pgfqpoint{0.000000in}{0.000000in}}{%
\pgfpathmoveto{\pgfqpoint{0.000000in}{0.000000in}}%
\pgfpathlineto{\pgfqpoint{0.000000in}{-0.048611in}}%
\pgfusepath{stroke,fill}%
}%
\begin{pgfscope}%
\pgfsys@transformshift{1.410397in}{0.288642in}%
\pgfsys@useobject{currentmarker}{}%
\end{pgfscope}%
\end{pgfscope}%
\begin{pgfscope}%
\definecolor{textcolor}{rgb}{0.000000,0.000000,0.000000}%
\pgfsetstrokecolor{textcolor}%
\pgfsetfillcolor{textcolor}%
\pgftext[x=1.410397in,y=0.191419in,,top]{\color{textcolor}\fontsize{7.000000}{8.400000}\selectfont EFPA}%
\end{pgfscope}%
\begin{pgfscope}%
\pgfsetbuttcap%
\pgfsetroundjoin%
\definecolor{currentfill}{rgb}{0.000000,0.000000,0.000000}%
\pgfsetfillcolor{currentfill}%
\pgfsetlinewidth{0.803000pt}%
\definecolor{currentstroke}{rgb}{0.000000,0.000000,0.000000}%
\pgfsetstrokecolor{currentstroke}%
\pgfsetdash{}{0pt}%
\pgfsys@defobject{currentmarker}{\pgfqpoint{0.000000in}{-0.048611in}}{\pgfqpoint{0.000000in}{0.000000in}}{%
\pgfpathmoveto{\pgfqpoint{0.000000in}{0.000000in}}%
\pgfpathlineto{\pgfqpoint{0.000000in}{-0.048611in}}%
\pgfusepath{stroke,fill}%
}%
\begin{pgfscope}%
\pgfsys@transformshift{1.715179in}{0.288642in}%
\pgfsys@useobject{currentmarker}{}%
\end{pgfscope}%
\end{pgfscope}%
\begin{pgfscope}%
\definecolor{textcolor}{rgb}{0.000000,0.000000,0.000000}%
\pgfsetstrokecolor{textcolor}%
\pgfsetfillcolor{textcolor}%
\pgftext[x=1.715179in,y=0.191419in,,top]{\color{textcolor}\fontsize{7.000000}{8.400000}\selectfont LTL}%
\end{pgfscope}%
\begin{pgfscope}%
\pgfsetbuttcap%
\pgfsetroundjoin%
\definecolor{currentfill}{rgb}{0.000000,0.000000,0.000000}%
\pgfsetfillcolor{currentfill}%
\pgfsetlinewidth{0.803000pt}%
\definecolor{currentstroke}{rgb}{0.000000,0.000000,0.000000}%
\pgfsetstrokecolor{currentstroke}%
\pgfsetdash{}{0pt}%
\pgfsys@defobject{currentmarker}{\pgfqpoint{-0.048611in}{0.000000in}}{\pgfqpoint{-0.000000in}{0.000000in}}{%
\pgfpathmoveto{\pgfqpoint{-0.000000in}{0.000000in}}%
\pgfpathlineto{\pgfqpoint{-0.048611in}{0.000000in}}%
\pgfusepath{stroke,fill}%
}%
\begin{pgfscope}%
\pgfsys@transformshift{0.621012in}{0.288642in}%
\pgfsys@useobject{currentmarker}{}%
\end{pgfscope}%
\end{pgfscope}%
\begin{pgfscope}%
\definecolor{textcolor}{rgb}{0.000000,0.000000,0.000000}%
\pgfsetstrokecolor{textcolor}%
\pgfsetfillcolor{textcolor}%
\pgftext[x=0.468427in, y=0.254884in, left, base]{\color{textcolor}\fontsize{7.000000}{8.400000}\selectfont \(\displaystyle {0}\)}%
\end{pgfscope}%
\begin{pgfscope}%
\pgfsetbuttcap%
\pgfsetroundjoin%
\definecolor{currentfill}{rgb}{0.000000,0.000000,0.000000}%
\pgfsetfillcolor{currentfill}%
\pgfsetlinewidth{0.803000pt}%
\definecolor{currentstroke}{rgb}{0.000000,0.000000,0.000000}%
\pgfsetstrokecolor{currentstroke}%
\pgfsetdash{}{0pt}%
\pgfsys@defobject{currentmarker}{\pgfqpoint{-0.048611in}{0.000000in}}{\pgfqpoint{-0.000000in}{0.000000in}}{%
\pgfpathmoveto{\pgfqpoint{-0.000000in}{0.000000in}}%
\pgfpathlineto{\pgfqpoint{-0.048611in}{0.000000in}}%
\pgfusepath{stroke,fill}%
}%
\begin{pgfscope}%
\pgfsys@transformshift{0.621012in}{0.629218in}%
\pgfsys@useobject{currentmarker}{}%
\end{pgfscope}%
\end{pgfscope}%
\begin{pgfscope}%
\definecolor{textcolor}{rgb}{0.000000,0.000000,0.000000}%
\pgfsetstrokecolor{textcolor}%
\pgfsetfillcolor{textcolor}%
\pgftext[x=0.302338in, y=0.595460in, left, base]{\color{textcolor}\fontsize{7.000000}{8.400000}\selectfont \(\displaystyle {5000}\)}%
\end{pgfscope}%
\begin{pgfscope}%
\pgfsetbuttcap%
\pgfsetroundjoin%
\definecolor{currentfill}{rgb}{0.000000,0.000000,0.000000}%
\pgfsetfillcolor{currentfill}%
\pgfsetlinewidth{0.803000pt}%
\definecolor{currentstroke}{rgb}{0.000000,0.000000,0.000000}%
\pgfsetstrokecolor{currentstroke}%
\pgfsetdash{}{0pt}%
\pgfsys@defobject{currentmarker}{\pgfqpoint{-0.048611in}{0.000000in}}{\pgfqpoint{-0.000000in}{0.000000in}}{%
\pgfpathmoveto{\pgfqpoint{-0.000000in}{0.000000in}}%
\pgfpathlineto{\pgfqpoint{-0.048611in}{0.000000in}}%
\pgfusepath{stroke,fill}%
}%
\begin{pgfscope}%
\pgfsys@transformshift{0.621012in}{0.969794in}%
\pgfsys@useobject{currentmarker}{}%
\end{pgfscope}%
\end{pgfscope}%
\begin{pgfscope}%
\definecolor{textcolor}{rgb}{0.000000,0.000000,0.000000}%
\pgfsetstrokecolor{textcolor}%
\pgfsetfillcolor{textcolor}%
\pgftext[x=0.246975in, y=0.936036in, left, base]{\color{textcolor}\fontsize{7.000000}{8.400000}\selectfont \(\displaystyle {10000}\)}%
\end{pgfscope}%
\begin{pgfscope}%
\pgfsetbuttcap%
\pgfsetroundjoin%
\definecolor{currentfill}{rgb}{0.000000,0.000000,0.000000}%
\pgfsetfillcolor{currentfill}%
\pgfsetlinewidth{0.803000pt}%
\definecolor{currentstroke}{rgb}{0.000000,0.000000,0.000000}%
\pgfsetstrokecolor{currentstroke}%
\pgfsetdash{}{0pt}%
\pgfsys@defobject{currentmarker}{\pgfqpoint{-0.048611in}{0.000000in}}{\pgfqpoint{-0.000000in}{0.000000in}}{%
\pgfpathmoveto{\pgfqpoint{-0.000000in}{0.000000in}}%
\pgfpathlineto{\pgfqpoint{-0.048611in}{0.000000in}}%
\pgfusepath{stroke,fill}%
}%
\begin{pgfscope}%
\pgfsys@transformshift{0.621012in}{1.310370in}%
\pgfsys@useobject{currentmarker}{}%
\end{pgfscope}%
\end{pgfscope}%
\begin{pgfscope}%
\definecolor{textcolor}{rgb}{0.000000,0.000000,0.000000}%
\pgfsetstrokecolor{textcolor}%
\pgfsetfillcolor{textcolor}%
\pgftext[x=0.246975in, y=1.276613in, left, base]{\color{textcolor}\fontsize{7.000000}{8.400000}\selectfont \(\displaystyle {15000}\)}%
\end{pgfscope}%
\begin{pgfscope}%
\definecolor{textcolor}{rgb}{0.000000,0.000000,0.000000}%
\pgfsetstrokecolor{textcolor}%
\pgfsetfillcolor{textcolor}%
\pgftext[x=0.191419in,y=0.841821in,,bottom,rotate=90.000000]{\color{textcolor}\fontsize{7.000000}{8.400000}\selectfont Number of operations}%
\end{pgfscope}%
\begin{pgfscope}%
\pgfsetrectcap%
\pgfsetmiterjoin%
\pgfsetlinewidth{0.803000pt}%
\definecolor{currentstroke}{rgb}{0.000000,0.000000,0.000000}%
\pgfsetstrokecolor{currentstroke}%
\pgfsetdash{}{0pt}%
\pgfpathmoveto{\pgfqpoint{0.621012in}{0.288642in}}%
\pgfpathlineto{\pgfqpoint{0.621012in}{1.395000in}}%
\pgfusepath{stroke}%
\end{pgfscope}%
\begin{pgfscope}%
\pgfsetrectcap%
\pgfsetmiterjoin%
\pgfsetlinewidth{0.803000pt}%
\definecolor{currentstroke}{rgb}{0.000000,0.000000,0.000000}%
\pgfsetstrokecolor{currentstroke}%
\pgfsetdash{}{0pt}%
\pgfpathmoveto{\pgfqpoint{1.895000in}{0.288642in}}%
\pgfpathlineto{\pgfqpoint{1.895000in}{1.395000in}}%
\pgfusepath{stroke}%
\end{pgfscope}%
\begin{pgfscope}%
\pgfsetrectcap%
\pgfsetmiterjoin%
\pgfsetlinewidth{0.803000pt}%
\definecolor{currentstroke}{rgb}{0.000000,0.000000,0.000000}%
\pgfsetstrokecolor{currentstroke}%
\pgfsetdash{}{0pt}%
\pgfpathmoveto{\pgfqpoint{0.621012in}{0.288642in}}%
\pgfpathlineto{\pgfqpoint{1.895000in}{0.288642in}}%
\pgfusepath{stroke}%
\end{pgfscope}%
\begin{pgfscope}%
\pgfsetrectcap%
\pgfsetmiterjoin%
\pgfsetlinewidth{0.803000pt}%
\definecolor{currentstroke}{rgb}{0.000000,0.000000,0.000000}%
\pgfsetstrokecolor{currentstroke}%
\pgfsetdash{}{0pt}%
\pgfpathmoveto{\pgfqpoint{0.621012in}{1.395000in}}%
\pgfpathlineto{\pgfqpoint{1.895000in}{1.395000in}}%
\pgfusepath{stroke}%
\end{pgfscope}%
\end{pgfpicture}%
\makeatother%
\endgroup%

%% file: graphics/Atlas_12D.pgf
%% Creator: Matplotlib, PGF backend
%%
%% To include the figure in your LaTeX document, write
%%   \input{<filename>.pgf}
%%
%% Make sure the required packages are loaded in your preamble
%%   \usepackage{pgf}
%%
%% Figures using additional raster images can only be included by \input if
%% they are in the same directory as the main LaTeX file. For loading figures
%% from other directories you can use the `import` package
%% and then include the figures with
%%   \import{<path to file>}{<filename>.pgf}
%%
%% Matplotlib used the following preamble
%%
\begingroup%
\makeatletter%
\begin{pgfpicture}%
\pgfpathrectangle{\pgfpointorigin}{\pgfqpoint{2.000000in}{1.500000in}}%
\pgfusepath{use as bounding box, clip}%
\begin{pgfscope}%
\pgfsetbuttcap%
\pgfsetmiterjoin%
\definecolor{currentfill}{rgb}{1.000000,1.000000,1.000000}%
\pgfsetfillcolor{currentfill}%
\pgfsetlinewidth{0.000000pt}%
\definecolor{currentstroke}{rgb}{1.000000,1.000000,1.000000}%
\pgfsetstrokecolor{currentstroke}%
\pgfsetdash{}{0pt}%
\pgfpathmoveto{\pgfqpoint{0.000000in}{0.000000in}}%
\pgfpathlineto{\pgfqpoint{2.000000in}{0.000000in}}%
\pgfpathlineto{\pgfqpoint{2.000000in}{1.500000in}}%
\pgfpathlineto{\pgfqpoint{0.000000in}{1.500000in}}%
\pgfpathclose%
\pgfusepath{fill}%
\end{pgfscope}%
\begin{pgfscope}%
\pgfsetbuttcap%
\pgfsetmiterjoin%
\definecolor{currentfill}{rgb}{1.000000,1.000000,1.000000}%
\pgfsetfillcolor{currentfill}%
\pgfsetlinewidth{0.000000pt}%
\definecolor{currentstroke}{rgb}{0.000000,0.000000,0.000000}%
\pgfsetstrokecolor{currentstroke}%
\pgfsetstrokeopacity{0.000000}%
\pgfsetdash{}{0pt}%
\pgfpathmoveto{\pgfqpoint{0.621012in}{0.288642in}}%
\pgfpathlineto{\pgfqpoint{1.895000in}{0.288642in}}%
\pgfpathlineto{\pgfqpoint{1.895000in}{1.395000in}}%
\pgfpathlineto{\pgfqpoint{0.621012in}{1.395000in}}%
\pgfpathclose%
\pgfusepath{fill}%
\end{pgfscope}%
\begin{pgfscope}%
\pgfpathrectangle{\pgfqpoint{0.621012in}{0.288642in}}{\pgfqpoint{1.273988in}{1.106358in}}%
\pgfusepath{clip}%
\pgfsetbuttcap%
\pgfsetmiterjoin%
\definecolor{currentfill}{rgb}{0.000000,1.000000,1.000000}%
\pgfsetfillcolor{currentfill}%
\pgfsetlinewidth{0.000000pt}%
\definecolor{currentstroke}{rgb}{0.000000,0.000000,0.000000}%
\pgfsetstrokecolor{currentstroke}%
\pgfsetstrokeopacity{0.000000}%
\pgfsetdash{}{0pt}%
\pgfpathmoveto{\pgfqpoint{0.678920in}{0.288642in}}%
\pgfpathlineto{\pgfqpoint{0.922746in}{0.288642in}}%
\pgfpathlineto{\pgfqpoint{0.922746in}{0.997968in}}%
\pgfpathlineto{\pgfqpoint{0.678920in}{0.997968in}}%
\pgfpathclose%
\pgfusepath{fill}%
\end{pgfscope}%
\begin{pgfscope}%
\pgfpathrectangle{\pgfqpoint{0.621012in}{0.288642in}}{\pgfqpoint{1.273988in}{1.106358in}}%
\pgfusepath{clip}%
\pgfsetbuttcap%
\pgfsetmiterjoin%
\definecolor{currentfill}{rgb}{0.000000,0.000000,1.000000}%
\pgfsetfillcolor{currentfill}%
\pgfsetlinewidth{0.000000pt}%
\definecolor{currentstroke}{rgb}{0.000000,0.000000,0.000000}%
\pgfsetstrokecolor{currentstroke}%
\pgfsetstrokeopacity{0.000000}%
\pgfsetdash{}{0pt}%
\pgfpathmoveto{\pgfqpoint{0.983702in}{0.288642in}}%
\pgfpathlineto{\pgfqpoint{1.227528in}{0.288642in}}%
\pgfpathlineto{\pgfqpoint{1.227528in}{1.054968in}}%
\pgfpathlineto{\pgfqpoint{0.983702in}{1.054968in}}%
\pgfpathclose%
\pgfusepath{fill}%
\end{pgfscope}%
\begin{pgfscope}%
\pgfpathrectangle{\pgfqpoint{0.621012in}{0.288642in}}{\pgfqpoint{1.273988in}{1.106358in}}%
\pgfusepath{clip}%
\pgfsetbuttcap%
\pgfsetmiterjoin%
\definecolor{currentfill}{rgb}{1.000000,0.752941,0.796078}%
\pgfsetfillcolor{currentfill}%
\pgfsetlinewidth{0.000000pt}%
\definecolor{currentstroke}{rgb}{0.000000,0.000000,0.000000}%
\pgfsetstrokecolor{currentstroke}%
\pgfsetstrokeopacity{0.000000}%
\pgfsetdash{}{0pt}%
\pgfpathmoveto{\pgfqpoint{1.288484in}{0.288642in}}%
\pgfpathlineto{\pgfqpoint{1.532310in}{0.288642in}}%
\pgfpathlineto{\pgfqpoint{1.532310in}{1.231606in}}%
\pgfpathlineto{\pgfqpoint{1.288484in}{1.231606in}}%
\pgfpathclose%
\pgfusepath{fill}%
\end{pgfscope}%
\begin{pgfscope}%
\pgfpathrectangle{\pgfqpoint{0.621012in}{0.288642in}}{\pgfqpoint{1.273988in}{1.106358in}}%
\pgfusepath{clip}%
\pgfsetbuttcap%
\pgfsetmiterjoin%
\definecolor{currentfill}{rgb}{0.000000,0.501961,0.000000}%
\pgfsetfillcolor{currentfill}%
\pgfsetlinewidth{0.000000pt}%
\definecolor{currentstroke}{rgb}{0.000000,0.000000,0.000000}%
\pgfsetstrokecolor{currentstroke}%
\pgfsetstrokeopacity{0.000000}%
\pgfsetdash{}{0pt}%
\pgfpathmoveto{\pgfqpoint{1.593266in}{0.288642in}}%
\pgfpathlineto{\pgfqpoint{1.837091in}{0.288642in}}%
\pgfpathlineto{\pgfqpoint{1.837091in}{1.342316in}}%
\pgfpathlineto{\pgfqpoint{1.593266in}{1.342316in}}%
\pgfpathclose%
\pgfusepath{fill}%
\end{pgfscope}%
\begin{pgfscope}%
\pgfsetbuttcap%
\pgfsetroundjoin%
\definecolor{currentfill}{rgb}{0.000000,0.000000,0.000000}%
\pgfsetfillcolor{currentfill}%
\pgfsetlinewidth{0.803000pt}%
\definecolor{currentstroke}{rgb}{0.000000,0.000000,0.000000}%
\pgfsetstrokecolor{currentstroke}%
\pgfsetdash{}{0pt}%
\pgfsys@defobject{currentmarker}{\pgfqpoint{0.000000in}{-0.048611in}}{\pgfqpoint{0.000000in}{0.000000in}}{%
\pgfpathmoveto{\pgfqpoint{0.000000in}{0.000000in}}%
\pgfpathlineto{\pgfqpoint{0.000000in}{-0.048611in}}%
\pgfusepath{stroke,fill}%
}%
\begin{pgfscope}%
\pgfsys@transformshift{0.800833in}{0.288642in}%
\pgfsys@useobject{currentmarker}{}%
\end{pgfscope}%
\end{pgfscope}%
\begin{pgfscope}%
\definecolor{textcolor}{rgb}{0.000000,0.000000,0.000000}%
\pgfsetstrokecolor{textcolor}%
\pgfsetfillcolor{textcolor}%
\pgftext[x=0.800833in,y=0.191419in,,top]{\color{textcolor}\fontsize{7.000000}{8.400000}\selectfont PV-OSIMr}%
\end{pgfscope}%
\begin{pgfscope}%
\pgfsetbuttcap%
\pgfsetroundjoin%
\definecolor{currentfill}{rgb}{0.000000,0.000000,0.000000}%
\pgfsetfillcolor{currentfill}%
\pgfsetlinewidth{0.803000pt}%
\definecolor{currentstroke}{rgb}{0.000000,0.000000,0.000000}%
\pgfsetstrokecolor{currentstroke}%
\pgfsetdash{}{0pt}%
\pgfsys@defobject{currentmarker}{\pgfqpoint{0.000000in}{-0.048611in}}{\pgfqpoint{0.000000in}{0.000000in}}{%
\pgfpathmoveto{\pgfqpoint{0.000000in}{0.000000in}}%
\pgfpathlineto{\pgfqpoint{0.000000in}{-0.048611in}}%
\pgfusepath{stroke,fill}%
}%
\begin{pgfscope}%
\pgfsys@transformshift{1.105615in}{0.288642in}%
\pgfsys@useobject{currentmarker}{}%
\end{pgfscope}%
\end{pgfscope}%
\begin{pgfscope}%
\definecolor{textcolor}{rgb}{0.000000,0.000000,0.000000}%
\pgfsetstrokecolor{textcolor}%
\pgfsetfillcolor{textcolor}%
\pgftext[x=1.105615in,y=0.191419in,,top]{\color{textcolor}\fontsize{7.000000}{8.400000}\selectfont PV}%
\end{pgfscope}%
\begin{pgfscope}%
\pgfsetbuttcap%
\pgfsetroundjoin%
\definecolor{currentfill}{rgb}{0.000000,0.000000,0.000000}%
\pgfsetfillcolor{currentfill}%
\pgfsetlinewidth{0.803000pt}%
\definecolor{currentstroke}{rgb}{0.000000,0.000000,0.000000}%
\pgfsetstrokecolor{currentstroke}%
\pgfsetdash{}{0pt}%
\pgfsys@defobject{currentmarker}{\pgfqpoint{0.000000in}{-0.048611in}}{\pgfqpoint{0.000000in}{0.000000in}}{%
\pgfpathmoveto{\pgfqpoint{0.000000in}{0.000000in}}%
\pgfpathlineto{\pgfqpoint{0.000000in}{-0.048611in}}%
\pgfusepath{stroke,fill}%
}%
\begin{pgfscope}%
\pgfsys@transformshift{1.410397in}{0.288642in}%
\pgfsys@useobject{currentmarker}{}%
\end{pgfscope}%
\end{pgfscope}%
\begin{pgfscope}%
\definecolor{textcolor}{rgb}{0.000000,0.000000,0.000000}%
\pgfsetstrokecolor{textcolor}%
\pgfsetfillcolor{textcolor}%
\pgftext[x=1.410397in,y=0.191419in,,top]{\color{textcolor}\fontsize{7.000000}{8.400000}\selectfont EFPA}%
\end{pgfscope}%
\begin{pgfscope}%
\pgfsetbuttcap%
\pgfsetroundjoin%
\definecolor{currentfill}{rgb}{0.000000,0.000000,0.000000}%
\pgfsetfillcolor{currentfill}%
\pgfsetlinewidth{0.803000pt}%
\definecolor{currentstroke}{rgb}{0.000000,0.000000,0.000000}%
\pgfsetstrokecolor{currentstroke}%
\pgfsetdash{}{0pt}%
\pgfsys@defobject{currentmarker}{\pgfqpoint{0.000000in}{-0.048611in}}{\pgfqpoint{0.000000in}{0.000000in}}{%
\pgfpathmoveto{\pgfqpoint{0.000000in}{0.000000in}}%
\pgfpathlineto{\pgfqpoint{0.000000in}{-0.048611in}}%
\pgfusepath{stroke,fill}%
}%
\begin{pgfscope}%
\pgfsys@transformshift{1.715179in}{0.288642in}%
\pgfsys@useobject{currentmarker}{}%
\end{pgfscope}%
\end{pgfscope}%
\begin{pgfscope}%
\definecolor{textcolor}{rgb}{0.000000,0.000000,0.000000}%
\pgfsetstrokecolor{textcolor}%
\pgfsetfillcolor{textcolor}%
\pgftext[x=1.715179in,y=0.191419in,,top]{\color{textcolor}\fontsize{7.000000}{8.400000}\selectfont LTL}%
\end{pgfscope}%
\begin{pgfscope}%
\pgfsetbuttcap%
\pgfsetroundjoin%
\definecolor{currentfill}{rgb}{0.000000,0.000000,0.000000}%
\pgfsetfillcolor{currentfill}%
\pgfsetlinewidth{0.803000pt}%
\definecolor{currentstroke}{rgb}{0.000000,0.000000,0.000000}%
\pgfsetstrokecolor{currentstroke}%
\pgfsetdash{}{0pt}%
\pgfsys@defobject{currentmarker}{\pgfqpoint{-0.048611in}{0.000000in}}{\pgfqpoint{-0.000000in}{0.000000in}}{%
\pgfpathmoveto{\pgfqpoint{-0.000000in}{0.000000in}}%
\pgfpathlineto{\pgfqpoint{-0.048611in}{0.000000in}}%
\pgfusepath{stroke,fill}%
}%
\begin{pgfscope}%
\pgfsys@transformshift{0.621012in}{0.288642in}%
\pgfsys@useobject{currentmarker}{}%
\end{pgfscope}%
\end{pgfscope}%
\begin{pgfscope}%
\definecolor{textcolor}{rgb}{0.000000,0.000000,0.000000}%
\pgfsetstrokecolor{textcolor}%
\pgfsetfillcolor{textcolor}%
\pgftext[x=0.468427in, y=0.254884in, left, base]{\color{textcolor}\fontsize{7.000000}{8.400000}\selectfont \(\displaystyle {0}\)}%
\end{pgfscope}%
\begin{pgfscope}%
\pgfsetbuttcap%
\pgfsetroundjoin%
\definecolor{currentfill}{rgb}{0.000000,0.000000,0.000000}%
\pgfsetfillcolor{currentfill}%
\pgfsetlinewidth{0.803000pt}%
\definecolor{currentstroke}{rgb}{0.000000,0.000000,0.000000}%
\pgfsetstrokecolor{currentstroke}%
\pgfsetdash{}{0pt}%
\pgfsys@defobject{currentmarker}{\pgfqpoint{-0.048611in}{0.000000in}}{\pgfqpoint{-0.000000in}{0.000000in}}{%
\pgfpathmoveto{\pgfqpoint{-0.000000in}{0.000000in}}%
\pgfpathlineto{\pgfqpoint{-0.048611in}{0.000000in}}%
\pgfusepath{stroke,fill}%
}%
\begin{pgfscope}%
\pgfsys@transformshift{0.621012in}{0.523596in}%
\pgfsys@useobject{currentmarker}{}%
\end{pgfscope}%
\end{pgfscope}%
\begin{pgfscope}%
\definecolor{textcolor}{rgb}{0.000000,0.000000,0.000000}%
\pgfsetstrokecolor{textcolor}%
\pgfsetfillcolor{textcolor}%
\pgftext[x=0.302338in, y=0.489838in, left, base]{\color{textcolor}\fontsize{7.000000}{8.400000}\selectfont \(\displaystyle {5000}\)}%
\end{pgfscope}%
\begin{pgfscope}%
\pgfsetbuttcap%
\pgfsetroundjoin%
\definecolor{currentfill}{rgb}{0.000000,0.000000,0.000000}%
\pgfsetfillcolor{currentfill}%
\pgfsetlinewidth{0.803000pt}%
\definecolor{currentstroke}{rgb}{0.000000,0.000000,0.000000}%
\pgfsetstrokecolor{currentstroke}%
\pgfsetdash{}{0pt}%
\pgfsys@defobject{currentmarker}{\pgfqpoint{-0.048611in}{0.000000in}}{\pgfqpoint{-0.000000in}{0.000000in}}{%
\pgfpathmoveto{\pgfqpoint{-0.000000in}{0.000000in}}%
\pgfpathlineto{\pgfqpoint{-0.048611in}{0.000000in}}%
\pgfusepath{stroke,fill}%
}%
\begin{pgfscope}%
\pgfsys@transformshift{0.621012in}{0.758550in}%
\pgfsys@useobject{currentmarker}{}%
\end{pgfscope}%
\end{pgfscope}%
\begin{pgfscope}%
\definecolor{textcolor}{rgb}{0.000000,0.000000,0.000000}%
\pgfsetstrokecolor{textcolor}%
\pgfsetfillcolor{textcolor}%
\pgftext[x=0.246975in, y=0.724792in, left, base]{\color{textcolor}\fontsize{7.000000}{8.400000}\selectfont \(\displaystyle {10000}\)}%
\end{pgfscope}%
\begin{pgfscope}%
\pgfsetbuttcap%
\pgfsetroundjoin%
\definecolor{currentfill}{rgb}{0.000000,0.000000,0.000000}%
\pgfsetfillcolor{currentfill}%
\pgfsetlinewidth{0.803000pt}%
\definecolor{currentstroke}{rgb}{0.000000,0.000000,0.000000}%
\pgfsetstrokecolor{currentstroke}%
\pgfsetdash{}{0pt}%
\pgfsys@defobject{currentmarker}{\pgfqpoint{-0.048611in}{0.000000in}}{\pgfqpoint{-0.000000in}{0.000000in}}{%
\pgfpathmoveto{\pgfqpoint{-0.000000in}{0.000000in}}%
\pgfpathlineto{\pgfqpoint{-0.048611in}{0.000000in}}%
\pgfusepath{stroke,fill}%
}%
\begin{pgfscope}%
\pgfsys@transformshift{0.621012in}{0.993504in}%
\pgfsys@useobject{currentmarker}{}%
\end{pgfscope}%
\end{pgfscope}%
\begin{pgfscope}%
\definecolor{textcolor}{rgb}{0.000000,0.000000,0.000000}%
\pgfsetstrokecolor{textcolor}%
\pgfsetfillcolor{textcolor}%
\pgftext[x=0.246975in, y=0.959746in, left, base]{\color{textcolor}\fontsize{7.000000}{8.400000}\selectfont \(\displaystyle {15000}\)}%
\end{pgfscope}%
\begin{pgfscope}%
\pgfsetbuttcap%
\pgfsetroundjoin%
\definecolor{currentfill}{rgb}{0.000000,0.000000,0.000000}%
\pgfsetfillcolor{currentfill}%
\pgfsetlinewidth{0.803000pt}%
\definecolor{currentstroke}{rgb}{0.000000,0.000000,0.000000}%
\pgfsetstrokecolor{currentstroke}%
\pgfsetdash{}{0pt}%
\pgfsys@defobject{currentmarker}{\pgfqpoint{-0.048611in}{0.000000in}}{\pgfqpoint{-0.000000in}{0.000000in}}{%
\pgfpathmoveto{\pgfqpoint{-0.000000in}{0.000000in}}%
\pgfpathlineto{\pgfqpoint{-0.048611in}{0.000000in}}%
\pgfusepath{stroke,fill}%
}%
\begin{pgfscope}%
\pgfsys@transformshift{0.621012in}{1.228458in}%
\pgfsys@useobject{currentmarker}{}%
\end{pgfscope}%
\end{pgfscope}%
\begin{pgfscope}%
\definecolor{textcolor}{rgb}{0.000000,0.000000,0.000000}%
\pgfsetstrokecolor{textcolor}%
\pgfsetfillcolor{textcolor}%
\pgftext[x=0.246975in, y=1.194700in, left, base]{\color{textcolor}\fontsize{7.000000}{8.400000}\selectfont \(\displaystyle {20000}\)}%
\end{pgfscope}%
\begin{pgfscope}%
\definecolor{textcolor}{rgb}{0.000000,0.000000,0.000000}%
\pgfsetstrokecolor{textcolor}%
\pgfsetfillcolor{textcolor}%
\pgftext[x=0.191419in,y=0.841821in,,bottom,rotate=90.000000]{\color{textcolor}\fontsize{7.000000}{8.400000}\selectfont Number of operations}%
\end{pgfscope}%
\begin{pgfscope}%
\pgfsetrectcap%
\pgfsetmiterjoin%
\pgfsetlinewidth{0.803000pt}%
\definecolor{currentstroke}{rgb}{0.000000,0.000000,0.000000}%
\pgfsetstrokecolor{currentstroke}%
\pgfsetdash{}{0pt}%
\pgfpathmoveto{\pgfqpoint{0.621012in}{0.288642in}}%
\pgfpathlineto{\pgfqpoint{0.621012in}{1.395000in}}%
\pgfusepath{stroke}%
\end{pgfscope}%
\begin{pgfscope}%
\pgfsetrectcap%
\pgfsetmiterjoin%
\pgfsetlinewidth{0.803000pt}%
\definecolor{currentstroke}{rgb}{0.000000,0.000000,0.000000}%
\pgfsetstrokecolor{currentstroke}%
\pgfsetdash{}{0pt}%
\pgfpathmoveto{\pgfqpoint{1.895000in}{0.288642in}}%
\pgfpathlineto{\pgfqpoint{1.895000in}{1.395000in}}%
\pgfusepath{stroke}%
\end{pgfscope}%
\begin{pgfscope}%
\pgfsetrectcap%
\pgfsetmiterjoin%
\pgfsetlinewidth{0.803000pt}%
\definecolor{currentstroke}{rgb}{0.000000,0.000000,0.000000}%
\pgfsetstrokecolor{currentstroke}%
\pgfsetdash{}{0pt}%
\pgfpathmoveto{\pgfqpoint{0.621012in}{0.288642in}}%
\pgfpathlineto{\pgfqpoint{1.895000in}{0.288642in}}%
\pgfusepath{stroke}%
\end{pgfscope}%
\begin{pgfscope}%
\pgfsetrectcap%
\pgfsetmiterjoin%
\pgfsetlinewidth{0.803000pt}%
\definecolor{currentstroke}{rgb}{0.000000,0.000000,0.000000}%
\pgfsetstrokecolor{currentstroke}%
\pgfsetdash{}{0pt}%
\pgfpathmoveto{\pgfqpoint{0.621012in}{1.395000in}}%
\pgfpathlineto{\pgfqpoint{1.895000in}{1.395000in}}%
\pgfusepath{stroke}%
\end{pgfscope}%
\end{pgfpicture}%
\makeatother%
\endgroup%

%% file: graphics/SR.pgf
%% Creator: Matplotlib, PGF backend
%%
%% To include the figure in your LaTeX document, write
%%   \input{<filename>.pgf}
%%
%% Make sure the required packages are loaded in your preamble
%%   \usepackage{pgf}
%%
%% Figures using additional raster images can only be included by \input if
%% they are in the same directory as the main LaTeX file. For loading figures
%% from other directories you can use the `import` package
%% and then include the figures with
%%   \import{<path to file>}{<filename>.pgf}
%%
%% Matplotlib used the following preamble
%%
\begingroup%
\makeatletter%
\begin{pgfpicture}%
\pgfpathrectangle{\pgfpointorigin}{\pgfqpoint{2.000000in}{1.500000in}}%
\pgfusepath{use as bounding box, clip}%
\begin{pgfscope}%
\pgfsetbuttcap%
\pgfsetmiterjoin%
\definecolor{currentfill}{rgb}{1.000000,1.000000,1.000000}%
\pgfsetfillcolor{currentfill}%
\pgfsetlinewidth{0.000000pt}%
\definecolor{currentstroke}{rgb}{1.000000,1.000000,1.000000}%
\pgfsetstrokecolor{currentstroke}%
\pgfsetdash{}{0pt}%
\pgfpathmoveto{\pgfqpoint{0.000000in}{0.000000in}}%
\pgfpathlineto{\pgfqpoint{2.000000in}{0.000000in}}%
\pgfpathlineto{\pgfqpoint{2.000000in}{1.500000in}}%
\pgfpathlineto{\pgfqpoint{0.000000in}{1.500000in}}%
\pgfpathclose%
\pgfusepath{fill}%
\end{pgfscope}%
\begin{pgfscope}%
\pgfsetbuttcap%
\pgfsetmiterjoin%
\definecolor{currentfill}{rgb}{1.000000,1.000000,1.000000}%
\pgfsetfillcolor{currentfill}%
\pgfsetlinewidth{0.000000pt}%
\definecolor{currentstroke}{rgb}{0.000000,0.000000,0.000000}%
\pgfsetstrokecolor{currentstroke}%
\pgfsetstrokeopacity{0.000000}%
\pgfsetdash{}{0pt}%
\pgfpathmoveto{\pgfqpoint{0.565649in}{0.288642in}}%
\pgfpathlineto{\pgfqpoint{1.895000in}{0.288642in}}%
\pgfpathlineto{\pgfqpoint{1.895000in}{1.395000in}}%
\pgfpathlineto{\pgfqpoint{0.565649in}{1.395000in}}%
\pgfpathclose%
\pgfusepath{fill}%
\end{pgfscope}%
\begin{pgfscope}%
\pgfpathrectangle{\pgfqpoint{0.565649in}{0.288642in}}{\pgfqpoint{1.329351in}{1.106358in}}%
\pgfusepath{clip}%
\pgfsetbuttcap%
\pgfsetmiterjoin%
\definecolor{currentfill}{rgb}{0.000000,1.000000,1.000000}%
\pgfsetfillcolor{currentfill}%
\pgfsetlinewidth{0.000000pt}%
\definecolor{currentstroke}{rgb}{0.000000,0.000000,0.000000}%
\pgfsetstrokecolor{currentstroke}%
\pgfsetstrokeopacity{0.000000}%
\pgfsetdash{}{0pt}%
\pgfpathmoveto{\pgfqpoint{0.626074in}{0.288642in}}%
\pgfpathlineto{\pgfqpoint{0.880495in}{0.288642in}}%
\pgfpathlineto{\pgfqpoint{0.880495in}{1.081738in}}%
\pgfpathlineto{\pgfqpoint{0.626074in}{1.081738in}}%
\pgfpathclose%
\pgfusepath{fill}%
\end{pgfscope}%
\begin{pgfscope}%
\pgfpathrectangle{\pgfqpoint{0.565649in}{0.288642in}}{\pgfqpoint{1.329351in}{1.106358in}}%
\pgfusepath{clip}%
\pgfsetbuttcap%
\pgfsetmiterjoin%
\definecolor{currentfill}{rgb}{0.000000,0.000000,1.000000}%
\pgfsetfillcolor{currentfill}%
\pgfsetlinewidth{0.000000pt}%
\definecolor{currentstroke}{rgb}{0.000000,0.000000,0.000000}%
\pgfsetstrokecolor{currentstroke}%
\pgfsetstrokeopacity{0.000000}%
\pgfsetdash{}{0pt}%
\pgfpathmoveto{\pgfqpoint{0.944101in}{0.288642in}}%
\pgfpathlineto{\pgfqpoint{1.198522in}{0.288642in}}%
\pgfpathlineto{\pgfqpoint{1.198522in}{1.133047in}}%
\pgfpathlineto{\pgfqpoint{0.944101in}{1.133047in}}%
\pgfpathclose%
\pgfusepath{fill}%
\end{pgfscope}%
\begin{pgfscope}%
\pgfpathrectangle{\pgfqpoint{0.565649in}{0.288642in}}{\pgfqpoint{1.329351in}{1.106358in}}%
\pgfusepath{clip}%
\pgfsetbuttcap%
\pgfsetmiterjoin%
\definecolor{currentfill}{rgb}{1.000000,0.752941,0.796078}%
\pgfsetfillcolor{currentfill}%
\pgfsetlinewidth{0.000000pt}%
\definecolor{currentstroke}{rgb}{0.000000,0.000000,0.000000}%
\pgfsetstrokecolor{currentstroke}%
\pgfsetstrokeopacity{0.000000}%
\pgfsetdash{}{0pt}%
\pgfpathmoveto{\pgfqpoint{1.262127in}{0.288642in}}%
\pgfpathlineto{\pgfqpoint{1.516548in}{0.288642in}}%
\pgfpathlineto{\pgfqpoint{1.516548in}{1.342316in}}%
\pgfpathlineto{\pgfqpoint{1.262127in}{1.342316in}}%
\pgfpathclose%
\pgfusepath{fill}%
\end{pgfscope}%
\begin{pgfscope}%
\pgfpathrectangle{\pgfqpoint{0.565649in}{0.288642in}}{\pgfqpoint{1.329351in}{1.106358in}}%
\pgfusepath{clip}%
\pgfsetbuttcap%
\pgfsetmiterjoin%
\definecolor{currentfill}{rgb}{0.000000,0.501961,0.000000}%
\pgfsetfillcolor{currentfill}%
\pgfsetlinewidth{0.000000pt}%
\definecolor{currentstroke}{rgb}{0.000000,0.000000,0.000000}%
\pgfsetstrokecolor{currentstroke}%
\pgfsetstrokeopacity{0.000000}%
\pgfsetdash{}{0pt}%
\pgfpathmoveto{\pgfqpoint{1.580154in}{0.288642in}}%
\pgfpathlineto{\pgfqpoint{1.834575in}{0.288642in}}%
\pgfpathlineto{\pgfqpoint{1.834575in}{1.322892in}}%
\pgfpathlineto{\pgfqpoint{1.580154in}{1.322892in}}%
\pgfpathclose%
\pgfusepath{fill}%
\end{pgfscope}%
\begin{pgfscope}%
\pgfsetbuttcap%
\pgfsetroundjoin%
\definecolor{currentfill}{rgb}{0.000000,0.000000,0.000000}%
\pgfsetfillcolor{currentfill}%
\pgfsetlinewidth{0.803000pt}%
\definecolor{currentstroke}{rgb}{0.000000,0.000000,0.000000}%
\pgfsetstrokecolor{currentstroke}%
\pgfsetdash{}{0pt}%
\pgfsys@defobject{currentmarker}{\pgfqpoint{0.000000in}{-0.048611in}}{\pgfqpoint{0.000000in}{0.000000in}}{%
\pgfpathmoveto{\pgfqpoint{0.000000in}{0.000000in}}%
\pgfpathlineto{\pgfqpoint{0.000000in}{-0.048611in}}%
\pgfusepath{stroke,fill}%
}%
\begin{pgfscope}%
\pgfsys@transformshift{0.753285in}{0.288642in}%
\pgfsys@useobject{currentmarker}{}%
\end{pgfscope}%
\end{pgfscope}%
\begin{pgfscope}%
\definecolor{textcolor}{rgb}{0.000000,0.000000,0.000000}%
\pgfsetstrokecolor{textcolor}%
\pgfsetfillcolor{textcolor}%
\pgftext[x=0.753285in,y=0.191419in,,top]{\color{textcolor}\fontsize{7.000000}{8.400000}\selectfont PV-OSIMr}%
\end{pgfscope}%
\begin{pgfscope}%
\pgfsetbuttcap%
\pgfsetroundjoin%
\definecolor{currentfill}{rgb}{0.000000,0.000000,0.000000}%
\pgfsetfillcolor{currentfill}%
\pgfsetlinewidth{0.803000pt}%
\definecolor{currentstroke}{rgb}{0.000000,0.000000,0.000000}%
\pgfsetstrokecolor{currentstroke}%
\pgfsetdash{}{0pt}%
\pgfsys@defobject{currentmarker}{\pgfqpoint{0.000000in}{-0.048611in}}{\pgfqpoint{0.000000in}{0.000000in}}{%
\pgfpathmoveto{\pgfqpoint{0.000000in}{0.000000in}}%
\pgfpathlineto{\pgfqpoint{0.000000in}{-0.048611in}}%
\pgfusepath{stroke,fill}%
}%
\begin{pgfscope}%
\pgfsys@transformshift{1.071311in}{0.288642in}%
\pgfsys@useobject{currentmarker}{}%
\end{pgfscope}%
\end{pgfscope}%
\begin{pgfscope}%
\definecolor{textcolor}{rgb}{0.000000,0.000000,0.000000}%
\pgfsetstrokecolor{textcolor}%
\pgfsetfillcolor{textcolor}%
\pgftext[x=1.071311in,y=0.191419in,,top]{\color{textcolor}\fontsize{7.000000}{8.400000}\selectfont PV}%
\end{pgfscope}%
\begin{pgfscope}%
\pgfsetbuttcap%
\pgfsetroundjoin%
\definecolor{currentfill}{rgb}{0.000000,0.000000,0.000000}%
\pgfsetfillcolor{currentfill}%
\pgfsetlinewidth{0.803000pt}%
\definecolor{currentstroke}{rgb}{0.000000,0.000000,0.000000}%
\pgfsetstrokecolor{currentstroke}%
\pgfsetdash{}{0pt}%
\pgfsys@defobject{currentmarker}{\pgfqpoint{0.000000in}{-0.048611in}}{\pgfqpoint{0.000000in}{0.000000in}}{%
\pgfpathmoveto{\pgfqpoint{0.000000in}{0.000000in}}%
\pgfpathlineto{\pgfqpoint{0.000000in}{-0.048611in}}%
\pgfusepath{stroke,fill}%
}%
\begin{pgfscope}%
\pgfsys@transformshift{1.389338in}{0.288642in}%
\pgfsys@useobject{currentmarker}{}%
\end{pgfscope}%
\end{pgfscope}%
\begin{pgfscope}%
\definecolor{textcolor}{rgb}{0.000000,0.000000,0.000000}%
\pgfsetstrokecolor{textcolor}%
\pgfsetfillcolor{textcolor}%
\pgftext[x=1.389338in,y=0.191419in,,top]{\color{textcolor}\fontsize{7.000000}{8.400000}\selectfont EFPA}%
\end{pgfscope}%
\begin{pgfscope}%
\pgfsetbuttcap%
\pgfsetroundjoin%
\definecolor{currentfill}{rgb}{0.000000,0.000000,0.000000}%
\pgfsetfillcolor{currentfill}%
\pgfsetlinewidth{0.803000pt}%
\definecolor{currentstroke}{rgb}{0.000000,0.000000,0.000000}%
\pgfsetstrokecolor{currentstroke}%
\pgfsetdash{}{0pt}%
\pgfsys@defobject{currentmarker}{\pgfqpoint{0.000000in}{-0.048611in}}{\pgfqpoint{0.000000in}{0.000000in}}{%
\pgfpathmoveto{\pgfqpoint{0.000000in}{0.000000in}}%
\pgfpathlineto{\pgfqpoint{0.000000in}{-0.048611in}}%
\pgfusepath{stroke,fill}%
}%
\begin{pgfscope}%
\pgfsys@transformshift{1.707364in}{0.288642in}%
\pgfsys@useobject{currentmarker}{}%
\end{pgfscope}%
\end{pgfscope}%
\begin{pgfscope}%
\definecolor{textcolor}{rgb}{0.000000,0.000000,0.000000}%
\pgfsetstrokecolor{textcolor}%
\pgfsetfillcolor{textcolor}%
\pgftext[x=1.707364in,y=0.191419in,,top]{\color{textcolor}\fontsize{7.000000}{8.400000}\selectfont LTL}%
\end{pgfscope}%
\begin{pgfscope}%
\pgfsetbuttcap%
\pgfsetroundjoin%
\definecolor{currentfill}{rgb}{0.000000,0.000000,0.000000}%
\pgfsetfillcolor{currentfill}%
\pgfsetlinewidth{0.803000pt}%
\definecolor{currentstroke}{rgb}{0.000000,0.000000,0.000000}%
\pgfsetstrokecolor{currentstroke}%
\pgfsetdash{}{0pt}%
\pgfsys@defobject{currentmarker}{\pgfqpoint{-0.048611in}{0.000000in}}{\pgfqpoint{-0.000000in}{0.000000in}}{%
\pgfpathmoveto{\pgfqpoint{-0.000000in}{0.000000in}}%
\pgfpathlineto{\pgfqpoint{-0.048611in}{0.000000in}}%
\pgfusepath{stroke,fill}%
}%
\begin{pgfscope}%
\pgfsys@transformshift{0.565649in}{0.288642in}%
\pgfsys@useobject{currentmarker}{}%
\end{pgfscope}%
\end{pgfscope}%
\begin{pgfscope}%
\definecolor{textcolor}{rgb}{0.000000,0.000000,0.000000}%
\pgfsetstrokecolor{textcolor}%
\pgfsetfillcolor{textcolor}%
\pgftext[x=0.413064in, y=0.254884in, left, base]{\color{textcolor}\fontsize{7.000000}{8.400000}\selectfont \(\displaystyle {0}\)}%
\end{pgfscope}%
\begin{pgfscope}%
\pgfsetbuttcap%
\pgfsetroundjoin%
\definecolor{currentfill}{rgb}{0.000000,0.000000,0.000000}%
\pgfsetfillcolor{currentfill}%
\pgfsetlinewidth{0.803000pt}%
\definecolor{currentstroke}{rgb}{0.000000,0.000000,0.000000}%
\pgfsetstrokecolor{currentstroke}%
\pgfsetdash{}{0pt}%
\pgfsys@defobject{currentmarker}{\pgfqpoint{-0.048611in}{0.000000in}}{\pgfqpoint{-0.000000in}{0.000000in}}{%
\pgfpathmoveto{\pgfqpoint{-0.000000in}{0.000000in}}%
\pgfpathlineto{\pgfqpoint{-0.048611in}{0.000000in}}%
\pgfusepath{stroke,fill}%
}%
\begin{pgfscope}%
\pgfsys@transformshift{0.565649in}{0.532972in}%
\pgfsys@useobject{currentmarker}{}%
\end{pgfscope}%
\end{pgfscope}%
\begin{pgfscope}%
\definecolor{textcolor}{rgb}{0.000000,0.000000,0.000000}%
\pgfsetstrokecolor{textcolor}%
\pgfsetfillcolor{textcolor}%
\pgftext[x=0.246975in, y=0.499214in, left, base]{\color{textcolor}\fontsize{7.000000}{8.400000}\selectfont \(\displaystyle {2000}\)}%
\end{pgfscope}%
\begin{pgfscope}%
\pgfsetbuttcap%
\pgfsetroundjoin%
\definecolor{currentfill}{rgb}{0.000000,0.000000,0.000000}%
\pgfsetfillcolor{currentfill}%
\pgfsetlinewidth{0.803000pt}%
\definecolor{currentstroke}{rgb}{0.000000,0.000000,0.000000}%
\pgfsetstrokecolor{currentstroke}%
\pgfsetdash{}{0pt}%
\pgfsys@defobject{currentmarker}{\pgfqpoint{-0.048611in}{0.000000in}}{\pgfqpoint{-0.000000in}{0.000000in}}{%
\pgfpathmoveto{\pgfqpoint{-0.000000in}{0.000000in}}%
\pgfpathlineto{\pgfqpoint{-0.048611in}{0.000000in}}%
\pgfusepath{stroke,fill}%
}%
\begin{pgfscope}%
\pgfsys@transformshift{0.565649in}{0.777302in}%
\pgfsys@useobject{currentmarker}{}%
\end{pgfscope}%
\end{pgfscope}%
\begin{pgfscope}%
\definecolor{textcolor}{rgb}{0.000000,0.000000,0.000000}%
\pgfsetstrokecolor{textcolor}%
\pgfsetfillcolor{textcolor}%
\pgftext[x=0.246975in, y=0.743545in, left, base]{\color{textcolor}\fontsize{7.000000}{8.400000}\selectfont \(\displaystyle {4000}\)}%
\end{pgfscope}%
\begin{pgfscope}%
\pgfsetbuttcap%
\pgfsetroundjoin%
\definecolor{currentfill}{rgb}{0.000000,0.000000,0.000000}%
\pgfsetfillcolor{currentfill}%
\pgfsetlinewidth{0.803000pt}%
\definecolor{currentstroke}{rgb}{0.000000,0.000000,0.000000}%
\pgfsetstrokecolor{currentstroke}%
\pgfsetdash{}{0pt}%
\pgfsys@defobject{currentmarker}{\pgfqpoint{-0.048611in}{0.000000in}}{\pgfqpoint{-0.000000in}{0.000000in}}{%
\pgfpathmoveto{\pgfqpoint{-0.000000in}{0.000000in}}%
\pgfpathlineto{\pgfqpoint{-0.048611in}{0.000000in}}%
\pgfusepath{stroke,fill}%
}%
\begin{pgfscope}%
\pgfsys@transformshift{0.565649in}{1.021633in}%
\pgfsys@useobject{currentmarker}{}%
\end{pgfscope}%
\end{pgfscope}%
\begin{pgfscope}%
\definecolor{textcolor}{rgb}{0.000000,0.000000,0.000000}%
\pgfsetstrokecolor{textcolor}%
\pgfsetfillcolor{textcolor}%
\pgftext[x=0.246975in, y=0.987875in, left, base]{\color{textcolor}\fontsize{7.000000}{8.400000}\selectfont \(\displaystyle {6000}\)}%
\end{pgfscope}%
\begin{pgfscope}%
\pgfsetbuttcap%
\pgfsetroundjoin%
\definecolor{currentfill}{rgb}{0.000000,0.000000,0.000000}%
\pgfsetfillcolor{currentfill}%
\pgfsetlinewidth{0.803000pt}%
\definecolor{currentstroke}{rgb}{0.000000,0.000000,0.000000}%
\pgfsetstrokecolor{currentstroke}%
\pgfsetdash{}{0pt}%
\pgfsys@defobject{currentmarker}{\pgfqpoint{-0.048611in}{0.000000in}}{\pgfqpoint{-0.000000in}{0.000000in}}{%
\pgfpathmoveto{\pgfqpoint{-0.000000in}{0.000000in}}%
\pgfpathlineto{\pgfqpoint{-0.048611in}{0.000000in}}%
\pgfusepath{stroke,fill}%
}%
\begin{pgfscope}%
\pgfsys@transformshift{0.565649in}{1.265963in}%
\pgfsys@useobject{currentmarker}{}%
\end{pgfscope}%
\end{pgfscope}%
\begin{pgfscope}%
\definecolor{textcolor}{rgb}{0.000000,0.000000,0.000000}%
\pgfsetstrokecolor{textcolor}%
\pgfsetfillcolor{textcolor}%
\pgftext[x=0.246975in, y=1.232205in, left, base]{\color{textcolor}\fontsize{7.000000}{8.400000}\selectfont \(\displaystyle {8000}\)}%
\end{pgfscope}%
\begin{pgfscope}%
\definecolor{textcolor}{rgb}{0.000000,0.000000,0.000000}%
\pgfsetstrokecolor{textcolor}%
\pgfsetfillcolor{textcolor}%
\pgftext[x=0.191419in,y=0.841821in,,bottom,rotate=90.000000]{\color{textcolor}\fontsize{7.000000}{8.400000}\selectfont Number of operations}%
\end{pgfscope}%
\begin{pgfscope}%
\pgfsetrectcap%
\pgfsetmiterjoin%
\pgfsetlinewidth{0.803000pt}%
\definecolor{currentstroke}{rgb}{0.000000,0.000000,0.000000}%
\pgfsetstrokecolor{currentstroke}%
\pgfsetdash{}{0pt}%
\pgfpathmoveto{\pgfqpoint{0.565649in}{0.288642in}}%
\pgfpathlineto{\pgfqpoint{0.565649in}{1.395000in}}%
\pgfusepath{stroke}%
\end{pgfscope}%
\begin{pgfscope}%
\pgfsetrectcap%
\pgfsetmiterjoin%
\pgfsetlinewidth{0.803000pt}%
\definecolor{currentstroke}{rgb}{0.000000,0.000000,0.000000}%
\pgfsetstrokecolor{currentstroke}%
\pgfsetdash{}{0pt}%
\pgfpathmoveto{\pgfqpoint{1.895000in}{0.288642in}}%
\pgfpathlineto{\pgfqpoint{1.895000in}{1.395000in}}%
\pgfusepath{stroke}%
\end{pgfscope}%
\begin{pgfscope}%
\pgfsetrectcap%
\pgfsetmiterjoin%
\pgfsetlinewidth{0.803000pt}%
\definecolor{currentstroke}{rgb}{0.000000,0.000000,0.000000}%
\pgfsetstrokecolor{currentstroke}%
\pgfsetdash{}{0pt}%
\pgfpathmoveto{\pgfqpoint{0.565649in}{0.288642in}}%
\pgfpathlineto{\pgfqpoint{1.895000in}{0.288642in}}%
\pgfusepath{stroke}%
\end{pgfscope}%
\begin{pgfscope}%
\pgfsetrectcap%
\pgfsetmiterjoin%
\pgfsetlinewidth{0.803000pt}%
\definecolor{currentstroke}{rgb}{0.000000,0.000000,0.000000}%
\pgfsetstrokecolor{currentstroke}%
\pgfsetdash{}{0pt}%
\pgfpathmoveto{\pgfqpoint{0.565649in}{1.395000in}}%
\pgfpathlineto{\pgfqpoint{1.895000in}{1.395000in}}%
\pgfusepath{stroke}%
\end{pgfscope}%
\end{pgfpicture}%
\makeatother%
\endgroup%

%% file: graphics/atlas_left_hand.pgf
%% Creator: Matplotlib, PGF backend
%%
%% To include the figure in your LaTeX document, write
%%   \input{<filename>.pgf}
%%
%% Make sure the required packages are loaded in your preamble
%%   \usepackage{pgf}
%%
%% Figures using additional raster images can only be included by \input if
%% they are in the same directory as the main LaTeX file. For loading figures
%% from other directories you can use the `import` package
%% and then include the figures with
%%   \import{<path to file>}{<filename>.pgf}
%%
%% Matplotlib used the following preamble
%%
\begingroup%
\makeatletter%
\begin{pgfpicture}%
\pgfpathrectangle{\pgfpointorigin}{\pgfqpoint{2.000000in}{1.500000in}}%
\pgfusepath{use as bounding box, clip}%
\begin{pgfscope}%
\pgfsetbuttcap%
\pgfsetmiterjoin%
\definecolor{currentfill}{rgb}{1.000000,1.000000,1.000000}%
\pgfsetfillcolor{currentfill}%
\pgfsetlinewidth{0.000000pt}%
\definecolor{currentstroke}{rgb}{1.000000,1.000000,1.000000}%
\pgfsetstrokecolor{currentstroke}%
\pgfsetdash{}{0pt}%
\pgfpathmoveto{\pgfqpoint{0.000000in}{0.000000in}}%
\pgfpathlineto{\pgfqpoint{2.000000in}{0.000000in}}%
\pgfpathlineto{\pgfqpoint{2.000000in}{1.500000in}}%
\pgfpathlineto{\pgfqpoint{0.000000in}{1.500000in}}%
\pgfpathclose%
\pgfusepath{fill}%
\end{pgfscope}%
\begin{pgfscope}%
\pgfsetbuttcap%
\pgfsetmiterjoin%
\definecolor{currentfill}{rgb}{1.000000,1.000000,1.000000}%
\pgfsetfillcolor{currentfill}%
\pgfsetlinewidth{0.000000pt}%
\definecolor{currentstroke}{rgb}{0.000000,0.000000,0.000000}%
\pgfsetstrokecolor{currentstroke}%
\pgfsetstrokeopacity{0.000000}%
\pgfsetdash{}{0pt}%
\pgfpathmoveto{\pgfqpoint{0.621012in}{0.288642in}}%
\pgfpathlineto{\pgfqpoint{1.895000in}{0.288642in}}%
\pgfpathlineto{\pgfqpoint{1.895000in}{1.395000in}}%
\pgfpathlineto{\pgfqpoint{0.621012in}{1.395000in}}%
\pgfpathclose%
\pgfusepath{fill}%
\end{pgfscope}%
\begin{pgfscope}%
\pgfpathrectangle{\pgfqpoint{0.621012in}{0.288642in}}{\pgfqpoint{1.273988in}{1.106358in}}%
\pgfusepath{clip}%
\pgfsetbuttcap%
\pgfsetmiterjoin%
\definecolor{currentfill}{rgb}{0.000000,1.000000,1.000000}%
\pgfsetfillcolor{currentfill}%
\pgfsetlinewidth{0.000000pt}%
\definecolor{currentstroke}{rgb}{0.000000,0.000000,0.000000}%
\pgfsetstrokecolor{currentstroke}%
\pgfsetstrokeopacity{0.000000}%
\pgfsetdash{}{0pt}%
\pgfpathmoveto{\pgfqpoint{0.678920in}{0.288642in}}%
\pgfpathlineto{\pgfqpoint{0.922746in}{0.288642in}}%
\pgfpathlineto{\pgfqpoint{0.922746in}{0.644942in}}%
\pgfpathlineto{\pgfqpoint{0.678920in}{0.644942in}}%
\pgfpathclose%
\pgfusepath{fill}%
\end{pgfscope}%
\begin{pgfscope}%
\pgfpathrectangle{\pgfqpoint{0.621012in}{0.288642in}}{\pgfqpoint{1.273988in}{1.106358in}}%
\pgfusepath{clip}%
\pgfsetbuttcap%
\pgfsetmiterjoin%
\definecolor{currentfill}{rgb}{0.000000,0.000000,1.000000}%
\pgfsetfillcolor{currentfill}%
\pgfsetlinewidth{0.000000pt}%
\definecolor{currentstroke}{rgb}{0.000000,0.000000,0.000000}%
\pgfsetstrokecolor{currentstroke}%
\pgfsetstrokeopacity{0.000000}%
\pgfsetdash{}{0pt}%
\pgfpathmoveto{\pgfqpoint{0.983702in}{0.288642in}}%
\pgfpathlineto{\pgfqpoint{1.227528in}{0.288642in}}%
\pgfpathlineto{\pgfqpoint{1.227528in}{0.774349in}}%
\pgfpathlineto{\pgfqpoint{0.983702in}{0.774349in}}%
\pgfpathclose%
\pgfusepath{fill}%
\end{pgfscope}%
\begin{pgfscope}%
\pgfpathrectangle{\pgfqpoint{0.621012in}{0.288642in}}{\pgfqpoint{1.273988in}{1.106358in}}%
\pgfusepath{clip}%
\pgfsetbuttcap%
\pgfsetmiterjoin%
\definecolor{currentfill}{rgb}{1.000000,0.752941,0.796078}%
\pgfsetfillcolor{currentfill}%
\pgfsetlinewidth{0.000000pt}%
\definecolor{currentstroke}{rgb}{0.000000,0.000000,0.000000}%
\pgfsetstrokecolor{currentstroke}%
\pgfsetstrokeopacity{0.000000}%
\pgfsetdash{}{0pt}%
\pgfpathmoveto{\pgfqpoint{1.288484in}{0.288642in}}%
\pgfpathlineto{\pgfqpoint{1.532310in}{0.288642in}}%
\pgfpathlineto{\pgfqpoint{1.532310in}{0.899607in}}%
\pgfpathlineto{\pgfqpoint{1.288484in}{0.899607in}}%
\pgfpathclose%
\pgfusepath{fill}%
\end{pgfscope}%
\begin{pgfscope}%
\pgfpathrectangle{\pgfqpoint{0.621012in}{0.288642in}}{\pgfqpoint{1.273988in}{1.106358in}}%
\pgfusepath{clip}%
\pgfsetbuttcap%
\pgfsetmiterjoin%
\definecolor{currentfill}{rgb}{0.000000,0.501961,0.000000}%
\pgfsetfillcolor{currentfill}%
\pgfsetlinewidth{0.000000pt}%
\definecolor{currentstroke}{rgb}{0.000000,0.000000,0.000000}%
\pgfsetstrokecolor{currentstroke}%
\pgfsetstrokeopacity{0.000000}%
\pgfsetdash{}{0pt}%
\pgfpathmoveto{\pgfqpoint{1.593266in}{0.288642in}}%
\pgfpathlineto{\pgfqpoint{1.837091in}{0.288642in}}%
\pgfpathlineto{\pgfqpoint{1.837091in}{1.342316in}}%
\pgfpathlineto{\pgfqpoint{1.593266in}{1.342316in}}%
\pgfpathclose%
\pgfusepath{fill}%
\end{pgfscope}%
\begin{pgfscope}%
\pgfsetbuttcap%
\pgfsetroundjoin%
\definecolor{currentfill}{rgb}{0.000000,0.000000,0.000000}%
\pgfsetfillcolor{currentfill}%
\pgfsetlinewidth{0.803000pt}%
\definecolor{currentstroke}{rgb}{0.000000,0.000000,0.000000}%
\pgfsetstrokecolor{currentstroke}%
\pgfsetdash{}{0pt}%
\pgfsys@defobject{currentmarker}{\pgfqpoint{0.000000in}{-0.048611in}}{\pgfqpoint{0.000000in}{0.000000in}}{%
\pgfpathmoveto{\pgfqpoint{0.000000in}{0.000000in}}%
\pgfpathlineto{\pgfqpoint{0.000000in}{-0.048611in}}%
\pgfusepath{stroke,fill}%
}%
\begin{pgfscope}%
\pgfsys@transformshift{0.800833in}{0.288642in}%
\pgfsys@useobject{currentmarker}{}%
\end{pgfscope}%
\end{pgfscope}%
\begin{pgfscope}%
\definecolor{textcolor}{rgb}{0.000000,0.000000,0.000000}%
\pgfsetstrokecolor{textcolor}%
\pgfsetfillcolor{textcolor}%
\pgftext[x=0.800833in,y=0.191419in,,top]{\color{textcolor}\fontsize{7.000000}{8.400000}\selectfont PV-OSIMr}%
\end{pgfscope}%
\begin{pgfscope}%
\pgfsetbuttcap%
\pgfsetroundjoin%
\definecolor{currentfill}{rgb}{0.000000,0.000000,0.000000}%
\pgfsetfillcolor{currentfill}%
\pgfsetlinewidth{0.803000pt}%
\definecolor{currentstroke}{rgb}{0.000000,0.000000,0.000000}%
\pgfsetstrokecolor{currentstroke}%
\pgfsetdash{}{0pt}%
\pgfsys@defobject{currentmarker}{\pgfqpoint{0.000000in}{-0.048611in}}{\pgfqpoint{0.000000in}{0.000000in}}{%
\pgfpathmoveto{\pgfqpoint{0.000000in}{0.000000in}}%
\pgfpathlineto{\pgfqpoint{0.000000in}{-0.048611in}}%
\pgfusepath{stroke,fill}%
}%
\begin{pgfscope}%
\pgfsys@transformshift{1.105615in}{0.288642in}%
\pgfsys@useobject{currentmarker}{}%
\end{pgfscope}%
\end{pgfscope}%
\begin{pgfscope}%
\definecolor{textcolor}{rgb}{0.000000,0.000000,0.000000}%
\pgfsetstrokecolor{textcolor}%
\pgfsetfillcolor{textcolor}%
\pgftext[x=1.105615in,y=0.191419in,,top]{\color{textcolor}\fontsize{7.000000}{8.400000}\selectfont PV}%
\end{pgfscope}%
\begin{pgfscope}%
\pgfsetbuttcap%
\pgfsetroundjoin%
\definecolor{currentfill}{rgb}{0.000000,0.000000,0.000000}%
\pgfsetfillcolor{currentfill}%
\pgfsetlinewidth{0.803000pt}%
\definecolor{currentstroke}{rgb}{0.000000,0.000000,0.000000}%
\pgfsetstrokecolor{currentstroke}%
\pgfsetdash{}{0pt}%
\pgfsys@defobject{currentmarker}{\pgfqpoint{0.000000in}{-0.048611in}}{\pgfqpoint{0.000000in}{0.000000in}}{%
\pgfpathmoveto{\pgfqpoint{0.000000in}{0.000000in}}%
\pgfpathlineto{\pgfqpoint{0.000000in}{-0.048611in}}%
\pgfusepath{stroke,fill}%
}%
\begin{pgfscope}%
\pgfsys@transformshift{1.410397in}{0.288642in}%
\pgfsys@useobject{currentmarker}{}%
\end{pgfscope}%
\end{pgfscope}%
\begin{pgfscope}%
\definecolor{textcolor}{rgb}{0.000000,0.000000,0.000000}%
\pgfsetstrokecolor{textcolor}%
\pgfsetfillcolor{textcolor}%
\pgftext[x=1.410397in,y=0.191419in,,top]{\color{textcolor}\fontsize{7.000000}{8.400000}\selectfont EFPA}%
\end{pgfscope}%
\begin{pgfscope}%
\pgfsetbuttcap%
\pgfsetroundjoin%
\definecolor{currentfill}{rgb}{0.000000,0.000000,0.000000}%
\pgfsetfillcolor{currentfill}%
\pgfsetlinewidth{0.803000pt}%
\definecolor{currentstroke}{rgb}{0.000000,0.000000,0.000000}%
\pgfsetstrokecolor{currentstroke}%
\pgfsetdash{}{0pt}%
\pgfsys@defobject{currentmarker}{\pgfqpoint{0.000000in}{-0.048611in}}{\pgfqpoint{0.000000in}{0.000000in}}{%
\pgfpathmoveto{\pgfqpoint{0.000000in}{0.000000in}}%
\pgfpathlineto{\pgfqpoint{0.000000in}{-0.048611in}}%
\pgfusepath{stroke,fill}%
}%
\begin{pgfscope}%
\pgfsys@transformshift{1.715179in}{0.288642in}%
\pgfsys@useobject{currentmarker}{}%
\end{pgfscope}%
\end{pgfscope}%
\begin{pgfscope}%
\definecolor{textcolor}{rgb}{0.000000,0.000000,0.000000}%
\pgfsetstrokecolor{textcolor}%
\pgfsetfillcolor{textcolor}%
\pgftext[x=1.715179in,y=0.191419in,,top]{\color{textcolor}\fontsize{7.000000}{8.400000}\selectfont LTL}%
\end{pgfscope}%
\begin{pgfscope}%
\pgfsetbuttcap%
\pgfsetroundjoin%
\definecolor{currentfill}{rgb}{0.000000,0.000000,0.000000}%
\pgfsetfillcolor{currentfill}%
\pgfsetlinewidth{0.803000pt}%
\definecolor{currentstroke}{rgb}{0.000000,0.000000,0.000000}%
\pgfsetstrokecolor{currentstroke}%
\pgfsetdash{}{0pt}%
\pgfsys@defobject{currentmarker}{\pgfqpoint{-0.048611in}{0.000000in}}{\pgfqpoint{-0.000000in}{0.000000in}}{%
\pgfpathmoveto{\pgfqpoint{-0.000000in}{0.000000in}}%
\pgfpathlineto{\pgfqpoint{-0.048611in}{0.000000in}}%
\pgfusepath{stroke,fill}%
}%
\begin{pgfscope}%
\pgfsys@transformshift{0.621012in}{0.288642in}%
\pgfsys@useobject{currentmarker}{}%
\end{pgfscope}%
\end{pgfscope}%
\begin{pgfscope}%
\definecolor{textcolor}{rgb}{0.000000,0.000000,0.000000}%
\pgfsetstrokecolor{textcolor}%
\pgfsetfillcolor{textcolor}%
\pgftext[x=0.468427in, y=0.254884in, left, base]{\color{textcolor}\fontsize{7.000000}{8.400000}\selectfont \(\displaystyle {0}\)}%
\end{pgfscope}%
\begin{pgfscope}%
\pgfsetbuttcap%
\pgfsetroundjoin%
\definecolor{currentfill}{rgb}{0.000000,0.000000,0.000000}%
\pgfsetfillcolor{currentfill}%
\pgfsetlinewidth{0.803000pt}%
\definecolor{currentstroke}{rgb}{0.000000,0.000000,0.000000}%
\pgfsetstrokecolor{currentstroke}%
\pgfsetdash{}{0pt}%
\pgfsys@defobject{currentmarker}{\pgfqpoint{-0.048611in}{0.000000in}}{\pgfqpoint{-0.000000in}{0.000000in}}{%
\pgfpathmoveto{\pgfqpoint{-0.000000in}{0.000000in}}%
\pgfpathlineto{\pgfqpoint{-0.048611in}{0.000000in}}%
\pgfusepath{stroke,fill}%
}%
\begin{pgfscope}%
\pgfsys@transformshift{0.621012in}{0.532666in}%
\pgfsys@useobject{currentmarker}{}%
\end{pgfscope}%
\end{pgfscope}%
\begin{pgfscope}%
\definecolor{textcolor}{rgb}{0.000000,0.000000,0.000000}%
\pgfsetstrokecolor{textcolor}%
\pgfsetfillcolor{textcolor}%
\pgftext[x=0.246975in, y=0.498909in, left, base]{\color{textcolor}\fontsize{7.000000}{8.400000}\selectfont \(\displaystyle {10000}\)}%
\end{pgfscope}%
\begin{pgfscope}%
\pgfsetbuttcap%
\pgfsetroundjoin%
\definecolor{currentfill}{rgb}{0.000000,0.000000,0.000000}%
\pgfsetfillcolor{currentfill}%
\pgfsetlinewidth{0.803000pt}%
\definecolor{currentstroke}{rgb}{0.000000,0.000000,0.000000}%
\pgfsetstrokecolor{currentstroke}%
\pgfsetdash{}{0pt}%
\pgfsys@defobject{currentmarker}{\pgfqpoint{-0.048611in}{0.000000in}}{\pgfqpoint{-0.000000in}{0.000000in}}{%
\pgfpathmoveto{\pgfqpoint{-0.000000in}{0.000000in}}%
\pgfpathlineto{\pgfqpoint{-0.048611in}{0.000000in}}%
\pgfusepath{stroke,fill}%
}%
\begin{pgfscope}%
\pgfsys@transformshift{0.621012in}{0.776691in}%
\pgfsys@useobject{currentmarker}{}%
\end{pgfscope}%
\end{pgfscope}%
\begin{pgfscope}%
\definecolor{textcolor}{rgb}{0.000000,0.000000,0.000000}%
\pgfsetstrokecolor{textcolor}%
\pgfsetfillcolor{textcolor}%
\pgftext[x=0.246975in, y=0.742934in, left, base]{\color{textcolor}\fontsize{7.000000}{8.400000}\selectfont \(\displaystyle {20000}\)}%
\end{pgfscope}%
\begin{pgfscope}%
\pgfsetbuttcap%
\pgfsetroundjoin%
\definecolor{currentfill}{rgb}{0.000000,0.000000,0.000000}%
\pgfsetfillcolor{currentfill}%
\pgfsetlinewidth{0.803000pt}%
\definecolor{currentstroke}{rgb}{0.000000,0.000000,0.000000}%
\pgfsetstrokecolor{currentstroke}%
\pgfsetdash{}{0pt}%
\pgfsys@defobject{currentmarker}{\pgfqpoint{-0.048611in}{0.000000in}}{\pgfqpoint{-0.000000in}{0.000000in}}{%
\pgfpathmoveto{\pgfqpoint{-0.000000in}{0.000000in}}%
\pgfpathlineto{\pgfqpoint{-0.048611in}{0.000000in}}%
\pgfusepath{stroke,fill}%
}%
\begin{pgfscope}%
\pgfsys@transformshift{0.621012in}{1.020716in}%
\pgfsys@useobject{currentmarker}{}%
\end{pgfscope}%
\end{pgfscope}%
\begin{pgfscope}%
\definecolor{textcolor}{rgb}{0.000000,0.000000,0.000000}%
\pgfsetstrokecolor{textcolor}%
\pgfsetfillcolor{textcolor}%
\pgftext[x=0.246975in, y=0.986958in, left, base]{\color{textcolor}\fontsize{7.000000}{8.400000}\selectfont \(\displaystyle {30000}\)}%
\end{pgfscope}%
\begin{pgfscope}%
\pgfsetbuttcap%
\pgfsetroundjoin%
\definecolor{currentfill}{rgb}{0.000000,0.000000,0.000000}%
\pgfsetfillcolor{currentfill}%
\pgfsetlinewidth{0.803000pt}%
\definecolor{currentstroke}{rgb}{0.000000,0.000000,0.000000}%
\pgfsetstrokecolor{currentstroke}%
\pgfsetdash{}{0pt}%
\pgfsys@defobject{currentmarker}{\pgfqpoint{-0.048611in}{0.000000in}}{\pgfqpoint{-0.000000in}{0.000000in}}{%
\pgfpathmoveto{\pgfqpoint{-0.000000in}{0.000000in}}%
\pgfpathlineto{\pgfqpoint{-0.048611in}{0.000000in}}%
\pgfusepath{stroke,fill}%
}%
\begin{pgfscope}%
\pgfsys@transformshift{0.621012in}{1.264741in}%
\pgfsys@useobject{currentmarker}{}%
\end{pgfscope}%
\end{pgfscope}%
\begin{pgfscope}%
\definecolor{textcolor}{rgb}{0.000000,0.000000,0.000000}%
\pgfsetstrokecolor{textcolor}%
\pgfsetfillcolor{textcolor}%
\pgftext[x=0.246975in, y=1.230983in, left, base]{\color{textcolor}\fontsize{7.000000}{8.400000}\selectfont \(\displaystyle {40000}\)}%
\end{pgfscope}%
\begin{pgfscope}%
\definecolor{textcolor}{rgb}{0.000000,0.000000,0.000000}%
\pgfsetstrokecolor{textcolor}%
\pgfsetfillcolor{textcolor}%
\pgftext[x=0.191419in,y=0.841821in,,bottom,rotate=90.000000]{\color{textcolor}\fontsize{7.000000}{8.400000}\selectfont Number of operations}%
\end{pgfscope}%
\begin{pgfscope}%
\pgfsetrectcap%
\pgfsetmiterjoin%
\pgfsetlinewidth{0.803000pt}%
\definecolor{currentstroke}{rgb}{0.000000,0.000000,0.000000}%
\pgfsetstrokecolor{currentstroke}%
\pgfsetdash{}{0pt}%
\pgfpathmoveto{\pgfqpoint{0.621012in}{0.288642in}}%
\pgfpathlineto{\pgfqpoint{0.621012in}{1.395000in}}%
\pgfusepath{stroke}%
\end{pgfscope}%
\begin{pgfscope}%
\pgfsetrectcap%
\pgfsetmiterjoin%
\pgfsetlinewidth{0.803000pt}%
\definecolor{currentstroke}{rgb}{0.000000,0.000000,0.000000}%
\pgfsetstrokecolor{currentstroke}%
\pgfsetdash{}{0pt}%
\pgfpathmoveto{\pgfqpoint{1.895000in}{0.288642in}}%
\pgfpathlineto{\pgfqpoint{1.895000in}{1.395000in}}%
\pgfusepath{stroke}%
\end{pgfscope}%
\begin{pgfscope}%
\pgfsetrectcap%
\pgfsetmiterjoin%
\pgfsetlinewidth{0.803000pt}%
\definecolor{currentstroke}{rgb}{0.000000,0.000000,0.000000}%
\pgfsetstrokecolor{currentstroke}%
\pgfsetdash{}{0pt}%
\pgfpathmoveto{\pgfqpoint{0.621012in}{0.288642in}}%
\pgfpathlineto{\pgfqpoint{1.895000in}{0.288642in}}%
\pgfusepath{stroke}%
\end{pgfscope}%
\begin{pgfscope}%
\pgfsetrectcap%
\pgfsetmiterjoin%
\pgfsetlinewidth{0.803000pt}%
\definecolor{currentstroke}{rgb}{0.000000,0.000000,0.000000}%
\pgfsetstrokecolor{currentstroke}%
\pgfsetdash{}{0pt}%
\pgfpathmoveto{\pgfqpoint{0.621012in}{1.395000in}}%
\pgfpathlineto{\pgfqpoint{1.895000in}{1.395000in}}%
\pgfusepath{stroke}%
\end{pgfscope}%
\end{pgfpicture}%
\makeatother%
\endgroup%

%% file: graphics/atlas_both_hands.pgf
%% Creator: Matplotlib, PGF backend
%%
%% To include the figure in your LaTeX document, write
%%   \input{<filename>.pgf}
%%
%% Make sure the required packages are loaded in your preamble
%%   \usepackage{pgf}
%%
%% Figures using additional raster images can only be included by \input if
%% they are in the same directory as the main LaTeX file. For loading figures
%% from other directories you can use the `import` package
%% and then include the figures with
%%   \import{<path to file>}{<filename>.pgf}
%%
%% Matplotlib used the following preamble
%%
\begingroup%
\makeatletter%
\begin{pgfpicture}%
\pgfpathrectangle{\pgfpointorigin}{\pgfqpoint{2.000000in}{1.500000in}}%
\pgfusepath{use as bounding box, clip}%
\begin{pgfscope}%
\pgfsetbuttcap%
\pgfsetmiterjoin%
\definecolor{currentfill}{rgb}{1.000000,1.000000,1.000000}%
\pgfsetfillcolor{currentfill}%
\pgfsetlinewidth{0.000000pt}%
\definecolor{currentstroke}{rgb}{1.000000,1.000000,1.000000}%
\pgfsetstrokecolor{currentstroke}%
\pgfsetdash{}{0pt}%
\pgfpathmoveto{\pgfqpoint{0.000000in}{0.000000in}}%
\pgfpathlineto{\pgfqpoint{2.000000in}{0.000000in}}%
\pgfpathlineto{\pgfqpoint{2.000000in}{1.500000in}}%
\pgfpathlineto{\pgfqpoint{0.000000in}{1.500000in}}%
\pgfpathclose%
\pgfusepath{fill}%
\end{pgfscope}%
\begin{pgfscope}%
\pgfsetbuttcap%
\pgfsetmiterjoin%
\definecolor{currentfill}{rgb}{1.000000,1.000000,1.000000}%
\pgfsetfillcolor{currentfill}%
\pgfsetlinewidth{0.000000pt}%
\definecolor{currentstroke}{rgb}{0.000000,0.000000,0.000000}%
\pgfsetstrokecolor{currentstroke}%
\pgfsetstrokeopacity{0.000000}%
\pgfsetdash{}{0pt}%
\pgfpathmoveto{\pgfqpoint{0.621012in}{0.288642in}}%
\pgfpathlineto{\pgfqpoint{1.895000in}{0.288642in}}%
\pgfpathlineto{\pgfqpoint{1.895000in}{1.395000in}}%
\pgfpathlineto{\pgfqpoint{0.621012in}{1.395000in}}%
\pgfpathclose%
\pgfusepath{fill}%
\end{pgfscope}%
\begin{pgfscope}%
\pgfpathrectangle{\pgfqpoint{0.621012in}{0.288642in}}{\pgfqpoint{1.273988in}{1.106358in}}%
\pgfusepath{clip}%
\pgfsetbuttcap%
\pgfsetmiterjoin%
\definecolor{currentfill}{rgb}{0.000000,1.000000,1.000000}%
\pgfsetfillcolor{currentfill}%
\pgfsetlinewidth{0.000000pt}%
\definecolor{currentstroke}{rgb}{0.000000,0.000000,0.000000}%
\pgfsetstrokecolor{currentstroke}%
\pgfsetstrokeopacity{0.000000}%
\pgfsetdash{}{0pt}%
\pgfpathmoveto{\pgfqpoint{0.678920in}{0.288642in}}%
\pgfpathlineto{\pgfqpoint{0.922746in}{0.288642in}}%
\pgfpathlineto{\pgfqpoint{0.922746in}{0.670084in}}%
\pgfpathlineto{\pgfqpoint{0.678920in}{0.670084in}}%
\pgfpathclose%
\pgfusepath{fill}%
\end{pgfscope}%
\begin{pgfscope}%
\pgfpathrectangle{\pgfqpoint{0.621012in}{0.288642in}}{\pgfqpoint{1.273988in}{1.106358in}}%
\pgfusepath{clip}%
\pgfsetbuttcap%
\pgfsetmiterjoin%
\definecolor{currentfill}{rgb}{0.000000,0.000000,1.000000}%
\pgfsetfillcolor{currentfill}%
\pgfsetlinewidth{0.000000pt}%
\definecolor{currentstroke}{rgb}{0.000000,0.000000,0.000000}%
\pgfsetstrokecolor{currentstroke}%
\pgfsetstrokeopacity{0.000000}%
\pgfsetdash{}{0pt}%
\pgfpathmoveto{\pgfqpoint{0.983702in}{0.288642in}}%
\pgfpathlineto{\pgfqpoint{1.227528in}{0.288642in}}%
\pgfpathlineto{\pgfqpoint{1.227528in}{0.793575in}}%
\pgfpathlineto{\pgfqpoint{0.983702in}{0.793575in}}%
\pgfpathclose%
\pgfusepath{fill}%
\end{pgfscope}%
\begin{pgfscope}%
\pgfpathrectangle{\pgfqpoint{0.621012in}{0.288642in}}{\pgfqpoint{1.273988in}{1.106358in}}%
\pgfusepath{clip}%
\pgfsetbuttcap%
\pgfsetmiterjoin%
\definecolor{currentfill}{rgb}{1.000000,0.752941,0.796078}%
\pgfsetfillcolor{currentfill}%
\pgfsetlinewidth{0.000000pt}%
\definecolor{currentstroke}{rgb}{0.000000,0.000000,0.000000}%
\pgfsetstrokecolor{currentstroke}%
\pgfsetstrokeopacity{0.000000}%
\pgfsetdash{}{0pt}%
\pgfpathmoveto{\pgfqpoint{1.288484in}{0.288642in}}%
\pgfpathlineto{\pgfqpoint{1.532310in}{0.288642in}}%
\pgfpathlineto{\pgfqpoint{1.532310in}{0.885890in}}%
\pgfpathlineto{\pgfqpoint{1.288484in}{0.885890in}}%
\pgfpathclose%
\pgfusepath{fill}%
\end{pgfscope}%
\begin{pgfscope}%
\pgfpathrectangle{\pgfqpoint{0.621012in}{0.288642in}}{\pgfqpoint{1.273988in}{1.106358in}}%
\pgfusepath{clip}%
\pgfsetbuttcap%
\pgfsetmiterjoin%
\definecolor{currentfill}{rgb}{0.000000,0.501961,0.000000}%
\pgfsetfillcolor{currentfill}%
\pgfsetlinewidth{0.000000pt}%
\definecolor{currentstroke}{rgb}{0.000000,0.000000,0.000000}%
\pgfsetstrokecolor{currentstroke}%
\pgfsetstrokeopacity{0.000000}%
\pgfsetdash{}{0pt}%
\pgfpathmoveto{\pgfqpoint{1.593266in}{0.288642in}}%
\pgfpathlineto{\pgfqpoint{1.837091in}{0.288642in}}%
\pgfpathlineto{\pgfqpoint{1.837091in}{1.342316in}}%
\pgfpathlineto{\pgfqpoint{1.593266in}{1.342316in}}%
\pgfpathclose%
\pgfusepath{fill}%
\end{pgfscope}%
\begin{pgfscope}%
\pgfsetbuttcap%
\pgfsetroundjoin%
\definecolor{currentfill}{rgb}{0.000000,0.000000,0.000000}%
\pgfsetfillcolor{currentfill}%
\pgfsetlinewidth{0.803000pt}%
\definecolor{currentstroke}{rgb}{0.000000,0.000000,0.000000}%
\pgfsetstrokecolor{currentstroke}%
\pgfsetdash{}{0pt}%
\pgfsys@defobject{currentmarker}{\pgfqpoint{0.000000in}{-0.048611in}}{\pgfqpoint{0.000000in}{0.000000in}}{%
\pgfpathmoveto{\pgfqpoint{0.000000in}{0.000000in}}%
\pgfpathlineto{\pgfqpoint{0.000000in}{-0.048611in}}%
\pgfusepath{stroke,fill}%
}%
\begin{pgfscope}%
\pgfsys@transformshift{0.800833in}{0.288642in}%
\pgfsys@useobject{currentmarker}{}%
\end{pgfscope}%
\end{pgfscope}%
\begin{pgfscope}%
\definecolor{textcolor}{rgb}{0.000000,0.000000,0.000000}%
\pgfsetstrokecolor{textcolor}%
\pgfsetfillcolor{textcolor}%
\pgftext[x=0.800833in,y=0.191419in,,top]{\color{textcolor}\fontsize{7.000000}{8.400000}\selectfont PV-OSIMr}%
\end{pgfscope}%
\begin{pgfscope}%
\pgfsetbuttcap%
\pgfsetroundjoin%
\definecolor{currentfill}{rgb}{0.000000,0.000000,0.000000}%
\pgfsetfillcolor{currentfill}%
\pgfsetlinewidth{0.803000pt}%
\definecolor{currentstroke}{rgb}{0.000000,0.000000,0.000000}%
\pgfsetstrokecolor{currentstroke}%
\pgfsetdash{}{0pt}%
\pgfsys@defobject{currentmarker}{\pgfqpoint{0.000000in}{-0.048611in}}{\pgfqpoint{0.000000in}{0.000000in}}{%
\pgfpathmoveto{\pgfqpoint{0.000000in}{0.000000in}}%
\pgfpathlineto{\pgfqpoint{0.000000in}{-0.048611in}}%
\pgfusepath{stroke,fill}%
}%
\begin{pgfscope}%
\pgfsys@transformshift{1.105615in}{0.288642in}%
\pgfsys@useobject{currentmarker}{}%
\end{pgfscope}%
\end{pgfscope}%
\begin{pgfscope}%
\definecolor{textcolor}{rgb}{0.000000,0.000000,0.000000}%
\pgfsetstrokecolor{textcolor}%
\pgfsetfillcolor{textcolor}%
\pgftext[x=1.105615in,y=0.191419in,,top]{\color{textcolor}\fontsize{7.000000}{8.400000}\selectfont PV}%
\end{pgfscope}%
\begin{pgfscope}%
\pgfsetbuttcap%
\pgfsetroundjoin%
\definecolor{currentfill}{rgb}{0.000000,0.000000,0.000000}%
\pgfsetfillcolor{currentfill}%
\pgfsetlinewidth{0.803000pt}%
\definecolor{currentstroke}{rgb}{0.000000,0.000000,0.000000}%
\pgfsetstrokecolor{currentstroke}%
\pgfsetdash{}{0pt}%
\pgfsys@defobject{currentmarker}{\pgfqpoint{0.000000in}{-0.048611in}}{\pgfqpoint{0.000000in}{0.000000in}}{%
\pgfpathmoveto{\pgfqpoint{0.000000in}{0.000000in}}%
\pgfpathlineto{\pgfqpoint{0.000000in}{-0.048611in}}%
\pgfusepath{stroke,fill}%
}%
\begin{pgfscope}%
\pgfsys@transformshift{1.410397in}{0.288642in}%
\pgfsys@useobject{currentmarker}{}%
\end{pgfscope}%
\end{pgfscope}%
\begin{pgfscope}%
\definecolor{textcolor}{rgb}{0.000000,0.000000,0.000000}%
\pgfsetstrokecolor{textcolor}%
\pgfsetfillcolor{textcolor}%
\pgftext[x=1.410397in,y=0.191419in,,top]{\color{textcolor}\fontsize{7.000000}{8.400000}\selectfont EFPA}%
\end{pgfscope}%
\begin{pgfscope}%
\pgfsetbuttcap%
\pgfsetroundjoin%
\definecolor{currentfill}{rgb}{0.000000,0.000000,0.000000}%
\pgfsetfillcolor{currentfill}%
\pgfsetlinewidth{0.803000pt}%
\definecolor{currentstroke}{rgb}{0.000000,0.000000,0.000000}%
\pgfsetstrokecolor{currentstroke}%
\pgfsetdash{}{0pt}%
\pgfsys@defobject{currentmarker}{\pgfqpoint{0.000000in}{-0.048611in}}{\pgfqpoint{0.000000in}{0.000000in}}{%
\pgfpathmoveto{\pgfqpoint{0.000000in}{0.000000in}}%
\pgfpathlineto{\pgfqpoint{0.000000in}{-0.048611in}}%
\pgfusepath{stroke,fill}%
}%
\begin{pgfscope}%
\pgfsys@transformshift{1.715179in}{0.288642in}%
\pgfsys@useobject{currentmarker}{}%
\end{pgfscope}%
\end{pgfscope}%
\begin{pgfscope}%
\definecolor{textcolor}{rgb}{0.000000,0.000000,0.000000}%
\pgfsetstrokecolor{textcolor}%
\pgfsetfillcolor{textcolor}%
\pgftext[x=1.715179in,y=0.191419in,,top]{\color{textcolor}\fontsize{7.000000}{8.400000}\selectfont LTL}%
\end{pgfscope}%
\begin{pgfscope}%
\pgfsetbuttcap%
\pgfsetroundjoin%
\definecolor{currentfill}{rgb}{0.000000,0.000000,0.000000}%
\pgfsetfillcolor{currentfill}%
\pgfsetlinewidth{0.803000pt}%
\definecolor{currentstroke}{rgb}{0.000000,0.000000,0.000000}%
\pgfsetstrokecolor{currentstroke}%
\pgfsetdash{}{0pt}%
\pgfsys@defobject{currentmarker}{\pgfqpoint{-0.048611in}{0.000000in}}{\pgfqpoint{-0.000000in}{0.000000in}}{%
\pgfpathmoveto{\pgfqpoint{-0.000000in}{0.000000in}}%
\pgfpathlineto{\pgfqpoint{-0.048611in}{0.000000in}}%
\pgfusepath{stroke,fill}%
}%
\begin{pgfscope}%
\pgfsys@transformshift{0.621012in}{0.288642in}%
\pgfsys@useobject{currentmarker}{}%
\end{pgfscope}%
\end{pgfscope}%
\begin{pgfscope}%
\definecolor{textcolor}{rgb}{0.000000,0.000000,0.000000}%
\pgfsetstrokecolor{textcolor}%
\pgfsetfillcolor{textcolor}%
\pgftext[x=0.468427in, y=0.254884in, left, base]{\color{textcolor}\fontsize{7.000000}{8.400000}\selectfont \(\displaystyle {0}\)}%
\end{pgfscope}%
\begin{pgfscope}%
\pgfsetbuttcap%
\pgfsetroundjoin%
\definecolor{currentfill}{rgb}{0.000000,0.000000,0.000000}%
\pgfsetfillcolor{currentfill}%
\pgfsetlinewidth{0.803000pt}%
\definecolor{currentstroke}{rgb}{0.000000,0.000000,0.000000}%
\pgfsetstrokecolor{currentstroke}%
\pgfsetdash{}{0pt}%
\pgfsys@defobject{currentmarker}{\pgfqpoint{-0.048611in}{0.000000in}}{\pgfqpoint{-0.000000in}{0.000000in}}{%
\pgfpathmoveto{\pgfqpoint{-0.000000in}{0.000000in}}%
\pgfpathlineto{\pgfqpoint{-0.048611in}{0.000000in}}%
\pgfusepath{stroke,fill}%
}%
\begin{pgfscope}%
\pgfsys@transformshift{0.621012in}{0.554872in}%
\pgfsys@useobject{currentmarker}{}%
\end{pgfscope}%
\end{pgfscope}%
\begin{pgfscope}%
\definecolor{textcolor}{rgb}{0.000000,0.000000,0.000000}%
\pgfsetstrokecolor{textcolor}%
\pgfsetfillcolor{textcolor}%
\pgftext[x=0.246975in, y=0.521115in, left, base]{\color{textcolor}\fontsize{7.000000}{8.400000}\selectfont \(\displaystyle {20000}\)}%
\end{pgfscope}%
\begin{pgfscope}%
\pgfsetbuttcap%
\pgfsetroundjoin%
\definecolor{currentfill}{rgb}{0.000000,0.000000,0.000000}%
\pgfsetfillcolor{currentfill}%
\pgfsetlinewidth{0.803000pt}%
\definecolor{currentstroke}{rgb}{0.000000,0.000000,0.000000}%
\pgfsetstrokecolor{currentstroke}%
\pgfsetdash{}{0pt}%
\pgfsys@defobject{currentmarker}{\pgfqpoint{-0.048611in}{0.000000in}}{\pgfqpoint{-0.000000in}{0.000000in}}{%
\pgfpathmoveto{\pgfqpoint{-0.000000in}{0.000000in}}%
\pgfpathlineto{\pgfqpoint{-0.048611in}{0.000000in}}%
\pgfusepath{stroke,fill}%
}%
\begin{pgfscope}%
\pgfsys@transformshift{0.621012in}{0.821103in}%
\pgfsys@useobject{currentmarker}{}%
\end{pgfscope}%
\end{pgfscope}%
\begin{pgfscope}%
\definecolor{textcolor}{rgb}{0.000000,0.000000,0.000000}%
\pgfsetstrokecolor{textcolor}%
\pgfsetfillcolor{textcolor}%
\pgftext[x=0.246975in, y=0.787345in, left, base]{\color{textcolor}\fontsize{7.000000}{8.400000}\selectfont \(\displaystyle {40000}\)}%
\end{pgfscope}%
\begin{pgfscope}%
\pgfsetbuttcap%
\pgfsetroundjoin%
\definecolor{currentfill}{rgb}{0.000000,0.000000,0.000000}%
\pgfsetfillcolor{currentfill}%
\pgfsetlinewidth{0.803000pt}%
\definecolor{currentstroke}{rgb}{0.000000,0.000000,0.000000}%
\pgfsetstrokecolor{currentstroke}%
\pgfsetdash{}{0pt}%
\pgfsys@defobject{currentmarker}{\pgfqpoint{-0.048611in}{0.000000in}}{\pgfqpoint{-0.000000in}{0.000000in}}{%
\pgfpathmoveto{\pgfqpoint{-0.000000in}{0.000000in}}%
\pgfpathlineto{\pgfqpoint{-0.048611in}{0.000000in}}%
\pgfusepath{stroke,fill}%
}%
\begin{pgfscope}%
\pgfsys@transformshift{0.621012in}{1.087334in}%
\pgfsys@useobject{currentmarker}{}%
\end{pgfscope}%
\end{pgfscope}%
\begin{pgfscope}%
\definecolor{textcolor}{rgb}{0.000000,0.000000,0.000000}%
\pgfsetstrokecolor{textcolor}%
\pgfsetfillcolor{textcolor}%
\pgftext[x=0.246975in, y=1.053576in, left, base]{\color{textcolor}\fontsize{7.000000}{8.400000}\selectfont \(\displaystyle {60000}\)}%
\end{pgfscope}%
\begin{pgfscope}%
\pgfsetbuttcap%
\pgfsetroundjoin%
\definecolor{currentfill}{rgb}{0.000000,0.000000,0.000000}%
\pgfsetfillcolor{currentfill}%
\pgfsetlinewidth{0.803000pt}%
\definecolor{currentstroke}{rgb}{0.000000,0.000000,0.000000}%
\pgfsetstrokecolor{currentstroke}%
\pgfsetdash{}{0pt}%
\pgfsys@defobject{currentmarker}{\pgfqpoint{-0.048611in}{0.000000in}}{\pgfqpoint{-0.000000in}{0.000000in}}{%
\pgfpathmoveto{\pgfqpoint{-0.000000in}{0.000000in}}%
\pgfpathlineto{\pgfqpoint{-0.048611in}{0.000000in}}%
\pgfusepath{stroke,fill}%
}%
\begin{pgfscope}%
\pgfsys@transformshift{0.621012in}{1.353565in}%
\pgfsys@useobject{currentmarker}{}%
\end{pgfscope}%
\end{pgfscope}%
\begin{pgfscope}%
\definecolor{textcolor}{rgb}{0.000000,0.000000,0.000000}%
\pgfsetstrokecolor{textcolor}%
\pgfsetfillcolor{textcolor}%
\pgftext[x=0.246975in, y=1.319807in, left, base]{\color{textcolor}\fontsize{7.000000}{8.400000}\selectfont \(\displaystyle {80000}\)}%
\end{pgfscope}%
\begin{pgfscope}%
\definecolor{textcolor}{rgb}{0.000000,0.000000,0.000000}%
\pgfsetstrokecolor{textcolor}%
\pgfsetfillcolor{textcolor}%
\pgftext[x=0.191419in,y=0.841821in,,bottom,rotate=90.000000]{\color{textcolor}\fontsize{7.000000}{8.400000}\selectfont Number of operations}%
\end{pgfscope}%
\begin{pgfscope}%
\pgfsetrectcap%
\pgfsetmiterjoin%
\pgfsetlinewidth{0.803000pt}%
\definecolor{currentstroke}{rgb}{0.000000,0.000000,0.000000}%
\pgfsetstrokecolor{currentstroke}%
\pgfsetdash{}{0pt}%
\pgfpathmoveto{\pgfqpoint{0.621012in}{0.288642in}}%
\pgfpathlineto{\pgfqpoint{0.621012in}{1.395000in}}%
\pgfusepath{stroke}%
\end{pgfscope}%
\begin{pgfscope}%
\pgfsetrectcap%
\pgfsetmiterjoin%
\pgfsetlinewidth{0.803000pt}%
\definecolor{currentstroke}{rgb}{0.000000,0.000000,0.000000}%
\pgfsetstrokecolor{currentstroke}%
\pgfsetdash{}{0pt}%
\pgfpathmoveto{\pgfqpoint{1.895000in}{0.288642in}}%
\pgfpathlineto{\pgfqpoint{1.895000in}{1.395000in}}%
\pgfusepath{stroke}%
\end{pgfscope}%
\begin{pgfscope}%
\pgfsetrectcap%
\pgfsetmiterjoin%
\pgfsetlinewidth{0.803000pt}%
\definecolor{currentstroke}{rgb}{0.000000,0.000000,0.000000}%
\pgfsetstrokecolor{currentstroke}%
\pgfsetdash{}{0pt}%
\pgfpathmoveto{\pgfqpoint{0.621012in}{0.288642in}}%
\pgfpathlineto{\pgfqpoint{1.895000in}{0.288642in}}%
\pgfusepath{stroke}%
\end{pgfscope}%
\begin{pgfscope}%
\pgfsetrectcap%
\pgfsetmiterjoin%
\pgfsetlinewidth{0.803000pt}%
\definecolor{currentstroke}{rgb}{0.000000,0.000000,0.000000}%
\pgfsetstrokecolor{currentstroke}%
\pgfsetdash{}{0pt}%
\pgfpathmoveto{\pgfqpoint{0.621012in}{1.395000in}}%
\pgfpathlineto{\pgfqpoint{1.895000in}{1.395000in}}%
\pgfusepath{stroke}%
\end{pgfscope}%
\end{pgfpicture}%
\makeatother%
\endgroup%

%% file: graphics/atlas_standing_both_hands.pgf
%% Creator: Matplotlib, PGF backend
%%
%% To include the figure in your LaTeX document, write
%%   \input{<filename>.pgf}
%%
%% Make sure the required packages are loaded in your preamble
%%   \usepackage{pgf}
%%
%% Figures using additional raster images can only be included by \input if
%% they are in the same directory as the main LaTeX file. For loading figures
%% from other directories you can use the `import` package
%% and then include the figures with
%%   \import{<path to file>}{<filename>.pgf}
%%
%% Matplotlib used the following preamble
%%
\begingroup%
\makeatletter%
\begin{pgfpicture}%
\pgfpathrectangle{\pgfpointorigin}{\pgfqpoint{2.000000in}{1.500000in}}%
\pgfusepath{use as bounding box, clip}%
\begin{pgfscope}%
\pgfsetbuttcap%
\pgfsetmiterjoin%
\definecolor{currentfill}{rgb}{1.000000,1.000000,1.000000}%
\pgfsetfillcolor{currentfill}%
\pgfsetlinewidth{0.000000pt}%
\definecolor{currentstroke}{rgb}{1.000000,1.000000,1.000000}%
\pgfsetstrokecolor{currentstroke}%
\pgfsetdash{}{0pt}%
\pgfpathmoveto{\pgfqpoint{0.000000in}{0.000000in}}%
\pgfpathlineto{\pgfqpoint{2.000000in}{0.000000in}}%
\pgfpathlineto{\pgfqpoint{2.000000in}{1.500000in}}%
\pgfpathlineto{\pgfqpoint{0.000000in}{1.500000in}}%
\pgfpathclose%
\pgfusepath{fill}%
\end{pgfscope}%
\begin{pgfscope}%
\pgfsetbuttcap%
\pgfsetmiterjoin%
\definecolor{currentfill}{rgb}{1.000000,1.000000,1.000000}%
\pgfsetfillcolor{currentfill}%
\pgfsetlinewidth{0.000000pt}%
\definecolor{currentstroke}{rgb}{0.000000,0.000000,0.000000}%
\pgfsetstrokecolor{currentstroke}%
\pgfsetstrokeopacity{0.000000}%
\pgfsetdash{}{0pt}%
\pgfpathmoveto{\pgfqpoint{0.621012in}{0.288642in}}%
\pgfpathlineto{\pgfqpoint{1.895000in}{0.288642in}}%
\pgfpathlineto{\pgfqpoint{1.895000in}{1.395000in}}%
\pgfpathlineto{\pgfqpoint{0.621012in}{1.395000in}}%
\pgfpathclose%
\pgfusepath{fill}%
\end{pgfscope}%
\begin{pgfscope}%
\pgfpathrectangle{\pgfqpoint{0.621012in}{0.288642in}}{\pgfqpoint{1.273988in}{1.106358in}}%
\pgfusepath{clip}%
\pgfsetbuttcap%
\pgfsetmiterjoin%
\definecolor{currentfill}{rgb}{0.000000,1.000000,1.000000}%
\pgfsetfillcolor{currentfill}%
\pgfsetlinewidth{0.000000pt}%
\definecolor{currentstroke}{rgb}{0.000000,0.000000,0.000000}%
\pgfsetstrokecolor{currentstroke}%
\pgfsetstrokeopacity{0.000000}%
\pgfsetdash{}{0pt}%
\pgfpathmoveto{\pgfqpoint{0.678920in}{0.288642in}}%
\pgfpathlineto{\pgfqpoint{0.922746in}{0.288642in}}%
\pgfpathlineto{\pgfqpoint{0.922746in}{0.777870in}}%
\pgfpathlineto{\pgfqpoint{0.678920in}{0.777870in}}%
\pgfpathclose%
\pgfusepath{fill}%
\end{pgfscope}%
\begin{pgfscope}%
\pgfpathrectangle{\pgfqpoint{0.621012in}{0.288642in}}{\pgfqpoint{1.273988in}{1.106358in}}%
\pgfusepath{clip}%
\pgfsetbuttcap%
\pgfsetmiterjoin%
\definecolor{currentfill}{rgb}{0.000000,0.000000,1.000000}%
\pgfsetfillcolor{currentfill}%
\pgfsetlinewidth{0.000000pt}%
\definecolor{currentstroke}{rgb}{0.000000,0.000000,0.000000}%
\pgfsetstrokecolor{currentstroke}%
\pgfsetstrokeopacity{0.000000}%
\pgfsetdash{}{0pt}%
\pgfpathmoveto{\pgfqpoint{0.983702in}{0.288642in}}%
\pgfpathlineto{\pgfqpoint{1.227528in}{0.288642in}}%
\pgfpathlineto{\pgfqpoint{1.227528in}{0.841809in}}%
\pgfpathlineto{\pgfqpoint{0.983702in}{0.841809in}}%
\pgfpathclose%
\pgfusepath{fill}%
\end{pgfscope}%
\begin{pgfscope}%
\pgfpathrectangle{\pgfqpoint{0.621012in}{0.288642in}}{\pgfqpoint{1.273988in}{1.106358in}}%
\pgfusepath{clip}%
\pgfsetbuttcap%
\pgfsetmiterjoin%
\definecolor{currentfill}{rgb}{1.000000,0.752941,0.796078}%
\pgfsetfillcolor{currentfill}%
\pgfsetlinewidth{0.000000pt}%
\definecolor{currentstroke}{rgb}{0.000000,0.000000,0.000000}%
\pgfsetstrokecolor{currentstroke}%
\pgfsetstrokeopacity{0.000000}%
\pgfsetdash{}{0pt}%
\pgfpathmoveto{\pgfqpoint{1.288484in}{0.288642in}}%
\pgfpathlineto{\pgfqpoint{1.532310in}{0.288642in}}%
\pgfpathlineto{\pgfqpoint{1.532310in}{0.973138in}}%
\pgfpathlineto{\pgfqpoint{1.288484in}{0.973138in}}%
\pgfpathclose%
\pgfusepath{fill}%
\end{pgfscope}%
\begin{pgfscope}%
\pgfpathrectangle{\pgfqpoint{0.621012in}{0.288642in}}{\pgfqpoint{1.273988in}{1.106358in}}%
\pgfusepath{clip}%
\pgfsetbuttcap%
\pgfsetmiterjoin%
\definecolor{currentfill}{rgb}{0.000000,0.501961,0.000000}%
\pgfsetfillcolor{currentfill}%
\pgfsetlinewidth{0.000000pt}%
\definecolor{currentstroke}{rgb}{0.000000,0.000000,0.000000}%
\pgfsetstrokecolor{currentstroke}%
\pgfsetstrokeopacity{0.000000}%
\pgfsetdash{}{0pt}%
\pgfpathmoveto{\pgfqpoint{1.593266in}{0.288642in}}%
\pgfpathlineto{\pgfqpoint{1.837091in}{0.288642in}}%
\pgfpathlineto{\pgfqpoint{1.837091in}{1.342316in}}%
\pgfpathlineto{\pgfqpoint{1.593266in}{1.342316in}}%
\pgfpathclose%
\pgfusepath{fill}%
\end{pgfscope}%
\begin{pgfscope}%
\pgfsetbuttcap%
\pgfsetroundjoin%
\definecolor{currentfill}{rgb}{0.000000,0.000000,0.000000}%
\pgfsetfillcolor{currentfill}%
\pgfsetlinewidth{0.803000pt}%
\definecolor{currentstroke}{rgb}{0.000000,0.000000,0.000000}%
\pgfsetstrokecolor{currentstroke}%
\pgfsetdash{}{0pt}%
\pgfsys@defobject{currentmarker}{\pgfqpoint{0.000000in}{-0.048611in}}{\pgfqpoint{0.000000in}{0.000000in}}{%
\pgfpathmoveto{\pgfqpoint{0.000000in}{0.000000in}}%
\pgfpathlineto{\pgfqpoint{0.000000in}{-0.048611in}}%
\pgfusepath{stroke,fill}%
}%
\begin{pgfscope}%
\pgfsys@transformshift{0.800833in}{0.288642in}%
\pgfsys@useobject{currentmarker}{}%
\end{pgfscope}%
\end{pgfscope}%
\begin{pgfscope}%
\definecolor{textcolor}{rgb}{0.000000,0.000000,0.000000}%
\pgfsetstrokecolor{textcolor}%
\pgfsetfillcolor{textcolor}%
\pgftext[x=0.800833in,y=0.191419in,,top]{\color{textcolor}\fontsize{7.000000}{8.400000}\selectfont PV-OSIMr}%
\end{pgfscope}%
\begin{pgfscope}%
\pgfsetbuttcap%
\pgfsetroundjoin%
\definecolor{currentfill}{rgb}{0.000000,0.000000,0.000000}%
\pgfsetfillcolor{currentfill}%
\pgfsetlinewidth{0.803000pt}%
\definecolor{currentstroke}{rgb}{0.000000,0.000000,0.000000}%
\pgfsetstrokecolor{currentstroke}%
\pgfsetdash{}{0pt}%
\pgfsys@defobject{currentmarker}{\pgfqpoint{0.000000in}{-0.048611in}}{\pgfqpoint{0.000000in}{0.000000in}}{%
\pgfpathmoveto{\pgfqpoint{0.000000in}{0.000000in}}%
\pgfpathlineto{\pgfqpoint{0.000000in}{-0.048611in}}%
\pgfusepath{stroke,fill}%
}%
\begin{pgfscope}%
\pgfsys@transformshift{1.105615in}{0.288642in}%
\pgfsys@useobject{currentmarker}{}%
\end{pgfscope}%
\end{pgfscope}%
\begin{pgfscope}%
\definecolor{textcolor}{rgb}{0.000000,0.000000,0.000000}%
\pgfsetstrokecolor{textcolor}%
\pgfsetfillcolor{textcolor}%
\pgftext[x=1.105615in,y=0.191419in,,top]{\color{textcolor}\fontsize{7.000000}{8.400000}\selectfont PV}%
\end{pgfscope}%
\begin{pgfscope}%
\pgfsetbuttcap%
\pgfsetroundjoin%
\definecolor{currentfill}{rgb}{0.000000,0.000000,0.000000}%
\pgfsetfillcolor{currentfill}%
\pgfsetlinewidth{0.803000pt}%
\definecolor{currentstroke}{rgb}{0.000000,0.000000,0.000000}%
\pgfsetstrokecolor{currentstroke}%
\pgfsetdash{}{0pt}%
\pgfsys@defobject{currentmarker}{\pgfqpoint{0.000000in}{-0.048611in}}{\pgfqpoint{0.000000in}{0.000000in}}{%
\pgfpathmoveto{\pgfqpoint{0.000000in}{0.000000in}}%
\pgfpathlineto{\pgfqpoint{0.000000in}{-0.048611in}}%
\pgfusepath{stroke,fill}%
}%
\begin{pgfscope}%
\pgfsys@transformshift{1.410397in}{0.288642in}%
\pgfsys@useobject{currentmarker}{}%
\end{pgfscope}%
\end{pgfscope}%
\begin{pgfscope}%
\definecolor{textcolor}{rgb}{0.000000,0.000000,0.000000}%
\pgfsetstrokecolor{textcolor}%
\pgfsetfillcolor{textcolor}%
\pgftext[x=1.410397in,y=0.191419in,,top]{\color{textcolor}\fontsize{7.000000}{8.400000}\selectfont EFPA}%
\end{pgfscope}%
\begin{pgfscope}%
\pgfsetbuttcap%
\pgfsetroundjoin%
\definecolor{currentfill}{rgb}{0.000000,0.000000,0.000000}%
\pgfsetfillcolor{currentfill}%
\pgfsetlinewidth{0.803000pt}%
\definecolor{currentstroke}{rgb}{0.000000,0.000000,0.000000}%
\pgfsetstrokecolor{currentstroke}%
\pgfsetdash{}{0pt}%
\pgfsys@defobject{currentmarker}{\pgfqpoint{0.000000in}{-0.048611in}}{\pgfqpoint{0.000000in}{0.000000in}}{%
\pgfpathmoveto{\pgfqpoint{0.000000in}{0.000000in}}%
\pgfpathlineto{\pgfqpoint{0.000000in}{-0.048611in}}%
\pgfusepath{stroke,fill}%
}%
\begin{pgfscope}%
\pgfsys@transformshift{1.715179in}{0.288642in}%
\pgfsys@useobject{currentmarker}{}%
\end{pgfscope}%
\end{pgfscope}%
\begin{pgfscope}%
\definecolor{textcolor}{rgb}{0.000000,0.000000,0.000000}%
\pgfsetstrokecolor{textcolor}%
\pgfsetfillcolor{textcolor}%
\pgftext[x=1.715179in,y=0.191419in,,top]{\color{textcolor}\fontsize{7.000000}{8.400000}\selectfont LTL}%
\end{pgfscope}%
\begin{pgfscope}%
\pgfsetbuttcap%
\pgfsetroundjoin%
\definecolor{currentfill}{rgb}{0.000000,0.000000,0.000000}%
\pgfsetfillcolor{currentfill}%
\pgfsetlinewidth{0.803000pt}%
\definecolor{currentstroke}{rgb}{0.000000,0.000000,0.000000}%
\pgfsetstrokecolor{currentstroke}%
\pgfsetdash{}{0pt}%
\pgfsys@defobject{currentmarker}{\pgfqpoint{-0.048611in}{0.000000in}}{\pgfqpoint{-0.000000in}{0.000000in}}{%
\pgfpathmoveto{\pgfqpoint{-0.000000in}{0.000000in}}%
\pgfpathlineto{\pgfqpoint{-0.048611in}{0.000000in}}%
\pgfusepath{stroke,fill}%
}%
\begin{pgfscope}%
\pgfsys@transformshift{0.621012in}{0.288642in}%
\pgfsys@useobject{currentmarker}{}%
\end{pgfscope}%
\end{pgfscope}%
\begin{pgfscope}%
\definecolor{textcolor}{rgb}{0.000000,0.000000,0.000000}%
\pgfsetstrokecolor{textcolor}%
\pgfsetfillcolor{textcolor}%
\pgftext[x=0.468427in, y=0.254884in, left, base]{\color{textcolor}\fontsize{7.000000}{8.400000}\selectfont \(\displaystyle {0}\)}%
\end{pgfscope}%
\begin{pgfscope}%
\pgfsetbuttcap%
\pgfsetroundjoin%
\definecolor{currentfill}{rgb}{0.000000,0.000000,0.000000}%
\pgfsetfillcolor{currentfill}%
\pgfsetlinewidth{0.803000pt}%
\definecolor{currentstroke}{rgb}{0.000000,0.000000,0.000000}%
\pgfsetstrokecolor{currentstroke}%
\pgfsetdash{}{0pt}%
\pgfsys@defobject{currentmarker}{\pgfqpoint{-0.048611in}{0.000000in}}{\pgfqpoint{-0.000000in}{0.000000in}}{%
\pgfpathmoveto{\pgfqpoint{-0.000000in}{0.000000in}}%
\pgfpathlineto{\pgfqpoint{-0.048611in}{0.000000in}}%
\pgfusepath{stroke,fill}%
}%
\begin{pgfscope}%
\pgfsys@transformshift{0.621012in}{0.529149in}%
\pgfsys@useobject{currentmarker}{}%
\end{pgfscope}%
\end{pgfscope}%
\begin{pgfscope}%
\definecolor{textcolor}{rgb}{0.000000,0.000000,0.000000}%
\pgfsetstrokecolor{textcolor}%
\pgfsetfillcolor{textcolor}%
\pgftext[x=0.246975in, y=0.495391in, left, base]{\color{textcolor}\fontsize{7.000000}{8.400000}\selectfont \(\displaystyle {20000}\)}%
\end{pgfscope}%
\begin{pgfscope}%
\pgfsetbuttcap%
\pgfsetroundjoin%
\definecolor{currentfill}{rgb}{0.000000,0.000000,0.000000}%
\pgfsetfillcolor{currentfill}%
\pgfsetlinewidth{0.803000pt}%
\definecolor{currentstroke}{rgb}{0.000000,0.000000,0.000000}%
\pgfsetstrokecolor{currentstroke}%
\pgfsetdash{}{0pt}%
\pgfsys@defobject{currentmarker}{\pgfqpoint{-0.048611in}{0.000000in}}{\pgfqpoint{-0.000000in}{0.000000in}}{%
\pgfpathmoveto{\pgfqpoint{-0.000000in}{0.000000in}}%
\pgfpathlineto{\pgfqpoint{-0.048611in}{0.000000in}}%
\pgfusepath{stroke,fill}%
}%
\begin{pgfscope}%
\pgfsys@transformshift{0.621012in}{0.769656in}%
\pgfsys@useobject{currentmarker}{}%
\end{pgfscope}%
\end{pgfscope}%
\begin{pgfscope}%
\definecolor{textcolor}{rgb}{0.000000,0.000000,0.000000}%
\pgfsetstrokecolor{textcolor}%
\pgfsetfillcolor{textcolor}%
\pgftext[x=0.246975in, y=0.735899in, left, base]{\color{textcolor}\fontsize{7.000000}{8.400000}\selectfont \(\displaystyle {40000}\)}%
\end{pgfscope}%
\begin{pgfscope}%
\pgfsetbuttcap%
\pgfsetroundjoin%
\definecolor{currentfill}{rgb}{0.000000,0.000000,0.000000}%
\pgfsetfillcolor{currentfill}%
\pgfsetlinewidth{0.803000pt}%
\definecolor{currentstroke}{rgb}{0.000000,0.000000,0.000000}%
\pgfsetstrokecolor{currentstroke}%
\pgfsetdash{}{0pt}%
\pgfsys@defobject{currentmarker}{\pgfqpoint{-0.048611in}{0.000000in}}{\pgfqpoint{-0.000000in}{0.000000in}}{%
\pgfpathmoveto{\pgfqpoint{-0.000000in}{0.000000in}}%
\pgfpathlineto{\pgfqpoint{-0.048611in}{0.000000in}}%
\pgfusepath{stroke,fill}%
}%
\begin{pgfscope}%
\pgfsys@transformshift{0.621012in}{1.010164in}%
\pgfsys@useobject{currentmarker}{}%
\end{pgfscope}%
\end{pgfscope}%
\begin{pgfscope}%
\definecolor{textcolor}{rgb}{0.000000,0.000000,0.000000}%
\pgfsetstrokecolor{textcolor}%
\pgfsetfillcolor{textcolor}%
\pgftext[x=0.246975in, y=0.976406in, left, base]{\color{textcolor}\fontsize{7.000000}{8.400000}\selectfont \(\displaystyle {60000}\)}%
\end{pgfscope}%
\begin{pgfscope}%
\pgfsetbuttcap%
\pgfsetroundjoin%
\definecolor{currentfill}{rgb}{0.000000,0.000000,0.000000}%
\pgfsetfillcolor{currentfill}%
\pgfsetlinewidth{0.803000pt}%
\definecolor{currentstroke}{rgb}{0.000000,0.000000,0.000000}%
\pgfsetstrokecolor{currentstroke}%
\pgfsetdash{}{0pt}%
\pgfsys@defobject{currentmarker}{\pgfqpoint{-0.048611in}{0.000000in}}{\pgfqpoint{-0.000000in}{0.000000in}}{%
\pgfpathmoveto{\pgfqpoint{-0.000000in}{0.000000in}}%
\pgfpathlineto{\pgfqpoint{-0.048611in}{0.000000in}}%
\pgfusepath{stroke,fill}%
}%
\begin{pgfscope}%
\pgfsys@transformshift{0.621012in}{1.250671in}%
\pgfsys@useobject{currentmarker}{}%
\end{pgfscope}%
\end{pgfscope}%
\begin{pgfscope}%
\definecolor{textcolor}{rgb}{0.000000,0.000000,0.000000}%
\pgfsetstrokecolor{textcolor}%
\pgfsetfillcolor{textcolor}%
\pgftext[x=0.246975in, y=1.216913in, left, base]{\color{textcolor}\fontsize{7.000000}{8.400000}\selectfont \(\displaystyle {80000}\)}%
\end{pgfscope}%
\begin{pgfscope}%
\definecolor{textcolor}{rgb}{0.000000,0.000000,0.000000}%
\pgfsetstrokecolor{textcolor}%
\pgfsetfillcolor{textcolor}%
\pgftext[x=0.191419in,y=0.841821in,,bottom,rotate=90.000000]{\color{textcolor}\fontsize{7.000000}{8.400000}\selectfont Number of operations}%
\end{pgfscope}%
\begin{pgfscope}%
\pgfsetrectcap%
\pgfsetmiterjoin%
\pgfsetlinewidth{0.803000pt}%
\definecolor{currentstroke}{rgb}{0.000000,0.000000,0.000000}%
\pgfsetstrokecolor{currentstroke}%
\pgfsetdash{}{0pt}%
\pgfpathmoveto{\pgfqpoint{0.621012in}{0.288642in}}%
\pgfpathlineto{\pgfqpoint{0.621012in}{1.395000in}}%
\pgfusepath{stroke}%
\end{pgfscope}%
\begin{pgfscope}%
\pgfsetrectcap%
\pgfsetmiterjoin%
\pgfsetlinewidth{0.803000pt}%
\definecolor{currentstroke}{rgb}{0.000000,0.000000,0.000000}%
\pgfsetstrokecolor{currentstroke}%
\pgfsetdash{}{0pt}%
\pgfpathmoveto{\pgfqpoint{1.895000in}{0.288642in}}%
\pgfpathlineto{\pgfqpoint{1.895000in}{1.395000in}}%
\pgfusepath{stroke}%
\end{pgfscope}%
\begin{pgfscope}%
\pgfsetrectcap%
\pgfsetmiterjoin%
\pgfsetlinewidth{0.803000pt}%
\definecolor{currentstroke}{rgb}{0.000000,0.000000,0.000000}%
\pgfsetstrokecolor{currentstroke}%
\pgfsetdash{}{0pt}%
\pgfpathmoveto{\pgfqpoint{0.621012in}{0.288642in}}%
\pgfpathlineto{\pgfqpoint{1.895000in}{0.288642in}}%
\pgfusepath{stroke}%
\end{pgfscope}%
\begin{pgfscope}%
\pgfsetrectcap%
\pgfsetmiterjoin%
\pgfsetlinewidth{0.803000pt}%
\definecolor{currentstroke}{rgb}{0.000000,0.000000,0.000000}%
\pgfsetstrokecolor{currentstroke}%
\pgfsetdash{}{0pt}%
\pgfpathmoveto{\pgfqpoint{0.621012in}{1.395000in}}%
\pgfpathlineto{\pgfqpoint{1.895000in}{1.395000in}}%
\pgfusepath{stroke}%
\end{pgfscope}%
\end{pgfpicture}%
\makeatother%
\endgroup%

%% file: conclusions.tex
We presented PV-OSIMr, an algorithm with the lowest known computational complexity of $O(n + m^2)$ to compute the Delassus matrix or the inverse OSIM matrix, which is a fundamental quantity with various simulation and control applications in a computationally demanding setting. Our algorithm is conceived by re-ordering and eliminating unnessary computations from the PV-OSIM algorithm. The PV-OSIMr algorithm not only has the lowest computational complexity, but is also found to lead to up to 24\% improvement over practical examples involving the Atlas robot. Furthermore, since computing the inverse OSIM matrix requires considering all the $n$ joints and populating a dense matrix of size $m \times m$, the achieved computational complexity of $O(n + m^2)$ can be considered asymptotically optimal.